# GEOMETRIC PRIMITIVE FEATURE EXTRACTION – CONCEPTS, ALGORITHMS, AND APPLICATIONS

## DILIP KUMAR PRASAD

## School of Computer Engineering

A Thesis submitted to the Nanyang Technological University
in fulfillment of the requirement for the degree of
Doctor of Philosophy

2012

# Abstract

This thesis presents important insights and concepts related to the topic of the extraction of geometric primitives from the edge contours of digital images. Three specific problems related to this topic have been studied, viz., polygonal approximation of digital curves, tangent estimation of digital curves, and ellipse fitting anddetection from digital curves. For the problem of polygonal approximation, two fundamental problems have been addressed. First, the nature of the performance evaluation metrics in relation to the local and global fitting characteristics has been studied. Second, an explicit error bound of the error introduced by digitizing a continuous line segment has been derived and used to propose a generic non-heuristic parameter independent framework which can be used in several dominant point detection methods. For the problem of tangent estimation for digital curves, a simple method of tangent estimation has been proposed. It is shown that the method has a definite upper bound of the error for conic digital curves. It has been shown that the method performs better than almost all (seventy two) existing tangent estimation methods for conic as well as several non-conic digital curves. For the problem of fitting ellipses on digital curves, a geometric distance minimization model has been considered. An unconstrained, linear, non-iterative, and numerically stable ellipse fitting method has been proposed and it has been shown that the proposed method has better selectivity for elliptic digital curves (high true positive and low false positive) as compared to several other ellipse fitting methods. For the problem of detecting ellipses in a set of digital curves, several innovative and fast pre-processing, grouping, and hypotheses evaluation concepts applicable for digital curves have been proposed and combined to form an ellipse detection method. Performance of the proposed ellipse detection method is better than several recent ellipse detection methods and close to the ideal case. All algorithms presented in this thesis have been developed using detailed mathematical analysis of the discrete geometry involved. The validity of these methods has been verified using rigorous mathematical analysis, numerical experiments in various difficult scenario, and extensive testing on large image datasets of practical importance. The utility of these algorithms has been shown using three practical applications related to image processing for robotics, medical image processing, and object and face detection.

# Acknowledgements

कर्मण्येवाधिकारस्ते मा फलेषु कदाचन।
मा कर्मफलहेतुर्भूर्मा ते सङ्गोऽस्त्वकर्मणि।

*"Yours is the right to perform, not the right to expect (the fruits of your actions). Do not let the fruit be the purpose of your actions. By being driven by the fruits of your actions, do not lead yourself to inaction."*

*-Bhagvad Gita (Chapter 2.47)*

In this proud moment of my life, when I am writing the acknowledgement for my PhD dissertation, I pay my respects first to all the wise people in my life like my parents, my teachers, and the Bhagvad Gita for reiterating the above quote time and again in my life. I pay my humble respects to my teachers. Specifically, I recall the teachers who left indelible marks in my life: my parents, my elder sister Kaushalya, Master Ji, Ratan Maharaj (little monk) from Ramakrishna Ashram, Mrs. Gopa Gupta, Mrs. Tapati Das, Mrs. S. Mukherjee, Mr. Lal Bahadur Shastri, Dr. Subrata Bose, Mr. Arup Pal, Dr. Munindra Prakash, Dr. A Chattopadhayay, Dr. D.P Mukherjee, Dr. Maylor K.H. Leung, Dr. Janusz Starzyk, Dr. Wendy Torrance, and Dr. Chai Quek (in the chronological order of their first impact in my life). Life is a great teacher, and I salute to the provider of life and my little Ganesha.

I especially thank Mrs. Gopa Gupta, Dr. Subrata Bose, Mr. Girish Mishra, Dr. Janusz Starzyk, Dr. Wendy Torrance, Dr. Maylor Leung, and Dr. Chai Quek, for mentoring a vagabond and stubborn person as me. I am very grateful to Dr. Maylor Leung for giving me the independence to pursue independent research on the topic of discrete geometry and for scrutinizing several lengthy derivations very patiently and persistently. I respect his philosophy of never compromising on the quality of work and this has motivated me to keep high expectation from myself in terms of the quality of work.

I take this opportunity to thank Nanyang Technological University and Ministry of Education, Singapore for providing me with the necessary infrastructure and funding for my doctoral studies. I thank Dr. Jiang Xudong, Dr. Chai Quek, Dr. Wolfgang Mueller, Dr. Janusz Starzyk, Dr. Philip Fu, and Dr. Mao Kezhi for conducting exceptional graduate modules. I thank Centre of Life Sciences, National University of Singapore for allowing me to fulfill my desire to acquire more knowledge about the field of Neuroscience. I thank Dr. Dale Purves, Dr. George Augustine, Dr. Shih-Cheng Yen, Dr. Thorsten Wohland, Dr. Ji Hui,Dr. Soong Tuck Wah, Dr. Antonius VanDongen, Dr. Marc Fivaz, Dr. Hongyan Wang, Dr. Fengwei Yu, Dr. Eyleen Goh, and Dr. Suresh Jesuthasan for conducting the post graduate modules on Vision and Perception, Neuronal Signaling, Developmental Neurobiology and Neurological and Behavioural Disorder and making me an active participant of these modules. I thank Dr. Janusz Starzyk and Dr. Krishna Agarwal for being a constant source of inspiration for me. I thank Dr. Siu-Yeung Cho and Mrs. Christina Lee for providing me with a good and hassle-free laboratory environment.

I had the privilege of having great friends in my life, who make the life a rich and happy experience. I begin with the friends who helped me directly or indirectly with my doctoral studies. I thank Dr. Krishna Agarwal for motivating and helping me to join the doctoral program. I thank Dr. Raj Gupta, Deepak Subramanian, Dr. Ashis Mallick, and Dr. Ranjan Das for being and remaining very good friends in odd and even times.

Other friends in my research lab, who deserve special mention, are Dr. Tang Chaoying, Dr. Hengyi Zhang, Dr. Atiqur Rahman, Dr. Pravin Kakar, and Ha Thanh. My Ph.D. life would not have been as much fun without Melanie, Sulley Goh, Ngyuen Chi, Dr. Balaji Gokulan, Dr. Dwarikanath Mohapatra, Mimi, Kabita bhabhi, and fellow ISMites (Abhishek Seth, Dr. Amit Sachan, Dr. Amit Singh, Arpita Singh, Kushal Anand, Dr. A.V. Subramanyam, Dr. Alok Prakash, and Neha Tripathi) in Singapore. I also thank my friends and ex-colleagues, who are too many to count. In a non-exclusive list, I mention Vijay, Abhishek Awasthi, Srikanth, Pillai, Sridhar, Ramesh Babu, Atul Jha, Chandrashekar, Vineet Singh, Himanshu Pati, Kambli, Ashu, Dubi, and Kuseswar Prasad. I recall fondly Sanju, Sunita, Nupur, Loet, Erik, David Gernaat, Leila, and others. My start-up team Bharti, Prashant, Dishant, and Mohit Bansal understood my commitment towards my thesis and supported me throughout last 2 years. I am thankful to them too.

I thank the 20 volunteers who participated in the generation of ground truth for section 6.1. They are Dr. A.V. Subramanyam, Dr. Alok Prakash, Dr. Amit Sachan, Dr. Amit Singh, Dr. Atiqur Rahman, Dr. Balaji Gokulan, Dr. Dwarikanath Mohapatra, Dr. Hengyi Zhang, Dr. Krishna Agarwal, Dr. Raj Kumar Gupta, Abhishek Seth, Deepak Subramanian, Kushal Anand, Neha Tripathi, Ngyuen Chi,  K. Prasad Sir, Dr. Manish Narwaria, Bharti, Prashant, and Dishant.

I acknowledge the support received from Dr. Alex Chia during my first month of Ph.D. and providing me with his insight about the problem of ellipse detection and sharing his source code for generating synthetic data for ellipse detection experiment. I acknowledge Dr. P. Kovesi, Dr. F. Mai, Dr. A. Opelt, Dr. R McLaughlin, Dr. Partha Bhowmick, and Dr. J.-O. Lachaud for sharing their insights or source code or executables for generating the results of their methods.

I convey love and gratitude to my sisters, Kaushalya and Sangita for keeping me on track of my studies right from childhood. I would like to give special thanks to each and every person of my neighborhood during my childhood who indirectly took special care of me by protecting me from all the evil things and negative influences around me. I would like to thank my best friend Bhaskar and teenage friends Deepak, Narayan, Sulendra, Rajendra, Subodh, Mantu, Pintu, Bablu, Shankar da, Dadu, and others for protecting me from wrong and dark career paths. I like to thank Deepika for being indirect source of inspiration to study harder during senior school days.

I thank Kaushalya Gupta and Sangita Gupta for taking care of my family in the difficult times and ensuring that my parents do not miss my presence. I fall short of words when trying to acknowledge my loving wife, Krishna. She has excelled and helped me in every possible way to pursue my study.

I dedicate this important document of my life to my family and everyone related to me directly or indirectly in my life. My achievements are collective efforts of each and everyone in my life.

I hope that this thesis will inspire a few future researchers to understand and tap the potential of discrete geometry in image processing applications.

Dilip Prasad

# Table of contents

# Table of figures

# List of tables

# List of abbreviations

| Abbreviation | Detail | Abbreviation | Detail |
|---|---|---|---|
| PA | Polygonal Approximation | AEV | Associated Error Value |
| HT | Hough Transform | DSS | Digitally Straight Segment |
| SHT | Simplified Hough Transform | TE | Tangent estimation/estimator |
| RHT | Randomized Hough Transform | DEB | Definite Error Bounded |
| ISE | Integral Square Error | LR | Linear Regression |
| MD, $E_{max}$ | Maximum Deviation (also represented as $max(d_i)$ ) | IPF | Implicit Parabolic Fitting |
| CR | Compression Ratio | EPF | Explicit Parabolic Fitting |
| DR | Dimensionality Reduction | ICIPF | Independent Coordinate IPF |
| FOM | Figure of Merit | GD | Gaussian Derivative |
| PV | Perez & Vidal | λMST | Lambda-Maximal Segment Tangent |
| PRO | Precision Reliability Optimization | λMSG | Lambda-Maximal Segment Gaussian |
| PRO-LS | PRO Least Squares | NSAF | Numerically Stable Algebraic Fitting |
| PRO-DP | PRO Dominant Point | ElliFit | Ellipse Fitting |
| RDP | Ramer-Douglas-Peucker | ED | Ellipse Detection |
| RDP-mod | RDP Modified | ECC | Edge Curvature and Convexity |
| Masood-mod | Masood Modified | EH | Elliptic Hypothesis |
| Carmona | Method by Carmon-Poyato | RANSAC | Random Sample Consensus |
| Carmona-mod | Carmona Modified | ADR | Average dimensionality reduction |
| DP | Dominant Point | RP-AUC | Recall-Precision Area under curve |
| | | RP-EER | Recall Precision Equal Error Rate |

# Chapter 1 : Introduction

## 1.1 Background

Most man-made objects and structures are defined by polygonal and quadric surfaces. Besides irregular and fractal surfaces encountered in nature, quadric surfaces are commonly encountered. Thus, it is not surprising that mankind has a long standing (more than 5000 years) fascination and inspiration to study the geometry of polygons and conics. From ancient civilizations of Aryans, Egyptians, Babylonians, etc. to Greek stalwarts like Thales, Pythagoras, Euclid, and Archimedes to the modern day engineers and mathematicians, geometry has triggered the imagination of many and served as the tool for several practical engineering innovations.

In the context of image processing, the role of geometry begun since computers could read the image as an intensity matrix and mathematicians could then manipulate the intensity matrix to derive some geometric properties and patterns from the image using the computers. It can be said that the study of geometric features or properties of images is as old as the field of digital image processing, which begun approximately in the decade of 1960s.

Images are the projections of the light emanating from the objects in front of the sensor. Thus, images contain two-dimensional projections of the three-dimensional shapes of the objects. Thus, the images are replete with geometric shapes like lines, polygons, conics, etc. Ideally, if there are no artifacts due to illumination conditions, sensor defects (aberrations, digital sensor grid, etc.), noise, etc., the shapes in the image are ideal projections of the shapes of the objects in the image. Here, the effect of the point spread function has been ignored and it is assumed that ray optics is valid. If the shapes in the images can be correlated to their corresponding objects, such tool is very helpful in image analysis for diverse applications. This is the main motivation in the continued interest of the image processing research community on the topic of geometry in image processing.

Several geometric properties of interest are found in the images. Some of them include finding the edges representing the boundaries of shapes, determining shapes, angles, and curvatures, computing tangents at the boundaries, etc. All these topics have interested researchers for more than 50 years already [5-15]. The research on edge detection is quite mature already [5-10] with pioneering works reported in [16-18]. However, *given the edge map of the image*, the problems of shape detection, curvature estimation, and tangent computations continue to be active research topics [19-24]. This thesis specifically deals with the following three geometric problems involving the edge contours which have wide application in several image processing problems:

1. Representing shapes of the digital curves using approximate polygons
2. Estimating tangents in digital curves
3. Finding elliptic shapes in images using digital curves.

While the progress in image processing and computer vision has been fast and successfully applied in complex applications like face detection, object detection, etc., these fundamental problems experienced in image processing are often neglected despite their significant influence in these high end applications [19, 25-30]. The main reason behind this neglect is the technological challenges related to the geometric features. The major technical issues are presented briefly in a few paragraphs below.

The first issue is the effect of digitization in the images. Problems of estimating tangents, finding geometric shape features, or representing geometric shapes are generally easy to deal with in the continuous space. This is because in the continuous space, these geometric properties are governed by analytical equations whose solutions may lie in the continuous space. These problems become significantly difficult in the digitized or quantized pixel space of images because the analytical equations may not take any continuous solution for this case. The chosen solution is almost always an approximate integer solution nearest to the actual solution of the analytic equations. Digitization introduces a non-linear corruption in the continuous curve which cannot be analyzed using equations [20, 29-31]. A very simple example is presented in Figure 1.1-1. It demonstrates how the continuous shape of a semicircle gets corrupted due to digitization.

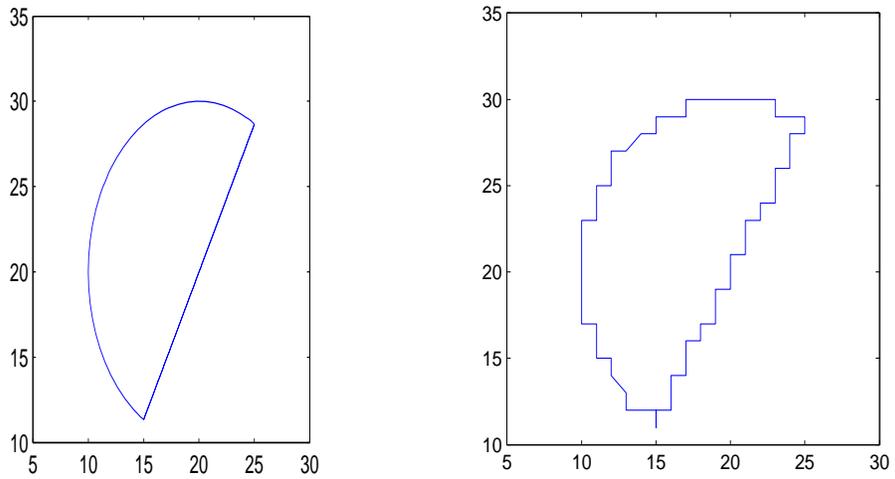

(a) A shape in continuous domain      (b) Digitized curve corresponding to (a)

**Figure 1.1-1: Illustration of the effect of digitization on continuous curves.**

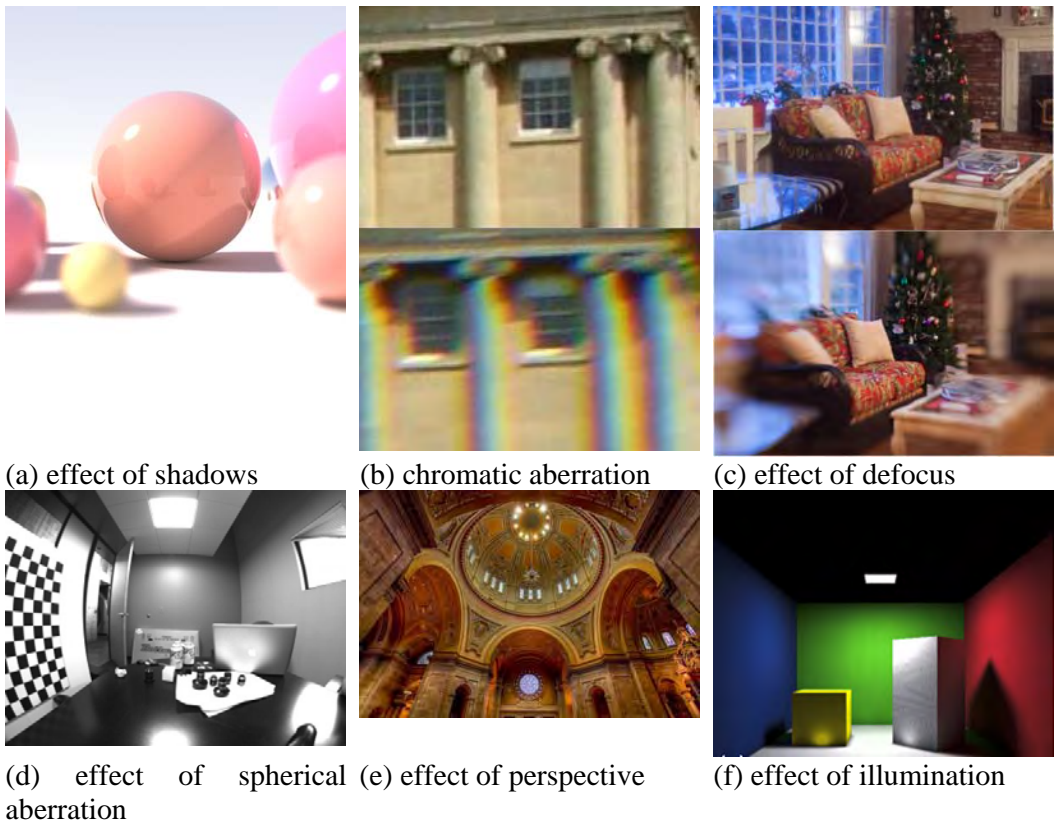

(a) effect of shadows     (b) chromatic aberration     (c) effect of defocus

(d) effect of spherical aberration     (e) effect of perspective     (f) effect of illumination

**Figure 1.1-2: Various optical effects that non-linearly distort the shapes.**

The second issue is the corruptive distortion of the shapes due to optical effects like illumination, perspective, aberration, chromatism, high numerical aperture, etc. While in general the three-dimensional shapes are expected to be linearly projected in the image, such optical effects introduce non-linearity into the projection. Thus, for example, a sphere may be projected as an ellipse or a line may distort to a hyperbole. Though these distortions are easy to predict and some mathematical models may completely explain them, several other forms of distortions (like aberrations, motion blurs, etc) are quite complicated and difficult to invert or interpret. Some examples of distortions due to optical effects (taken from internet) are presented in Figure 1.1-2.

The third issue is that of the sensor noise or numerical noise. Due to the presence of noise in images, the edges become quibbled or broken into separate contours. As a result the local properties get severely affected and relating a broken edge curve to the original shape becomes difficult. Even worse, it is possible that the broken edge fragment of one shape can be confused as a part of another shape. The effects of noise are demonstrated in Figure 1.1-3.

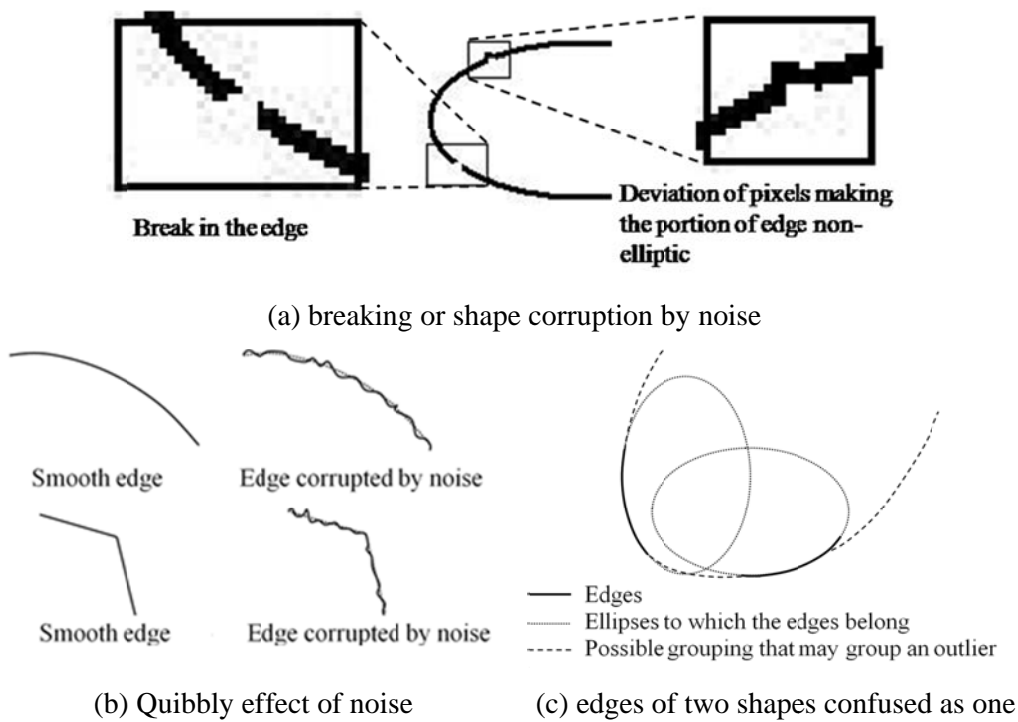

(a) breaking or shape corruption by noise

(b) Quibbly effect of noise          (c) edges of two shapes confused as one

**Figure 1.1-3: Effect of sensor noise or numerical noise.**

Another issue is that the shapes often appear incomplete or merges with other shapes due to overlap and occlusion of the objects in the images. The edges of the shapes often become broken small fragments and it becomes difficult to relate the small broken edges to their original shapes. See Figure 1.1-4 for an example.

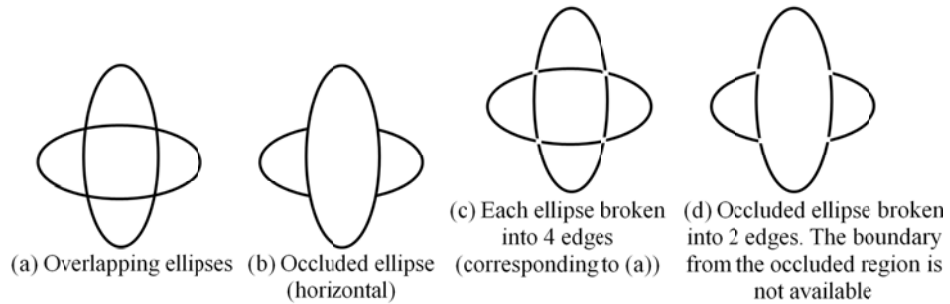

(a) Overlapping ellipses (b) Occluded ellipse (horizontal) (c) Each ellipse broken into 4 edges (corresponding to (a)) (d) Occluded ellipse broken into 2 edges. The boundary from the occluded region is not available

**Figure 1.1-4: Illustration of the presence of overlapping and occluded ellipses**

These technical challenges make it very difficult to deal with geometric primitives and the research in these topics has been consistent but the progress has been limited. As a consequence of these challenges, very complicated algorithms with lots of heuristically chosen control parameters are used in practice. As a consequence, the applications are very sensitive to the choice of control parameters [32-37] related to geometric primitives. The research presented in this thesis is based on the philosophy that a better treatment of geometric primitives would have significant impact on advanced image processing applications. The basic principle behind the work presented in this thesis is that for each kind of geometric primitive, at least the effect of digitization can be modeled with sufficient accuracy. In most cases, explicit error bounds for the error introduced due to digitization can be derived. The error bounds for the digitization problem can be used for improving the existing algorithms. In other cases, instead of using purely numerical framework for applying least squares or computing the inflexion points (which are sensitive to digitization and other noise), unconventional means based on geometry in the two-dimensional discrete plane can be used to obtain better solutions to the difficult problems.

In the next few sections, each of the following topics is discussed from the following research perspectives:

1. Representing shapes of the digital curves using approximating polygons,

2. Estimating tangents in digital curves,

3. Primitive ellipse fitting using edge pixels, and

4. Ellipse detection for practical scenario using hybrid approaches.

Each of the above topics is significantly different from each other. For each of these topics, the challenges related to it, the background literature review, and a gist of the work presented in this thesis is presented.

## 1.2 Polygonal approximation of digital curves

In several image processing applications [38-45], it is desired to express the boundaries of shapes (edges) using approximate polygons made of a few representative pixels (called the dominant points) from the boundary itself. Through polygonal approximation (PA), it is sought to represent a digital curve using fewer points such that:

i. The representation is insensitive to the digitization noise of the digital curve.

ii. The properties of the curvature of the digital curve are retained, so that geometrical properties like inflexion points or concavities can be subsequently assessed.

iii. The time efficiency of higher level processing can be improved since the digital curve is represented by fewer points.

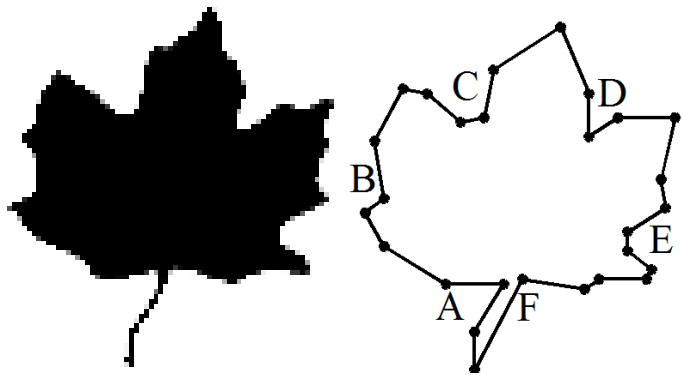

**Figure 1.2-1: Example of polygonal approximation of a digital shape.**

An example is presented in Figure 1.2-1. In Figure 1.2-1, a digital shape of a maple leaf is illustrated. The boundary of the shape is made of 244 pixels. A PA of this shape is shown in Figure 1.2-1. The maple leaf is represented using only 27 dominant points in this approximation and the concavities associated with the maple leaf are preserved (labeled A-F).

Despite being a very old problem of interest, this problem attracts significant attention even today in the research community. Some of the recent PA methods are proposed by Masood [36, 46], Carmona-Poyato [35], Ngyuen [37], Wu [47], Kolesnikov [40, 48], Bhowmick [49] and Marji [50] while a few older ones are found in [12, 13, 51-60]. These algorithms can be generally classified based upon the approach taken by them. For example, some used dynamic programming [40, 48, 52], while others used splitting [12, 13, 53], merging [54], digitally straight segments [37, 49], suppression of break points [35, 36, 46, 50], curvature and convexity [47, 51, 55, 58].

Among the older methods, The method of Teh and Chin [51] relies primarily on the accurate determination of the support region based on chord length and the perpendicular distance of the pixels from the chords to determine the dominant points. Ansari [58] proposed a method in which a support region is assigned to each boundary point based on its local properties. A combination of Gaussian filtering and a significance measure is used on each pixel for identifying the dominant points. Cronin's [59] method finds the support region for every pixel based on a non-uniform significance measure criterion calculated by locally determining the support region for each point. Ray and Ray [61] proposed a k-cosine-transform based method to determine the support region. Sarkar [62] proposed a chain code manipulation based method for determining the dominant points where the chain code is sufficient and the exact coordinates of the pixels are not necessary.

Regarding the recent methods like Masood [46], Carmona-Poyato [35], and Nguyen [37], these methods have already shown considerable improvements over earlier dominant point detection or PA methods. However, all of these methods except Carmona-Poyato [35] use local properties of fit like the maximum distance (deviation) of the pixels on the digital curve from the fitted polygon. On the other hand Carmona-Poyato [35] uses a ratio 'r' which incorporates the quality of the global fit instead of the local fit.

The reasons for the continued relevance of this problem in the current era are highlighted here:

1. The performance metrics for evaluating the performance of a PA method, and

2. The control parameter and optimization goal used in the PA method.

The technical details about these reasons are discussed in section 2.1. This thesis addresses both issues. Specifically, the work in this thesis has the following contributions to the PA problem:

i. Two simple metrics that relate to the global and local properties of fit of the polygon are proposed,

ii. It is shown using these metrics that the global and local properties of fit are always conflicting in the least squares fitting scenario and that most existing PA methods optimize either the local or the global properties of fit,

iii. A PA method based on simultaneous optimization of both the local and global properties of fit is proposed,

iv. An upper bound of the error in line fitting due to digitization is derived,

v. This upper bound is used to design a non-parametric framework which can be used to make most PA algorithms independent of control parameters and related heuristics,

vi. The applicability of the framework is shown for a very popular PA method and two recently proposed PA methods, and

vii. Experiments of various types are conducted to show the various practical aspects of PA problems and the performance of the proposed methods.

## 1.3 Tangent estimation for digital curves

Tangent estimation (TE) is important in many applications like shape and perimeter estimations, concavity analysis, segmentation, etc. Despite the significant influence of the error in TE, most researchers tend to use heuristics and algorithms tailored specifically to the application. Also, they typically use complex optimization or curve fitting based algorithms that are computationally intensive, parameter controlled, sensitive to the digitization error, noise, and distortion. The problem of tangent estimation for digital curves had long been considered old and saturated [19-24]. However, a few researchers have begun to address the problem of tangent estimation for digital curves, with the specific aim of proposing criteria for evaluating the performance of tangent estimators and the development of better tangent estimators for practical applications [20, 21].

The problem of tangent estimation (TE) for digitized curves faces the following conceptual challenges:

1. The tangent is typically defined on a point, though it is a property of the continuous curve to which the point belongs. Thus, it has the local as well as the global properties of the curve. Due to digitization, both these properties are affected and the nature or extent of the effect cannot be quantified or analyzed using simple mathematical tools. At best, some estimates of maximum error or localized precision may be developed.

2. Usually, while estimating the tangents, prior information about the nature of the curve is unavailable. Further, appropriate size of the local region around a point is also unknown. Hence, the choice of these parameters is mainly heuristically guided and non-robust.

One of the methods to find the tangent is to use continuous function (typically second order) to approximate the curvature of the digital curve in a local region around the point of interest [27, 63, 64]. The derivative of the continuous function is then used to determine the tangent. Such approach is restrictive in the choice of the nature of continuous function and the definition, shape, and dimension of the local region, etc. Further, there are applications where tangents need to be computed to fit a shape (for example ellipse) on the digital curve [25, 28, 45, 65]. In such cases, it is difficult to rely on a method that first fits a shape in the local region to estimate the tangent, and then uses the tangent to fit a shape to the whole curve.

In order to overcome the problem of choosing the continuous function, researchers sometimes use a Gaussian filter to smoothen the digital curve and obtain a smooth continuous curve. This Gaussian smoothened continuous curve is then used for estimating the tangents [44]. In essence, this is similar to applying a one-dimensional spatial Gaussian filter. A similar approach is taken in [66], where one-dimensional spatial median filtering is used. Another method is to consider a family of continuous curves of various types. The complete digital curve is approximated by one of the continuous curves in the family using a global optimization technique. The tangents are then computed on the curve chosen by optimization [67]. A different approach is to approximate the digital curves using line segments. Two main variations in this approach are in vogue. The first variation is based on the theory of maximal segments [20]. At the point of interest, the maximal line segments passing through it are found and weighted convex combination of their slopes is used to find the orientation of the tangent. This method is parameter-free, has asymptotic convergence, and incorporates convexity property. Though the theory of maximal segments is well-developed and fool-proof, the assumption that their weighted combination (and the value of the weights) is indeed a true representation of the curvature is based on heuristics, rather than analytic foundation. Despite that, to our knowledge, this is the first parameter-independent tangent estimation method (though involving heuristic choice of a function) that provides good properties in tangent estimation.

The second variation is to approximate the digital curve using small line segments such that the maximum deviation of any point on the digital curve with one of the fitted line segments is small; for example, below a threshold value of a few pixels [44, 68]. This procedure divides the curve into small sub-curves each corresponding to a fitted line segment. The slope of the tangent at the midpoint of each sub-curve is then considered to be the same as the slope of the corresponding line segment. The main restriction with this method is that the tangents are available only at some points of the digital curves, viz., the mid points of the digital sub-curves. In our opinion, such method is essentially similar to the concept of maximal segments [20], especially if the threshold of the maximum deviation is less than or equal to 1.414 pixel.

This thesis proposes a very simple and computationally efficient tangent estimator. It is shown that despite the simplicity of the method, the method has a definite upper bounded error for conics. Thus, this method does not suffer from the tangent estimation singularity experienced by other methods [31, 69]. The method uses a simple control parameter and a rule of thumb for choosing the control parameter is also provided. It is notable that the rule of thumb also uses the upper bound of the error in tangent estimation and thus it is less empirical than most other tangent estimation methods. Extensive numerical experiments validate the superiority of the proposed tangent estimator over almost all the existent tangent estimators. Further, the applicability of the proposed tangent estimator for non-conic curves with convex and concave curvatures is also exhibited.

## 1.4 Primitive ellipse fitting

Ellipses appear in many natural objects ranging from cells and nuclei to astronomical bodies. Further, ellipses also appears commonly in man-made objects from medicinal tablets to space ships. Thus, using simple mathematical framework to detect ellipses from edge data has inspired many researchers.

Initially, approaches like least squares fitting and Hough transform (HT) were the main approaches used by researchers. These approaches generally use the mathematical model of ellipse and the edge pixels for detecting the ellipses. Hough transform was introduced for ellipse detection in [14] in the form of simplified Hough transform (SHT). Modifications of Hough transform, randomized Hough transform (RHT) [70, 71] and probabilistic Hough transform [72, 73] were proposed to improve the performance of HT for non-linear problems like ellipse detection. HT based ellipse detection methods are usually more robust than least squares based ellipse detection when the edge contours are not smooth because they use pixels for detecting the ellipses instead of the edge contours (connected edge pixels) [74] while most least squares formulations for detecting ellipses use edge contours.

HT based ellipse detection methods have two main problems. The first problem is that HT is computation and memory intensive because it uses a five-dimensional parameter space. For solving the problem of five-dimensional parameter space, many methods were proposed that split the five-dimensional space into two or more subspaces with lesser dimensionality and deal with each of the subspaces in separate steps [28, 75-78]. The most popular approach in these methods was to find the centers of the ellipses using geometrical theorems and Hough transform in the first step and finding the remaining parameters of the ellipses in the second step. Second problem is that since the pixels used in HT need not belong to the same edge contour, number of samples required for detecting each ellipse is very high. Once an ellipse is detected, the pixels near the detected ellipses are not considered for the detection of the next ellipse. Due to this, HT based methods may be unable to detect all obvious ellipses (represented by edge contours) and the accuracy of such methods reduces with the increase in the number of ellipses. Some methods used edge contours (instead of edge pixels) and piece-wise linear approximation of edge contours to improve the performance of HT [76, 79].

Least squares based method was used for ellipse detection in [80-85]. Least squares based methods usually cast the ellipse fitting problem into a constrained matrix equation in which the solution should minimize the residue in the mathematical model. The choices of constraints and solution approaches for constrained matrix equations have an impact on the performance and selectivity of the ellipse detection methods [80, 83]. One problem with the least squares method is over-fitting of the data. Many pixels are used for ellipse detection and the elliptic hypothesis that minimizes the residue in the chosen mathematical equation and satisfies the constraint is generated. Thus, the chances of fitting an ellipse on a non-elliptic curve are high [84]. Second, though the amount of residue is used as the quality of fit and termination criterion, it may not necessarily be related directly to the quality of the detected ellipse.

The least squares methods currently in use are based on the fundamental work by Rosin [82, 83, 86-88] and Fitzgibbon [80]. They both employ the algebraic equation of general conics to define the minimization problem and additional numeric constraints are introduced in order to restrict the solutions to the elliptic curves. In other works [86-89], Rosin developed and tested several error metrics for quantifying the quality of fit. Fitzgibbon [80] solves the constrained minimization problem using generalized eigenvalue decomposition. It is shown in this thesis that the method of Fitzgibbon [80] is prone to the problem of numerical instability and a simple modification of the method can make it numerically more stable. All the methods discussed above employ algebraic equation of the ellipse and the algebraic distance of the points on the ellipse as the cost function. As opposed to these, Ahn's method [90] uses the geometric distances of the data points from the ellipse as the central quantity for fitting the ellipse. Ahn's method involves nested iterative non-linear optimization, which makes it computation intensive and prone to the problem of local minima. Beyond this problem, Ahn's method in general has a superior performance due to the fact that it uses minimization of the geometric distance instead of the algebraic distance which results in more physically relevant solutions.

In both HT and least squares methods, if the points chosen lie actually on the loci of interest, the error in retrieval is expected to be very low. In such case, a small bin size is able to provide good precision [91]. However, actual points on the loci are hardly available from images, since the digitization rounds the coordinates of the actual points to the nearest integers [92, 93]. Digitization effect may impose severe restrictions on methods based on analytical equations. For example, it is easily conceivable that such methods will result into large errors for small circles and ellipses because the round off error due to quantization is of the order of the ellipse itself in these cases. Another example is the error in highly eccentric ellipses. However, the error may be non-negligible in other cases as well and may depend upon the choice of points [94]. Despite the practical importance of the effect of quantization on the accuracy of such methods, this issue has received least attention [91-93]. If such issues and their impact on the accuracy of these methods are understood, better techniques may be designed, reliable constraints may be introduced, and better methods may be chosen for achieving an acceptable accuracy in practice.

This thesis proposes an ellipse fitting method based on least squares which uses the concept of minimizing the geometric distance of the data points from the ellipse to be fitted. However, as compared to Ahn's method [90], the proposed method uses a linear least squares model and shifts the non-linearity involved in the determination of the geometric parameters of ellipse into a non-linear, but unique and easily computable operator. Further, the method is unconstrained and non-iterative. The proposed method performs superior to the remaining methods for various experiments and gives less false positives as compared to other methods. As a consequence, it is more suitable than other methods for the problem of detecting ellipses in images.

## 1.5 Ellipse detection for practical scenario using hybrid approach

Ellipse detection in real images has been an open research problem for a long time. However, the performance is still poor for most real images like images of the Caltech 256 dataset [1]. Most hybrid approaches use a combination of pre-processing of edges for curvature correction, some or the other grouping scheme (usually based on edge continuity), primitive ellipse fitting algorithms (like discussed in section 1.4), elliptic hypotheses selection schemes, etc. Thus they are usually grouped under the umbrella term 'hybrid ellipse detection methods'.

While there are some sophisticated algorithms for detecting ellipses from images, most of these algorithms cannot be used in real time applications. Further, there is generally a least squares method at the core of these algorithms [80, 82, 83, 86, 87] and the other pre-processing and post-processing steps are used to increase the selectivity of the algorithms to only elliptic shapes and reduce the false positives [25, 34, 68, 95, 96]. For instance, least squares methods typically need a fraction of a second, while extra processing steps used for improving selectivity may require a few seconds [34] to a several few minutes per image [33]. Furthermore, the selectivity of the ellipses can be poor and ellipses may be fitted in non-elliptic curves as well. For real time applications such as pupil tracking, where time is a critical parameter, a large number of processing steps are deemed undesirable and generally least squares method has to be used alone. Even for non-real time applications, it is desirable to have a least squares method that is inherently selective and good at reducing the false positives for non-elliptic curves, so that the burden on extra processing can be reduced.

Researchers commonly use edge following methods for this problem, in which the connectivity of the edge pixels in the form of edge contours and the continuity of edge contours were used in addition to the mathematical model of ellipse [33, 34, 68, 81, 92, 97-99]. These methods work well for simple real images typically containing one or two ellipses in the foreground but fail to perform well in more complicated scenario such as presented in Figure 1.5-1.

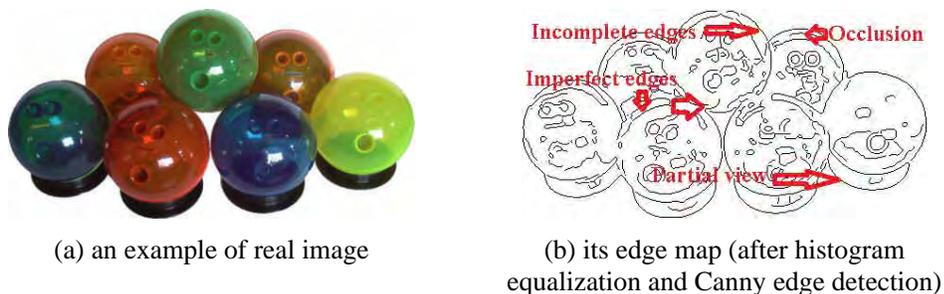

(a) an example of real image

(b) its edge map (after histogram equalization and Canny edge detection)

**Figure 1.5-1**: **An example of a real image and the problems in detection of ellipses in real images**

Some of the most notable articles are briefly discussed here. Mai [34] proposed a modified RANSAC (RANdom SAmple Consensus) [100] based ellipse detection method, which first extracts the fitted line segments from the edge data of the image. It follows an edge in terms of its continuity to group the edge with other edges. Finally, RANSAC based ellipse fitting is performed on these grouped arc segments. This method shows good performance in terms of accuracy and computational efficiency over many existing methods. However, the performance of this method is highly dependent upon the choice of the two thresholds – proximity distance and angular curvature. Split and merge detector proposed by Chia [101] in essence is the same. The performance of both these methods deteriorates in the presence of occluded ellipses.

Hahn [102] proposed an ellipse detection method based on grouping points on elliptic contour. Whether some curved segments belong to the same ellipse or not, are tested by comparing the parameters of candidate ellipses that are made by the curve segments. This method can reduce the total execution time because it estimates the ellipse parameters in the curve segment level not in the individual edge pixel level. However, the performance of this method deteriorates for complex real images.

Kawaguchi [103, 104] groups adjacent edge pixels with similar gradient orientations into the regions called as line-support regions, which are subsequently used for ellipse detection

Ji [105] proposed a grouping scheme to pair the arc segments belonging to the same ellipse as an improvement over [81]. While [81] groups the edges based on a scale invariant statistical geometric criterion which can be verified either in parametric space or in residual error space, Ji [105] takes proximity and direction of arc segments (clockwise and counter clockwise) into account.

Kim [68] proposed a grouping scheme based on three curvature and proximity based conditions as follows: an arc should be a neighbor in eight group classification [68], the arcs follow a convexity relationship as proposed in [78], and the inner angle between two edge pixel on an ellipse should not exceed 90 degree [68]. If the arcs satisfy these three constraints they represent circular arcs. In order to determine ellipses from the list of circular arc (obtained by merging three circular arc segments), it first finds the center of ellipse by center finding method [28]. Two arcs belong to one ellipse if they satisfy the parameters obtained by least squares method and the three constraints. These arcs are then merged and remaining elliptical parameters are extracted.

One of the problems faced by any ellipse detection algorithm is that in attempts to detect all the ellipses actually present in real images in the absence of prior knowledge of the number of ellipses, they have to compromise on the accuracy of the algorithm. Any ellipse detection method, including the proposed method, suffers from the problem of reliability and precision uncertainty. Due to this, the ellipse detection methods have to compromise on either the reliability or the precision. The popular choice is to compromise on the precision, as the quantization already limits the precision and due to the absence of a priori information on the elliptic objects present in the image, it is often important to be reliable in detecting the elliptic shapes. Due to this, often, the ellipse detection algorithms generate numerous elliptic hypotheses, not all of which correspond to actually present elliptic objects. Sometimes, many elliptic hypotheses are generated for a single elliptic object, and at other times, the hypotheses do not correspond to any actually present elliptic object in the image. Thus, it becomes important to evaluate the possibility of an elliptic hypothesis actually corresponding to an elliptic object. In other words, a method is needed to identify the elliptic hypotheses which are more likely to correspond to an object in image. Researchers use the 'saliency' or 'distinctiveness' scores for quantifying the reliability of a hypothesis [74, 97, 98, 106-110] and select the more salient elliptic hypotheses.

Wang [109] presents some important work in this regard. But the measures suggested are neither simple nor computationally efficient. Further, the applicability of this method is restricted to Hough transform related ellipse detection methods and edge following methods. Basca [107] proposed a similarity measure that compares two elliptic hypotheses using the Euclidean distance between them in the Hough parametric space. Due to the absence of normalization, this method is sensitive to scale changes. Qiao [98] proposed ellipse detection method based on saliency of an arc. This method explores the relationship between spatial connectivity and the incremental point angle of elliptic inliers. This relationship can be used to detect elliptic arc end points. The angle subtended by the elliptic arc can then be used as ellipse validation criteria. This method depends on too many thresholds which are application dependent. This limits its use for wide range of application. The performance of this method depends on the proper choice of these threshold parameters.

Another popular measure is the pixel count feature in [106]. Elmowafy [111] verified the elliptic hypotheses by checking the ratio of count of pixel on elliptic curve and approximated circumference of the ellipse. There are two major concerns regarding such scheme [97]. First, the pixels are an approximate (quantized) representation of the elliptic hypothesis. Due to this, in case of a complete elliptic edge, the number of pixels is more than the actual perimeter of the ellipse. Even if there is an incomplete edge, the edge represents a fraction of perimeter of the ellipse using more pixels than the actual length of the fraction. Second, the determination of perimeter of an ellipse is a classical problem in mathematics for which no closed form analytic solutions are available. The perimeter used is typically one among the various numerical approximations provided by scientists [112]. It is notable that all these approximations are subject to some assumptions that may not be generally valid for all the elliptic hypotheses generated.

Also noteworthy are some of the steps related to edge data pre-processing, which is instrumental in enabling the detection of ellipses in real images. These include forming edge contours, removing junctions and branches [113], detecting inflexion points [114-118], obtaining edges of smooth curvature [33], etc. Among these, detection of the inflexion points is the most difficult problem. The reason is that inflexion points (or saddle points) are essentially defined by mathematical derivative in continuous space domain only. Continuous shape space (CSS) [115] based techniques like [117, 118] are noteworthy. These typically involve the convolution of the digital curve with a Gaussian kernel in order to transfer the curve into the CSS space and then use the persistence properties of the inflexion points in the convoluted CSS space to detect the inflexion points. Though the concept is interesting and apparently quite effective, it is complicated and involves significant amount of computation. Relatively simpler approach is taken in [116] but it depends strongly on three control parameters. Further, it is not specific to the inflexion points only and detects points with large curvature changes also since its aim is to detect the corner points and not specifically inflexion points. On the other hand, though Bai [114] is simple, it does not deal with all the possible situatiosn of inflexion points and its performance is control-parameter dependent.

The hybrid ellipse detection method proposed in this thesis deals with almost all the issues raised above using simple approaches for each issue. Only a few control parameters are used in the proposed algorithm and the algorithmic design makes it less sensitive to the choice of control parameters. Specific contributions of the proposed method are listed below:

i.    Simple inflexion point detection method based on curvature changes is proposed. The inflexion point detection method deals with more cases than those dealt with Bai [114], is simpler than the inflexion point detection using curvature shape space [115, 118], and is parameter independent unlike [114, 116].

ii.    Grouping scheme which is not based on edge following method but still based on the curvature of the edges. Two simple schemes are used. The first is based on the search region for identifying the candidate edges for grouping. Second is based on the concept of associated convexity. For both of the scheme, simple and easily computable steps are needed.

iii.    Proposition of a non-linear trust score which gives more selectivity as compared to histogram score to Yuen's method [28] for detecting centers of ellipses.

iv.    A simple similarity measure based on Jaccard index to determine similar ellipses.

v.    Three easily computable saliency criteria for dealing with various aspects of the quality of elliptic hypotheses.

vi.    A non-heuristic parameter-independent scheme for selecting ellipses that performs well for images of various kinds.

## 1.6 Practical applications

In order to show the impact of the geometric primitives on practical image processing applications, three practical image processing problems are presented. The three applications are ellipse detection in real images, detection of cell organelles in extremely low contrast microscopy images, and object detection using polygonal and elliptic features only. For each of the applications, the core of the proposed algorithms is the geometric features.

For detection of ellipses in real images, the hybrid ellipse detection method proposed in Chapter 5 is directly applied on images in the Caltech dataset [1]. It is notable that the method in Chapter 5 uses the PA methods and TE method proposed in Chapter 2 and Chapter 3. It also uses the improvement of Fitzgibbon's method proposed in Chapter 4.

For the application of detection of cell organelles, a scheme of pre-processing to improve the contrast and enhance the regions of interest is proposed. The least squares ellipse fitting proposed in Chapter 4 is then used for fast and accurate detection of the elliptic cell organelles. The method is easily parallelizable and it is possible to reduce the processing time to values close to the fluorescence microscopy acquisition time. Thus, it can be easily integrated in the microscopes.

For the application of object detection in images, instead of conventionally used edge features, patch features, texture features, etc., the geometric features, i.e., *PA of the edge contours and ellipses are used as the only features*. In addition, a hierarchical template is used for object representation. The result of object detection for categories which have consistent geometric features is reasonably good. Indeed the performance can be enhanced by using other features like patch and texture features and this shall comprise the future work for this application.

## 1.7 Research not covered in the thesis

While working on the doctoral topic, several related research topics were also explored. The research work related to these topics is briefly presented here. For instance, in the preliminary stage of the doctoral work, works on image composition analysis [119], a simpler ellipse detection scheme (as compared to the one presented in Chapter 5) [96], and a simple clustering scheme for ellipses were studied [120]. For the problem of object detection, contextual object detection for conscious machines [121] was studied. In addition to these works (directly related to the doctoral thesis), works related to cloud computing [122], embedded systems [123, 124], machine consciousness [121, 125-127], machine learning [128, 129], education data analysis [130], etc. were also done.

## 1.8 Outline and highlights of the thesis

The thesis comprises of seven chapters. Chapter 1 and Chapter 7 present the introduction of the thesis and the conclusions respectively. Chapter 7 also presents some of the potential future directions. For the remaining chapters, the detailed outline is presented below.

**Chapter 2:**

Chapter 2 discusses the problem of polygonal approximation (PA). The contents of Chapter 2 have been reported in [93, 131-136]. After introducing the problem in section 2.1, the global and local properties of polygonal fitting upon edge curves are discussed in section 2.2. Specifically precision metric for local quality of fit and reliability metric for global quality of fit are proposed in section 2.2.1. For least square fitting, it is shown in section 2.2.2 that the precision and reliability are always at conflict. This thus corroborates with the fact discussed in section 2.1.1 that the available metrics are not singly sufficient for PA. In this context, section 2.2.3 discussed whether the conventionally used performance metrics represent the global or local qualities of fit. Section 2.2.4 generalizes the precision and reliability metrics so that they can be applied to an edge curve, or an image with several edge curves, or a dataset with several images.

Section 2.3 proposes a PA method that uses simultaneous optimization of both the precision and reliability metrics for an edge curve. The algorithm, called precision and reliability optimization (PRO), is proposed in section 2.3.1. Numerical experiments are presented in section 2.3.2. It is shown that the control parameter can be easily tweaked to obtain the desired performance of PRO.

Section 2.4 presents a very important contribution of this chapter. It presents the analytical derivation of the error due to digitization of a continuous line segment in sections 2.4.1 and 2.4.2. Section 2.4.2 also proposes the non-parametric framework for PA. The error bound can be analytically derived and depends upon both the size and the orientation of the continuous line segment. This error bound is compared against other simpler error bounds proposed earlier by other researchers in section 2.4.3.

Based on section 2.4.2, section 2.5 adapts three existing methods into the non-parametric framework and makes them parameter-independent. Section 2.5.1 makes the popular PA method proposed by Ramer-Douglas-Peucker [12, 13] parameter independent [132, 135]. Sections 2.5.2 and 2.5.3 make the recently proposed methods of Masood [46] and Carmona [35] parameter-independent [132, 133]. All the three methods are quite different from each other and yet the non-parametric framework can be used for them. The numerical results show that the non-parametric versions of these methods perform either better than or similar to the original versions of these methods.

Even though other PA methods have not been adapted in the non-parametric framework, at least their optimization goals and update schemes are analyzed in the context of precision, reliability, and the analytical error bound in section 2.6. Various interesting and practical aspects of PA methods are analyzed in section 2.7. In this section, the performance of the methods proposed in sections 2.3 and 2.5 are compared for scalability, noisy curves, non-digitized curves, and performance over large datasets of practical use. Some recommendations regarding the choice of PA methods in practical applications are provided in section 2.7.5. The chapter is concluded in section 2.8.

**Chapter 3:**

Chapter 3 considers the problem of tangent estimation (TE) for digital curves. The content of this chapter have been reported in [31, 69]. After providing the background of the problem in section 3.1, an example is shown in section 3.2 which highlights the importance of reducing the error in TE. The tangent estimation method proposed in this thesis appears in section 3.3 in which the concept, algorithm, and the computation complexity of the method are discussed. There is only one control parameter which can be chosen using a thumb rule presented in section 3.4.3.

Section 3.4 is an important highlight of this chapter. The error in tangent estimation has been analytically derived for the continuous conics in section 3.4.1. The error in tangent estimation due to digitization of the continuous conics and the total error bound (sum of error bounds for continuous and digital curves) are presented in section 3.4.2. Section 3.4.3 discusses the choice of control parameter which is based upon the total error bound and the smallest circle of interest. In the same section, the multigrid performance of the presented method is also discussed. Section 3.5 presents several numerical simulations to illustrate the error bound for conics of various types.

Section 3.6 also presents a highlight of this chapter. In this section, the performance of the proposed method is compared against almost all the existing tangent estimation methods and it is seen that the proposed method perform superior to all of them. Besides considering circular and elliptic geometries, some non-conic curves are also considered and it is seen that the proposed method performs quite well for all the geometries. The chapter is concluded in section 3.7.

**Chapter 4:**

Chapter 4 considers the problem of least squares fitting of ellipses. The contents of this chapter have been reported in [25, 137-139]. After introducing the problem of least squares fitting of ellipses and the state of the art in section 4.1, the conventionally used algebraic fitting of ellipses is presented in section 4.2. Specifically, the popular method of Fitzgibbon [80] is considered in this section and the problem of numerical instability of this method is discussed. Based on the discussion, a simple solution for dealing with this numerical instability is proposed in section 4.2.2.

The concept of geometric distance minimization is discussed in section 4.3. The geometric equation of a general ellipse, its simplified form, and the distance of a data point from the ellipse to be fit is presented here. These expressions are useful for the theory presented in section 4.4.

Section 4.4 is the highlight of this chapter. For the subsequent use of linear unconstrained least squares method, a modification of the minimization function (the geometric distance) is proposed in section 4.4.1. The mathematical model of the method is proposed in section 4.4.2. The model of fitting involves using two operators and an intermediate unknown parameters space such that one operator is linear and the other operator is non-linear. By the properties of the linear operator and the definition of the intermediate parameter space, linear unconstrained least squares minimization of the residue can be used for determining the intermediate variables. The non-linear operator, its injective mapping, and inversion are presented in section 4.4.3. The numerical stabilization of the linear operator is discussed in section 4.4.4. Finally the computational complexity of the proposed method is presented in section 4.4.5.

Several numerical simulations and comparison with other least squares methods are presented in section 4.5. The proposed method performs better than the other least squares method for variety of cases like digital incomplete curves, noisy clusters of data points, as well as images with several incomplete ellipses. The proposed method is also tested for non-elliptic conics and non-conics as well and it is seen that the proposed method has lower false positives as compared to other least squares based methods. The chapter is concluded in section 4.6.

**Chapter 5:**

Chapter 5 proposes a hybrid advanced ellipse detection method. The contents of this chapter have been reported in [45, 95, 96, 139]. The background introduction to the proposed method is presented in sections 5.1 and 5.2 respectively. Preprocessing of edge curves is proposed in section 5.3. One of the important contributions of this chapter is the method for finding inflexion points in digital curves, which is described in section 5.3.2.

Grouping and elliptic hypotheses generation is proposed in section 5.4. The concepts of search region and associated convexities proposed in sections 5.4.1 and 5.4.2 are some highlights of this section. A non-linear relationship score for grouping and ranking the edges in a group is proposed in section 5.4.4. This relationship score is more selective than the usual histogram score and thus provides more robust grouping and ranking of the edges.

Section 5.5 is an important contribution of this chapter. Section 5.5.1 proposes a simple and elegant method of detecting similar elliptic hypotheses using Jaccard index. Section 5.5.2 presents three saliency score, which can be easily computed and represent different aspects of the quality of the elliptic hypotheses. Section 5.5.3 presents several ways of combining the saliency scores and studies the issues with each combination. Section 5.5.4 proposes a hypotheses selection scheme which does not require user specified thresholds or filtering parameters. The scheme is non-heuristic and determines the suitable values of threshold from the image itself. The scheme works well for most images.

Several numerical evaluations and comparison with other methods are presented in section 5.6. It is seen that the proposed method performs much better than several other hybrid methods. The chapter is concluded in section 5.7.

**Chapter 6:**

Chapter 6 presents three practical applications in which the geometric primitives discussed in Chapter 2 to Chapter 5 are used. The first application – detection of elliptic shapes in real images of Caltech 256 dataset [1] – is presented in section 6.1. The contents of this section are presented in [45]. It is noted that the method presented in Chapter 5 performs better than most other hybrid methods and provides reasonable performance for use in practical applications.

Section 6.2 presents the biomedical application of detecting cell organelles from microscopy images. Its contents has been reported in [140]. The highlight of the method proposed for this problem in section 6.2.1 is the preprocessing. The discussion in section 6.2.1.1 shows how the gray scale histogram for all the images in the dataset can be used for improving the contrast of the images (which have extremely poor contrast). The method shows very good performance. Further, it is shown that the time taken by the method for each image can be pushed to below 1 second using parallel computing and more efficient openCV and C++ implementations.

Section 6.3 presents the application of object detection in images. Its contents has been reported in [141]. The object detection method presented in this section uses *only approximating polygons and detected ellipses as the features of the object*. The hierarchical object representation template with generative and discriminative capabilities presented in section 6.3.1 is a highlight of this method. The performance of the object detection method is shown for 16 object categories spread over two datasets.

**Chapter 7 and Appendix:**

Chapter 7 presents the conclusion and potential future directions of this thesis. There are six appendices labeled Appendix A - F. Appendix A provides an important proof for a theorem used in Chapter 2. Appendices B - D provide important derivations of the expressions in Chapter 3. Appendices E and F concern Chapter 6. Appendix E provides the details of the tool used for generating the ground truth for section 6.1. Appendix F provides a brief summary on the various aspects of object detection methods.

# Chapter 2 : Polygonal approximation of digital curves

## 2.1 Background

In this chapter, the problem of representing a digital curve as a polygon is considered. The problem of polygonal approximation (PA) of digital curves is often cast as either a min-# problem or a min-$\varepsilon$ problem [142]. Both the problems are essentially minimization problems. However, the former's aim is to find the minimum number of points such that the value of a particular error function is below a certain threshold, while the latter's aim is to find a fixed number of dominant points such that the error function has minimum value. Recently, several methods have been proposed to obtain the PA of digital curves in the framework of min-# problem. This is because it is difficult to determine the fixed number of points in min-$\varepsilon$ problem suitably for many shapes, while if the error function in min-# is related to the quality of fit, it is easier to use heuristics to determine a generally acceptable threshold for the error function.

A rigorous literature review of the problem reveals that the problem is quite old. Some of the initial works in this topic are [12, 52, 55, 61, 62, 142-149]. Despite several years of research and apparent simplicity of the problem, it remains relevant in the current era as well [35-37, 40, 42, 46-50, 57, 60, 64, 150-159]. The reasons for the continued relevance of this problem in the current era are highlighted here:

1. The performance metrics for evaluating the performance of a PA method.
2. The control parameter and optimization goal used in the PA method.

### 2.1.1 Performance metrics

First, even though many algorithms were developed for PA of digital curves, the metrics used to compare and benchmark various algorithms were not effective, as shown in [149, 159]. Researchers tried using absolute performance metrics such as compression ratio, integral square error, figure of merit, zero norm, infinity norm, etc. Studies proved that such metrics fail to represent the quality of fit in one manner or another [151, 159]. The reason for this was not understood fully. This chapter shows that there is a perennial conflict in the quality of fit in the local scale (precision at the level of a few pixels) and global scale (reliability at the level of complete curve) [93, 134]. This duality of precision and reliability of a fit haunts most fitting algorithms. This aspect is highlighted in section 2.2.2. Due to this reason, most absolute metrics fail in properly quantifying the quality of fit. The existing absolute measures in the context of local and global quality of fits are presented in section 2.2.3. It is shown that the absolute metrics either concentrate on the local or the global natures of fit. Several PA methods are studied in the context of the local and global natures of fitting in section 2.6.

Researchers tried relative measures like fidelity, efficiency, and merit to quantify the quality of fit [149, 159]. In relative measures, a so-called optimal algorithm is considered as the reference for comparing the performance of the algorithm being tested. The method proposed by Perez and Vidal [52] based on dynamic programming is generally used by the researchers as the reference algorithm. This is because it targets $\min-\varepsilon$ and $\min-\#$ such that the fitting error is minimized for a certain number of points ($\min-\varepsilon$) or the number of points for fitting is minimized for a given value of fitting error ($\min-\#$). It is logical that there is no way of determining an optimal value for the fixed number of points ($\min-\varepsilon$) or the fixed value of fitting error ($\min-\#$), because such a value depends upon the nature of the digital curve for which PA is sought. One PA method that uses the precision and reliability metrics for PA is proposed in section 2.2.4.

### 2.1.2 Control parameter and optimization goal

As mentioned above, heuristics are involved in choosing the threshold (for min-# problem) or the fixed number of points (in min-$\varepsilon$ problem). The control parameters used in most PA methods are often related to the maximum deviation of the pixels on the digital curve segment between adjacent dominant points from the line segment connecting the dominant points. When the control parameters are related to the maximum deviation, the allowable or tolerable maximum deviation is chosen heuristically as a threshold value. Although the threshold is generally chosen to be a constant, a suitable value of the threshold varies from one digital curve to another and even within the digital curve. However, no specific rules are available for choosing either the constant threshold value or an adaptive threshold value depending upon the digital curve.

This thesis proposes a non-parametric framework for the automatic and adaptive determination of the threshold for the min-# problem. A theoretical bound for the maximum deviation of a set of pixels by digitizing a line segment is first derived in section 2.2.4. This explicit and analytically defined bound is related to the length and the slope of the line segment. This bound is compared to other theoretical bounds related to the digitization of continuous line segments and it is shown that this bound is more versatile than the other bounds.

The maximum possible deviation due to digitization can be computed for any line segment using the proposed error bound. Thus, the computed maximum possible deviation can be easily used as a threshold for each individual edge of the polygon. In this sense, it serves as a natural benchmark for any dominant point detection method. The application of the proposed non-parametric framework in the dominant point detection methods is demonstrated in section 2.4.3. Three popular PA methods by Ramer, Douglas, and Peucker [12, 13] (referred to as RDP), Masood [46] (referred to as Masood), and Carmona-Poyato [35] (referred to as Carmona for brevity) are adapted in the proposed non-parametric framework. The modified non-parametric versions of the methods are called RDP-mod, Masood-mod, and Carmona-mod respectively.

The adapted versions of the methods are control parameter independent and do not require user specified inputs (thus making these algorithms free from heuristics). The modified methods show balanced performance over their original version despite being control parameter independent. The bound based non-parametric framework can be easily integrated in other PA or Dominant point detection methods as well and can be used to make them non-heuristic (or less heuristic) and self-adaptive. It is highlighted that in the proposed framework, the basic construct and the nature of the algorithms remain unchanged, while only optimization or termination condition is altered in order to make the method non-parametric.

### 2.1.3 Experiments for PA methods

For investigating the performance of the four PA methods proposed in this thesis (Precision Reliability Optimization (PRO), RDP-mod, Masood-mod, Carmona-mod), quantitative and qualitative analysis of the methods for 18 standard digital shaped are used to assess PA methods. In addition to this, some interesting and important experiments that reflect interesting and useful properties are considered in section 2.6. The first experiment studies the performance of PA methods for digital curves corresponding to scaled versions of the same shape. The second experiment considers the performance of the PA methods for noisy digital curves. The third experiment investigates if the PA methods can be applied for semi-digitized or non-digitized curves represented by discrete points with real valued coordinates. In the last experiment, the performance of the PA methods over large datasets of real images is considered. In general, the performance metrics are defined for a single digital curve. However, for the last experiment, it is important to find generalized versions of the performance metrics which can represent the performance of PA methods for a set of images, each containing several digital curves.

### 2.1.4: Outline and major contributions

The global and local properties of polygonal fitting upon edge curves are discussed in section 2.2. Such an analysis of the global and local properties of fit has been done for the first time in this doctoral. Precision metric for local quality of fit and reliability metric for global quality of fit are proposed in section 2.2.1. For least square fitting, it is shown in section 2.2.2 that the precision and reliability are always at conflict. This thus corroborates with the fact discussed in section 2.1.1 that the available metrics are not singly sufficient for PA. Section 2.2.3 discussed whether the conventionally used performance metrics represent the global or local qualities of fit. Section 2.2.4 generalizes the precision and reliability metrics so that they can be applied to an edge curve, or an image with several edge curves, or a dataset with several images.

Section 2.3 proposes a novel PA method that uses simultaneous optimization of both the precision and reliability metrics for an edge curve. The algorithm, called precision and reliability optimization (PRO), is proposed in section 2.3.1. Numerical experiments are presented in section 2.3.2. It is shown that the control parameter can be easily tweaked to obtain the desired performance of PRO.

Section 2.4 presents a very important contribution of this chapter. It presents the analytical derivation of the error due to digitization of a continuous line segment in sections 2.4.1 and 2.4.2. Section 2.4.2 also proposes the non-parametric framework for PA. The error bound can be analytically derived and depends upon both the size and the orientation of the continuous line segment. This error bound is compared against other simpler error bounds proposed earlier by other researchers in section 2.4.3.

Section 2.5 adapts three existing methods into a novel non-parametric framework and makes them parameter-independent. The three methods are quite different from each other and yet the non-parametric framework can be used for them successfully. The numerical results show that the non-parametric versions of these methods perform either better than or similar to the original versions of these methods.

The optimization goals and update schemes of several methods are analyzed in the context of precision, reliability, and the analytical error bound in section 2.6. Various interesting and practical aspects of PA methods are analyzed in section 2.7. In this section, the performance of the methods proposed in sections 2.3 and 2.5 are compared for scalability, noisy curves, non-digitized curves, and performance over large datasets of practical use. Some recommendations regarding the choice of PA methods in practical applications are provided in section 2.7.5. The chapter is concluded in section 2.8.

## 2.2 Local and global qualities of fit

### 2.2.1 Precision and reliability metrics

This section presents the recently proposed precision and reliability criteria which represent the local and global nature of fitting. Suppose a digital curve is represented using a sequence of connected pixels $S = \left\{ P_i \left( x_i, y_i \right) \right\}$, $i = 1$ to $M$ and it is approximated using a line $ax + by = 1$.

For considering the local characteristics of fitting, it is useful to compute the algebraic distance of the pixels $P_i \left( x_i, y_i \right)$ from the line $ax + by = 1$. Thus, the precision metric of fitting can be modeled using the normalized residue given in eqn. (2-1):

$$\varepsilon_p' = \frac{\left\| \mathbf{X}\overline{\mathbf{A}} - \overline{\mathbf{J}} \right\|_2}{\left\| \overline{\mathbf{A}} \right\|_2} \tag{2-1}$$

where $\mathbf{X} = \left[ \begin{bmatrix} x_1 & x_2 & \cdots & x_M \end{bmatrix}^T \quad \begin{bmatrix} y_1 & y_2 & \cdots & y_M \end{bmatrix}^T \right]$, $\overline{\mathbf{A}} = \begin{bmatrix} a & b \end{bmatrix}^T$, and $\overline{\mathbf{J}}$ is a column matrix containing $M$ rows, the every element of which is 1. The superscript $T$ denotes the transpose operation and $\left\| \bullet \right\|_p$ represents the $p-$norm of vectors. Since $\varepsilon'_p$ considers the residue for each pixel, it is characteristic of the local nature of fitting alone.

On the other hand for the global characteristics of fit, reliability of a fit refers to how well the fit is expected to satisfy at least two conditions:

i. The fit should be valid for a sufficiently large region (or in this case a long curve).

ii. It should not be sensitive to occasional spurious large deviations in the edge.

A combination of both these properties can be sought by defining a reliability metric as shown in eqn. (2-2).

$$\varepsilon_r = \frac{\left\| \mathbf{X}\overline{\mathbf{A}} - \overline{\mathbf{J}} \right\|_1}{s_{\max}} \tag{2-2}$$

where $s_{\max}$ is the maximum Euclidean distance between any two pair of pixels [93]. Here, $|\bullet|$ represents the magnitude or the absolute value.

### 2.2.2 Duality of precision and reliability

If a line $ax + by = 1$ has to be fitted on the digital curve $S$, then the coefficients of the line, $a$ and $b$, can be determined by solving the matrix equation [160] of eqn. (2-3) for the vector $\overline{\mathbf{A}}$:

$$\mathbf{X}\overline{\mathbf{A}} = \overline{\mathbf{J}}, \tag{2-3}$$

The above equation is clearly an over-determined equation. Further, due to the discretization, the integer co-ordinates (pixels) closest to the required line may not exactly lie on the line. According to the consistency criterion, equation (2-3) has a unique solution only when the rank of $\mathbf{X}$ and $\left[ \mathbf{X} \vdots \overline{\mathbf{J}} \right]$ is same [161]. It is notable that due to the discretization in pixels, $x_i$ and $y_i$ being integers, the consistency criterion is satisfied only in four cases: a perfectly horizontal line, a perfectly vertical line, and the lines with slopes $\pm 1$. These cases are shown in Figure 2.2-1(a-d). For these cases, there are exactly two independent rows in $\mathbf{X}$ and the solution is unique.

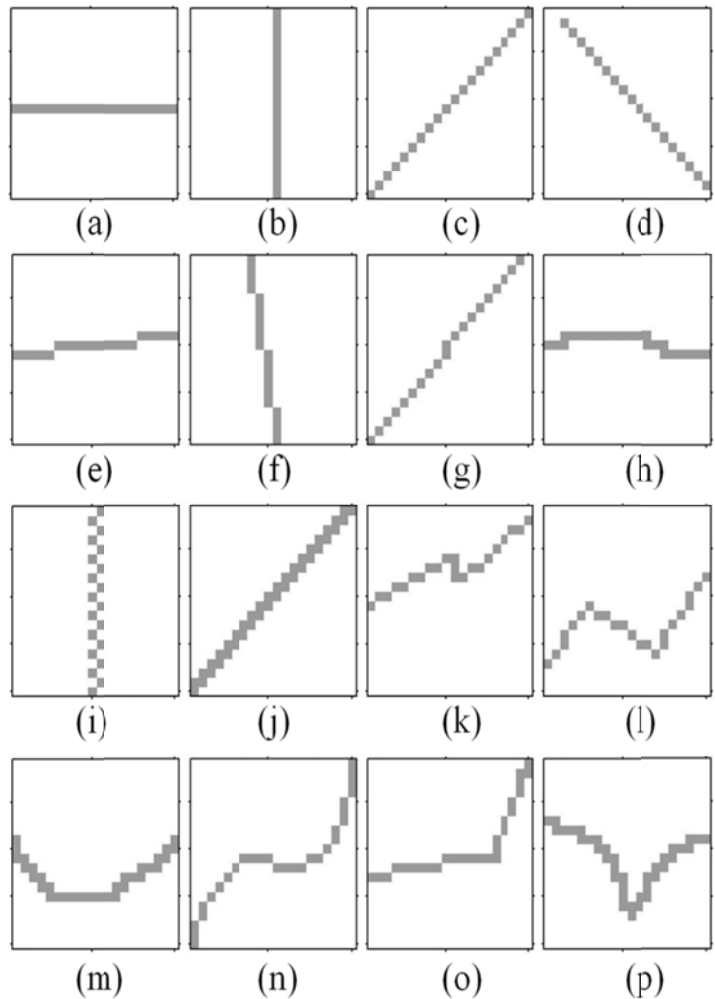

**Figure 2.2-1: Examples of digital curves (lines) for the precision-reliability duality.**

For all the other cases, using least squares method, the above equation can be solved for minimum residue as in eqn. (2-4) below [160, 161]:

$$\overline{\mathbf{A}} = \left(\mathbf{X}^{\mathrm{T}}\mathbf{X}\right)^{\mathbf{-1}} \mathbf{X}^{\mathrm{T}}\overline{\mathbf{J}} \,. \tag{2-4}$$

Now, using the above solution in eqn. (2-1) and assigning $\mathbf{B} = \mathbf{X}\left(\mathbf{X}^{\mathrm{T}}\mathbf{X}\right)^{-1}\mathbf{X}^{\mathrm{T}}$, $\mathbf{C} = \mathbf{X}\left(\mathbf{X}^{\mathrm{T}}\mathbf{X}\right)^{-2}\mathbf{X}^{\mathrm{T}}$, the precision metric can be written as in eqn. (2-5) below:

$$\varepsilon_p' = \frac{\sqrt{\left(\mathbf{B}\overline{\mathbf{J}}-\overline{\mathbf{J}}\right)^T \left(\mathbf{B}\overline{\mathbf{J}}-\overline{\mathbf{J}}\right)}}{\sqrt{\overline{\mathbf{A}}^T \overline{\mathbf{A}}}} = \frac{\sqrt{\overline{\mathbf{J}}^T \left(\mathbf{B}-\mathbf{I}\right)^T \left(\mathbf{B}-\mathbf{I}\right)\overline{\mathbf{J}}}}{\sqrt{\overline{\mathbf{J}}^T \mathbf{C}\overline{\mathbf{J}}}} = \frac{\sqrt{\overline{\mathbf{J}}^T \left(\mathbf{B}^T\mathbf{B}-\mathbf{B}-\mathbf{B}^T+\mathbf{I}\right)\overline{\mathbf{J}}}}{\sqrt{\overline{\mathbf{J}}^T \mathbf{C}\overline{\mathbf{J}}}}$$

$$= \frac{\sqrt{\overline{\mathbf{J}}^T \left(\mathbf{I}-\mathbf{B}\right)\overline{\mathbf{J}}}}{\sqrt{\overline{\mathbf{J}}^T \mathbf{C}\overline{\mathbf{J}}}} = \sqrt{\frac{\sum\left(\mathbf{I}-\mathbf{B}\right)}{\sum\mathbf{C}}} = \sqrt{\frac{M-\sum\mathbf{B}}{\sum\mathbf{C}}}$$

(2-5)

where $\mathbf{I}$ is the identity matrix of size $M$, $\sum\mathbf{A}$ represents the sum of all the elements of the matrix $\mathbf{A}$ and the properties of matrix $\mathbf{B}$ and vector $\overline{\mathbf{J}}$ in eqn. (2-6) and (2-7) respectively are used:

$$\mathbf{B} = \mathbf{B}^{\mathrm{T}} = \mathbf{B}^{\mathrm{T}}\mathbf{B} \tag{2-6}$$

$$\overline{\mathbf{J}}^T \mathbf{A}\overline{\mathbf{J}} = \sum\mathbf{A} \tag{2-7}$$

Further, it is proven in Appendix A that eqn. (2-8) below is valid:

$$\sum\mathbf{B} \le M \tag{2-8}$$

Thus, it is evident from eqn. (2-5) that the precision metric $\varepsilon_p'$ can be reduced by choosing lesser number of pixels $M$. It should be noted that with the decrease in the number of pixels, $\mathbf{X}$ and consequently $\mathbf{B}$ and $\mathbf{C}$ change. This in effect amount to saying that fitting the line in a smaller local region is more precise than a large region.

From eqn. (2-2), since a sequence of pixels is considered, $s_{\max}$ can be high if and only if the number of pixels $M$ is high. Thus, there is always a contradiction between precision and reliability. In order to increase the precision, smaller regions should be considered for fitting, whereas for increasing the reliability, larger regions need to be considered (largest region being the region spanned by the connected edge pixels under consideration).

Indeed the contradiction does not occur in ideal lines as shown in Figure 2.2-1 (a-d). Further, the contradiction is not an issue if the lines are in general smooth, so that the precision within a large region is already very high, such that reliability and precision are already sufficiently high and there is no practical need to increase the precision or reliability. Some such examples are presented in Figure 2.2-1(e-g) and the lack of contradiction in such cases is seen in Figure 2.2-2 (e-g) and Table 2.2-1.

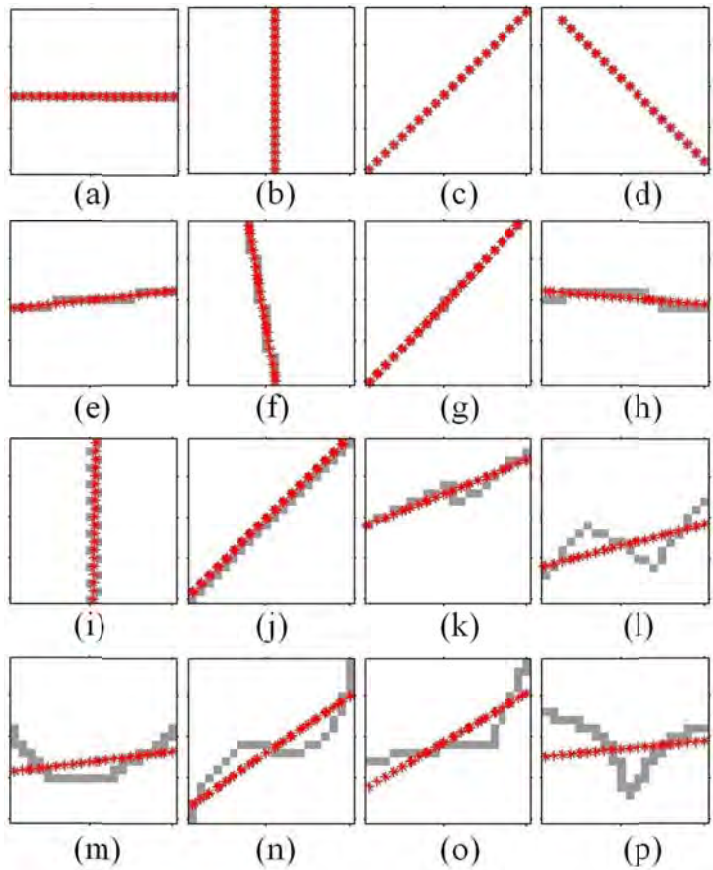

**Figure 2.2-2: Fitted line (red asterisks) on the digital curves in
Figure 2.2-1 using the least squares approach.**

**Table 2.2-1: Precision and reliability metrics for Figure 2.2-2.**

| Digital curve | Figure 2.2-2(a) | Figure 2.2-2(b) | Figure 2.2-2(c) | Figure 2.2-2(d) |
|---|---|---|---|---|
| Precision metric $\varepsilon'_p$ | 0.000 | 0.000 | 0.000 | 0.000 |
| Reliability metric $\varepsilon_r$ | 0.000 | 0.000 | 0.000 | 0.000 |
| Digital curve | Figure 2.2-2(e) | Figure 2.2-2(f) | Figure 2.2-2(g) | Figure 2.2-2(h) |
| Precision metric $\varepsilon'_p$ | 0.024 | 0.035 | 0.013 | 0.071 |
| Reliability metric $\varepsilon_r$ | 0.022 | 0.031 | 0.008 | 0.066 |
| Digital curve | Figure 2.2-2(i) | Figure 2.2-2(j) | Figure 2.2-2(k) | Figure 2.2-2(l) |
| Precision metric $\varepsilon'_p$ | 0.047 | 0.015 | 0.072 | 0.121 |
| Reliability metric $\varepsilon_r$ | 0.048 | 0.007 | 0.057 | 0.114 |
| Digital curve | Figure 2.2-2(m) | Figure 2.2-2(n) | Figure 2.2-2(o) | Figure 2.2-2(p) |
| Precision metric $\varepsilon'_p$ | 0.150 | 0.111 | 0.121 | 0.233 |
| Reliability metric $\varepsilon_r$ | 0.135 | 0.107 | 0.115 | 0.248 |

Examples of more practical cases are shown in Figure 2.2-1(h)-(p). In such cases, the contradiction comes into picture strongly as seen in Figure 2.2-2(h-p) and Table 2.2-1 (values are shown till 3 decimal points.). For practical representation of the curves in these cases, a balance has to be achieved such that the fit is sufficiently reliable as well as precise.

### 2.2.3 Existing performance metric in the context of precision and reliability

In this section, the absolute performance metrics generally used for assessing the quality of fit are discussed in the context of the precision and reliability metrics (eqns. (2-1) and (2-2) respectively). The aim here is to investigate if the performance metric under consideration in this section represents local quality of fit or global quality of fit.

#### 2.2.3.1 Maximum deviation (MD, $\max(d_i)$ )

Suppose the minimum distance for each pixel on a digital curve from its approximate polygon is computed. This distance is called the deviation of the pixel from the polygon. The maximum deviation (MD) of a digital curve from its polygon then corresponds to the pixel for which the deviation is maximum. For convenience, it is referred to as MD or $\max(d_i)$. For a general line given by eqn. (2-3) fitted upon a digital curve (or its portion), $\max(d_i)$ is specified by eqn. (2-9) below:

$$\max(d_i) = \left\| \mathbf{X}\overline{\mathbf{A}} - \overline{\mathbf{J}} \right\|_\infty \Big/ \left\| \overline{\mathbf{A}} \right\|_2 \tag{2-9}$$

Since $\left\| \bullet \right\|_\infty \leq \left\| \bullet \right\|_2$ [112], $\max(d_i) \leq \varepsilon'_p$. Thus, it can be concluded that $\max(d_i)$ is a form of precision (local fit) measure.

#### 2.2.3.2 Integral square error (ISE)

Integral square error (ISE) is the sum of squares of deviations of the pixels from the approximate polygon is given by eqn. (2-10) below:

$$\text{ISE} = \left( \varepsilon'_p \right)^2 \tag{2-10}$$

Thus, effectively, ISE is also a precision (local fit) measure.

#### 2.2.3.3 Dimensionality reduction (DR) ratio or compression ratio (CR)

The compression ratio given in eqn. (2-11) is the ratio of number of pixels in the digital curve ( $N$ ) to the number of vertices of the approximate polygon ( $M$ ),

$$CR = N/M \tag{2-11}$$

Though this measure is not related to either precision or reliability, it is an important performance metric in practice. A larger value of CR is beneficial for the reduction of data and computational resources for further higher level processing. Instead of compression ratio, its reciprocal dimensionality reduction ratio given in eqn. (2-12) can be used as a minimization metric (i.e. the lesser, the better).

$$DR = CR^{-1} = M/N \tag{2-12}$$

### 2.2.3.4 Figure of merit (FOM)

Figure of merit (FOM) is given by eqn. (2-13):

$$FOM = CR/ISE \tag{2-13}$$

This is a maximization metric, i.e., larger value of FOM is preferred over a lower value. However, it is well known that FOM is biased towards ISE [35]. For example, if the break points of a digital curve [46] are considered as the dominant points, the ISE is zero and inconsequent of the CR, FOM is infinity. If a minimization metric is preferred, $WE^{1}$=1/FOM [50] may be used. It suffers with the same deficiency as FOM.

### 2.2.3.5 Discussion

It is notable that most important metrics use local quality of fit. There are hardly any metrics that use the global quality of fit. The logic that is probably applied is that if the fitting is locally good for individual curve segments, the quality of fit all over the digital curve is also good. However, as shown earlier in section 2.2.2, there is often a contradiction between the local and global qualities of fit, i.e., a very precisely fitted curve may not be a reliable fit.

### 2.2.4 Generalizing precision and reliability metrics for datasets

This section extends the precision and reliability metrics presented in equations (2-1) and (2-2) to the cases of digital curves, images, and datasets. Suppose $J$ line segments are fitted upon a digital curve. Also, suppose that an image contains $K$ digital curves and a dataset contains $L$ images.

### 2.2.4.1 Precision metric

The net precision metric for the digital curve is defined as follows in eqn. (2-14) [134]:

$$\varepsilon_p' = \text{mean}\left(\varepsilon_p'^{\,j}; j = 1 \text{ to } J\right), \qquad (2\text{-}14)$$

where $\varepsilon_p'^{\,j}$ is the precision metric of the $j$th line segment, defined using eqn. (2-1). Since an algorithm is applied on one digital curve in one execution, hence taking the mean in eqn. (2-14) is reasonable. Further, since precision is a local property of fit, it can be defined only for the individual line segments and therefore the net precision accounts for the precision of every line segment. The precision metric of an image is defined as in eqn. (2-15):

$$\varepsilon_p' = \max\left(\varepsilon_p'^{\,k}; k = 1 \text{ to } K\right), \qquad (2\text{-}15)$$

where $\varepsilon_p'^{\,k}$ is the precision metric of the $k$th digital curve, defined using eqn. (2-14). It should be noted that instead of taking the mean as in eqn. (2-14), max is used for computing the metric for an image in eqn. (2-15). This is because one curve may differ from another curve in an image due to various factors like the shapes, lighting, noise, etc. in the image. The precision metric of a dataset of images is defined as in eqn. (2-16):

$$\varepsilon_p' = \text{mean}\left(\varepsilon_p'^{\,l}; l = 1 \text{ to } L\right), \qquad (2\text{-}16)$$

where $\varepsilon_p'^{\,l}$ is the precision metric of the $l$th digital curve, defined using eqn. (2-15). Taking the mean as the metric for an image is reasonable.

### 2.2.4.2 Reliability metric

The net reliability measure of the digital curve is defined as follows in eqn. (2-17) [134]:

$$\varepsilon_r = \sum_{j=1}^{J} \left\| \mathbf{X}_j \, \overline{\mathbf{A}}_j - \overline{\mathbf{J}} \right\|_1 \Bigg/ \sum_{j=1}^{J} s_{\max}^j , \qquad (2\text{-}17)$$

where $\mathbf{X}_j$, $\overline{\mathbf{A}}_j$, and $s_{\max}^j$ correspond to $\mathbf{X}$, $\overline{\mathbf{A}}$, and $s_{\max}$ in eqn. (2-2) for the $j$th line segment. The above definition in the context of a digital curve is consistent with the concept of reliability metric as a metric of global fit. Thus, the net reliability measure does not consider the reliability measure of individual segment. Rather, the individual constituents of reliability measure (the numerator and denominator in eqn. (2-17)) are computed using all the segments.

### 2.2.4.3 Other performance metrics

Following four performance metrics are also generalized for an image or a dataset.

1. Compression ratio (CR): Compression ratio for an image is computed as the average CR of all the digital curves in an image. Similarly, the compression ratio of a dataset is computed as the average CR of all the images in the dataset.

2. Maximum deviation (MD): MD for an image is computed as the average MD of all the digital curves in an image. Similarly, MD of a dataset is computed as the average MD of all the images in the dataset.

3. Integral square error (ISE): ISE for an image is computed as the average ISE of all the digital curves in an image. Similarly, ISE of a dataset is computed as the average ISE of all the images in the dataset.

4. Figure of merit (FOM): FOM for an image is computed as the average FOM of all the digital curves in an image. Similarly, FOM of a dataset is computed as the average FOM of all the images in the dataset.

## 2.3 Algorithm based upon precision and reliability optimization (PRO)

In this section, a method designed to optimize both the precision and reliability is proposed. Since this method optimizes both the precision and reliability metrics, it is referred to as the precision and reliability based optimization (PRO) method. For the ease of further reference, a distance function is defined in eqn. (2-18):

$$d\left(a,b,p\right) = \frac{\left| x_p\left(y_a - y_b\right) + y_p\left(x_b - x_a\right) + y_b x_a - y_a x_b \right|}{\sqrt{\left(x_b - x_a\right)^2 + \left(y_a - y_b\right)^2}}. \tag{2-18}$$

This distance function denotes the distance of a point $P_p\left(x_p, y_p\right)$ from the line passing through two points $P_a(x_a, y_a)$ and $P_b(x_b, y_b)$. This distance function shall be used several times in this chapter and has been pre-defined for the ease and simplicity of reference. This function applies to points as well as pixels and the arguments $a$, $b$, and $p$ are the subscripts of the points in consideration.

### 2.3.1 Precision and Reliability based Optimization (PRO)

Consider a digital curve $S = \{P_1 \quad P_2 \quad ... \quad P_N\}$, where $P_i$ is the $i$th edge pixel in the digital curve $e$. The line passing through a pair of pixels $P_a(x_a, y_a)$ and $P_b(x_b, y_b)$ is given by eqn. (2-19):

$$x(y_a - y_b) + y(x_b - x_a) + y_b x_a - y_a x_b = 0.$$  (2-19)

The deviation $d_i$ of a pixel $P_i(x_i, y_i) \in S$ from the line passing through the pair $\{P_1, P_N\}$ is then given by eqn. (2-20):

$$d_i = d(1, N, i).$$  (2-20)

Accordingly, the pixel with maximum deviation (MD) can be found. Let it be denoted as $P_{max}$. Considering the pairs $\{P_1, P_{max}\}$ and $\{P_{max}, P_N\}$, two new pixels of maximum deviations are found from $S$ using the concept in the eqns. (2-19) and (2-20). It is evident that the MD goes on decreasing as newer pixels of MD between a pair are chosen.

This process is repeated until the condition in inequality (2-21) is satisfied by all the line segments.

$$\max(\varepsilon'_p, \varepsilon_r) < \varepsilon_0,$$  (2-21)

Typically, $\varepsilon_0$ is the chosen tolerance value and is typically less than 1.The pseudocode for the PRO algorithm is provided in Figure 2.3-1.

Since the optimization goal in PRO is based on precision and reliability measures $\varepsilon'_p$ and $\varepsilon_r$ (use of eqn. (2-21) as the optimization goal), the users get larger freedom in determining the nature of fit. For example, using $\varepsilon_0$ close to 0 (like 0.1 or 0.2) results into a very close fit on the curve, where the lines follow even small deviations in the curvature. This may be used to achieve high precision fit, where one needs to retain even the smallest deviations in the data, or where one studies the nature of noise in the curve itself. Using $\varepsilon_0$ close to 1 (like 0.9 or 1) results into line fit that smoothens over small spurious deviations and retains all significant curvature changes in the line fit.

```
Function DP=PRO ({P_1, P_2, ..., P_N})
{       DP=NULL; % DP contains the dominant points
%step 1: line and its parameters
Fit a line l using P_1 and P_N.
% step 2: maximum deviation
Find deviation {d_1, ..., d_N} of pixels {P_1, ..., P_N} from the line l.
Find d_max = max{d_1, ..., d_N} and point P_max corresponding to d_max.
Find ε'_p and ε_r.
%step 3: termination/recursion condition
If (ε'_p ≤ ε_0) AND (ε_r ≤ ε_0)
DP={DP,   P_1, P_N}
Else
{       DP={DP,   PRO (P_1, P_max)}.
DP={DP,   PRO (P_max, P_N)}.
}
End
Remove redundant points in DP.
Return(DP).}
```

**Figure 2.3-1: Pseudocode for PRO algorithm.**

### 2.3.2 Numerical examples

Some numerical examples of digital curves (18 curves considered in [35]) and the result of PRO for these curves are considered here. The digital curves and the approximate polygons are shown in Figure 2.3-2 and the performance metrics are tabulated in Table 2.3-1. The control parameter $\varepsilon_0$ used in PRO is listed in the parentheses in Table 2.3-1. The table lists the number of pixels in the digital sequence ($M$), number of dominant points (the number of vertices in the approximated polygon is $N$), maximum deviation $\max(d_i)$, integral square error (ISE), figure of merit (FOM), precision metric ($\varepsilon'_p$), reliability metric ($\varepsilon_r$), and compression ratio $CR = M/N$.

It is seen that PRO(0.2), and PRO(0.4) choose more number of pixels as the vertices of the approximate polygon. The fit is quite close to the curve in the local sense and does not represent the global curvature properties quite well. This is because both the precision and reliability values are optimized to be below 0.2 or 0.4, which is verified in the values of precision and reliability metrics in Table 2.3-1. In such event, the algorithm find it easiest to reduce $\left\|\mathbf{X}\overline{\mathbf{A}} - \overline{\mathbf{J}}\right\|_p$ which appears in both the precision and reliability metrics (eqns. (2-1) and (2-2) respectively). On the other hand, it is seen in Figure 2.3-2 that PRO(1.5) also does not fit the polygons on the digital curves very well. This happens because PRO is a splitting method which begins with the maximum value of $s_{\max}$ in eqn. (2-2). Thus, though the local fit may yet be poor, the method terminates as soon as $\varepsilon_p'$ and $\varepsilon_r$ fall below $\varepsilon_0 = 1.5$. It is noted in general that PRO(0.8) and PRO(1.0) give a reasonable tradeoff in the quality of fit for most curves as well as good compression ratio (CR in Table 2.3-1).

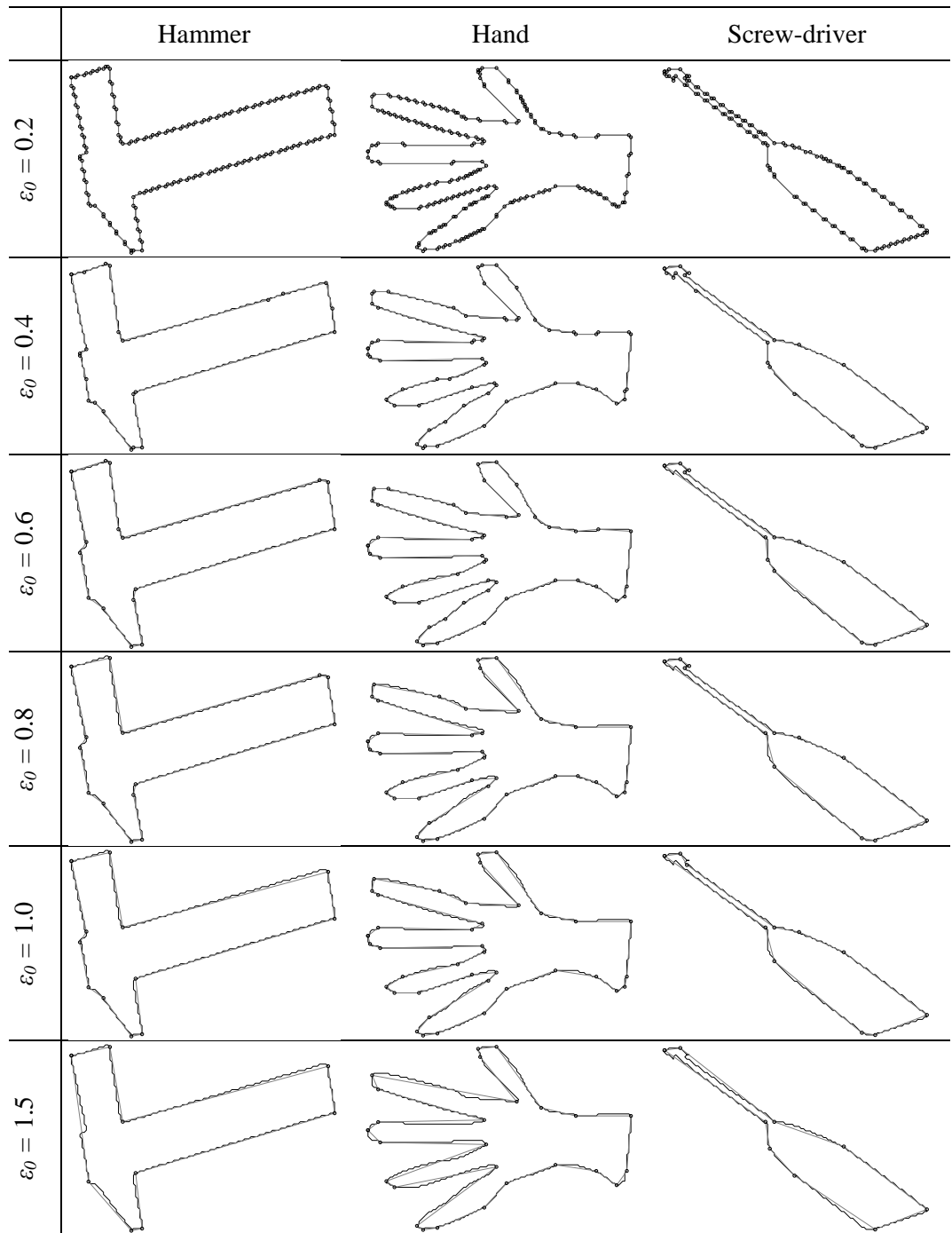

**Figure 2.3-2: Performance of PRO for selected digital curves.**

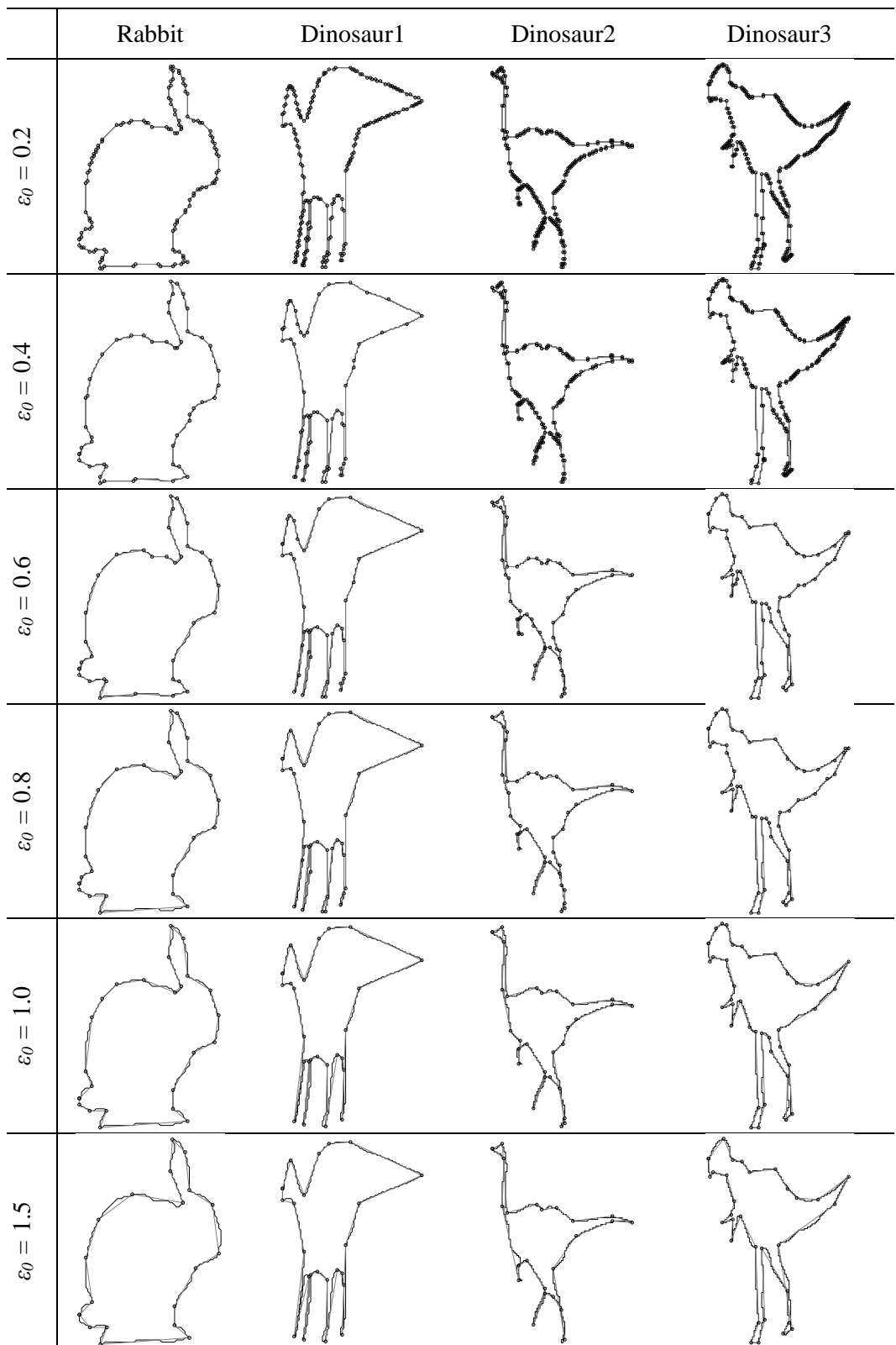

**Figure 2.3-2: Performance of PRO for selected digital curves... contd.**

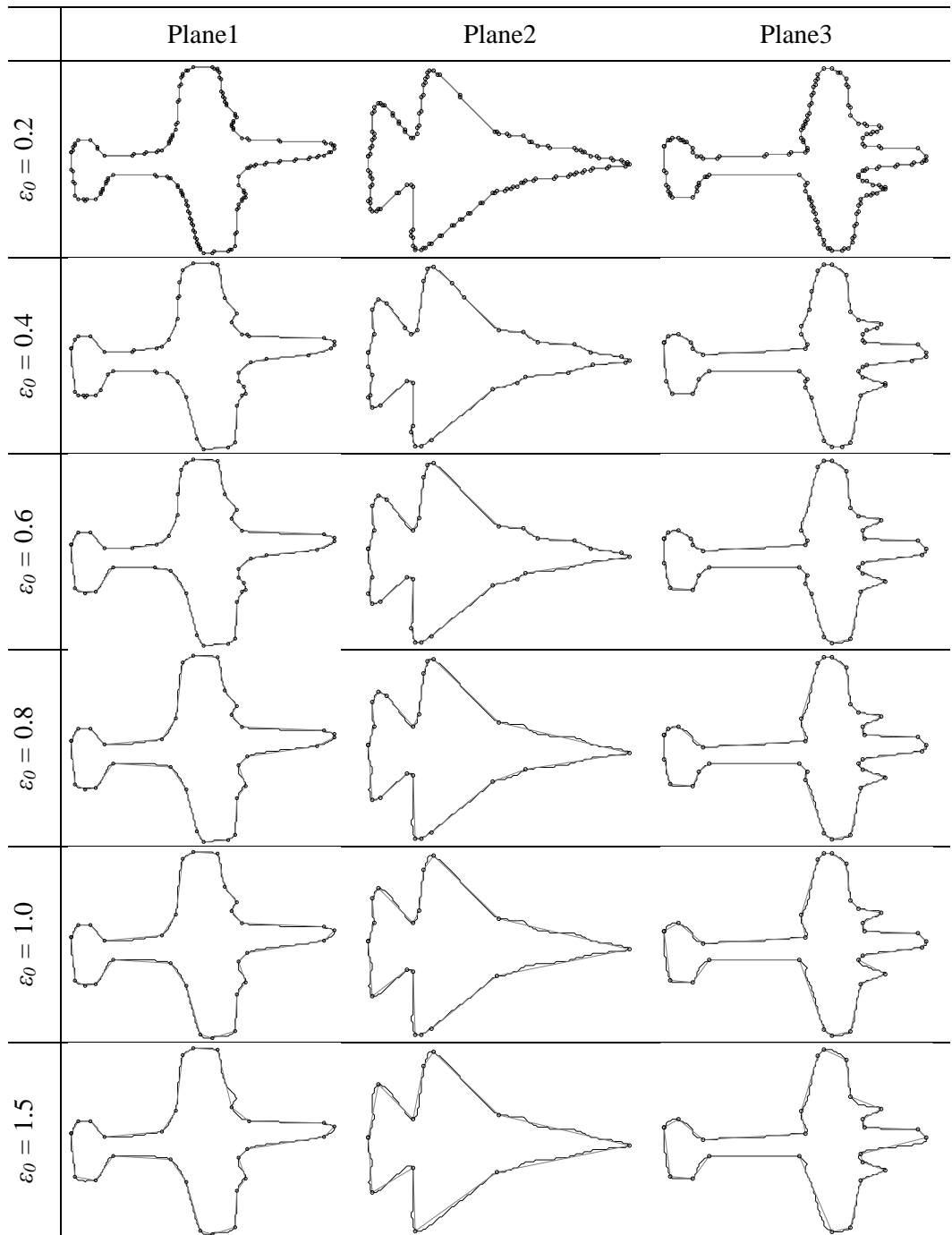

**Figure 2.3-2: Performance of PRO for selected digital curves... contd.**

| | Plane 4 | Plane 5 | Tinopener | Turtle |
|---|---|---|---|---|
| $\varepsilon_0 = 0.2$ | 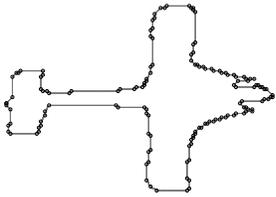 | 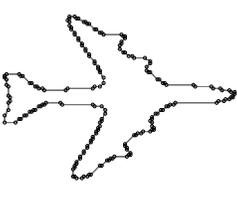 | 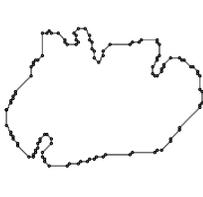 | 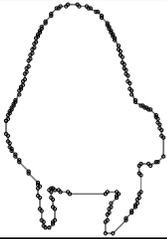 |
| $\varepsilon_0 = 0.4$ | 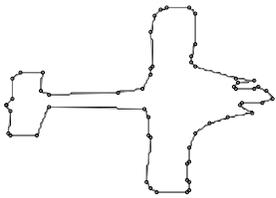 | 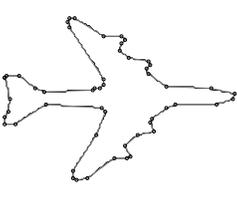 | 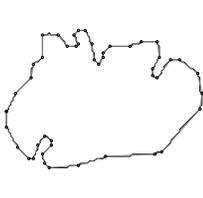 | 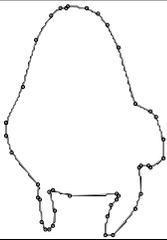 |
| $\varepsilon_0 = 0.6$ | 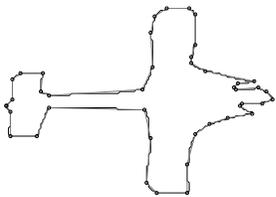 | 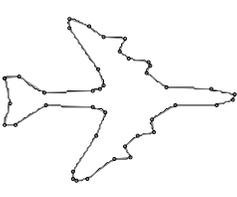 | 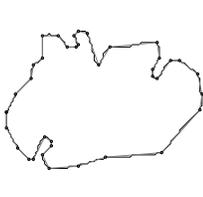 | 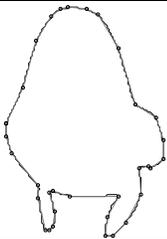 |
| $\varepsilon_0 = 0.8$ | 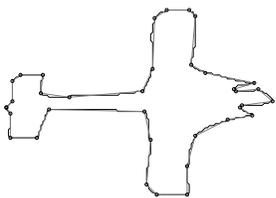 | 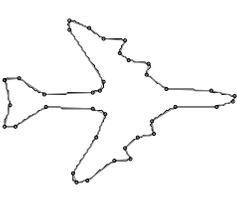 | 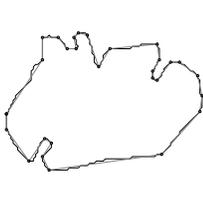 | 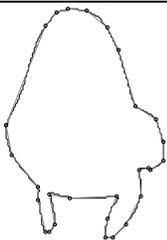 |
| $\varepsilon_0 = 1.0$ | 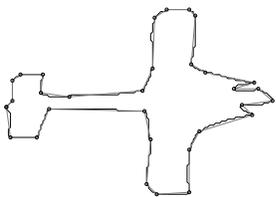 | 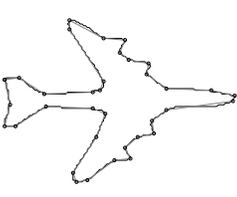 | 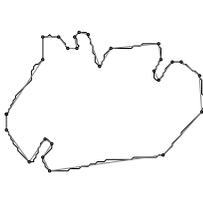 | 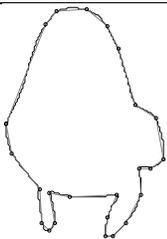 |
| $\varepsilon_0 = 1.5$ | 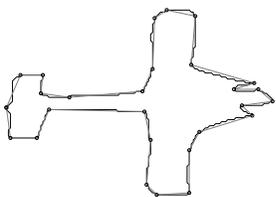 | 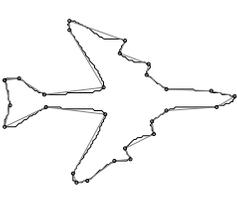 | 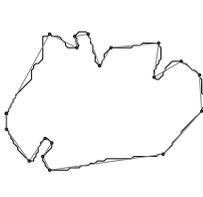 | 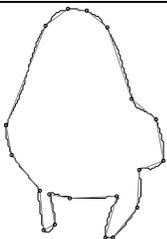 |

**Figure 2.3-2: Performance of PRO for selected digital curves... contd.**

| | Africa | Maple leaf | Sword fish | Dog |
|---|---|---|---|---|
| $\varepsilon_0 = 0.2$ | 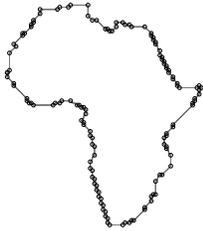 | 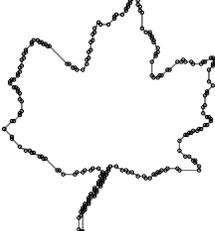 | 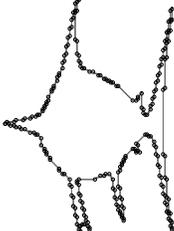 | 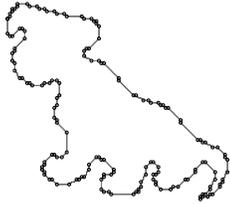 |
| $\varepsilon_0 = 0.4$ | 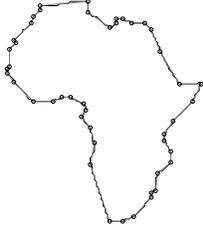 | 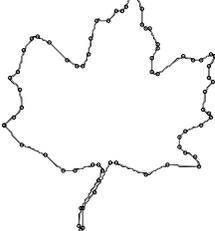 | 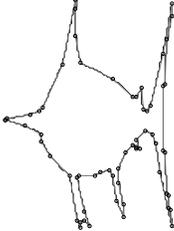 | 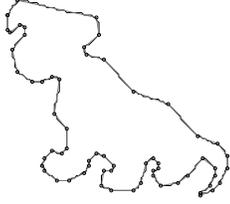 |
| $\varepsilon_0 = 0.6$ | 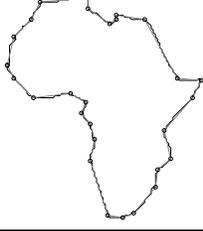 | 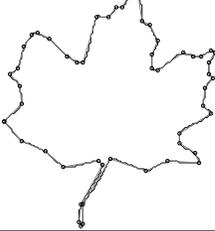 | 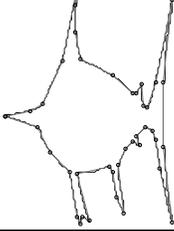 | 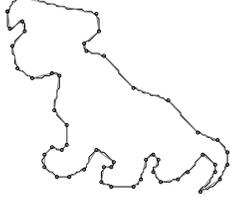 |
| $\varepsilon_0 = 0.8$ | 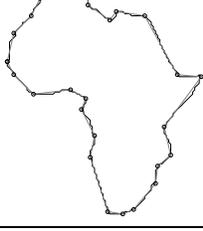 | 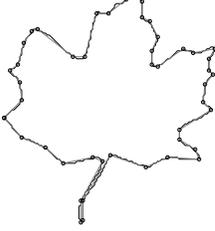 | 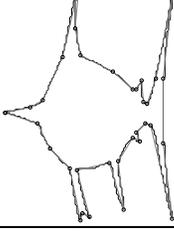 | 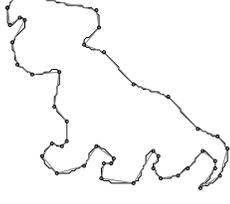 |
| $\varepsilon_0 = 1.0$ | 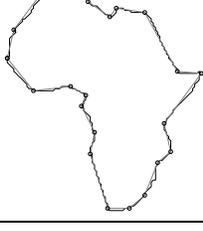 | 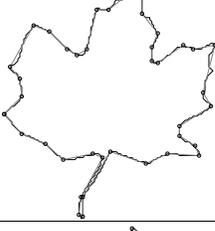 | 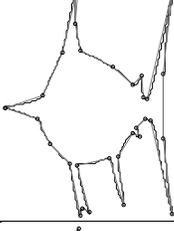 | 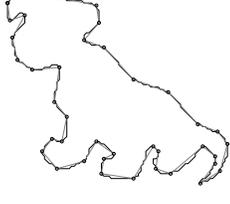 |
| $\varepsilon_0 = 1.5$ | 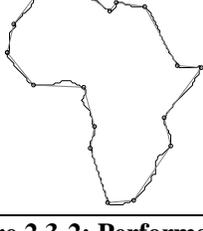 | 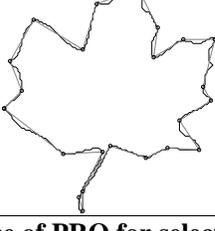 | 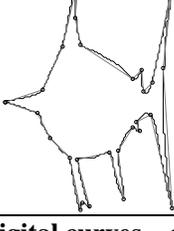 | 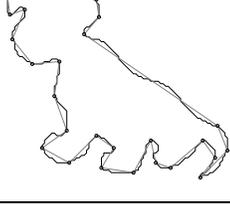 |

**Figure 2.3-2: Performance of PRO for selected digital curves... contd.**

**Table 2.3-1 (a): Quantitative performance of PRO**

**Quantitative comparison for the first six digital curves in Figure 2.3-2.**

| | No. of pixels (M) | Dominant points (N) | MD max($d_i$) | ISE | FOM | Precision $\varepsilon_p'$ | Reliability $\varepsilon_r$ | CR |
|---|---|---|---|---|---|---|---|---|
| **Hammer** | | | | | | | | |
| PRO(0.2) | 388 | 192 | 0.28 | 0.31 | 6.45 | 0.00 | 0.00 | 2.02 |
| PRO(0.4) | 388 | 24 | 0.81 | 47.98 | 0.33 | 0.25 | 0.27 | 16.17 |
| PRO(0.6) | 388 | 17 | 1.03 | 75.27 | 0.30 | 0.41 | 0.35 | 22.82 |
| PRO(0.8) | 388 | 15 | 1.41 | 94.25 | 0.27 | 0.47 | 0.39 | 25.87 |
| PRO(1.0) | 388 | 13 | 2.01 | 155.76 | 0.19 | 0.54 | 0.48 | 29.85 |
| PRO(1.5) | 388 | 10 | 2.50 | 236.67 | 0.16 | 0.71 | 0.59 | 38.80 |
| **Hand** | | | | | | | | |
| PRO(0.2) | 642 | 257 | 0.28 | 1.85 | 1.35 | 0.01 | 0.01 | 2.50 |
| PRO(0.4) | 642 | 64 | 0.82 | 58.24 | 0.17 | 0.16 | 0.20 | 10.03 |
| PRO(0.6) | 642 | 50 | 1.19 | 87.41 | 0.15 | 0.30 | 0.26 | 12.84 |
| PRO(0.8) | 642 | 41 | 1.96 | 183.97 | 0.08 | 0.41 | 0.37 | 15.66 |
| PRO(1.0) | 642 | 40 | 1.71 | 188.09 | 0.08 | 0.44 | 0.39 | 16.05 |
| PRO(1.5) | 642 | 27 | 2.54 | 610.11 | 0.04 | 0.71 | 0.71 | 23.78 |
| **Screw Driver** | | | | | | | | |
| PRO(0.2) | 253 | 134 | 0.28 | 0.46 | 4.09 | 0.00 | 0.01 | 1.89 |
| PRO(0.4) | 253 | 21 | 0.89 | 24.22 | 0.50 | 0.27 | 0.20 | 12.05 |
| PRO(0.6) | 253 | 14 | 1.27 | 46.80 | 0.39 | 0.38 | 0.29 | 18.07 |
| PRO(0.8) | 253 | 13 | 1.61 | 53.47 | 0.36 | 0.41 | 0.30 | 19.46 |
| PRO(1.0) | 253 | 12 | 1.89 | 60.04 | 0.35 | 0.48 | 0.32 | 21.08 |
| PRO(1.5) | 253 | 10 | 2.50 | 169.51 | 0.15 | 0.63 | 0.53 | 25.30 |
| **Rabbit** | | | | | | | | |
| PRO(0.2) | 293 | 127 | 0.28 | 1.38 | 1.66 | 0.01 | 0.02 | 2.31 |
| PRO(0.4) | 293 | 50 | 0.69 | 20.37 | 0.29 | 0.18 | 0.16 | 5.86 |
| PRO(0.6) | 293 | 35 | 1.17 | 47.51 | 0.18 | 0.30 | 0.27 | 8.37 |
| PRO(0.8) | 293 | 27 | 1.50 | 84.38 | 0.13 | 0.41 | 0.37 | 10.85 |
| PRO(1.0) | 293 | 25 | 1.60 | 115.08 | 0.10 | 0.50 | 0.45 | 11.72 |
| PRO(1.5) | 293 | 20 | 2.60 | 225.50 | 0.06 | 0.75 | 0.63 | 14.65 |
| **Dinosaur1** | | | | | | | | |
| PRO(0.2) | 587 | 213 | 0.28 | 1.69 | 1.64 | 0.01 | 0.01 | 2.76 |
| PRO(0.4) | 587 | 65 | 0.89 | 47.11 | 0.19 | 0.18 | 0.18 | 9.03 |
| PRO(0.6) | 587 | 41 | 1.22 | 96.62 | 0.15 | 0.30 | 0.27 | 14.32 |
| PRO(0.8) | 587 | 33 | 1.41 | 152.98 | 0.12 | 0.40 | 0.36 | 17.79 |
| PRO(1.0) | 587 | 26 | 2.16 | 305.28 | 0.07 | 0.54 | 0.55 | 22.58 |
| PRO(1.5) | 587 | 24 | 2.91 | 374.89 | 0.07 | 0.63 | 0.60 | 24.46 |
| **Dinosaur2** | | | | | | | | |
| PRO(0.2) | 446 | 266 | 0.00 | 0.00 | $\infty$ | 0.00 | 0.00 | 1.68 |
| PRO(0.4) | 446 | 181 | 0.78 | 32.66 | 0.09 | 0.07 | 0.16 | 2.46 |
| PRO(0.6) | 446 | 54 | 1.15 | 107.02 | 0.09 | 0.35 | 0.41 | 8.26 |
| PRO(0.8) | 446 | 41 | 2.00 | 133.83 | 0.10 | 0.41 | 0.46 | 10.88 |
| PRO(1.0) | 446 | 30 | 2.14 | 236.26 | 0.08 | 0.54 | 0.62 | 14.87 |
| PRO(1.5) | 446 | 24 | 4.01 | 370.47 | 0.06 | 0.65 | 0.75 | 18.58 |

### Table 2.5-3 (b): Quantitative performance of PRO

**Quantitative comparison for the next six digital curves in Figure 2.3-2.**

| | No. of pixels (M) | Dominant points (N) | MD max($d_i$) | ISE | FOM | Precision $\varepsilon'_p$ | Reliability $\varepsilon_r$ | CR |
|---|---|---|---|---|---|---|---|---|
| **Dinosaur3** | | | | | | | | |
| PRO(0.2) | 528 | 357 | 0.00 | 0.00 | $\infty$ | 0.00 | 0.00 | 1.48 |
| PRO(0.4) | 528 | 240 | 1.00 | 52.18 | 0.06 | 0.07 | 0.19 | 2.20 |
| PRO(0.6) | 528 | 60 | 1.32 | 152.93 | 0.08 | 0.38 | 0.45 | 8.80 |
| PRO(0.8) | 528 | 51 | 1.66 | 190.31 | 0.07 | 0.41 | 0.50 | 10.35 |
| PRO(1.0) | 528 | 39 | 2.09 | 307.77 | 0.06 | 0.55 | 0.63 | 13.54 |
| PRO(1.5) | 528 | 32 | 3.19 | 587.66 | 0.04 | 0.66 | 0.85 | 16.50 |
| **Plane1** | | | | | | | | |
| PRO(0.2) | 462 | 158 | 0.28 | 0.92 | 3.14 | 0.01 | 0.01 | 0.34 |
| PRO(0.4) | 462 | 56 | 0.80 | 34.54 | 0.24 | 0.17 | 0.18 | 0.12 |
| PRO(0.6) | 462 | 41 | 1.28 | 68.67 | 0.16 | 0.30 | 0.27 | 0.09 |
| PRO(0.8) | 462 | 32 | 1.58 | 133.77 | 0.11 | 0.40 | 0.40 | 0.07 |
| PRO(1.0) | 462 | 30 | 1.69 | 151.64 | 0.10 | 0.47 | 0.43 | 0.06 |
| PRO(1.5) | 462 | 27 | 4.09 | 233.97 | 0.07 | 0.57 | 0.48 | 0.06 |
| **Plane2** | | | | | | | | |
| PRO(0.2) | 365 | 135 | 0.28 | 0.31 | 8.71 | 0.00 | 0.00 | 2.70 |
| PRO(0.4) | 365 | 40 | 0.75 | 35.73 | 0.25 | 0.27 | 0.22 | 9.12 |
| PRO(0.6) | 365 | 31 | 1.28 | 62.49 | 0.19 | 0.34 | 0.30 | 11.77 |
| PRO(0.8) | 365 | 22 | 1.77 | 121.01 | 0.14 | 0.42 | 0.41 | 16.59 |
| PRO(1.0) | 365 | 18 | 2.16 | 191.07 | 0.11 | 0.53 | 0.51 | 20.28 |
| PRO(1.5) | 365 | 12 | 2.40 | 317.84 | 0.09 | 0.91 | 0.71 | 30.42 |
| **Plane3** | | | | | | | | |
| PRO(0.2) | 431 | 146 | 0.28 | 1.54 | 1.89 | 0.01 | 0.01 | 2.95 |
| PRO(0.4) | 431 | 50 | 0.89 | 37.08 | 0.23 | 0.19 | 0.20 | 8.62 |
| PRO(0.6) | 431 | 39 | 1.11 | 53.69 | 0.20 | 0.28 | 0.25 | 11.05 |
| PRO(0.8) | 431 | 35 | 1.56 | 80.82 | 0.15 | 0.32 | 0.29 | 12.31 |
| PRO(1.0) | 431 | 29 | 2.44 | 139.37 | 0.10 | 0.43 | 0.39 | 14.86 |
| PRO(1.5) | 431 | 25 | 2.48 | 248.29 | 0.07 | 0.61 | 0.52 | 17.24 |
| **Plane4** | | | | | | | | |
| PRO(0.2) | 450 | 146 | 0.28 | 0.77 | 4.06 | 0.01 | 0.01 | 3.08 |
| PRO(0.4) | 450 | 58 | 0.83 | 37.70 | 0.21 | 0.17 | 0.18 | 7.76 |
| PRO(0.6) | 450 | 43 | 1.21 | 70.72 | 0.15 | 0.28 | 0.26 | 10.47 |
| PRO(0.8) | 450 | 36 | 1.68 | 115.78 | 0.11 | 0.38 | 0.35 | 12.50 |
| PRO(1.0) | 450 | 32 | 1.68 | 160.73 | 0.09 | 0.47 | 0.43 | 14.06 |
| PRO(1.5) | 450 | 28 | 2.46 | 256.77 | 0.06 | 0.63 | 0.55 | 16.07 |
| **Plane5** | | | | | | | | |
| PRO(0.2) | 431 | 199 | 0.28 | 0.77 | 2.81 | 0.00 | 0.01 | 2.17 |
| PRO(0.4) | 431 | 53 | 0.85 | 36.63 | 0.22 | 0.20 | 0.20 | 8.13 |
| PRO(0.6) | 431 | 41 | 1.27 | 59.94 | 0.17 | 0.33 | 0.26 | 10.51 |
| PRO(0.8) | 431 | 38 | 1.49 | 67.08 | 0.17 | 0.37 | 0.28 | 11.34 |
| PRO(1.0) | 431 | 36 | 1.77 | 108.17 | 0.11 | 0.41 | 0.34 | 11.97 |
| PRO(1.5) | 431 | 28 | 3.07 | 402.19 | 0.04 | 0.70 | 0.69 | 15.39 |

**Table 2.5-3 (c): Quantitative performance of PRO**

**Quantitative comparison for the last six digital curves in Figure 2.3-2.**

| | No. of pixels (M) | Dominant points (N) | MD | ISE | FOM | Precision $\varepsilon'_p$ | Reliability $\varepsilon_r$ | CR |
|---|---|---|---|---|---|---|---|---|
| **Tin Opener** | | | | | | | | |
| PRO(0.2) | 278 | 116 | 0.28 | 0.77 | 3.09 | 0.01 | 0.01 | 2.40 |
| PRO(0.4) | 278 | 47 | 1.00 | 20.74 | 0.28 | 0.17 | 0.16 | 5.91 |
| PRO(0.6) | 278 | 38 | 1.21 | 35.96 | 0.20 | 0.24 | 0.23 | 7.32 |
| PRO(0.8) | 278 | 29 | 1.66 | 76.04 | 0.13 | 0.35 | 0.35 | 9.59 |
| PRO(1.0) | 278 | 27 | 1.66 | 90.16 | 0.11 | 0.40 | 0.38 | 10.30 |
| PRO(1.5) | 278 | 20 | 2.61 | 189.46 | 0.07 | 0.66 | 0.58 | 13.90 |
| **Turtle** | | | | | | | | |
| PRO(0.2) | 354 | 176 | 0.28 | 0.92 | 2.18 | 0.01 | 0.01 | 2.01 |
| PRO(0.4) | 354 | 43 | 1.00 | 37.86 | 0.22 | 0.21 | 0.22 | 8.23 |
| PRO(0.6) | 354 | 34 | 1.14 | 54.43 | 0.19 | 0.29 | 0.26 | 10.41 |
| PRO(0.8) | 354 | 29 | 1.48 | 95.46 | 0.13 | 0.36 | 0.36 | 12.21 |
| PRO(1.0) | 354 | 25 | 1.79 | 141.22 | 0.10 | 0.51 | 0.45 | 14.16 |
| PRO(1.5) | 354 | 20 | 2.24 | 258.88 | 0.07 | 0.76 | 0.64 | 17.70 |
| **Africa** | | | | | | | | |
| PRO(0.2) | 291 | 135 | 0.28 | 0.92 | 2.34 | 0.01 | 0.01 | 2.16 |
| PRO(0.4) | 291 | 47 | 1.00 | 24.03 | 0.26 | 0.17 | 0.16 | 6.19 |
| PRO(0.6) | 291 | 29 | 1.20 | 56.40 | 0.18 | 0.36 | 0.31 | 10.03 |
| PRO(0.8) | 291 | 26 | 1.54 | 70.55 | 0.16 | 0.42 | 0.34 | 11.19 |
| PRO(1.0) | 291 | 23 | 2.04 | 110.23 | 0.11 | 0.52 | 0.44 | 12.65 |
| PRO(1.5) | 291 | 19 | 2.74 | 195.69 | 0.08 | 0.67 | 0.56 | 15.32 |
| **Maple leaf** | | | | | | | | |
| PRO(0.2) | 424 | 224 | 0.28 | 1.38 | 1.36 | 0.01 | 0.01 | 1.89 |
| PRO(0.4) | 424 | 69 | 1.00 | 36.88 | 0.17 | 0.21 | 0.19 | 6.14 |
| PRO(0.6) | 424 | 53 | 1.14 | 65.52 | 0.12 | 0.28 | 0.26 | 8.00 |
| PRO(0.8) | 424 | 47 | 1.49 | 90.74 | 0.10 | 0.36 | 0.30 | 9.02 |
| PRO(1.0) | 424 | 41 | 2.18 | 131.65 | 0.08 | 0.43 | 0.36 | 10.34 |
| PRO(1.5) | 424 | 25 | 2.77 | 370.44 | 0.05 | 0.79 | 0.65 | 16.96 |
| **Sword Fish** | | | | | | | | |
| PRO(0.2) | 627 | 249 | 0.28 | 1.69 | 1.47 | 0.01 | 0.01 | 2.52 |
| PRO(0.4) | 627 | 64 | 0.78 | 55.87 | 0.17 | 0.20 | 0.20 | 9.80 |
| PRO(0.6) | 627 | 42 | 1.19 | 86.67 | 0.17 | 0.31 | 0.27 | 14.93 |
| PRO(0.8) | 627 | 34 | 1.57 | 148.94 | 0.12 | 0.40 | 0.36 | 18.44 |
| PRO(1.0) | 627 | 31 | 1.69 | 219.49 | 0.09 | 0.47 | 0.43 | 20.23 |
| PRO(1.5) | 627 | 28 | 2.34 | 388.73 | 0.06 | 0.64 | 0.57 | 22.39 |
| **Dog** | | | | | | | | |
| PRO(0.2) | 343 | 169 | 0.28 | 1.38 | 1.46 | 0.01 | 0.01 | 2.03 |
| PRO(0.4) | 343 | 64 | 0.85 | 28.13 | 0.19 | 0.19 | 0.18 | 5.36 |
| PRO(0.6) | 343 | 52 | 1.00 | 42.14 | 0.16 | 0.28 | 0.23 | 6.60 |
| PRO(0.8) | 343 | 44 | 1.51 | 78.06 | 0.10 | 0.36 | 0.32 | 7.80 |
| PRO(1.0) | 343 | 39 | 1.58 | 112.29 | 0.08 | 0.47 | 0.39 | 8.79 |
| PRO(1.5) | 343 | 27 | 2.69 | 384.93 | 0.03 | 0.92 | 0.79 | 12.70 |

## 2.4 Continuous and digital lines – error bound due to digitization

### 2.4.1 Error bound of the slope due to digitization

This section considers the effect of digitization on the slope of a line segment connecting two points (which may or may not be pixels). An upper bound for the deviation of the pixels obtained by the digitization of the line segment is derived. Due to digitization in the case of digital images, a general point $P(x, y)$ is approximated by a pixel $P'(x', y')$ as in eqn. (2-22):

$$x' = \text{round}(x); \quad y' = \text{round}(y) \tag{2-22}$$

where $\text{round}(x)$ denotes the rounding of the value of real number $x$ to its nearest integer. $P'(x', y')$ satisfy the conditions (2-23) - (2-25):

$$x', y' \in \mathbb{Z} \tag{2-23}$$

$$x' = x + \Delta x; \quad y' = y + \Delta y \tag{2-24}$$

$$-0.5 \leq \Delta x \leq 0.5, \quad -0.5 \leq \Delta y \leq 0.5 \tag{2-25}$$

Let the slope of the line $P_1 P_2$ (actual line) be denoted as $m$ and the slope of the line $P_1' P_2'$ (digital line) be denoted as $m'$. The slopes $m$ and $m'$ are then given by eqns. (2-26) and (2-27) below.

$$m = \tan \phi = \frac{y_2 - y_1}{x_2 - x_1} \tag{2-26}$$

$$m' = \frac{y_2' - y_1'}{x_2' - x_1'} = \left( m + \frac{\Delta y_2 - \Delta y_1}{x_2 - x_1} \right) \Big/ \left( 1 + \frac{\Delta x_2 - \Delta x_1}{x_2 - x_1} \right) \tag{2-27}$$

The angular difference between the numeric tangent and the digital tangent is used as the estimate of the error. This angular difference is given by eqn. (2-28):

$$\partial \phi = \left| \tan^{-1}(m) - \tan^{-1}(m') \right| = \left| \tan^{-1} \left( \frac{m - m'}{1 + mm'} \right) \right| \tag{2-28}$$

Eqn. (2-29) is obtained by substituting the eqn. (2-27) in the eqn. (2-28).

$$\partial\phi = \left| \tan^{-1}\left( \frac{\left(1+\dfrac{\Delta x_2 - \Delta x_1}{x_2 - x_1}\right)m - \left(m + \dfrac{\Delta y_2 - \Delta y_1}{x_2 - x_1}\right)}{\left(1+\dfrac{\Delta x_2 - \Delta x_1}{x_2 - x_1}\right) + m\left(m + \dfrac{\Delta y_2 - \Delta y_1}{x_2 - x_1}\right)}\right)\right|$$

$$= \left| \tan^{-1}\left( \frac{\left(\dfrac{\Delta x_2 - \Delta x_1}{x_2 - x_1}\right)m - \left(\dfrac{\Delta y_2 - \Delta y_1}{x_2 - x_1}\right)}{\left(1+m^2\right) + \left(\dfrac{\Delta x_2 - \Delta x_1}{x_2 - x_1}\right) + m\left(\dfrac{\Delta y_2 - \Delta y_1}{x_2 - x_1}\right)}\right)\right| \qquad (2\text{-}29)$$

$$= \left| \tan^{-1}\left( \frac{m\left(\Delta x_2 - \Delta x_1\right) - \left(\Delta y_2 - \Delta y_1\right)}{\left(1+m^2\right)\left(x_2 - x_1\right) + \left(\Delta x_2 - \Delta x_1\right) + m\left(\Delta y_2 - \Delta y_1\right)}\right)\right|$$

For convenience, $s$ and $t$ are defined as in eqns. (2-30) and (2-31), respectively.

$$s = \sqrt{\left(x_2 - x_1\right)^2 + \left(y_2 - y_1\right)^2} \qquad (2\text{-}30)$$

$$t = \frac{\left(\Delta x_2 - \Delta x_1\right)\left(x_2 - x_1\right)}{s^2} + \frac{\left(\Delta y_2 - \Delta y_1\right)\left(y_2 - y_1\right)}{s^2} \qquad (2\text{-}31)$$

Using eqns. (2-26), (2-30), and (2-31) in eqn. (2-29), the expression of $\partial\phi$ can be written as in eqn. (2-32) below.

$$\partial\phi = \left| \tan^{-1}\left( \left(\frac{x_2 - x_1}{s^2}\right)\left(1+t\right)^{-1}\left(m\left(\Delta x_2 - \Delta x_1\right) - \left(\Delta y_2 - \Delta y_1\right)\right)\right)\right| \qquad (2\text{-}32)$$

Now the following facts are together used in eqn. (2-32) in order to derive the analytical error bound.

- Due to eqn. (2-25), the maximum value of $\left|\Delta x_2 - \Delta x_1\right|$ and $\left|\Delta y_2 - \Delta y_1\right|$ is 1.

- $\left|\left(x_2 - x_1\right)/s\right|$ and $\left|\left(y_2 - y_1\right)/s\right|$ are both less than or equal to 1 due to the definition of $s$ in eqn. (2-30).

- For any digital line made of more than 3 pixels for 4-connected digital curve and more than 2 pixels for 8-connected digital curve, $s$ is always more that $\sqrt{2}$.

- As a consequence, $|t| < 1$.

- In general, the exact values of $|\Delta x_2 - \Delta x_1|$ and $|\Delta y_2 - \Delta y_1|$ are not known, which implies that the value of $t$ is not known except for the above mentioned fact that $|t| < 1$.

Thus, in order to derive the bound, infinite geometric series expansion is used in the eqn. (2-32) and $\partial \phi$ can be written as in eqn. (2-33):

$$\partial \phi = \left| \tan^{-1} \left( \left( \frac{x_2 - x_1}{s^2} \right) \left( m \left( \Delta x_2 - \Delta x_1 \right) - \left( \Delta y_2 - \Delta y_1 \right) \right) \left( \sum_{n=0}^{\infty} (-t)^n \right) \right) \right| \qquad (2\text{-}33)$$

Further it is noted that $\partial \phi$ has a maximum value when $|\Delta x_2 - \Delta x_1| = |\Delta y_2 - \Delta y_1| = 1$. Thus, using the definition of $\phi$ in eqn. (2-26), the maximum value of $t$ is given by eqn. (2-34):

$$t_{max} = \left( \frac{1}{s} \right) \left( |\cos \phi| + |\sin \phi| \right) \qquad (2\text{-}34)$$

Thus, the maximum value of $\partial \phi$ is gives by eqn. (2-35):

$$\partial \phi_{max} = \max \left( \tan^{-1} \left\{ \frac{1}{s} \left( |\sin \phi \pm \cos \phi| \right) \left( \left| \sum_{n=0}^{\infty} (-t_{max})^n \right| \right) \right\} \right) \qquad (2\text{-}35)$$

Since $|t| < 1$, $t_{max} \leq 1$ and eqn. (2-35) is bounded. Truncating the series by retaining up to second order terms only, eqn. (2-35) can be written as eqn. (2-36):

$$\partial \phi_{max} = \max \left( \tan^{-1} \left\{ \frac{1}{s} \left( |\sin \phi \pm \cos \phi| \right) \left( 1 - t_{max} + t_{max}^2 \right) \right\} \right) + O \left( s^{-1} t_{max}^3 \right) \qquad (2\text{-}36)$$

### 2.4.2 Error bound of maximum deviation and non-parametric framework for PA

Consider a line segment joining two points $P_1$ and $P_N$ and its corresponding digital line segment given by pixels $\{P_1', \ldots, P_N'\}$ (an illustration is shown in Figure 2.4-1(a)). Let the distances of the pixels $\{P_1', \ldots, P_N'\}$ from the line segment $P_1 P_N$ be denoted by $d_i; i = 1$ to $N$. For convenience, these distances shall be referred to as the deviations of the pixels $\{P_1', \ldots, P_N'\}$ from the line segment $P_1 P_N$. Using the eqn. (2-36) and assuming that only digitization is present, the distances should lie within $\left[0, d_{max}\right]$, where $d_{max}$ is given by eqn. (2-37) below.

$$d_{max} = s\partial\phi_{max} \approx \max\left( s\tan^{-1}\left\{\frac{1}{s}\left(\left|\sin\phi \pm \cos\phi\right|\right)\left(1 - t_{max} + t_{max}^2\right)\right\}\right) \qquad (2\text{-}37)$$

where $\phi$ corresponds to the slope $m$ of the continuous line segment $P_1 P_N$. For a digital curve, since the distance $s$ between any two points and the slope $m$ of the line passing through them can be computed, the upper bound of the deviations due to digitization alone $d_{max}$ can be computed using the eqn. (2-37). Sample plots of $d_{max}$ for various values of $s$ and $\phi$ are shown in Figure 2.4-1(b). This bound is the underlying concept in the proposed non-parametric framework for dominant point detection methods.

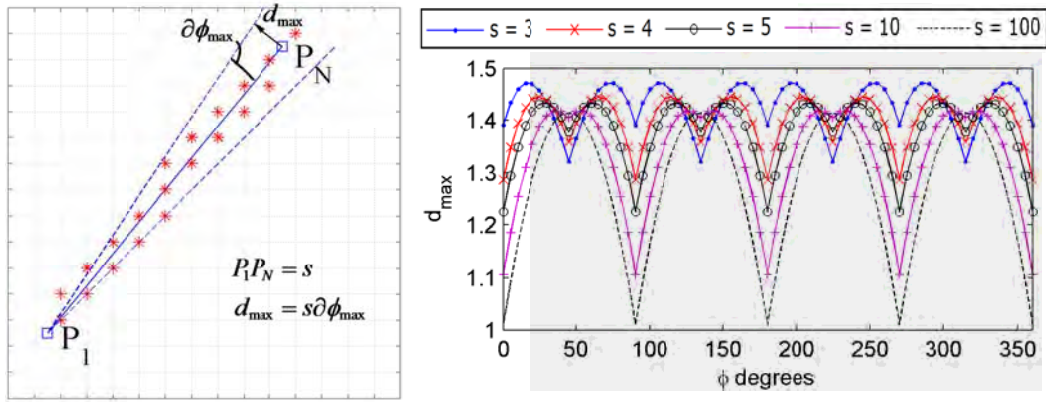

(a) illustration of the maximum
deviation

(b) plot of $d_{max}$

**Figure 2.4-1: Illustration of maximum deviation and values of the
upper bound.**

In the proposed framework, the deviations of the pixels from a line segment are compared with the bound of the eqn. (2-37) and use it either as an optimization goal or a termination condition in the three different types of dominant point detection methods. This concept is used as a framework in which most dominant point detection methods can be adapted and made parameter-free.

### 2.4.3 **Comparison with other bounds**

The bound derived in section 2.4.1 is not the only bound proposed for quantifying the effect of digitization. In this section two other bounds are presented and compared with the proposed error bound in section 2.4.1.

#### *2.4.3.1 Digital straight segments and associated error bound*

The concept of digital straight segments (DSS) was proposed in [15] and later elaborated upon in [162]. According to the concept, a digital line segment is specified by three integers $a$, $b$, and $\mu$ through the inequality (2-38):

$$\mu < ax' + by' < \mu + |a| + |b| \tag{2-38}$$

and for a digital line segment specified by $a$, $b$, and $\mu$, the pixels $P'(x', y')$ satisfying (2-38) are said to belong to that digital line. For convenience, a digital line segment is represented as $(a, b, \mu)$ since it is completely specified by these integers. Considering the equation of a line in eqn. (2-39):

$$ax + by = c \tag{2-39}$$

where $a$ and $b$ correspond to a digital line segment $(a, b, \mu)$ while the points $P(x, y)$ belong to the continuous two-dimensional space. For the pixels $P'(x', y')$ belonging to the digital straight segment $(a, b, \mu)$, if they are to satisfy (2-39), then $c$ has to satisfy inequality (2-40):

$$\mu < c < \mu + |a| + |b| \tag{2-40}$$

Thus, a digital line segment $(a, b, \mu)$ is representative of continuous line segments satisfying eqns. (2-39) and (2-40). For convenience, eqn. (2-38) is rewritten as in eqn. (2-41).

$$0 < \frac{ax' + by' - \mu}{\sqrt{a^2 + b^2}} < \frac{|a| + |b|}{\sqrt{a^2 + b^2}} \tag{2-41}$$

Eqn. (2-40) is rewritten as in eqn. (2-42):

$$c = \mu + \Delta\mu, \text{ where } 0 < \Delta\mu < |a| + |b| \tag{2-42}$$

Now, noting that the deviation of a point $P_i(x_i, y_i)$ from a line given by eqn. (2-39) is given by eqn. (2-43):

$$d_i = \frac{c - ax_i - by_i}{\sqrt{a^2 + b^2}} \tag{2-43}$$

and using eqns. (2-41) and (2-42), the inequality (2-44) is obtained.

$$0 < -d_i + \frac{\Delta\mu}{\sqrt{a^2 + b^2}} < \frac{|a| + |b|}{\sqrt{a^2 + b^2}} \tag{2-44}$$

The inequality (2-44) can be simplified to inequality (2-45) using the bound on $\Delta\mu$ given in eqn. (2-42).

$$0 < d_i < \frac{|a| + |b|}{\sqrt{a^2 + b^2}} \tag{2-45}$$

Now, expressing eqn. (2-45) in terms of the slope of the line segment $\theta = \tan^{-1}(a/b)$, the bound on the deviation can be written as in eqn. (2-46):

$$d_{\max} = |\sin\theta| + |\cos\theta| \tag{2-46}$$

For distinguishing this bound with the other bounds, this bound is referred to as $d_{\text{DSS}}$.

### 2.4.3.2 Simplified error in the slope estimation

In this section, instead of using the maximum error in the slope estimation derived in section 2.4.1, in this section a simpler expression of the error in slope estimation is presented. Consider a line segment joining two points $P_1(x_1, y_1)$ and $P_2(x_2, y_2)$, the slope of which is given by eqn. (2-47).

$$m = \tan\theta = (y_2 - y_1)/(x_2 - x_1) \tag{2-47}$$

Using the digitization model given by eqns. (2-22) and (2-25), the maximum distance between the points $P_1(x_1, y_1), P_2(x_2, y_2)$ and the corresponding pixels $P_1'(x_1', y_1')$, $P_2'(x_2', y_2')$ is $1/\sqrt{2}$. Thus, for any two points $P_1(x_1, y_1)$ and $P_2(x_2, y_2)$, it is evident that the pixels $P_1'(x_1', y_1')$ and $P_2'(x_2', y_2')$ are within the circles centered at $P_1(x_1, y_1)$ and $P_2(x_2, y_2)$ (respectively) with radii $1/\sqrt{2}$.

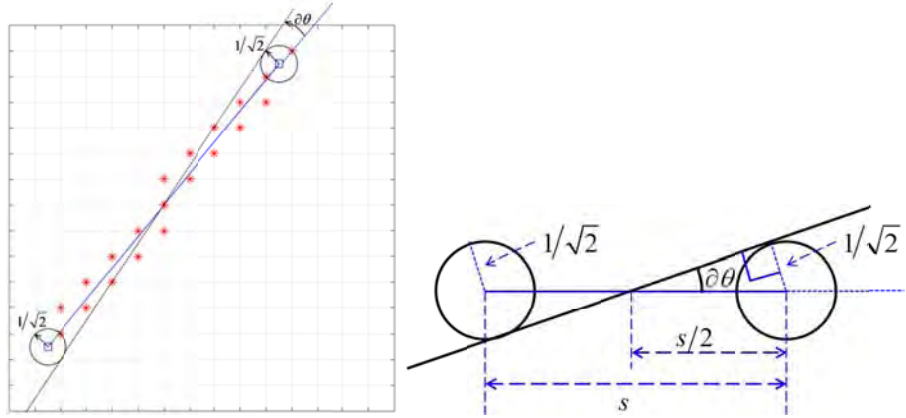

**Figure 2.4-2: Illustration of the simplified error in tangent estimation.**

Consider a line segment obtained by joining two points, one on the boundary of each circle. Geometrically, for non-intersecting circles, such a line segment makes a maximum angle with the line joining their centers if and only if the points belong to one of the inner-tangents of the pair of circles. In this case, if the distance $s$ between $P_1(x_1, y_1)$ and $P_2(x_2, y_2)$ is more than $\sqrt{2}$, the condition of non-intersecting circles is satisfied. Thus, using the simplified diagram presented in Figure 2.4-2, an estimate of $\partial\theta$ is computed using the fact that the two circles in this case are of equal radii, which implies that the inner tangent divides the line segment joining the centers into two equal halves. From Figure 2.4-2, the expression of $\partial\theta$ is written as in eqn. (2-48).

$$\partial\theta = \sin^{-1}\left(\frac{1/\sqrt{2}}{s/2}\right) = \sin^{-1}\left(\frac{\sqrt{2}}{s}\right) \tag{2-48}$$

Consequently, the bound on the deviation can be written as in eqn. (2-49):

$$d_{\max} = s\partial\theta = s\sin^{-1}\left(\frac{\sqrt{2}}{s}\right) \tag{2-49}$$

For distinguishing this bound from the other bounds in this section, this bound is referred to as $d_{\tan}$.

### 2.4.3.3 Comparison of the error bounds

For distinguishing the bound proposed in section 2.4.1 by the bounds in section 2.4.3.1 and section 2.4.3.2, the proposed bound is referred to as $d_{dig}$. Two interesting observations can be made by comparing the three bounds $d_{dig}$ (section 2.4.2), $d_{DSS}$ (section 2.4.3.1), and $d_{tan}$ (section 2.4.3.2). The first observation is regarding the dependence of the three bounds on the slope given by $\theta$ and length $s$ of the continuous line segment. It is noted that the bound $d_{DSS}$ has no dependence upon the length of the segment $s$, while the bound $d_{tan}$ does not have dependence upon the slope of the segment $\theta$. As opposed to both of these, the bound $d_{dig}$ has dependence upon both the slope given by $\theta$ and length $s$ of the continuous line segment. For explicit illustration, the bounds $d_{DSS}$ and $d_{tan}$ are plotted as functions of variables $\theta$ and $s$ respectively in Figure 2.4-3.

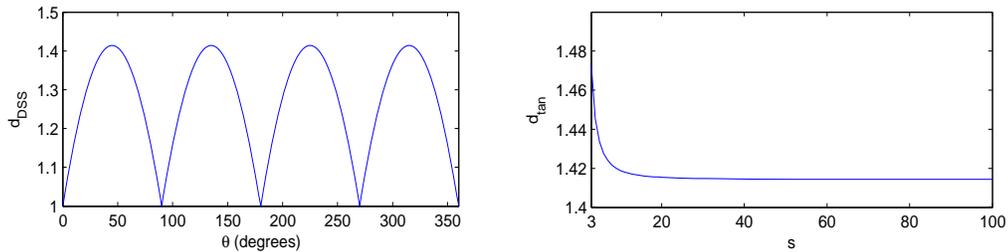

Digital error bound $d_{DSS}$ as a function of $\theta$    Digital error bound $d_{tan}$ as a function of $s$

**Figure 2.4-3: Plots of the error bounds as functions of $\theta$ and $s$.**

The second interesting point to note is the limiting case where all the three bounds converge. First, the bound $d_{dig}$ for the case of large $s$, i.e. $s \to \infty$, is considered. Eqn. (2-50) is obtained by applying this limit to the proposed error bound in eqn. (2-37).

$$\lim_{s \to \infty} d_{\text{dig}} = \lim_{s \to \infty} \max \left( s \tan^{-1} \left\{ \left| \sum_{n=0}^{\infty} (-t_{\max})^n \right| \left( \frac{|\sin\theta \pm \cos\theta|}{s} \right) \right\} \right)$$

$$= \max \left( \lim_{s \to \infty} s \tan^{-1} \left\{ \left| \sum_{n=0}^{\infty} (-t_{\max})^n \right| \left( \frac{|\sin\theta \pm \cos\theta|}{s} \right) \right\} \right) \qquad (2\text{-}50)$$

$$= \max \left( \lim_{s \to \infty} s \tan^{-1} \left( \frac{|\sin\theta \pm \cos\theta|}{s} \right) \right) = \max \left( \lim_{s \to \infty} s \left( \frac{|\sin\theta \pm \cos\theta|}{s} \right) \right)$$

$$= \max \left( |\sin\theta \pm \cos\theta| \right) = |\sin\theta| + |\cos\theta|$$

The following points have been used to obtain eqn. (2-50) above.

1. Since $s \to \infty$, $t_{\max} \to 0$ and $\lim_{t_{\max} \to 0} \left| \sum_{n=0}^{\infty} (-t_{\max})^n \right| = 1$.

2. Since $\lim_{x \to 0} \tan^{-1} x = x$, $\lim_{s \to \infty} \tan^{-1} \left( \frac{|\sin\theta \pm \cos\theta|}{s} \right) = \frac{|\sin\theta \pm \cos\theta|}{s}$

3. $\max |a \pm b| = |a| + |b|$.

Thus, it is evident that for $s \to \infty$, $d_{\text{dig}}$ converges to $d_{\text{DSS}}$. Now, consider the limiting case of $d_{\text{tan}}$ for large values of $s$. Applying the limit $s \to \infty$ on (2-49), the limit in eqn. (2-51) is obtained.

$$\lim_{s \to \infty} d_{\max} = \lim_{s \to \infty} s \sin^{-1} \left( \frac{\sqrt{2}}{s} \right) = \lim_{s \to \infty} s \left( \frac{\sqrt{2}}{s} \right) = \sqrt{2} \qquad (2\text{-}51)$$

where $\lim_{x \to 0} \sin^{-1} x = x$ has been used. This is also clearly evident in Figure 2.4-3(c), where it is seen that for large values of $s$, $d_{\text{tan}}$ indeed converges to $\sqrt{2}$. Finally, in the limiting case $s \to \infty$, if the dependence of $d_{\text{dig}}$ and $d_{\text{DSS}}$ on $\theta$ is removed by taking the maximum values of $d_{\text{dig}}$ and $d_{\text{DSS}}$ across all the values of $\theta$, eqn. (2-52) is obtained.

$$\max \left( d_{\text{DSS}}; \forall \theta \right) = \max \left( \lim_{s \to \infty} d_{\text{dig}}; \forall \theta \right) = \lim_{s \to \infty} d_{\text{tan}} = \sqrt{2} \qquad (2\text{-}52)$$

## 2.5 Algorithms with non-parametric framework for PA

### 2.5.1 Ramer-Douglas-Peucker's method

#### 2.5.1.1 Original method

Ramer, Douglas, and Peucker [12, 13] proposed a fast recursive method for computing the dominant points on digital curves. The method is described as follows. Consider a digital curve $S = \left\{ P_1 \quad P_2 \quad \ldots \quad P_N \right\}$, where $P_i$ is the $i$ th edge pixel in the digital curve $e$. The line passing through a pair of pixels $P_a(x_a, y_a)$ and $P_b(x_b, y_b)$ is given by eqn. (2-53):

$$x\left(y_a - y_b\right) + y\left(x_b - x_a\right) + y_b x_a - y_a x_b = 0.$$
(2-53)

The deviation $d_i$ of a pixel $P_i(x_i, y_i) \in S$ from the line passing through the pair $\left\{ P_1, P_N \right\}$ is then given by eqn. (2-54):

$$d_i = d\left(1, N, i\right).$$
(2-54)

Accordingly, the pixel with maximum deviation (MD) can be found. Let it be denoted as $P_{\max}$. Considering the pairs $\left\{ P_1, P_{\max} \right\}$ and $\left\{ P_{\max}, P_N \right\}$, two new pixels of maximum deviations are then found from $S$ using the concept in the eqns. (2-19) and (2-20). It is evident that the MD goes on decreasing as newer pixels of MD between a pair are chosen. This process can be repeated till a certain condition is satisfied by all the line segments. This condition shall be referred to as the optimization goal for the ease of reference.

The condition used by RDP [12, 13] is that for each line segment, the maximum deviation of the pixels contained in its corresponding edge segment is less than a certain tolerance value. This condition is specified in inequality (2-55):

$$\max(d_i) < d_{\text{tol}}.$$
(2-55)

where $d_{\text{tol}}$ is the chosen threshold and its value is typically a few pixels.

#### 2.5.1.2 Non-parametric adaptation of RDP's method

In the above method, at each step in the recursion, if the length of the most recently fitted line segment on the curve (or sub-curve) is $s$ and the slope of the line segment is $m$, then using the eqn. (2-37), $d_{\max}$ is computed and used in eqn. (2-37) as $d_{tol} = d_{\max}$. The pseudocodes of the original and the modified methods are given in Figure 2.5-1 and the changes are highlighted for the ease of comparison. As a consequence of the proposed modification, the original method does not require any control parameter and adaptively computes the suitable value of $d_{tol}$ automatically.

### 2.5.1.3 Comparison of the RDP original and RDP modified methods

Eighteen digital curves used in recent publications [35, 46] are considered. For comparison, two values of the control parameter $d_{tol}$ of the original method, $d_{tol} = 1$ and $d_{tol} = 2$, are used, and compared against the proposed modification which does not require user specified control parameter. The results are plotted in Figure 2.5-2 and quantitative comparisons are provided in Table 2.5-1. Figure 2.5-2 shows that the proposed modification provides good approximation to all the digital curves. In Table 2.5-1, the number of pixels $M$ in the digital curves, number of dominant points $N$ found by a method, the maximum deviation MD of the polygon from the digital curve, the integral square error (ISE), the figure of merit (FOM), the precision metric $\varepsilon'_p$, the reliability metric $\varepsilon_r$, and the compression ratio $\mathrm{CR} = M/N$ are listed.

The value of the maximum deviation $\max(d_m)$ for the modified method is between 1.20 to 1.53 while it varies from 0.95 to 2.00 for the original RDP with $d_{tol} = 1$ and $d_{tol} = 2$. The values of ISE, FOM, and CR for the modified RDP method are also between the values of these parameters for the original RDP. Thus, it can be concluded that the modified RDP gives balanced performance in comparison to the original RDP with $d_{tol} = 1$ and the original RDP with $d_{tol} = 2$.

The precision and reliability metrics show that precision and reliability of RDP(mod) is between RDP(1.0) and RDP(2.0). Thus, it is expected to give a more balanced and intermediate performance in comparison to RDP(1.0) and RDP(2.0). It is also noted that generally RPD(mod) has lower reliability metric as compared to the precision metric. This implies that RDP(mod) focuses more on reliability than on precision.

```
Function DP=RDP_original ({P_1,P_2,...,P_N}, d_tol)
{      DP=NULL; % DP contains the dominant points
%step 1: line
Fit a line l using P_1 and P_N.

% step 2: maximum deviation
Find deviation {d_1,...,d_N} of pixels {P_1,...,P_N} from the line l.

Find d_max = max{d_1,...,d_N} and point P_max corresponding to d_max.
%step 3: termination/recursion condition
If d_max ≤ d_tol

DP={DP,  P_1,P_N}

Else

{      DP={DP,  RDP_max (P_1,P_max)}.

DP={DP,  RDP_max (P_max,P_N)}.

}
End
Remove redundant points in DP.
Return(DP).
}
```

(a) Pseudocode for RDP (original)

```
Function DP=RDP_modified ({P_1,P_2,...,P_N})
{      DP=NULL; % DP contains the dominant points
%step 1: line and its parameters
Fit a line l using P_1 and P_N.

For the line, find distance s = |P_1P_N| and slope m.

Compute d_tol = s∂φ_max using eqn. (2-37).
% step 2: maximum deviation
Find deviation {d_1,...,d_N} of pixels {P_1,...,P_N} from the line l.

Find d_max = max{d_1,...,d_N} and point P_max corresponding to d_max.
%step 3: termination/recursion condition
If d_max ≤ d_tol

DP={DP,  P_1,P_N}

Else

{      DP={DP,  RDP_max (P_1,P_max)}.

DP={DP,  RDP_max (P_max,P_N)}.

}
End
Remove redundant points in DP.
Return(DP).
}
```

(b) Pseudocode for RDP (modified)

**Figure 2.5-1: Pseudocodes for RDP's – original and modified methods.**

| | Hammer | Hand | Screw-driver |
|---|---|---|---|
| $d_{tol} = 1$ | 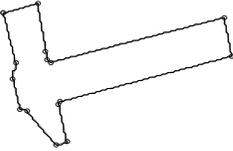 | 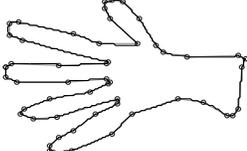 | 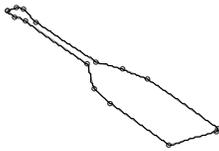 |
| $d_{tol} = 2$ | 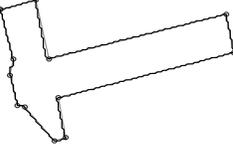 | 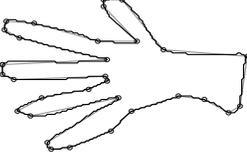 | 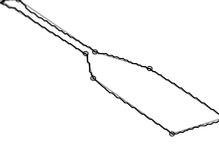 |
| modified | 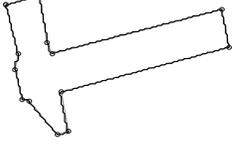 | 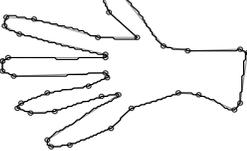 | 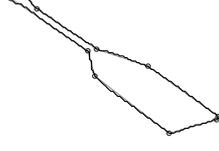 |

| | Rabbit | Dinosaur1 | Dinosaur2 | Dinosaur3 |
|---|---|---|---|---|
| $d_{tol} = 1$ | 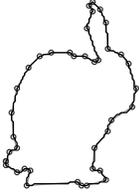 | 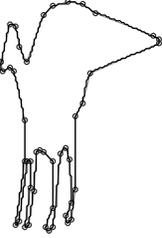 | 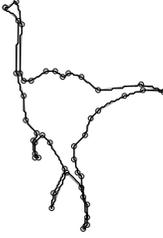 | 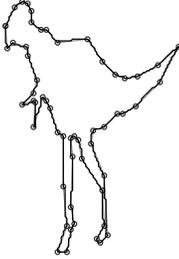 |
| $d_{tol} = 2$ | 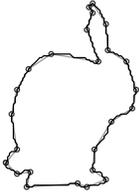 | 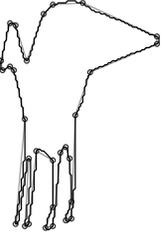 | 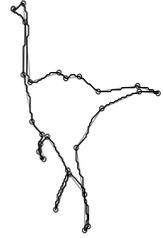 | 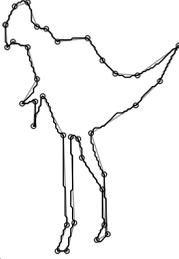 |
| modified | 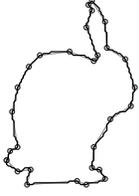 | 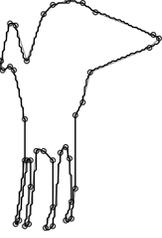 | 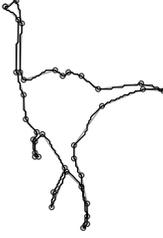 | 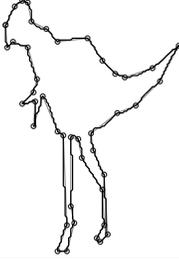 |

**Figure 2.5-2: Comparison of results of RDP (original and modified) for several digital curves.**

|  | Plane1 | Plane2 | Plane3 |
|---|---|---|---|
| $d_{tol} = 1$ | 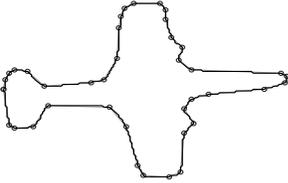 | 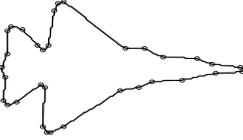 | 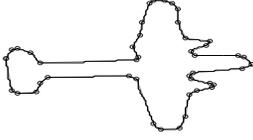 |
| $d_{tol} = 2$ | 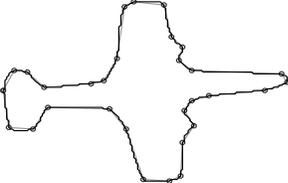 | 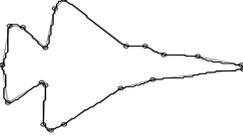 | 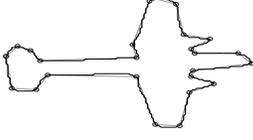 |
| modified | 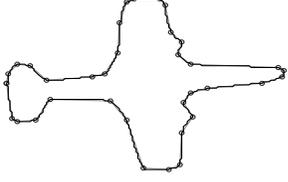 | 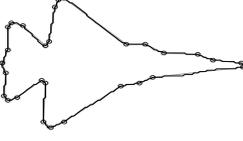 | 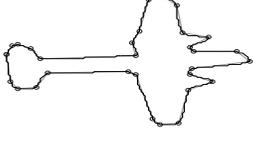 |

|  | Plane 4 | Plane 5 | Tinopener | |
|---|---|---|---|---|
| $d_{tol} = 1$ | 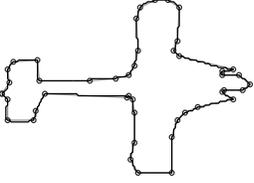 | 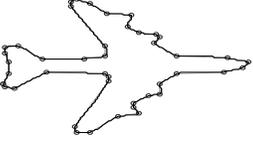 | 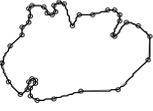 | 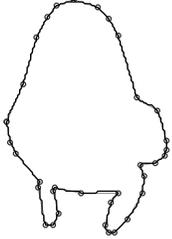 |
| $d_{tol} = 2$ | 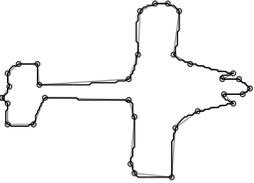 | 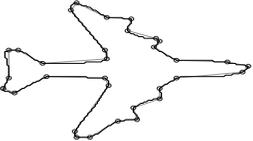 | 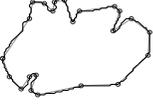 | 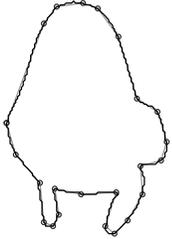 |
| modified | 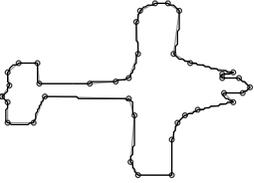 | 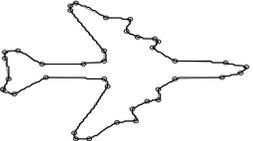 | 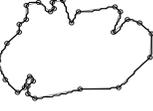 | 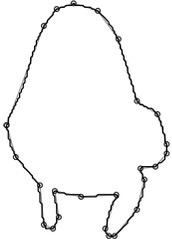 |

**Figure 2.5-2: Comparison of results of RDP (original and modified) for several digital curves... contd.**

| | Africa | Maple leaf | Sword fish | Dog |
|---|---|---|---|---|
| $d_{tol} = 1$ | 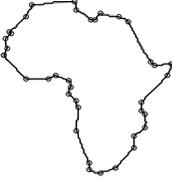 | 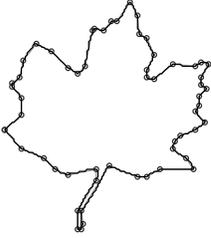 | 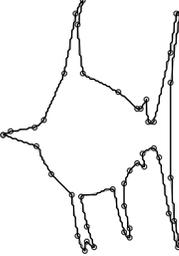 | 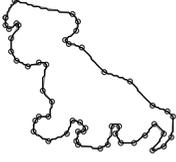 |
| $d_{tol} = 2$ | 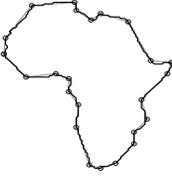 | 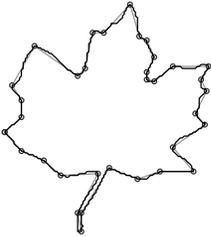 | 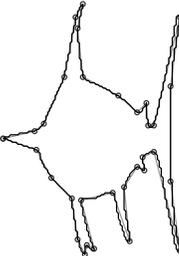 | 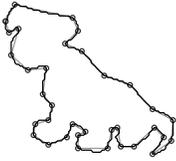 |
| modified | 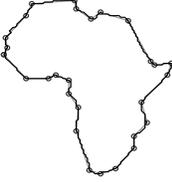 | 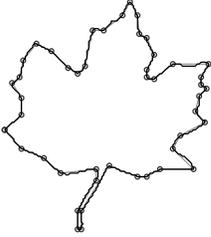 | 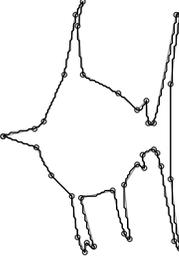 | 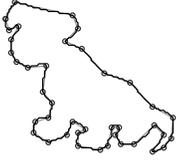 |

**Figure 2.5-2: Comparison of results of RDP (original and modified) for several digital curves... contd.**

**Table 2.5-1 (a): Original and modified methods of RDP**

**Quantitative comparison for the first nine digital curves in Figure 2.5-2.**

| | No. of pixels (M) | Dominant points (N) | MD | ISE | FOM | Precision $\varepsilon_p'$ | Reliability $\varepsilon_r$ | CR |
|---|---|---|---|---|---|---|---|---|
| **Hammer** | | | | | | | | |
| RDP(1.0) | 388 | 16 | 0.98 | 55.17 | 0.44 | 0.36 | 0.29 | 24.25 |
| RDP(2.0) | 388 | 13 | 1.65 | 92.94 | 0.32 | 0.64 | 0.49 | 29.85 |
| RDP(mod) | 388 | 14 | 1.30 | 67.90 | 0.41 | 0.43 | 0.32 | 27.71 |
| **Hand** | | | | | | | | |
| RDP(1.0) | 642 | 53 | 1.00 | 94.80 | 0.13 | 0.28 | 0.26 | 12.11 |
| RDP(2.0) | 642 | 32 | 1.99 | 314.65 | 0.06 | 0.58 | 0.71 | 20.06 |
| RDP(mod) | 642 | 39 | 1.40 | 178.80 | 0.09 | 0.43 | 0.37 | 16.46 |
| **Screw Driver** | | | | | | | | |
| RDP(1.0) | 253 | 16 | 0.95 | 34.52 | 0.46 | 0.32 | 0.25 | 15.81 |
| RDP(2.0) | 253 | 9 | 2.00 | 120.59 | 0.23 | 0.64 | 0.46 | 28.11 |
| RDP(mod) | 253 | 11 | 1.44 | 55.34 | 0.42 | 0.43 | 0.31 | 23.00 |
| **Rabbit** | | | | | | | | |
| RDP(1.0) | 293 | 40 | 0.97 | 35.98 | 0.20 | 0.27 | 0.24 | 7.33 |
| RDP(2.0) | 293 | 26 | 1.74 | 103.46 | 0.11 | 0.76 | 0.65 | 11.27 |
| RDP(mod) | 293 | 30 | 1.41 | 73.04 | 0.13 | 0.44 | 0.36 | 9.77 |
| **Dinosaur1** | | | | | | | | |
| RDP(1.0) | 587 | 46 | 1.00 | 71.49 | 0.18 | 0.28 | 0.24 | 12.76 |
| RDP(2.0) | 587 | 31 | 1.96 | 289.83 | 0.07 | 0.89 | 0.75 | 18.94 |
| RDP(mod) | 587 | 41 | 1.46 | 107.80 | 0.13 | 0.36 | 0.29 | 14.32 |
| **Dinosaur2** | | | | | | | | |
| RDP(1.0) | 409 | 46 | 0.99 | 56.34 | 0.16 | 0.27 | 0.25 | 8.89 |
| RDP(2.0) | 409 | 27 | 2.00 | 219.75 | 0.07 | 0.71 | 0.63 | 15.15 |
| RDP(mod) | 409 | 38 | 1.41 | 91.64 | 0.12 | 0.36 | 0.34 | 10.76 |
| **Dinosaur3** | | | | | | | | |
| RDP(1.0) | 528 | 57 | 0.99 | 76.40 | 0.12 | 0.27 | 0.26 | 9.26 |
| RDP(2.0) | 528 | 36 | 1.97 | 234.95 | 0.06 | 0.68 | 0.60 | 14.67 |
| RDP(mod) | 528 | 44 | 1.46 | 143.87 | 0.08 | 0.40 | 0.37 | 12.00 |
| **Plane1** | | | | | | | | |
| RDP(1.0) | 462 | 40 | 1.00 | 67.57 | 0.17 | 0.33 | 0.28 | 11.55 |
| RDP(2.0) | 462 | 31 | 1.79 | 148.71 | 0.10 | 0.77 | 0.72 | 14.90 |
| RDP(mod) | 462 | 36 | 1.20 | 87.51 | 0.15 | 0.39 | 0.32 | 12.83 |
| **Plane2** | | | | | | | | |
| RDP(1.0) | 365 | 35 | 0.95 | 40.48 | 0.26 | 0.27 | 0.23 | 10.43 |
| RDP(2.0) | 365 | 21 | 1.85 | 165.20 | 0.11 | 0.93 | 0.75 | 17.38 |
| RDP(mod) | 365 | 30 | 1.53 | 61.44 | 0.20 | 0.33 | 0.28 | 12.17 |

**Table 2.5-1 (b): Original and modified methods of RDP**

**Quantitative comparison for the last nine digital curves in Figure 2.5-2.**

| | No. of pixels (M) | Dominant points (N) | MD | ISE | FOM | Precision $\varepsilon_p'$ | Reliability $\varepsilon_r$ | CR |
|---|---|---|---|---|---|---|---|---|
| **Plane3** | | | | | | | | |
| RDP(1.0) | 431 | 45 | 0.99 | 56.46 | 0.17 | 0.29 | 0.26 | 9.58 |
| RDP(2.0) | 431 | 26 | 1.91 | 232.08 | 0.07 | 0.84 | 0.66 | 16.58 |
| RDP(mod) | 431 | 32 | 1.46 | 117.18 | 0.11 | 0.46 | 0.38 | 13.47 |
| **Plane4** | | | | | | | | |
| RDP(1.0) | 450 | 46 | 1.00 | 68.62 | 0.14 | 0.28 | 0.25 | 9.78 |
| RDP(2.0) | 450 | 33 | 1.80 | 228.60 | 0.06 | 1.26 | 1.11 | 13.64 |
| RDP(mod) | 450 | 39 | 1.41 | 101.99 | 0.11 | 0.37 | 0.31 | 11.54 |
| **Plane5** | | | | | | | | |
| RDP(1.0) | 431 | 44 | 1.00 | 57.52 | 0.17 | 0.28 | 0.25 | 9.80 |
| RDP(2.0) | 431 | 33 | 2.00 | 194.95 | 0.07 | 0.75 | 0.73 | 13.06 |
| RDP(mod) | 431 | 39 | 1.34 | 75.63 | 0.15 | 0.35 | 0.29 | 11.05 |
| **Tin Opener** | | | | | | | | |
| RDP(1.0) | 278 | 42 | 0.95 | 33.99 | 0.19 | 0.23 | 0.23 | 6.62 |
| RDP(2.0) | 278 | 24 | 1.83 | 127.99 | 0.09 | 0.82 | 0.71 | 11.58 |
| RDP(mod) | 278 | 31 | 1.48 | 77.84 | 0.12 | 0.39 | 0.35 | 8.97 |
| **Turtle** | | | | | | | | |
| RDP(1.0) | 354 | 35 | 1.00 | 49.99 | 0.20 | 0.29 | 0.26 | 10.11 |
| RDP(2.0) | 354 | 24 | 1.87 | 120.92 | 0.12 | 0.79 | 0.66 | 14.75 |
| RDP(mod) | 354 | 26 | 1.49 | 91.53 | 0.15 | 0.43 | 0.36 | 13.62 |
| **Africa** | | | | | | | | |
| RDP(1.0) | 291 | 38 | 1.00 | 37.02 | 0.21 | 0.25 | 0.23 | 7.66 |
| RDP(2.0) | 291 | 25 | 1.81 | 102.93 | 0.11 | 0.80 | 0.67 | 11.64 |
| RDP(mod) | 291 | 29 | 1.27 | 63.21 | 0.16 | 0.39 | 0.31 | 10.03 |
| **Maple leaf** | | | | | | | | |
| RDP(1.0) | 424 | 58 | 1.00 | 53.11 | 0.14 | 0.26 | 0.22 | 7.31 |
| RDP(2.0) | 424 | 35 | 1.98 | 202.30 | 0.06 | 0.87 | 0.74 | 12.11 |
| RDP(mod) | 424 | 47 | 1.41 | 93.97 | 0.10 | 0.35 | 0.29 | 9.02 |
| **Sword Fish** | | | | | | | | |
| RDP(1.0) | 627 | 46 | 1.00 | 80.21 | 0.17 | 0.33 | 0.25 | 13.63 |
| RDP(2.0) | 627 | 33 | 1.94 | 220.35 | 0.09 | 1.05 | 0.96 | 19.00 |
| RDP(mod) | 627 | 38 | 1.41 | 138.68 | 0.12 | 0.42 | 0.34 | 16.50 |
| **Dog** | | | | | | | | |
| RDP(1.0) | 343 | 52 | 1.00 | 44.81 | 0.15 | 0.26 | 0.24 | 6.60 |
| RDP(2.0) | 343 | 36 | 2.00 | 149.19 | 0.06 | 0.77 | 0.72 | 9.53 |
| RDP(mod) | 343 | 41 | 1.39 | 82.69 | 0.10 | 0.43 | 0.33 | 8.37 |

### 2.5.2 **Masood's method**

#### 2.5.2.1 *Original method*

As opposed to the method proposed by Ramer, Douglas, and Peuker [12, 13] (section 2.5.1.1), Masood [46] begins with the break points as the first list of dominant points and then iteratively removes one dominant point at a time till a termination condition is satisfied. In every iteration, when a dominant point is deleted, the remaining dominant points are re-optimized such that the dominant points after the optimization are such that the integral square error (ISE) is minimum. For convenience, the iteration number is denoted with $i$ and the number of dominant points in the $i$th iteration with $n = 1$ to $N$. The points in the digital curves are indexed from $m = 1$ to $M$. The list of dominant points in an iteration can then be specified using eqn. (2-56):

$$\mathbf{I}_i = \left\{ I_n ; n = 1 \text{ to } N \right\}_i \quad ; I_{\forall n} \in \left\{ 1, 2, \ldots, M \right\} \text{ and } I_n < I_{n+1} \tag{2-56}$$

where $I$ denotes the index of the point on the digital curve. Before beginning the optimization, the break points are taken as the initial set of dominant points, i.e., $\mathbf{I}_{i=0} = \left\{ I_n^{\text{break}} \right\}$. In an iteration, for each dominant point specified by $I_n$, an associated error value (AEV) is computed. This associated error value is calculated as follows – Considering the hypothesis that dominant point $I_n$ will be deleted, an optimization of the remaining dominant points, i.e., $\left( \mathbf{I}_i - \left\{ I_n \right\} \right)$, is performed to minimize the integral square error (ISE). This is done in two independent steps. In the first step, the indices of the dominant points before $I_n$, i.e., $\left\{ I_1, \ldots, I_{n-1} \right\}$, are optimized. For this, first it is checked that does changing $I_{n-1}$ within the range $\left( I_{n-2}, I_n \right)$ yields to a lower value of ISE. If this happens, the value of $I_{n-1}$ is changed to the optimal value within $\left( I_{n-2}, I_n \right)$ that yields the minimum ISE. In a similar manner, the index $I_{n-2}$ is then optimized. However, if changing $I_{n-1}$ within the range $\left( I_{n-2}, I_n \right)$ does not yield to a lower value of ISE, the optimization in this step is stopped. Using a similar approach, in the second step, the indices of the dominant points after $I_n$, i.e., $\left\{ I_{n+1}, \ldots I_N \right\}$, are optimized. After the optimization of both the steps, AEV can be computed in the following manner. For convenience, the set of dominant points obtained after both optimization steps are denoted as $\mathbf{I}_{i,n}^{\text{opt}}$. If the ISE corresponding to $\mathbf{I}_i$ is denoted by $ISE_i$ and the ISE corresponding to $\mathbf{I}_{i,n}^{\text{opt}}$ is denoted by $ISE_{i,n}$, then AEV of the $n$th point is given by eqn. (2-57):

$$AEV_{i,n} = ISE_i - ISE_{i,n} \qquad\qquad (2\text{-}57)$$

After computing the AEV for all the dominant points in the $i$th iteration, the point with the minimum value of AEV is removed and the list of optimal points corresponding to its removal is retained (or can be recomputed). The algorithm can be terminated by specifying a termination condition which may be based upon the maximum number of dominant points, or the maximum integral square error, or the maximum tolerable deviation (similar to $d_{tol}$ in the eqn. (2-55)). In [36], Masood proposed to use the condition (2-58) as the termination condition and the value of $d_{tol} = 0.9$:

$$\max\left(\left(d_m\right)^2\right) > d_{\text{tol}}. \qquad\qquad (2\text{-}58)$$

where $d_m$ is the deviation of the pixels on the digital curve from the polygon obtained by the dominant points.

### 2.5.2.2 Non-parametric adaptation of Masood's method

The method of Masood can be modified in the following manner. First, given a sequence of dominant points specified by the indices $\mathbf{I}_i = \left\{I_n; n = 1 \text{ to } N\right\}_i$, maximum deviation corresponding to the portion of digital curve corresponding to two consecutive dominant points is defined as in eqn. (2-59).

$$d_n = \max\left(d\left(n, n+1, m\right); m \in \left\{I_n, I_n+1, \ldots, I_{n+1}\right\}\right) \quad; \text{ for } n = 1 \text{ to } \left(N-1\right). \qquad (2\text{-}59)$$

For convenience, the set of these maximum deviations for the given set of dominant points $\mathbf{I}_i = \left\{I_n; n = 1 \text{ to } N\right\}_i$ is denoted as $\mathbf{D}\left(\mathbf{I}_i\right) = \left\{d_n; n = 1 \text{ to } \left(N-1\right)\right\}$. The definition of AEV in the proposed modification is given by eqn. (2-60):

$$AEV_{i,n} = \max\left(\mathbf{D}\left(\mathbf{I}_{i,n}^{\text{opt}}\right)\right) \qquad\qquad (2\text{-}60)$$

```
Global {P₁, P₂,..., P_M} .
Function I=Masood_original ( d_tol )
{        %step 1: break points
Compute the break points. Find their indices {I_n^break} ;

Assign i = 0 ;  I_i = {I_n^break} ;  flag_stop=FALSE
%step 2: iterations
do
{ Assign N = number of dominant points
%step 2.1: Compute AEV
For  n = 2 to (N-1) : Call (AEV(n), I^opt(n)) = compute_AEV ( I_i , n);
%step 2.2: Update or terminate
Compute d_m  (defined immediately after eqn. (2-58));

If  max((d_m)²) > d_tol , Then flag_stop=TRUE, Else { i = i+1 ;  I_i = I^opt(n')
} ;
} while(flag_stop=FALSE)
Return( I_i ).}
```

```
Function (AEV, I^opt ) = compute_AEV( I , n)
{        %Optimization step 1
Assign I^up = {I₁, I₂,..., I_{n-1}} ;  Assign k = 1 ;

Do { Assign I_min = I_{n-k-1}, I_max = I_{n-k+1} .

Find  I_opt ∈ (I_min, I_max) such that ∑_{m=I_min}^{I_max} (d_m)²  (i.e. ISE) is minimized.

If  I_opt = I_{n-k} . Then Flag_terminate_opt=TRUE, Else { I_{n-k} = I_opt ; k = k+1
;Flag_terminate_opt=FALSE};
}
While(Flag_terminate_opt=FALSE)
%Optimization step 2
Assign I^down = {I_{n+1}, I_{n+2},..., I_N} ;  Assign k = 1 ;

Do { Assign I_min = I_{n+k-1}, I_max = I_{n+k+1} .

Find  I_opt ∈ (I_min, I_max) such that ∑_{m=I_min}^{I_max} (d_m)²  (i.e. ISE) is minimized.

If  I_opt = I_{n+k} , Then Flag_terminate_opt=TRUE, Else { I_{n+k} = I_opt ; k = k+1
;Flag_terminate_opt=FALSE};
}
While(Flag_terminate_opt=FALSE)

%Computation of AEV
Compute AEV using eqn. (2-57);
Assign  I^opt = I^up ∪ I^down = {I₁, I₂,..., I_{n-1}, I_{n+1},..., I_N} ;

Return(AEV, I^opt ); }
```

**(a) Pseudocode for Masood (original)**

**Figure 2.5-3: Pseudocodes for the original and modified methods of Masood.**

```
Global {P₁, P₂, ..., Pₘ}.
```
Global $\{P_1, P_2, \ldots, P_M\}$.

```
Function I=Masood_modified
{        %step 1: break points
```

Compute the break points. Find their indices $\left\{I_n^{\text{break}}\right\}$;

Assign $i = 0$; $\mathbf{I}_i = \left\{I_n^{\text{break}}\right\}$; flag_stop=FALSE

**%step 2: iterations**
do { Assign $N$ = number of dominant points

**%step 2.1: Compute AEV**

For $n = 2$ to $(N-1)$: Call (AEV($n$), $\mathbf{I}^{\text{opt}}(n)$) =
compute_AEV ($\mathbf{I}_i$, $n$);

**%step 2.2: Update or terminate**

Find the index $n' = \arg\left(\min\left\{\text{AEV}_n; n = 2 \text{ to } N\right\}\right)$;

Compute $d_{n'-1,n'+1}^{\max}$ (using eqn. (2-62));

If AEV($n'$) > $d_{n'-1,n'+1}^{\max}$, Then flag_stop=TRUE, Else { $i = i+1$;

$\mathbf{I}_i = \mathbf{I}^{\text{opt}}(n')$ };

} while(flag_stop=FALSE)
Return($\mathbf{I}_i$).}

---

Function (AEV, $\mathbf{I}^{\text{opt}}$) = compute_AEV($\mathbf{I}$, $n$)
{        **%Optimization step 1**

Assign $\mathbf{I}^{\text{up}} = \left\{I_1, I_2, \ldots, I_{n-1}\right\}$; Assign $k = 1$;

Do { Assign $I_{\min} = I_{n-k-1}, I_{\max} = I_{n-k+1}$.

Find $I_{\text{opt}} \in \left(I_{\min}, I_{\max}\right)$ such that $d_{n-k}$ computed using eqn. (2-59) is minimized.

If $I_{\text{opt}} = I_{n-k}$. Then Flag_terminate_opt=TRUE, Else { $I_{n-k} = I_{\text{opt}}$; $k = k+1$
;Flag_terminate_opt=FALSE};
} While(Flag_terminate_opt=FALSE)

**%Optimization step 2**

Assign $\mathbf{I}^{\text{down}} = \left\{I_{n+1}, I_{n+2}, \ldots, I_N\right\}$; Assign $k = 1$;

Do { Assign $I_{\min} = I_{n+k-1}, I_{\max} = I_{n+k+1}$.

Find $I_{\text{opt}} \in \left(I_{\min}, I_{\max}\right)$ such that $d_{n+k-1}$ computed using eqn. (2-59) is minimized.

If $I_{\text{opt}} = I_{n+k}$, Then Flag_terminate_opt=TRUE, Else { $I_{n+k} = I_{\text{opt}}$; $k = k+1$
;Flag_terminate_opt=FALSE};
} While(Flag_terminate_opt=FALSE)

**%Computation of AEV**
Compute AEV using eqn. (2-60);

Assign $\mathbf{I}^{\text{opt}} = \mathbf{I}^{\text{up}} \cup \mathbf{I}^{\text{down}} = \left\{I_1, I_2, \ldots, I_{n-1}, I_{n+1}, \ldots, I_N\right\}$;

Return(AEV, $\mathbf{I}^{\text{opt}}$).}

**(b) Pseudocode for Masood (modified)**

**Figure 2.5-3: Pseudocodes for the original and modified methods of Masood... contd.**

Further, in the optimization step (for the hypothesis that dominant point $I_n$ shall be deleted) done for the sequence $\{I_1, \ldots, I_{n-1}\}$, instead of minimizing the integral square error (ISE), the goal is to minimize the maximum deviations $d_{n-1}, d_{n-2}$, and so on. Finally, the termination condition is modified as in inequality (2-61) and eqn. (2-62):

$$\min\left\{AEV_{i,n}; n = 2 \text{ to } (N-1)\right\} > d_{n-1,n+1}^{\max} \qquad (2\text{-}61)$$

$$d_{n-1,n+1}^{\max} = s_{n-1,n+1}\phi_{n-1,n+1} \qquad (2\text{-}62)$$

where $s_{n-1,n+1}$ is the length of the line segment formed by joining the dominant points $I_{n-1}$ and $I_{n+1}$ and $\phi_{n-1,n+1}$ is the angle made by the line segment with the $x-$axis. The pseudocodes of the original and proposed modifications of Masood are presented in Figure 2.5-3.

### 2.5.2.3 Comparison of the Masood original and Masood modified methods

Eighteen digital curves used in recent publications [35, 46] and in the section 2.5.1.3 are considered in this detailed benchmarking. For comparison, two values of the control parameter $d_{tol}$ of the original method of Masood, $d_{tol} = 0.9$ (recommended by Masood in [36]) and $d_{tol} = 1.2$ (taken as another control parameter for comparison) are used, and compared against the proposed modification which does not require user specified control parameter. The results are plotted in Figure 2.5-4 and quantitative comparisons are provided in Table 2.5-2. Figure 2.5-4 shows that the proposed modification provides good approximation to all the digital curves. In Table 2.5-2, the number of pixels $M$ in the digital curves, number of dominant points $N$ found by a method, the maximum deviation $\max(d_i)$ of the polygon from the digital curve, the integral square error (ISE), the figure of merit (FOM), the precision metric $\varepsilon_p'$, the reliability metric $\varepsilon_r$, and the compression ratio $CR = M/N$ are listed. In general it is desired that $\max(d_i)$ and ISE are as less as possible and FOM and CR are as large as possible [149].

The value of the maximum deviation MD for the modified method is between 0.79 to 1.27 while it varies from 0.90 to 1.54 for the original method of Masood with $d_{tol} = 0.9$ and $d_{tol} = 1.2$. In fact for each digital curve, the value of MD for the modified method is always lesser than the original method of Masood with $d_{tol} = 1.2$. On the other hand, ISE of the original method with $d_{tol} = 0.9$ and $d_{tol} = 1.2$ is lower than the modified method for 15 images. This is because the original method Masood focuses on the minimization of ISE in each iteration, while the modified method focuses upon $d_n$. The modified method has a better CR than the original method with $d_{tol} = 0.9$ for all the digital curves and the original method with $d_{tol} = 1.2$ for 9 digital curves.

Further, in Figure 2.5-4, the curves of turtle, Africa, maple leaf, and dinosaur 1 are brought to special notice. For turtle, it is seen that the modified method chooses much fewer dominant points ($N = 33$) than the original method with $d_{tol} = 0.9$ ($N = 38$) and $d_{tol} = 1.2$ ($N = 48$), while representing the digital curve effectively. Similar observations are noted for Africa and dinosaur 1. In maple leaf, though the number of dominant points obtained using the original method with $d_{tol} = 1.2$ and the modified method are same, the locations of the dominant points are different. Most differences occur in the concave regions of the digital curve and the locations where the curvature of the digital curve changes fast. Thus non-parametric adaptation of the method provide balanced the tradeoff in the performance.

The precision metrics and reliability metrics of Masood(0.9), Masood(1.2), and Masood(mod) are lesser in comparison with RDP. Also, as compared to RDP, Masood has lesser value of CR. This indicates that Masood focuses more on local fitting rather than global fitting.

|  | Hammer | Hand | Screw-driver |
|---|---|---|---|
| $d_{tol} = 0.9$ | 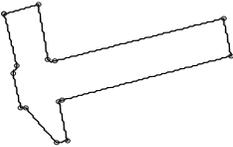 | 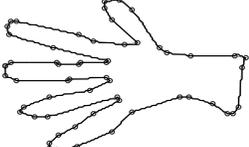 | 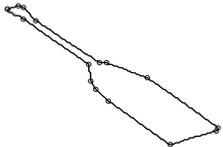 |
| $d_{tol} = 1.2$ | 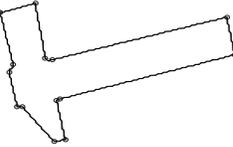 | 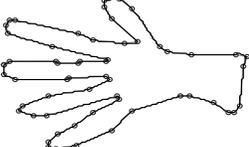 | 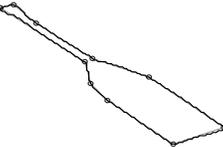 |
| modified | 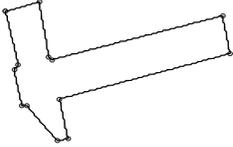 | 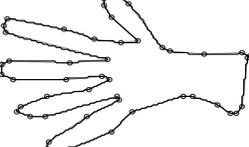 | 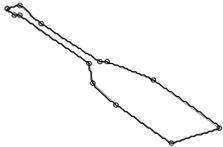 |

|  | Rabbit | Dinosaur1 | Dinosaur2 | Dinosaur3 |
|---|---|---|---|---|
| $d_{tol} = 0.9$ | 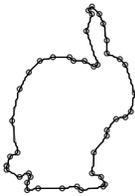 | 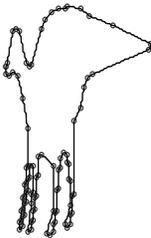 | 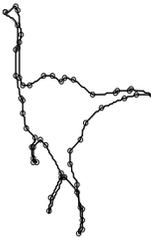 | 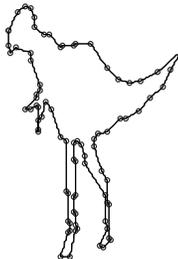 |
| $d_{tol} = 1.2$ | 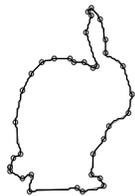 | 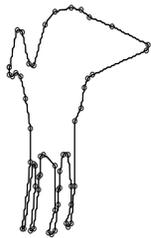 | 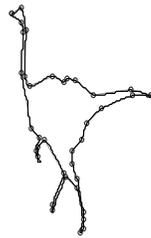 | 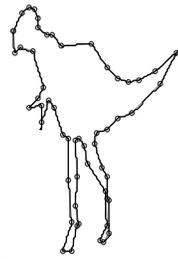 |
| modified | 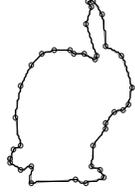 | 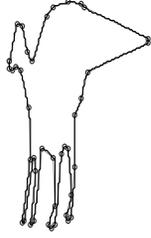 | 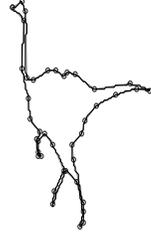 | 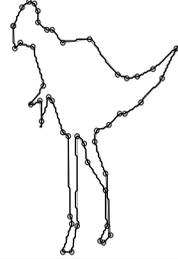 |

**Figure 2.5-4: Comparison of results of Masood (original and modified) for several digital curves.**

|  | Plane1 | Plane2 | Plane3 |
|---|---|---|---|
| $d_{tol} = 0.9$ | 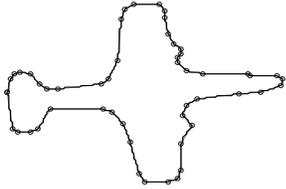 | 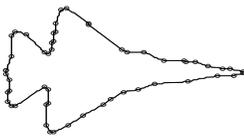 | 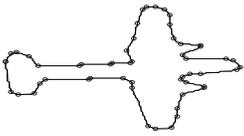 |
| $d_{tol} = 1.2$ | 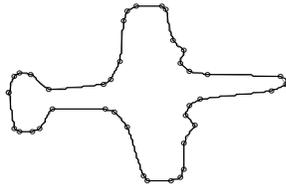 | 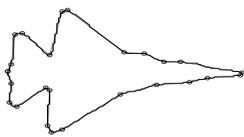 | 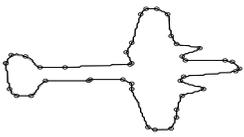 |
| modified | 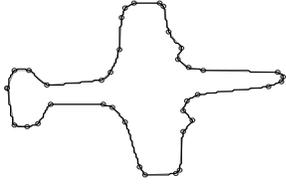 | 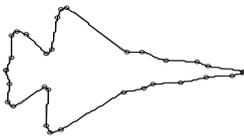 | 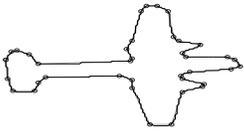 |

|  | Plane 4 | Plane 5 | Tinopener | Turtle |
|---|---|---|---|---|
| $d_{tol} = 0.9$ | 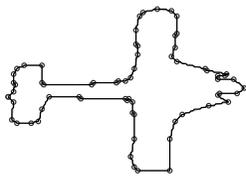 | 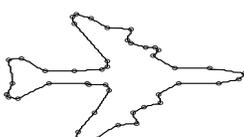 | 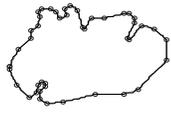 | 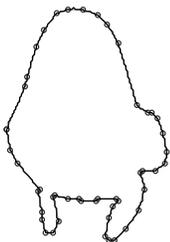 |
| $d_{tol} = 1.2$ | 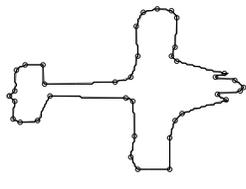 | 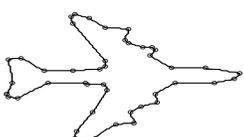 | 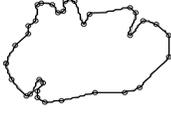 | 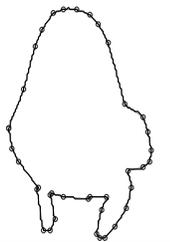 |
| modified | 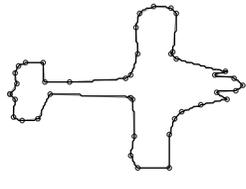 | 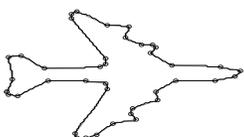 | 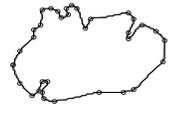 | 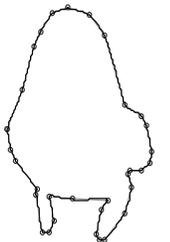 |

**Figure 2.5-4: Comparison of results of Masood (original and modified) for several digital curves... contd.**

| | Africa | Maple leaf | Sword fish | Dog |
|---|---|---|---|---|
| $d_{tol}$ = 0.9 | 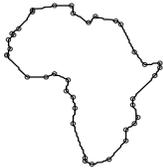 | 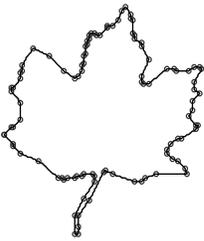 | 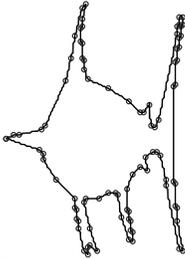 | 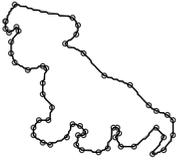 |
| $d_{tol}$ = 1.2 | 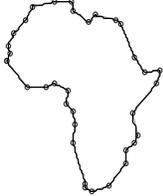 | 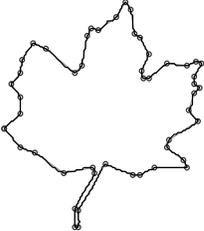 | 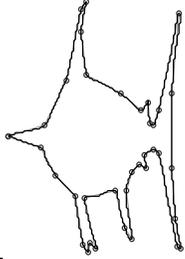 | 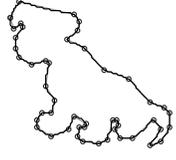 |
| modified | 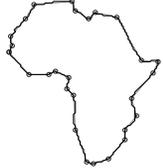 | 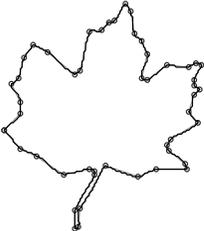 | 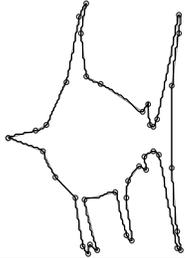 | 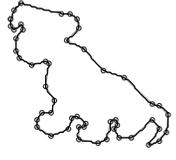 |

**Figure 2.5-4: Comparison of results of Masood (original and modified) for several digital curves... contd.**

**Table 2.5-2 (a): Original and modified methods of Masood**

**Quantitative comparison for the first nine digital curves in Figure 2.5-4.**

| | No. of pixels (M) | Dominant points (N) | MD | ISE | FOM | Precision $\varepsilon_p'$ | Reliability $\varepsilon_r$ | CR |
|---|---|---|---|---|---|---|---|---|
| **Hammer** | | | | | | | | |
| Masood(0.9) | 388 | 15 | 1.46 | 38.35 | 0.67 | 0.32 | 0.25 | 25.87 |
| Masood(1.2) | 388 | 15 | 1.46 | 38.35 | 0.67 | 0.32 | 0.25 | 25.87 |
| Masood (mod) | 388 | 14 | 0.79 | 48.86 | 0.57 | 0.28 | 0.28 | 27.71 |
| **Hand** | | | | | | | | |
| Masood(0.9) | 642 | 58 | 1.32 | 47.75 | 0.23 | 0.19 | 0.18 | 11.07 |
| Masood(1.2) | 642 | 58 | 1.32 | 47.75 | 0.23 | 0.19 | 0.18 | 11.07 |
| Masood (mod) | 642 | 48 | 1.14 | 97.50 | 0.14 | 0.29 | 0.27 | 13.38 |
| **Screw Driver** | | | | | | | | |
| Masood(0.9) | 253 | 16 | 1.11 | 24.68 | 0.64 | 0.26 | 0.20 | 15.81 |
| Masood(1.2) | 253 | 11 | 1.53 | 46.62 | 0.49 | 0.38 | 0.28 | 23.00 |
| Masood (mod) | 253 | 14 | 1.13 | 42.59 | 0.42 | 0.33 | 0.27 | 18.07 |
| **Rabbit** | | | | | | | | |
| Masood(0.9) | 293 | 45 | 0.93 | 21.67 | 0.30 | 0.19 | 0.18 | 6.51 |
| Masood(1.2) | 293 | 36 | 1.21 | 35.91 | 0.23 | 0.26 | 0.24 | 8.14 |
| Masood (mod) | 293 | 37 | 1.00 | 42.82 | 0.18 | 0.28 | 0.26 | 7.92 |
| **Dinosaur1** | | | | | | | | |
| Masood(0.9) | 587 | 96 | 1.00 | 24.09 | 0.25 | 0.10 | 0.11 | 6.11 |
| Masood(1.2) | 587 | 55 | 1.26 | 44.38 | 0.24 | 0.21 | 0.19 | 10.67 |
| Masood (mod) | 587 | 42 | 0.96 | 91.04 | 0.15 | 0.32 | 0.30 | 13.98 |
| **Dinosaur2** | | | | | | | | |
| Masood(0.9) | 409 | 56 | 1.00 | 30.36 | 0.24 | 0.17 | 0.16 | 7.30 |
| Masood(1.2) | 409 | 38 | 1.20 | 62.98 | 0.17 | 0.30 | 0.28 | 10.76 |
| Masood (mod) | 409 | 42 | 0.95 | 64.38 | 0.15 | 0.29 | 0.28 | 9.74 |
| **Dinosaur3** | | | | | | | | |
| Masood(0.9) | 528 | 84 | 1.00 | 22.63 | 0.28 | 0.12 | 0.11 | 6.29 |
| Masood(1.2) | 528 | 56 | 1.37 | 50.69 | 0.19 | 0.24 | 0.20 | 9.43 |
| Masood (mod) | 528 | 48 | 1.27 | 89.38 | 0.12 | 0.29 | 0.28 | 11.00 |
| **Plane1** | | | | | | | | |
| Masood(0.9) | 462 | 52 | 1.04 | 31.49 | 0.28 | 0.20 | 0.17 | 8.88 |
| Masood(1.2) | 462 | 40 | 1.54 | 50.59 | 0.23 | 0.27 | 0.22 | 11.55 |
| Masood (mod) | 462 | 38 | 1.11 | 64.81 | 0.19 | 0.33 | 0.27 | 12.16 |
| **Plane2** | | | | | | | | |
| Masood(0.9) | 365 | 54 | 0.95 | 18.54 | 0.36 | 0.13 | 0.13 | 6.76 |
| Masood(1.2) | 365 | 26 | 1.31 | 58.20 | 0.24 | 0.33 | 0.27 | 14.04 |
| Masood (mod) | 365 | 32 | 1.00 | 44.67 | 0.26 | 0.29 | 0.25 | 11.41 |

**Table 2.5-2(b): Original and modified methods of Masood**

**Quantitative comparison for the last nine digital curves in** Figure 2.5-4**.**

| | No. of pixels (M) | Dominant points (N) | MD | ISE | FOM | Precision $\varepsilon_p'$ | Reliability $\varepsilon_r$ | CR |
|---|---|---|---|---|---|---|---|---|
| Plane3 | | | | | | | | |
| Masood(0.9) | 431 | 54 | 0.90 | 32.06 | 0.25 | 0.19 | 0.17 | 7.98 |
| Masood(1.2) | 431 | 41 | 1.36 | 47.91 | 0.22 | 0.26 | 0.22 | 10.51 |
| Masood (mod) | 431 | 39 | 1.00 | 71.76 | 0.15 | 0.33 | 0.29 | 11.05 |
| Plane4 | | | | | | | | |
| Masood(0.9) | 450 | 68 | 1.00 | 16.92 | 0.39 | 0.11 | 0.09 | 6.62 |
| Masood(1.2) | 450 | 39 | 1.26 | 59.66 | 0.19 | 0.30 | 0.26 | 11.54 |
| Masood(mod) | 450 | 42 | 0.96 | 59.54 | 0.18 | 0.27 | 0.25 | 10.71 |
| Plane5 | | | | | | | | |
| Masood(0.9) | 431 | 45 | 1.21 | 36.86 | 0.26 | 0.23 | 0.19 | 9.58 |
| Masood(1.2) | 431 | 45 | 1.21 | 36.86 | 0.26 | 0.23 | 0.19 | 9.58 |
| Masood(mod) | 431 | 41 | 0.96 | 49.41 | 0.21 | 0.28 | 0.23 | 10.51 |
| Tin Opener | | | | | | | | |
| Masood(0.9) | 278 | 45 | 1.00 | 20.69 | 0.30 | 0.18 | 0.16 | 6.18 |
| Masood(1.2) | 278 | 34 | 1.40 | 40.37 | 0.20 | 0.30 | 0.24 | 8.18 |
| Masood(mod) | 278 | 35 | 1.02 | 51.71 | 0.15 | 0.39 | 0.30 | 7.94 |
| Turtle | | | | | | | | |
| Masood(0.9) | 354 | 48 | 1.00 | 23.48 | 0.31 | 0.16 | 0.17 | 7.38 |
| Masood(1.2) | 354 | 38 | 1.31 | 33.11 | 0.28 | 0.23 | 0.21 | 9.32 |
| Masood(mod) | 354 | 33 | 1.14 | 51.65 | 0.21 | 0.30 | 0.27 | 10.73 |
| Africa | | | | | | | | |
| Masood(0.9) | 291 | 39 | 0.97 | 25.89 | 0.29 | 0.22 | 0.19 | 7.46 |
| Masood(1.2) | 291 | 35 | 1.21 | 31.26 | 0.27 | 0.27 | 0.21 | 8.31 |
| Masood(mod) | 291 | 28 | 1.05 | 56.74 | 0.18 | 0.36 | 0.31 | 10.39 |
| Maple leaf | | | | | | | | |
| Masood(0.9) | 424 | 105 | 1.00 | 15.29 | 0.26 | 0.08 | 0.09 | 4.04 |
| Masood(1.2) | 424 | 50 | 1.23 | 46.34 | 0.18 | 0.26 | 0.22 | 8.48 |
| Masood(mod) | 424 | 50 | 1.20 | 64.45 | 0.13 | 0.31 | 0.26 | 8.48 |
| Sword Fish | | | | | | | | |
| Masood(0.9) | 627 | 91 | 1.00 | 33.14 | 0.21 | 0.11 | 0.14 | 6.89 |
| Masood(1.2) | 627 | 38 | 1.46 | 90.27 | 0.18 | 0.32 | 0.27 | 16.50 |
| Masood(mod) | 627 | 40 | 1.11 | 97.94 | 0.16 | 0.36 | 0.29 | 15.68 |
| Dog | | | | | | | | |
| Masood(0.9) | 343 | 54 | 0.91 | 32.51 | 0.20 | 0.22 | 0.20 | 6.35 |
| Masood(1.2) | 343 | 49 | 1.26 | 43.19 | 0.16 | 0.26 | 0.23 | 7.00 |
| Masood(mod) | 343 | 52 | 1.00 | 43.45 | 0.15 | 0.27 | 0.23 | 6.60 |

### 2.5.3 **Carmona's method**

#### *2.5.3.1 Original method*

Carmona-Poyato [35] (which is called Carmona for conciseness) is another method that begins with the list of break points as the initial set of dominant points (like Masood) and iteratively deletes points from the list of dominant points. However, beyond this initial similarity, the approach taken by Carmona is quite different. It is highlighted that there are two control parameters in Carmona's method [35], $d_{tol}$ and $r_{tol}$. However, only $r_{tol}$ is user specified and is used for termination condition only. On the other hand, $d_{tol}$ is an internal control parameter used for controlling the iterative process and as a condition for deleting the dominant points in an iteration. It begins with a small value and slowly increases with the iteration number. The method is now summarized below.

Let the sequence of dominant points in a particular iteration be denoted by $\left\{ P_n\left(x_n, y_n\right); n = 1 \text{ to } N \right\}_i$ where $i$ denotes the iteration number. For explaining the method of Carmona, it shall be handy to define a distance $d_n^{\text{C}}$ and a length $l_n$ as in eqns. (2-63) and (2-64), respectively:

$$d_n^{\text{C}} = d\left(n-1, n+1, n\right); n = 1 \text{ to } \left(N-1\right).$$ (2-63)

$$l_n = \sqrt{\left(x_n - x_{n-1}\right)^2 + \left(y_n - y_{n-1}\right)^2} + \sqrt{\left(x_{n+1} - x_n\right)^2 + \left(y_{n+1} - y_n\right)^2}$$
$$- \sqrt{\left(x_{n+1} - x_{n-1}\right)^2 + \left(y_{n+1} - y_{n-1}\right)^2}.$$ (2-64)

The distance $d_n^{\text{C}}$ is the distance of the $n$th dominant point from the line segment joining its adjacent dominant points and $l_n$ is the difference between the perimeter of the polygon formed by $\left\{ P_n\left(x_n, y_n\right); n = 1 \text{ to } N \right\}_i$ and the perimeter of the polygon formed by deleting the $n$th dominant point from $\left\{ P_n\left(x_n, y_n\right); n = 1 \text{ to } N \right\}_i$. The superscript C in $d_n^{\text{C}}$ denotes the method Carmona and is used to distinguish $d_n^{\text{C}}$ from $d_n$ in the eqn. (2-59).

Carmona recommends that if the digital curve is a closed curve, a most suitable initial dominant point $P_{n=1}$ should be chosen before beginning the iterative procedure. Assuming that the initial sequence of break points is $\left\{P_n^{\text{break}}\right\}$. For this sequence, the distances $d_n^C$ are computed and the point with the maximum value of $d_n^C$ is identified, i.e. $n' = \arg\left(\max\left(d_n^C\right)\right)$. Then, $P_{n=1} = P_{n'}^{\text{break}}$ and the remaining points in $\left\{P_n\left(x_n, y_n\right)\right\}_{i=0}$ follow the sequence.

The initial value of $d_{tol}(i=0)$ is set as zero. In each iteration, the value of $d_{tol}$ is increased by 0.5. Within an iteration, the first point for which $d_n^C < d_{tol}$ is deleted. This process is repeated for the newly obtained reduced sequence of dominant points, i.e., the first point for which $d_n^C < d_{tol}$ is deleted. This process of deletion is repeated till no point satisfies $d_n^C < d_{tol}$. At this point the current iteration is completed, the termination condition is checked and if the algorithm cannot be terminated then the next iteration is initiated. For the termination condition, a relative parameter $r_i$ is defined as in eqn. (2-65):

$$r_i = \frac{\max\left(\left\{l_n^{\text{deleted}}\right\}\right)}{\max\left(\left\{d_n\right\}\right)}, \tag{2-65}$$

where $d_n$ here is computed using the eqn. (2-59) and $l_n^{\text{deleted}}$ correspond to the dominant points deleted in the current iteration. If at the end of the $i$th iteration, $r_i < r_{tol}$, where $r_{tol}$ is a user specified control parameter, the algorithm is terminated.

### 2.5.3.2 Non-parametric adaptation of Carmona's method

Carmona can be made independent of user specified control parameter using the error bound in the eqn. (2-37) by modifying the termination condition of Carmona. For this, $d_n$ is computed using the eqn. (2-59). Further, a maximum error bound is defined as in eqn. (2-66).

$$d_n^{\text{max}} = s_n \phi_n \tag{2-66}$$

where $s_n$ is the length of the line segment formed by joining the dominant points $P_n$ and $P_{n+1}$ and $\phi_n$ is the angle made by the line segment with the $x-$axis. At the end of one step of iteration, if there is a dominant point such as specified in (2-67):

$$d_n > d_n^{\max} \tag{2-67}$$

then the algorithm is terminated. The pseudocodes of the original and modified methods are presented in Figure 2.5-5.

### 2.5.3.3 Comparison of the Carmona original and Carmona modified methods

Eighteen digital curves used in recent publications [35, 46] and in the section 2.5.1.3 are considered. For comparison, two values of the control parameter $r_{tol}$ of the original method of Carmona, $r_{tol} = 0.3$ and $r_{tol} = 0.7$ (recommended in [35] for these digital curves) are used, and compared against the proposed modification which does not require user specified control parameter. The results are plotted in Figure 2.5-6 and quantitative comparisons are provided in Table 2.5-3. Figure 2.5-6 shows that the proposed modification provides good approximation to all the digital curves. In Table 2.5-3, the number of pixels $M$ in the digital curves, number of dominant points $N$ found by a method, the maximum deviation $\max(d_i)$ of the polygon from the digital curve, the integral square error (ISE), the figure of merit (FOM), and the compression ratio $CR = M/N$ are listed. In general it is desired that $\max(d_i)$ and ISE are as less as possible and FOM and CR are as large as possible [149].

There are several interesting observations. First, see the results of Hammer in Table 2.5-3. It is seen that for $r_{tol} = 0.3$, the number of dominant points detected by the original method of Carmona is very large and the compression ratio is very poor. On the other hand, for $r_{tol} = 0.7$, the original method of Carmona misses some important features of the shape (like the top portion of the hammer). The modified method provides a better balance between precision and CR. Similar observations apply for Africa, plane 2, and plane 4. In such cases, the values of $\max(d_i)$, ISE, FOM, and CR for the modified method are in between the values of these parameters for the original method of Carmona with $r_{tol} = 0.3$ and $r_{tol} = 0.7$. Second, the curves of hand, maple leaf, dinosaur 3, sword fish, plane 3, and plane 5 are considered. For these curves, it is noted that the dominant points detected by the original method with $r_{tol} = 0.3$ are the same as the dominant points detected by the modified method. This is because for such curves, the value of one of the elements in $\{d_n\}$ (used in the eqns. (2-65) and (2-67)) is larger than $d_n^{\max}$ and is sufficiently high to reduce the value of $r_i$ below 0.3. Third, the curves of screw driver, tin-opener, turtle, rabbit, and plane 1 are considered in which the number of dominant points obtained by the modified method are larger than the original method with $r_{tol} = 0.3$ and $r_{tol} = 0.7$. For all these curves, the ISE for the modified method is significantly lower than the original method with $r_{tol} = 0.3$ and $r_{tol} = 0.7$ though the increase in the number of dominant points is not significant. Also, in most cases, the value of $\max(d_i)$ for the modified method is close to the value of $\max(d_i)$ for the original method with $r_{tol} = 0.3$. Fourth, the digital curves of dog, dinosaur 1, and dinosaur 2 are considered, for which the modified and the original methods with $r_{tol} = 0.3$ and $r_{tol} = 0.7$ perform result into exactly the same dominant points. For these curves, the dominant point deleted in the last iteration reduced the value of $r_i$ below 0.3 (from a value of $r_i$ more than 0.7 in the last iteration) as well as the one dominant point satisfying $d_n > d_n^{\max}$. In addition to the above, it is worth noting that the value of $\max(d_m)$ ranges from 0.49 to 6.75 for the original method and from 1.29 to 2.73 for the modified method. This indicates that the modified method provides a better balance of the maximum deviation $\max(d_i)$.

```
Function DP=Carmona_original ({P₁,P₂,...,Pₙ},r_tol)
{     DP=NULL; % DP contains the dominant points
%step 1: choosing initial dominant point
If   P₁ = Pₙ  Then
{     Compute {dₙᶜ} ; Find n'=arg(max(dₙᶜ)).
Assign DP={Pₙ',Pₙ'₊₁,...,Pₙ,P₂,P₃,...,Pₙ'}  }
%step 2: termination/updation
Assign d_tol = 0 ; i = 0 ; Flag_terminate = FALSE;
Do {Assign i = i+1; d_tol = d_tol +0.5 ;  n = 1 ;  lₙᵈᵉˡᵉᵗᵉᵈ =NULL;
While (n is not the last point in DP)
{ Compute dₙᶜ and lₙ using eqns. (2-63) and (2-64);
If  dₙᶜ < d_tol  Then
{     Append lₙᵈᵉˡᵉᵗᵉᵈ with lₙ ;
       Delete Pₙ from DP; }
Else Assign n = n+1
End; }
Compute {dₙ} and rᵢ using eqns. (2-59) and (2-65).
If rᵢ < r_tol Then Flag_terminate = TRUE }
While (Flag_terminate = FALSE);
Return(DP).  }
```

**(a) Pseudocode for Carmona (original)**

```
Function DP=Carmona_modified ({P₁,P₂,...,Pₙ})
{     DP=NULL; % DP contains the dominant points
%step 1: choosing initial dominant point
If   P₁ = Pₙ  Then
{     Compute {dₙᶜ} ; Find n'=arg(max(dₙᶜ)).
Assign DP={Pₙ',Pₙ'₊₁,...,Pₙ,P₂,P₃,...,Pₙ'}  }
%step 2: termination/updation
Assign d_tol = 0 ; i = 0 ; Flag_terminate = FALSE;
Do {     Assign i = i+1; d_tol = d_tol +0.5 ;  n = 1 ;
While (n is not the last point in DP) {
Compute dₙᶜ using eqn. (2-63);
If  dₙᶜ < d_tol  Then Delete Pₙ from DP Else Assign n = n+1 }
Compute {dₙ} and {dₙᵐᵃˣ} using eqns. (2-59) and (2-66).
If (for any n , dₙ > dₙᵐᵃˣ ) Then Flag_terminate = TRUE }
While (Flag_terminate = FALSE);
Return(DP).  }
```

**(b) Pseudocode for Carmona (modified)**

**Figure 2.5-5: Pseudocodes for the original and modified methods of Carmona.**

|  | Hammer | Hand | Screw-driver |
|---|---|---|---|
| $r_{tol} = 0.3$ | 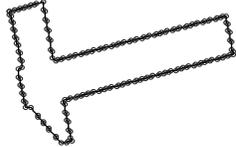 | 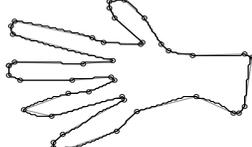 | 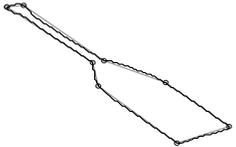 |
| $r_{tol} = 0.7$ | 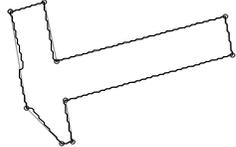 | 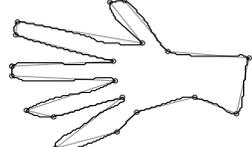 | 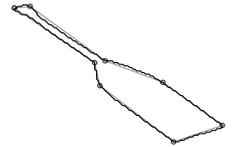 |
| modified | 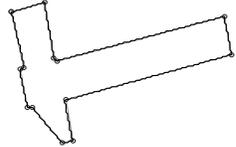 | 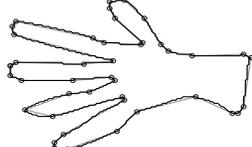 | 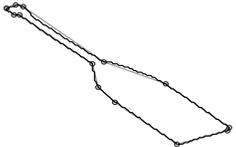 |

|  | Rabbit | Dinosaur1 | Dinosaur2 | Dinosaur3 |
|---|---|---|---|---|
| $r_{tol} = 0.3$ | 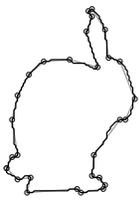 | 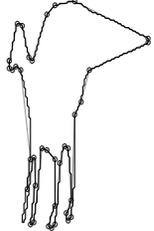 | 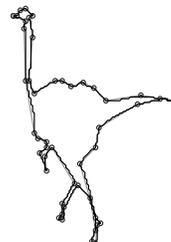 | 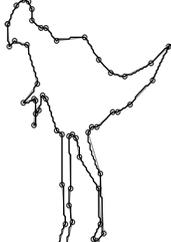 |
| $r_{tol} = 0.7$ | 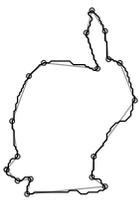 | 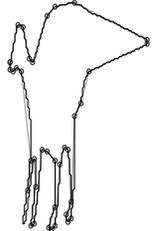 | 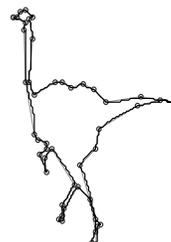 | 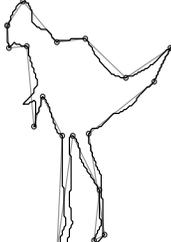 |
| modified | 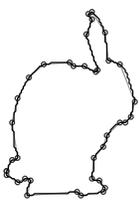 | 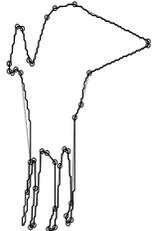 | 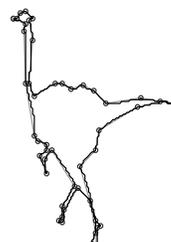 | 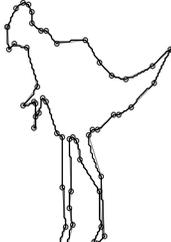 |

**Figure 2.5-6: Comparison of results of Carmona (original and modified) for several digital curves.**

| | Plane1 | Plane2 | Plane3 | |
|---|---|---|---|---|
| $r_{tol} = 0.3$ | 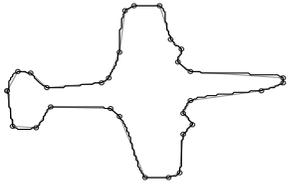 | 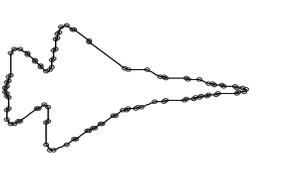 | 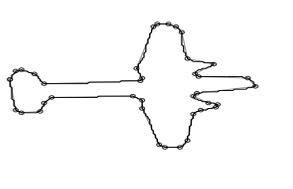 | |
| $r_{tol} = 0.7$ | 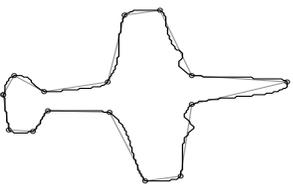 | 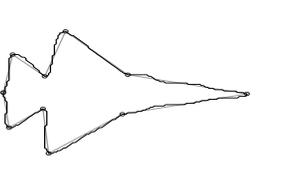 | 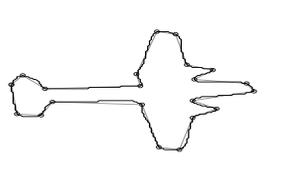 | |
| modified | 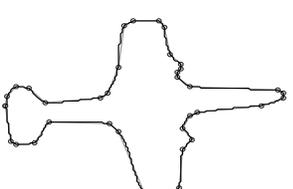 | 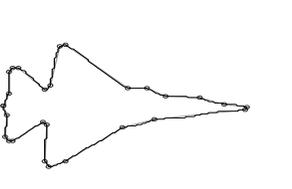 | 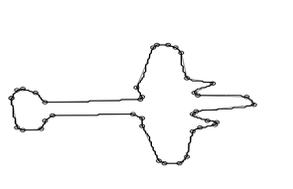 | |

| | Plane 4 | Plane 5 | Tinopener | Turtle |
|---|---|---|---|---|
| $r_{tol} = 0.3$ | 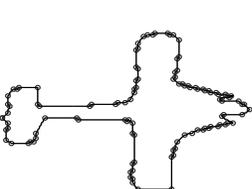 | 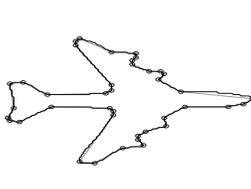 | 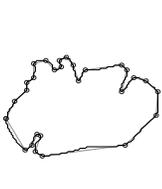 | 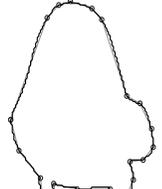 |
| $r_{tol} = 0.7$ | 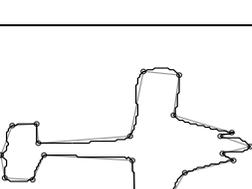 | 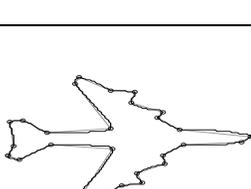 | 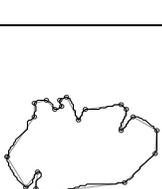 | 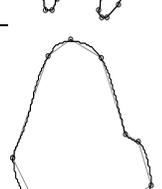 |
| modified | 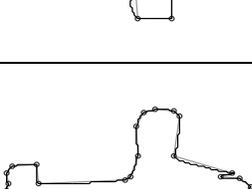 | 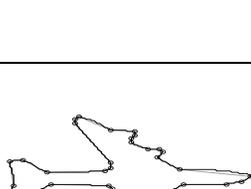 | 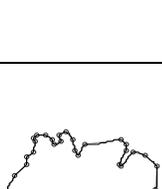 | 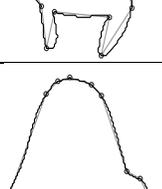 |

**Figure 2.5-6: Comparison of results of Carmona (original and modified) for several digital curves... contd.**

| | Africa | Maple leaf | Sword fish | Dog |
|---|---|---|---|---|
| $r_{tol} = 0.3$ | 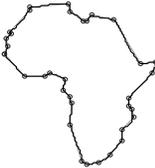 | 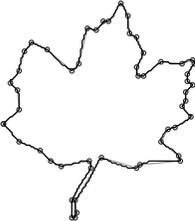 | 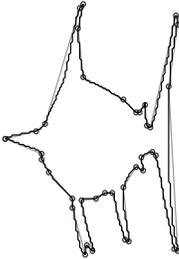 | 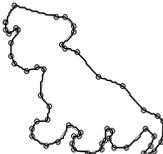 |
| $r_{tol} = 0.7$ | 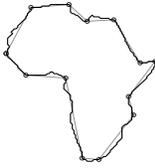 | 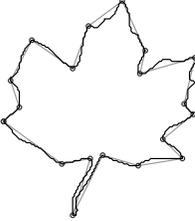 | 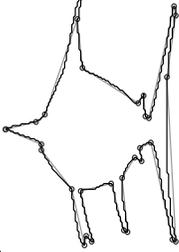 | 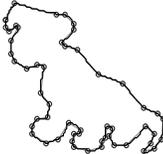 |
| modified | 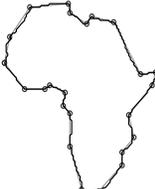 | 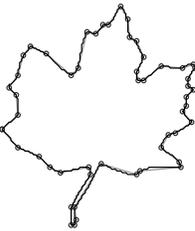 | 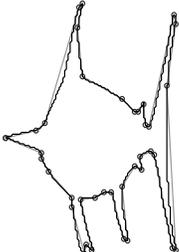 | 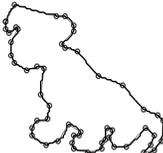 |

**Figure 2.5-6: Comparison of results of Carmona (original and modified) for several digital curves... contd.**

**Table 2.5-3 (a): Original and modified methods of Carmona**

**Quantitative comparison for the first nine digital curves in Figure 2.5-6.**

| | No. of pixels (M) | Dominant points (N) | MD | ISE | FOM | Precision $\varepsilon'_p$ | Reliability $\varepsilon_r$ | CR |
|---|---|---|---|---|---|---|---|---|
| Hammer | | | | | | | | |
| Carmona(0.3) | 388 | 194 | 0.49 | 1.32 | 1.52 | 0.01 | 0.01 | 2.00 |
| Carmona(0.7) | 388 | 10 | 2.32 | 177.40 | 0.22 | 0.60 | 0.48 | 38.80 |
| Carmona(mod) | 388 | 14 | 1.29 | 57.96 | 0.48 | 0.30 | 0.29 | 27.71 |
| Hand | | | | | | | | |
| Carmona(0.3) | 642 | 43 | 2.06 | 215.14 | 0.07 | 0.35 | 0.39 | 14.93 |
| Carmona(0.7) | 642 | 20 | 3.89 | 1111.60 | 0.03 | 1.04 | 0.97 | 32.10 |
| Carmona(mod) | 642 | 43 | 2.06 | 215.14 | 0.07 | 0.35 | 0.39 | 14.93 |
| Screw Driver | | | | | | | | |
| Carmona(0.3) | 253 | 9 | 2.08 | 179.47 | 0.16 | 0.75 | 0.56 | 28.11 |
| Carmona(0.7) | 253 | 9 | 2.08 | 179.47 | 0.16 | 0.75 | 0.56 | 28.11 |
| Carmona(mod) | 253 | 14 | 2.08 | 138.59 | 0.13 | 0.43 | 0.44 | 18.07 |
| Rabbit | | | | | | | | |
| Carmona(0.3) | 293 | 30 | 1.86 | 101.53 | 0.10 | 0.43 | 0.40 | 9.77 |
| Carmona(0.7) | 293 | 19 | 3.12 | 309.72 | 0.05 | 0.76 | 0.72 | 15.42 |
| Carmona(mod) | 293 | 38 | 1.86 | 66.64 | 0.12 | 0.31 | 0.32 | 7.71 |
| Dinosaur1 | | | | | | | | |
| Carmona(0.3) | 587 | 38 | 2.73 | 302.83 | 0.05 | 0.35 | 0.49 | 15.45 |
| Carmona(0.7) | 587 | 38 | 2.73 | 302.83 | 0.05 | 0.35 | 0.49 | 15.45 |
| Carmona(mod) | 587 | 38 | 2.73 | 302.83 | 0.05 | 0.35 | 0.49 | 15.45 |
| Dinosaur2 | | | | | | | | |
| Carmona(0.3) | 446 | 47 | 2.55 | 161.59 | 0.06 | 0.32 | 0.38 | 9.49 |
| Carmona(0.7) | 446 | 47 | 2.55 | 161.59 | 0.06 | 0.32 | 0.38 | 9.49 |
| Carmona(mod) | 446 | 47 | 2.55 | 161.59 | 0.06 | 0.32 | 0.38 | 9.49 |
| Dinosaur3 | | | | | | | | |
| Carmona(0.3) | 528 | 54 | 1.78 | 104.35 | 0.09 | 0.29 | 0.29 | 9.78 |
| Carmona(0.7) | 528 | 19 | 6.75 | 2115.16 | 0.01 | 1.37 | 1.43 | 27.79 |
| Carmona(mod) | 528 | 54 | 1.78 | 104.35 | 0.09 | 0.29 | 0.29 | 9.78 |
| Plane1 | | | | | | | | |
| Carmona(0.3) | 462 | 30 | 1.79 | 213.57 | 0.07 | 0.46 | 0.48 | 15.40 |
| Carmona(0.7) | 462 | 16 | 4.63 | 1105.78 | 0.03 | 1.25 | 1.14 | 28.88 |
| Carmona(mod) | 462 | 37 | 1.60 | 104.82 | 0.12 | 0.32 | 0.33 | 12.49 |
| Plane2 | | | | | | | | |
| Carmona(0.3) | 365 | 104 | 0.50 | 4.14 | 0.85 | 0.04 | 0.03 | 3.51 |
| Carmona(0.7) | 365 | 11 | 2.85 | 400.76 | 0.08 | 1.01 | 0.78 | 33.18 |
| Carmona(mod) | 365 | 28 | 1.44 | 80.54 | 0.16 | 0.33 | 0.32 | 13.04 |

**Table 2.5-3 (b): Original and modified methods of Carmona**

**Quantitative comparison for the last nine digital curves in** Figure 2.5-6**.**

| | No. of pixels (M) | Dominant points (N) | MD | ISE | FOM | Precision $\varepsilon_p'$ | Reliability $\varepsilon_r$ | CR |
|---|---|---|---|---|---|---|---|---|
| **Plane3** | | | | | | | | |
| Carmona(0.3) | 431 | 39 | 1.53 | 92.99 | 0.12 | 0.31 | 0.33 | 11.05 |
| Carmona(0.7) | 431 | 22 | 2.79 | 397.56 | 0.05 | 0.83 | 0.71 | 19.59 |
| Carmona(mod) | 431 | 39 | 1.53 | 92.99 | 0.12 | 0.31 | 0.33 | 11.05 |
| **Plane4** | | | | | | | | |
| Carmona(0.3) | 450 | 114 | 0.71 | 4.88 | 0.81 | 0.03 | 0.03 | 3.95 |
| Carmona(0.7) | 450 | 22 | 3.45 | 632.35 | 0.03 | 0.91 | 0.89 | 20.45 |
| Carmona(mod) | 450 | 41 | 2.00 | 127.93 | 0.09 | 0.32 | 0.36 | 10.98 |
| **Plane5** | | | | | | | | |
| Carmona(0.3) | 431 | 41 | 2.53 | 168.71 | 0.06 | 0.27 | 0.36 | 10.51 |
| Carmona(0.7) | 431 | 28 | 2.53 | 349.77 | 0.04 | 0.59 | 0.65 | 15.39 |
| Carmona(mod) | 431 | 41 | 2.53 | 168.71 | 0.06 | 0.27 | 0.36 | 10.51 |
| **Tin Opener** | | | | | | | | |
| Carmona(0.3) | 278 | 29 | 2.28 | 145.12 | 0.07 | 0.45 | 0.46 | 9.59 |
| Carmona(0.7) | 278 | 22 | 2.28 | 214.60 | 0.06 | 0.64 | 0.61 | 12.64 |
| Carmona(mod) | 278 | 39 | 2.19 | 73.08 | 0.10 | 0.24 | 0.30 | 7.13 |
| **Turtle** | | | | | | | | |
| Carmona(0.3) | 354 | 27 | 1.68 | 151.65 | 0.09 | 0.44 | 0.46 | 13.11 |
| Carmona(0.7) | 354 | 15 | 5.34 | 1134.83 | 0.02 | 1.54 | 1.31 | 23.60 |
| Carmona(mod) | 354 | 30 | 1.68 | 130.32 | 0.09 | 0.39 | 0.42 | 11.80 |
| **Africa** | | | | | | | | |
| Carmona(0.3) | 291 | 35 | 1.24 | 52.62 | 0.16 | 0.28 | 0.28 | 8.31 |
| Carmona(0.7) | 291 | 14 | 3.81 | 547.22 | 0.04 | 1.12 | 0.95 | 20.79 |
| Carmona(mod) | 291 | 26 | 1.80 | 93.62 | 0.12 | 0.41 | 0.37 | 11.19 |
| **Maple leaf** | | | | | | | | |
| Carmona(0.3) | 424 | 53 | 1.56 | 91.78 | 0.09 | 0.31 | 0.30 | 8.00 |
| Carmona(0.7) | 424 | 20 | 4.06 | 847.89 | 0.03 | 1.19 | 1.01 | 21.20 |
| Carmona(mod) | 424 | 53 | 1.56 | 91.78 | 0.09 | 0.31 | 0.30 | 8.00 |
| **Sword Fish** | | | | | | | | |
| Carmona(0.3) | 627 | 39 | 3.19 | 573.52 | 0.03 | 0.40 | 0.63 | 16.08 |
| Carmona(0.7) | 627 | 30 | 3.00 | 646.28 | 0.03 | 0.55 | 0.75 | 20.90 |
| Carmona(mod) | 627 | 39 | 3.19 | 573.52 | 0.03 | 0.40 | 0.63 | 16.08 |
| **Dog** | | | | | | | | |
| Carmona(0.3) | 343 | 54 | 1.39 | 54.26 | 0.12 | 0.27 | 0.25 | 6.35 |
| Carmona(0.7) | 343 | 54 | 1.39 | 54.26 | 0.12 | 0.27 | 0.25 | 6.35 |
| Carmona(mod) | 343 | 54 | 1.39 | 54.26 | 0.12 | 0.27 | 0.25 | 6.35 |

## 2.6 Existing PA methods in the context of precision, reliability, and the error bound

This section considers several popular or recent polygonal approximation methods and studies their optimization goal and update scheme. The optimization goal and the update scheme usually use the performance criteria mentioned in section 2.2.3.

### 2.6.1 Optimal polygonal representation of Perez and Vidal [52]

The algorithm proposed by Perez and Vidal (PV) [52] is by far the most popular algorithm used as a benchmark for comparing the performance of polygonal fitting algorithms. The reason for its popularity is twofold. For a given number of points $N \leq M$, where $M$ is the number of pixels in the digital curve, it computes the optimal choice of $N$ points from the digital curve such that some error metric is minimized. Since the error metric can be flexibly defined by a user, it is versatile in its use. Further, for the purpose of benchmarking, the designers of other algorithms can first perform the polygonal fitting using their own algorithms, obtain a value of $N$ as obtained by their own algorithms, use this value of $N$ in the algorithm by PV and simply compare the points obtained by their method against the optimal points obtained by PV.

Since PV can use any error metric to be minimized, it is interesting to note that it can either use the precision score or the reliability score as the error to be minimized. If precision score is used as the error function, PV fits segments such that all the line segments are of approximately the same length. If reliability score is used as the error function, PV fits segments that are combination of two types: first type are the small segments with small value of MD but with very small (close to zero) value of $\left\| \mathbf{X\bar{A}} - \mathbf{\bar{J}} \right\|_1$; the second type are long segments with comparatively larger values of $\left\| \mathbf{X\bar{A}} - \mathbf{\bar{J}} \right\|_1$ but significantly larger value of $s_{\max}$ such that the reliability score is also small valued.

On the other hand, PV does not guarantee that the maximum deviation of the pixels in curve is within the upper limit of the error due to digitization (discussed in section 2.4). If the value of $N$ is very large, it is likely that PV will fit the segments such that the maximum deviation is lesser than the upper bound. This means that the PA will over fit and be sensitive to the error due to digitization. On the other hand, if the value of $N$ is small, the maximum deviation of the fitted segments is larger than the upper bound, thus indicating under-fitting. In essence, this means that using a fixed value of $N$ or solving min-$\varepsilon$ problem is not suitable for optimal PA of the digital curves. In most practical applications, *apriori* knowledge of N is unknown so PV is not of a practical use.

### 2.6.2 **Ramer-Douglas-Peucker [12, 13] (RDP and RDP-mod)**

 Ramer-Douglas-Peucker [12, 13] is a splitting method in which the pixel of maximum deviation is found recursively till the maximum deviation (MD) of any edge pixel from the nearest line segment is less than a fixed value. Since this is a splitting algorithm, it begins with a very high value of MD, which reduces as the edge is split further. The algorithm stops at a point where the MD satisfies a minimum criterion. Thus, this algorithm focuses more on reliability and tends to barely satisfy a precision requirement.

In the sense of the upper bound, the original RDP gives a mixed performance. For a few segments, the chosen threshold may be below the upper bound in eqn. (2-37) and the result is an over-fitting for this segment. On the other hand, the chosen threshold may be above the upper bound for certain line segments, thus resulting in under-fitting for such segments. However, RDP-mod incorporates the upper bound inherently by the redefinition of the optimization goal.

### 2.6.3 **Lowe's method [53]**

Lowe's method [53] is quite similar to RDP's method with an important difference. Lowe uses $s/\mathrm{max}(d_i)$ for selecting the leaves from the binary tree created by the recursive splitting algorithm. However, Lowe is not completely non-parametric since Lowe enforces that $s \geq 4$ and uses it as the termination criterion for recursive splitting algorithm.

It is interesting to note that the parameter $s/\mathrm{max}(d_i)$ is quite similar to the reciprocal of the reliability metric in eqn. (2-2). Further the method uses splitting of edges as the approach, thus inherently focusing on reliability. Since both the approach and the goal focus upon reliability type of metrics, it can be concluded that Lowe's method is a PA method with focus on global quality of fit. Regarding the error bound in eqn. (2-37), this method does not take into account the upper bound of the error due to digitization.

### 2.6.4 **Precision and reliability based optimization (PRO)**

In this method, though the method of optimization is the same as the RDP method, the optimization goal is different than the goal in (2-55). Instead of (2-55), the optimization goal in eqn. (2-21) is used, where, $\varepsilon_0$ is the chosen heuristic parameter. Since this method explicitly uses the precision and reliability measures as the optimization functions, this method is expected to perform well for both precision and reliability measures. However, this method does not take into account the upper bound of the error due to digitization.

### 2.6.5 **Break point suppression method of Masood [46] (Masood and Masood-mod)**

Masood [46] begins with the sequence of the break points, i.e., the smallest set of line segments such that each pixel of the curve lies exactly on the line segments, which is considered as the initial set of dominant points. It then proceeds with recursively deleting one break point at a time such that removing it has a minimum impact in its immediate neighborhood and optimizing the locations of the dominant points for minimum precision score. Although the aim of optimization is to improve the global fit and thus indirectly improve the reliability, evidently, Masood's method is tailored for optimizing the precision and is expected to perform poorly in terms of reliability.

Since Masood begins with largest possible set of dominant points and removes the dominant points till a certain termination criterion is satisfied, if the termination criterion is not very relaxed, the maximum deviation is in general lesser than the upper bound. Thus, in essence, Masood's method is sensitive to the digitization effects and gives an unnecessarily close fit to the digital curve.

### 2.6.6 **Method of Carmona-Poyato [35] (Carmona and Carmona-mod)**

Like Masood [46], Carmona also begins with the sequence of break points and the initial set of dominant points. However, unlike Masood, Carmona recursively deletes the dominant points with minimum impact on the global fit of the line segments. Further, the numerator in the parameter $r_i$ used in the optimization goal in eqn. (2-65) is the maximum among the lengths defined in eqn. (2-64), while the denominator is related to the maximum deviation. Thus, in effect this metric $r_i$ is similar to the reciprocal of the reliability metric. Thus, inherently Carmona-Poyato focuses more on reliability than on precision. It is evident in the results reported in [35] that this method has a tendency to be lenient in the maximum allowable deviation in the favor of general shape representation for the whole curve. The original version of Carmona does not take into account the error bound in eqn. (2-37).

However, in the modified non-parametric version of Carmona, though the algorithmic structure is kept the same, the optimization goal is changed to the maximum deviation and uses the error bound in eqn. (2-37). Thus, the modified method incorporates both local and global qualities of fit.

### 2.6.7 Nguyen and Debled-Rennesson's PA method based on blurred segments

Nguyen and Debled-Rennesson [37] uses the concept of digital straight segments and considers the blurring of these segments for allowing their PA to represent noisy digital curves. Due to the concept of blurring, it inherently compromises on the precision to focus more on the global fitting characteristics. Further, it considers a control parameter $w$ which is the maximum width allowable for the blurred segments. This parameter is representative of the maximum allowable precision. Further, in the context of error bound in eqn. (2-37), if the parameter $w$ is chosen adaptively using eqn. (2-37), then it can incorporate the effect of digitization and be made non-parametric. However, since the focus of Nguyen and Debled-Rennesson's method is to deal with the problem of severely noisy curves, the value of $w$ is typically higher than the value of the error bound in eqn. (2-37).

## 2.7 Interesting properties – comparison among proposed algorithms

### 2.7.1 Experiment 1 – scaling of digital curves and impact on PA

In this section, the effect of scaling on the performance of the proposed methods is illustrated. Two curves – tin opener and Africa – are used for this purpose. The resolution of each curve is reduced by a factor of $1/2$, $1/3$, $1/4$, and $1/5$. PRO(1.0), RDP(mod), Masood(Mod), and Carmona(mod) are applied on these curves and the results are plotted in Figure 2.7-1, Figure 2.7-2, Figure 2.7-3, and Figure 2.7-4 respectively. The performance parameters are tabulated in Table 2.7-1, Table 2.7-2,

Table 2.7-3, and Table 2.7-4 respectively. The performance parameters indicate that when the curves are significantly de-scaled (for example by a factor $1/5$) and loose the details, the CR decreases for all the four methods. However, the behavior of MD, $\varepsilon'_p$, and $\varepsilon_r$ does not change significantly. This implies that the methods retain their natural fitting characteristics. It is also interesting to note that all the four methods represent the curves well.

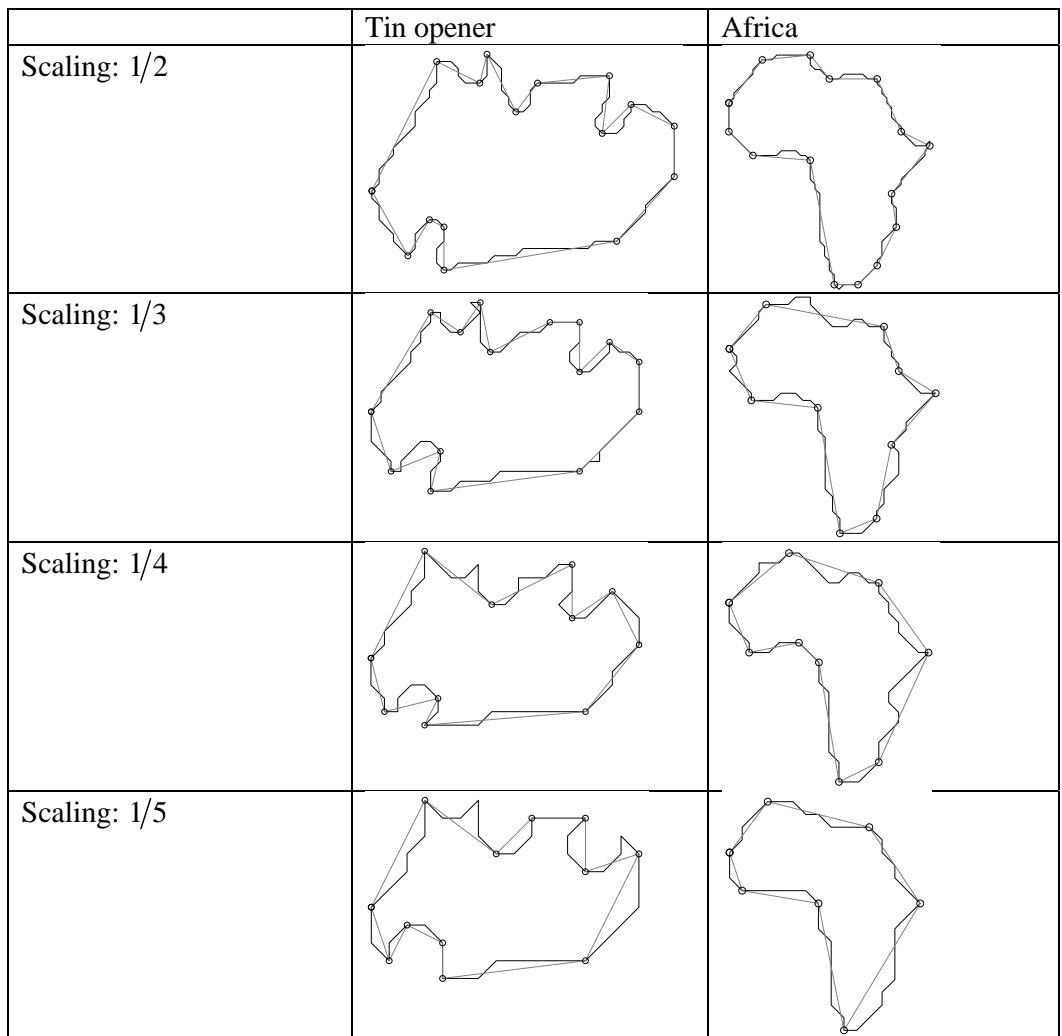

| | Tin opener | Africa |
|---|---|---|
| Scaling: 1/2 | | |
| Scaling: 1/3 | | |
| Scaling: 1/4 | | |
| Scaling: 1/5 | | |

**Figure 2.7-1: Effect of scaling on PRO(1.0).**

**Table 2.7-1: Scaling performance parameters of PRO(1.0).**

| | No. of pixels (M) | Dominant points (N) | MD | ISE | FOM | Precision $\varepsilon_p'$ | Reliability $\varepsilon_r$ | CR |
|---|---|---|---|---|---|---|---|---|
| Tin opener | | | | | | | | |
| Scaling 1 | 278 | 27 | 1.66 | 90.16 | 0.11 | 0.40 | 0.38 | 10.30 |
| Scaling 1/2 | 135 | 17 | 1.41 | 43.16 | 0.18 | 0.47 | 0.40 | 7.94 |
| Scaling 1/3 | 88 | 16 | 1.67 | 22.23 | 0.25 | 0.39 | 0.34 | 5.50 |
| Scaling 1/4 | 62 | 12 | 1.72 | 16.60 | 0.31 | 0.43 | 0.35 | 5.17 |
| Scaling 1/5 | 49 | 13 | 1.80 | 17.10 | 0.22 | 0.42 | 0.42 | 3.77 |
| Africa | | | | | | | | |
| Scaling 1 | 291 | 23 | 2.04 | 110.23 | 0.11 | 0.52 | 0.44 | 12.65 |
| Scaling 1/2 | 137 | 16 | 1.74 | 42.28 | 0.20 | 0.41 | 0.35 | 8.56 |
| Scaling 1/3 | 90 | 11 | 2.09 | 47.29 | 0.17 | 0.58 | 0.50 | 8.18 |
| Scaling 1/4 | 64 | 10 | 1.99 | 26.84 | 0.24 | 0.48 | 0.43 | 6.40 |
| Scaling 1/5 | 49 | 8 | 1.54 | 13.87 | 0.44 | 0.40 | 0.37 | 6.13 |

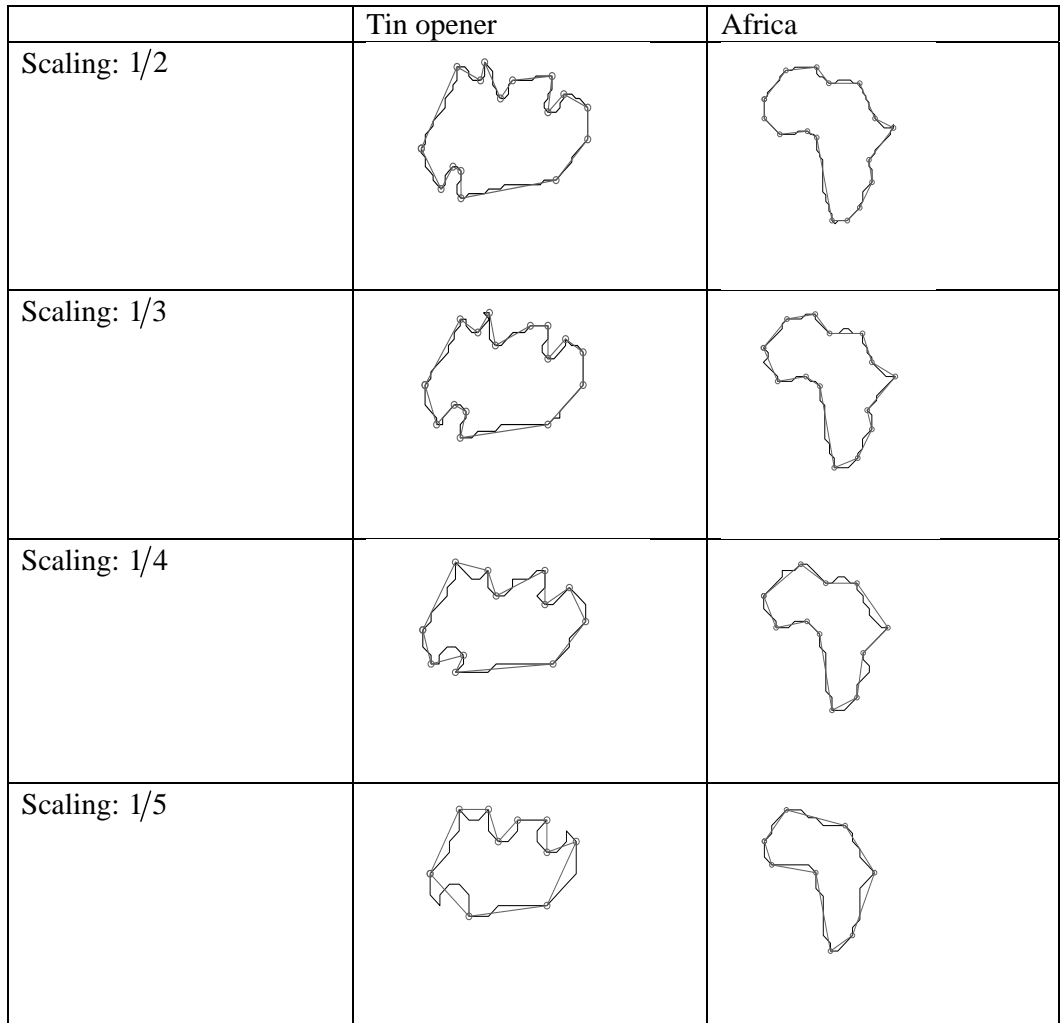

|  | Tin opener | Africa |
|---|---|---|
| Scaling: 1/2 | | |
| Scaling: 1/3 | | |
| Scaling: 1/4 | | |
| Scaling: 1/5 | | |

**Figure 2.7-2: Effect of scaling on RDP(mod).**

**Table 2.7-2: Scaling performance parameters of RDP(mod).**

| | No. of pixels (M) | Dominant points (N) | MD | ISE | FOM | Precision $\varepsilon'_p$ | Reliability $\varepsilon_r$ | CR |
|---|---|---|---|---|---|---|---|---|
| Tin opener | | | | | | | | |
| Scaling 1 | 278 | 31 | 1.48 | 77.84 | 0.12 | 0.39 | 0.35 | 8.97 |
| Scaling 1/2 | 135 | 17 | 1.41 | 43.16 | 0.18 | 0.47 | 0.40 | 7.94 |
| Scaling 1/3 | 88 | 17 | 1.06 | 16.76 | 0.31 | 0.34 | 0.29 | 5.18 |
| Scaling 1/4 | 62 | 13 | 1.46 | 17.15 | 0.28 | 0.44 | 0.36 | 4.77 |
| Scaling 1/5 | 49 | 11 | 1.41 | 24.22 | 0.18 | 0.55 | 0.56 | 4.46 |
| Africa | | | | | | | | |
| Scaling 1 | 291 | 29 | 1.27 | 63.21 | 0.16 | 0.39 | 0.31 | 10.03 |
| Scaling 1/2 | 137 | 17 | 1.34 | 30.85 | 0.26 | 0.35 | 0.30 | 8.06 |
| Scaling 1/3 | 90 | 15 | 1.18 | 17.00 | 0.35 | 0.38 | 0.30 | 6.00 |
| Scaling 1/4 | 64 | 12 | 1.41 | 17.13 | 0.31 | 0.39 | 0.33 | 5.33 |
| Scaling 1/5 | 49 | 9 | 1.17 | 8.08 | 0.67 | 0.35 | 0.29 | 5.44 |

| | Tin opener | Africa |
|---|---|---|
| Scaling: 1/2 | 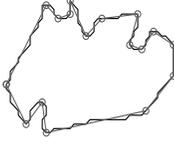 | 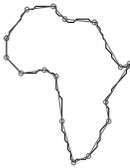 |
| Scaling: 1/3 | 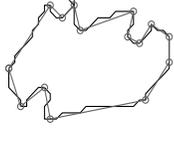 | 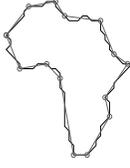 |
| Scaling: 1/4 | 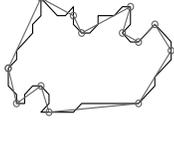 | 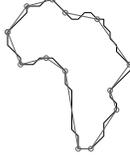 |
| Scaling: 1/5 | 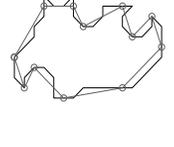 | 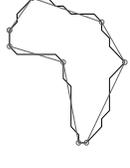 |

**Figure 2.7-3: Effect of scaling on Masood(mod)**

**Table 2.7-3: Scaling performance parameters of Masood(mod).**

| | No. of pixels (M) | Dominant points (N) | MD | ISE | FOM | Precision $\varepsilon'_p$ | Reliability $\varepsilon_r$ | CR |
|---|---|---|---|---|---|---|---|---|
| Tin opener | | | | | | | | |
| Scaling 1 | 278 | 35 | 1.02 | 51.71 | 0.15 | 0.39 | 0.30 | 7.94 |
| Scaling 1/2 | 135 | 18 | 1.09 | 28.10 | 0.27 | 0.38 | 0.33 | 7.50 |
| Scaling 1/3 | 88 | 16 | 0.98 | 16.92 | 0.33 | 0.34 | 0.33 | 5.50 |
| Scaling 1/4 | 62 | 14 | 0.95 | 11.56 | 0.38 | 0.35 | 0.32 | 4.43 |
| Scaling 1/5 | 49 | 13 | 1.00 | 12.22 | 0.31 | 0.42 | 0.40 | 3.77 |
| Africa | | | | | | | | |
| Scaling 1 | 291 | 28 | 1.05 | 56.74 | 0.18 | 0.36 | 0.31 | 10.39 |
| Scaling 1/2 | 137 | 17 | 0.99 | 27.31 | 0.30 | 0.36 | 0.32 | 8.06 |
| Scaling 1/3 | 90 | 15 | 1.18 | 15.40 | 0.39 | 0.35 | 0.30 | 6.00 |
| Scaling 1/4 | 64 | 14 | 0.81 | 8.86 | 0.52 | 0.29 | 0.26 | 4.57 |
| Scaling 1/5 | 49 | 9 | 0.89 | 8.00 | 0.68 | 0.28 | 0.29 | 5.44 |

| | Tin opener | Africa |
|---|---|---|
| Scaling: 1/2 | 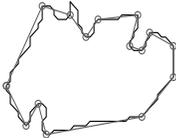 | 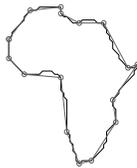 |
| Scaling: 1/3 | 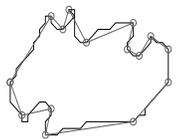 | 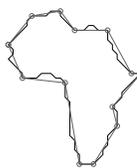 |
| Scaling: 1/4 | 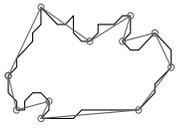 | 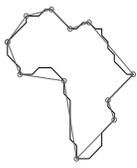 |
| Scaling: 1/5 | 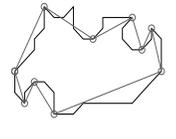 | 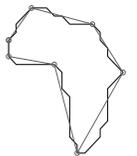 |

Figure 2.7-4: Effect of scaling on Carmona(mod)

**Table 2.7-4: Scaling performance parameters of Carmona(mod).**

| | No. of pixels (M) | Dominant points (N) | MD | ISE | FOM | Precision $\varepsilon'_p$ | Reliability $\varepsilon_r$ | CR |
|---|---|---|---|---|---|---|---|---|
| | | | | Tin opener | | | | |
| Scaling 1 | 278 | 39 | 2.19 | 73.08 | 0.10 | 0.24 | 0.30 | 7.13 |
| Scaling 1/2 | 135 | 16 | 2.19 | 62.50 | 0.14 | 0.53 | 0.48 | 8.44 |
| Scaling 1/3 | 88 | 16 | 1.37 | 26.79 | 0.21 | 0.43 | 0.39 | 5.50 |
| Scaling 1/4 | 62 | 12 | 1.46 | 19.67 | 0.26 | 0.48 | 0.40 | 5.17 |
| Scaling 1/5 | 49 | 11 | 1.79 | 27.16 | 0.16 | 0.58 | 0.59 | 4.45 |
| Africa | | | | | | | | |
| Scaling 1 | 291 | 26 | 1.80 | 93.62 | 0.12 | 0.41 | 0.37 | 11.19 |
| Scaling 1/2 | 137 | 17 | 1.74 | 44.19 | 0.18 | 0.40 | 0.38 | 8.06 |
| Scaling 1/3 | 90 | 14 | 1.66 | 25.27 | 0.25 | 0.38 | 0.34 | 6.43 |
| Scaling 1/4 | 64 | 12 | 1.70 | 21.55 | 0.25 | 0.38 | 0.35 | 5.33 |
| Scaling 1/5 | 49 | 8 | 1.54 | 14.70 | 0.42 | 0.37 | 0.38 | 6.13 |

**Table 2.7-5: Performance of the proposed methods for noisy digital curves.**

| | No. of Dominant | | | | | Precision | Reliability | |
|---|---|---|---|---|---|---|---|---|
| | pixels (M) | points (N) | MD | ISE | FOM | $\varepsilon'_p$ | $\varepsilon_r$ | CR |
| Tin opener | | | | | | | | |
| PRO(1.0) | 278 | 27 | 1.66 | 90.16 | 0.11 | 0.40 | 0.38 | 10.30 |
| PRO(1.0) noisy | 296 | 26 | 2.11 | 172.22 | 0.07 | 0.63 | 0.57 | 11.38 |
| RDP(mod) | 278 | 31 | 1.48 | 77.84 | 0.12 | 0.39 | 0.35 | 8.97 |
| RDP(mod) noisy | 296 | 31 | 1.40 | 98.20 | 0.10 | 0.48 | 0.40 | 9.55 |
| Masood(mod) | 278 | 35 | 1.02 | 51.71 | 0.15 | 0.39 | 0.30 | 7.94 |
| Masood(mod) noisy | 296 | 50 | 1.20 | 60.29 | 0.10 | 0.30 | 0.32 | 5.92 |
| Carmona(mod) | 278 | 39 | 2.19 | 73.08 | 0.10 | 0.24 | 0.30 | 7.13 |
| Carmona(mod) noisy | 296 | 29 | 3.41 | 237.69 | 0.04 | 0.58 | 0.61 | 10.21 |
| Africa | | | | | | | | |
| PRO(1.0) | 291 | 23 | 2.04 | 110.23 | 0.11 | 0.52 | 0.44 | 12.65 |
| PRO(1.0) noisy | 321 | 26 | 1.98 | 151.84 | 0.08 | 0.57 | 0.53 | 12.35 |
| RDP(mod) | 291 | 29 | 1.27 | 63.21 | 0.16 | 0.39 | 0.31 | 10.03 |
| RDP(mod) noisy | 321 | 37 | 1.36 | 93.35 | 0.09 | 0.46 | 0.40 | 8.68 |
| Masood(mod) | 291 | 28 | 1.05 | 56.74 | 0.18 | 0.36 | 0.31 | 10.39 |
| Masood(mod) noisy | 321 | 57 | 0.99 | 65.33 | 0.09 | 0.33 | 0.32 | 5.63 |
| Carmona(mod) | 291 | 26 | 1.80 | 93.62 | 0.12 | 0.41 | 0.37 | 11.19 |
| Carmona(mod) noisy | 321 | 28 | 2.04 | 166.37 | 0.07 | 0.58 | 0.32 | 11.46 |

### 2.7.2 Experiment 2 – noisy digital curves

This section presents the performance of the three modified algorithms (sections 2.5.1.2, 2.5.2.2, and 2.5.3.2) and PRO(1.0) for noisy digital curves. For this purpose, noise is added to the digital curves using the Kanungo model [163]. The parameters used for Kanungo noise model were $\alpha_0 = \beta_0 = 4, \alpha = \beta = 2, \eta_f = \eta_b = 0$. The performance of the four methods for two curves – tin opener and Africa – are presented in Figure 2.7-5 and tabulated in Table 2.7-5. Results for PRO(1.0), RDP(mod), Masood(mod) and Carmona(mod) in Table 2.7-5 correspond to the results from the Table 2.5-1, Table 2.5-2, and Table 2.5-3 respectively.

It is noted that the CR of PRO(1.0), RDP(mod), and Carmona(mod) remains approximately the same for noiseless and noisy curves, while the CR of Masood(mod) decreases significantly for noisy curves. This is consistent with the nature of Masood's method which supports very close fitting and consequently results in more dominant points in order to fit closely to the noise. On the other hand, RDP and Carmona both have relatively more smoothing effects as compared to Masood. It is also noted that the precision metric increases significantly for PRO(1.0), RDP(mod), and Carmona(mod) in the case of noisy curves. This shows that the precision is compromised in these methods for giving a good performance for noisy curves. The decrease in the precision metric results in the smoothing effect observed for these methods.

|  | Tin opener | Africa |
|---|---|---|
| Example of noisy boundaries | 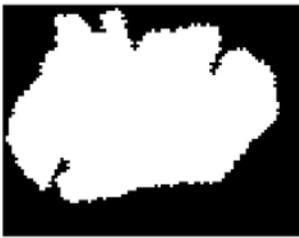 | 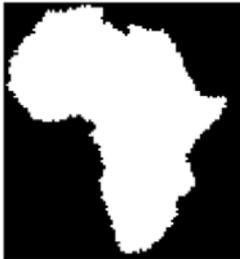 |
| PRO(1.0) | 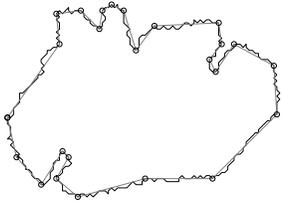 | 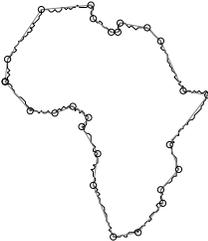 |
| RDP(mod) | 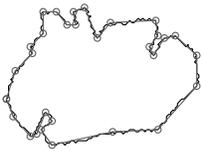 | 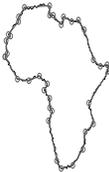 |
| Masood(mod) | 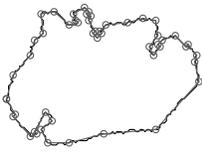 | 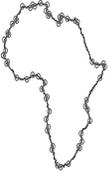 |
| Carmona(mod) | 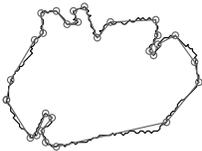 | 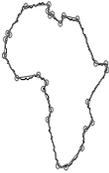 |

**Figure 2.7-5: Performance of the proposed methods for noisy digital curves.**

### 2.7.3 Experiment 3 – non-digitized and semi-digitized curves

In this section, the applicability of the PRO(1.0), RDP(mod), Masood(mod), and Carmona(mod) for non-digitized and semi-digitized curves is demonstrated. First, a set of non-digitized curves given by the parametric equations (2-68) and (2-69) is considered.

$$r = a\left(\left(\cos\left(\frac{m\theta}{4}\right)\right)^{n_2} + \left(\sin\left(\frac{m\theta}{4}\right)\right)^{n_3}\right)^{-\frac{1}{n_1}} \tag{2-68}$$

$$x = r\cos\left(\theta + \theta_0\right) + x_0 ; y = r\sin\left(\theta + \theta_0\right) + y_0 \tag{2-69}$$

where $a, m, n_1, n_2, n_3, \theta_0, x_0$, and $y_0$ are the parameters of the curve. Using equations (2-68) and (2-69), three curves with the set of parameters in Table 2.7-6 are generated. The values of the $x$ and $y$ coordinates given by (2-69) are kept in the form of double-precision floating numbers in Matlab. The curves are shown in Figure 2.7-6.

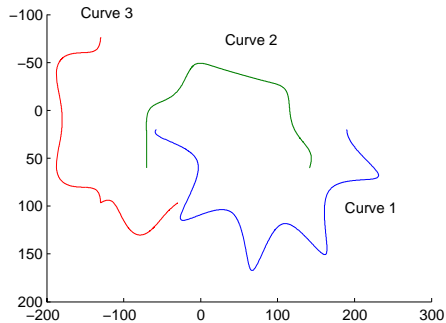

**Figure 2.7-6: Three non-digital curves.**

**Table 2.7-6: The parameters of three non-digital curves.**

| Curve | $a$ | $m$ | $n_1$ | $n_2$ | $n_3$ | $\theta_0$ | $x_0$ | $y_0$ | $\theta$ |
|---|---|---|---|---|---|---|---|---|---|
| Curve 1 | | 9 | 9 | 14 | 11 | 0 | 90 | 20 | $\theta = p\,\pi/1000$ |
| Curve 2 | 100 | 7 | 9 | 3 | 11 | $\pi$ | 30 | 60 | |
| Curve 3 | | 6 | 1 | 1 | 6 | $\pi/3$ | -80 | 10 | $p = 0$ to $1000$ |

For these curves, the result of RDP(mod), Masood (mod), and Carmona(mod) are given in Figure 2.7-7 and Table 2.7-7. The results clearly show that the modified methods based on the proposed non-parametric framework perform well even for non-digitized (real valued) curves.

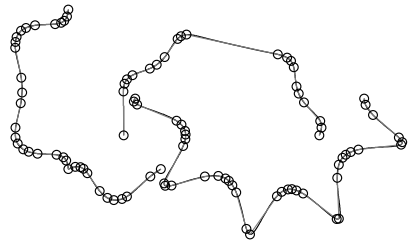

(a) Result of PRO(1.0)

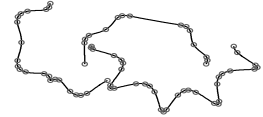

(b) Result of RDP(mod)

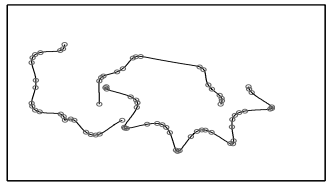

(c) Result of Masood(mod)

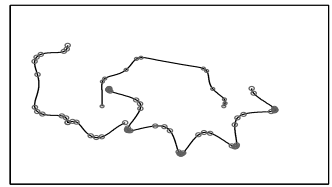

(d) Result of Carmona(mod)

**Figure 2.7-7: Performance of proposed methods for non-digital curves given in Figure 2.7-6.**

**Table 2.7-7: Performance of proposed methods for non-digital curves given in Figure 2.7-7.**

| | No. of data points (M) | Dominant points (N) | MD | ISE | FOM | Precision $\varepsilon'_p$ | Reliability $\varepsilon_r$ | CR |
|---|---|---|---|---|---|---|---|---|
| PRO(1.0) | 3003 | 96 | 1.93 | 234.65 | 0.13 | 0.22 | 0.52 | 31.28 |
| RDP(mod) | (1001 | 70 | 1.40 | 355.59 | 0.12 | 0.50 | 1.30 | 42.90 |
| Masood(mod) | per | 77 | 0.88 | 212.82 | 0.18 | 0.35 | 0.78 | 39.00 |
| Carmona(mod) | curve) | 76 | 4.04 | 1754.49 | 0.02 | 0.52 | 1.80 | 39.51 |

Now a semi-digitized curve is considered in which one axis ($x$ axis) is digitized (similar to data plots). A data plot given by the eqn. (2-70) and plotted in Figure 2.7-8(a) is considered as an example:

$$y = 100\frac{\sin(x/5)}{(x/5)}; \; x \in \{1, 2, \ldots, 100\} \tag{2-70}$$

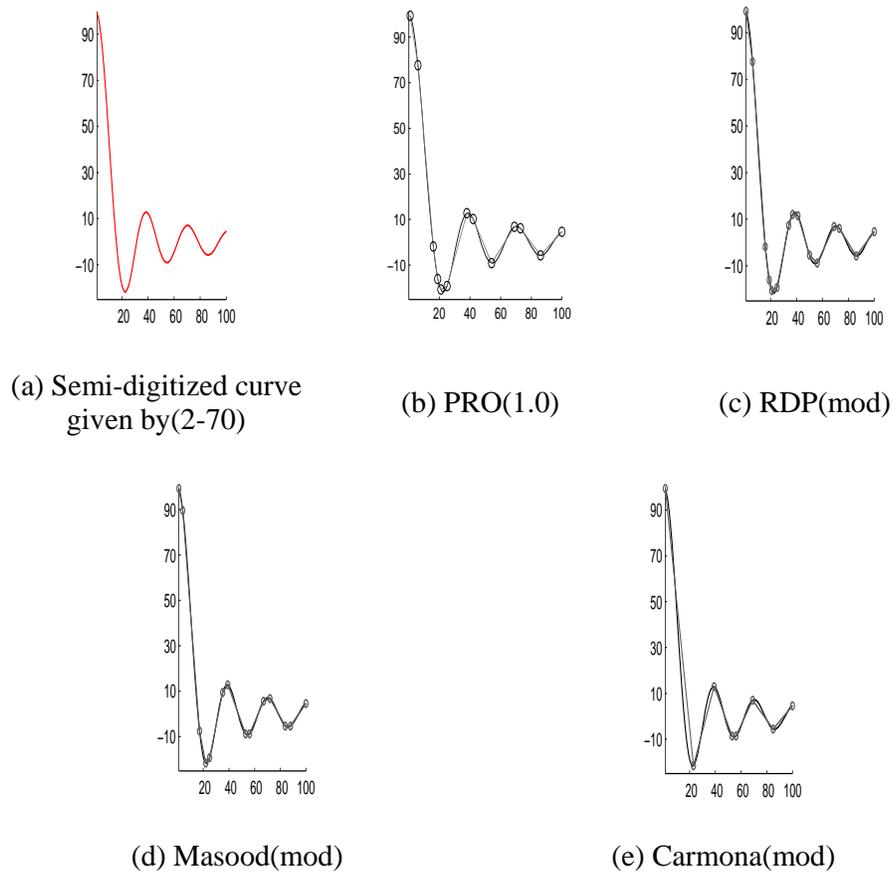

(a) Semi-digitized curve
given by(2-70)

(b) PRO(1.0)

(c) RDP(mod)

(d) Masood(mod)

(e) Carmona(mod)

Figure 2.7-8: Semi-digitized curve and performance of proposed methods.

**Table 2.7-8: Performance of proposed methods for semi-digitized curve.**

| | No. of data points (M) | Dominant points (N) | MD | ISE | FOM | Precision $\varepsilon_p'$ | Reliability $\varepsilon_r$ | CR |
|---|---|---|---|---|---|---|---|---|
| PRO(1.0) | | 13 | 1.64 | 60.14 | 0.13 | 0.58 | 0.24 | 7.69 |
| RDP(mod) | 100 | 15 | 1.52 | 34.21 | 0.20 | 0.45 | 0.17 | 6.67 |
| Masood(mod) | | 14 | 1.08 | 19.55 | 0.37 | 0.39 | 0.14 | 7.14 |
| Carmona(mod) | | 8 | 3.38 | 160.59 | 0.08 | 0.95 | 0.39 | 12.50 |

For this plot, the dominant points detected by RDP(mod), Masood (mod), and Carmona(mod) are given in Figure 2.7-8(b-e) respectively. The performance parameters are tabulated in Table 2.7-8. The results clearly demonstrate the applicability of the proposed framework for semi-digitized curves like data plots as well.

### 2.7.4 Experiment 4 – Performance over datasets

In this experiment, various datasets used in diverse image processing applications like segmentation, object detection, object recognition, object categorization, shape and contour based analysis, virtual reality, and gaming applications are considered. The datasets considered are Afreight dataset [164] (920 images), Google dataset[1] (5000 images), Berkley dataset [165] (300 images), Cars_inria dataset [166] (150 images), Caltech 101 dataset [167] (9149 images), Caltech 256 dataset [1] (30608 images), PASCAL 2007 dataset [168] (9963 images), PASCAL 2008 dataset [169] (10057 images), PASCAL 2009 dataset [170] (14743 images), and PASCAL 2010 dataset [171] (21738 images). The binary edge map of the image is passed to the various algorithms after canny edge operation with threshold ($T_L = 0.1$ and $T_H = 0.2$). For all these datasets the performance parameters are listed in Table 2.7-9.

Masood [46], and Carmona [35] have high values of $\varepsilon'_p$ and $\varepsilon_r$. PRO0.2 always has the lowest values of $\varepsilon'_p$, $\varepsilon_r$, $E_{max}$, and ISE and the highest values of ADR and FOM. It performs the best in terms of ADR but poorest among all for FOM (equaled with Carmona [35]). Since Masood [46] primarily focuses on the local nature of the fit, as compared to Carmona which concentrates on the global nature of fit, Masood has significantly low values of $E_{max}$, low value of ISE and higher value of FOM. However, qualitatively, the difference may not be significantly different. The performance of Masood and Carmona is similar in terms of $\varepsilon'_p$ and $\varepsilon_r$. This is because the precision and reliability metrics are better designed to balance the global and local natures of fit. Masood takes significantly longer time as compared to any other method and its time complexity increases with the increase in the size of the image (curves) as seen from Google dataset result in Table 2.7-9(i). Finally, the quantitative data of PRO1.0 indicates that PRO1.0 avoids both extremes for all the performance parameters and provides a good balance in terms of the conflicting requirements of dominant point detection methods.

---

[1] This dataset has been formed by selecting images of various sizes from a random Google search. In this dataset the minimum dimension (length or breadth) of the images are $25n$, $n = 1$ to $100$ and 50 images have been selected for each value of $n$. Thus, the smallest image is of size 25 pixels (length or breadth wise or both), while the largest image is of size 2500 pixels (length or breadth wise or both).

**Table 2.7-9: Performance of various methods for all the datasets.**

| (a) Precision metric $\varepsilon'_p$ | | | | | | |
|---|---|---|---|---|---|---|
| | PRO (1.0) | RDP (1.0) | RDP (mod) | Masood (0.9) | Masood (mod) | Carmon a (0.45) | Carmon a (mod) |
| Afreight | 0.44 | 0.29 | 0.41 | 0.89 | 0.49 | 0.87 | 0.53 |
| Google | 0.43 | 0.28 | 0.38 | 0.83 | 0.74 | 1.18 | 0.85 |
| Berkley | 0.21 | 0.11 | 0.39 | 1.17 | 0.52 | 1.19 | 0.36 |
| Cars_inria | 0.49 | 0.32 | 0.53 | 1.34 | 0.64 | 1.49 | 0.56 |
| Caltech 101 | 0.43 | 0.28 | 0.39 | 1.11 | 0.42 | 1.27 | 0.45 |
| Caltech 256 | 0.44 | 0.28 | 0.63 | 1.13 | 0.51 | 1.41 | 0.59 |
| Pascal 2007 | 0.43 | 0.28 | 0.44 | 1.30 | 0.46 | 1.49 | 0.62 |
| Pascal 2008 | 0.42 | 0.27 | 0.49 | 1.29 | 0.45 | 1.48 | 0.68 |
| Pascal 2009 | 0.41 | 0.27 | 0.51 | 1.29 | 0.49 | 1.48 | 0.65 |
| Pascal 2010 | 0.42 | 0.26 | 0.47 | 1.29 | 0.44 | 1.49 | 0.71 |
| **Mean** | **0.41** | **0.26** | **0.46** | **1.16** | **0.52** | **1.34** | **0.60** |

**Table 2.7-9: Performance of various methods for all the datasets.**

| (b) Reliability metric $\varepsilon_r$ | | | | | | |
|---|---|---|---|---|---|---|
| | PRO (1.0) | RDP (1.0) | RDP (mod) | Masood (0.9) | Masood (mod) | Carmon a (0.45) | Carmon a (mod) |
| Afreight | 0.37 | 0.25 | 0.35 | 0.86 | 0.51 | 0.84 | 0.49 |
| Google | 0.37 | 0.23 | 0.32 | 0.76 | 0.87 | 1.23 | 0.73 |
| Berkley | 0.17 | 0.09 | 0.32 | 1.26 | 0.46 | 1.13 | 0.39 |
| Cars_inria | 0.41 | 0.27 | 0.40 | 1.42 | 0.89 | 1.58 | 0.46 |
| Caltech 101 | 0.37 | 0.24 | 0.42 | 1.12 | 0.75 | 1.23 | 0.52 |
| Caltech 256 | 0.36 | 0.24 | 0.44 | 1.12 | 0.63 | 1.44 | 0.45 |
| Pascal 2007 | 0.36 | 0.24 | 0.36 | 1.32 | 0.73 | 1.5 | 0.51 |
| Pascal 2008 | 0.35 | 0.23 | 0.38 | 1.33 | 0.69 | 1.52 | 0.52 |
| Pascal 2009 | 0.35 | 0.23 | 0.39 | 1.32 | 0.75 | 1.5 | 0.49 |
| Pascal 2010 | 0.35 | 0.23 | 0.37 | 1.32 | 0.77 | 1.53 | 0.53 |
| **Mean** | **0.34** | **0.23** | **0.38** | **1.18** | **0.71** | **1.35** | **0.51** |

**Table 2.7-9: Performance of various methods for all the datasets.**

| (c) Average compression ratio (CR) | | | | | | |
|---|---|---|---|---|---|---|
| | PRO (1.0) | RDP (1.0) | RDP (mod) | Masood (0.9) | Masood (mod) | Carmon a (0.45) | Carmon a (mod) |
| Afreight | 7.69 | 6.25 | 7.18 | 6.25 | 6.10 | 4.55 | 6.21 |
| Google | 10.00 | 7.14 | 8.73 | 7.69 | 9.49 | 7.69 | 8.58 |
| Berkley | 5.88 | 4.76 | 5.86 | 5.26 | 6.21 | 4.17 | 5.96 |
| Cars_inria | 6.25 | 5.00 | 7.24 | 5.26 | 6.13 | 4.17 | 5.93 |
| Caltech 101 | 7.14 | 5.26 | 7.05 | 5.56 | 6.97 | 4.76 | 6.46 |
| Caltech 256 | 7.69 | 5.88 | 8.31 | 6.25 | 7.36 | 5.26 | 7.96 |
| Pascal 2007 | 11.11 | 8.33 | 12.34 | 5.88 | 8.74 | 4.76 | 10.52 |
| Pascal 2008 | 11.11 | 8.33 | 11.96 | 5.88 | 10.28 | 4.76 | 9.89 |
| Pascal 2009 | 12.50 | 8.33 | 11.68 | 5.88 | 8.72 | 4.76 | 9.98 |
| Pascal 2010 | 11.11 | 8.33 | 11.35 | 5.56 | 9.21 | 4.76 | 10.34 |
| **Mean** | **8.33** | **6.25** | **9.17** | **5.88** | **7.92** | **4.76** | **8.18** |

**Table 2.7-9: Performance of various methods for all the datasets.**

| **(d)** Maximum deviation (Emax) | PRO (1.0) | RDP (1.0) | RDP (mod) | Masood (0.9) | Masood (mod) | Carmon a (0.45) | Carmon a (mod) |
|---|---|---|---|---|---|---|---|
| Afreight | 2.06 | 1.00 | 1.35 | 2.22 | 1.43 | 2.05 | 1.38 |
| Google | 2.10 | 0.99 | 1.41 | 2.05 | 1.37 | 3.91 | 1.45 |
| Berkley | 2.31 | 1.00 | 1.31 | 0.94 | 1.28 | 2.63 | 1.39 |
| Cars_inria | 2.65 | 1.00 | 1.20 | 2.32 | 1.19 | 3.49 | 1.50 |
| Caltech 101 | 2.32 | 1.00 | 1.35 | 2.62 | 1.24 | 3.17 | 1.42 |
| Caltech 256 | 2.34 | 1.00 | 1.53 | 2.75 | 1.36 | 3.83 | 1.39 |
| Pascal 2007 | 2.53 | 1.00 | 1.45 | 3.08 | 1.29 | 3.59 | 1.41 |
| Pascal 2008 | 2.54 | 1.00 | 1.33 | 3.06 | 1.33 | 3.63 | 1.37 |
| Pascal 2009 | 2.55 | 1.00 | 1.39 | 3.11 | 1.27 | 3.57 | 1.45 |
| Pascal 2010 | 2.54 | 1.00 | 1.41 | 3.08 | 1.31 | 3.63 | 1.39 |
| **Mean** | **2.39** | **1.00** | **1.37** | **2.52** | **1.31** | **3.35** | **1.42** |

**Table 2.7-9: Performance of various methods for all the datasets.**

| **(e)** Integral square error (ISE) | PRO (1.0) | RDP (1.0) | RDP (mod) | Masood (0.9) | Masood (mod) | Carmon a (0.45) | Carmon a (mod) |
|---|---|---|---|---|---|---|---|
| Afreight | 59.15 | 27.03 | 30.33 | 39.77 | 35.32 | 71.73 | 41.25 |
| Google | 223.97 | 95.15 | 165.31 | 64.52 | 178.53 | 5171.01 | 196.48 |
| Berkley | 72.15 | 32.90 | 101.23 | 72.68 | 98.75 | 149.44 | 99.65 |
| Cars_inria | 180.79 | 69.67 | 203.51 | 81.58 | 185.91 | 754.50 | 195.73 |
| Caltech 101 | 128.52 | 53.39 | 192.64 | 54.97 | 164.57 | 417.71 | 186.43 |
| Caltech 256 | 155.97 | 62.14 | 185.67 | 64.17 | 173.68 | 1690.91 | 221.58 |
| Pascal 2007 | 146.07 | 57.64 | 181.59 | 83.32 | 154.83 | 639.90 | 265.31 |
| Pascal 2008 | 145.18 | 59.48 | 192.76 | 82.04 | 168.27 | 617.49 | 258.42 |
| Pascal 2009 | 146.00 | 60.48 | 182.83 | 86.37 | 165.39 | 600.87 | 251.39 |
| Pascal 2010 | 139.92 | 54.44 | 198.26 | 77.45 | 168.41 | 728.07 | 257.83 |
| **Mean** | **139.77** | **57.23** | **163.41** | **70.69** | **149.37** | **1084.16** | **197.41** |

**Table 2.7-9: Performance of various methods for all the datasets.**

| **(f)** Figure of merit (FOM) | PRO (1.0) | RDP (1.0) | RDP (mod) | Masood (0.9) | Masood (mod) | Carmon a (0.45) | Carmon a (mod) |
|---|---|---|---|---|---|---|---|
| Afreight | 0.19 | 0.32 | 0.31 | 0.28 | 0.32 | 0.15 | 0.28 |
| Google | 0.13 | 0.24 | 0.08 | 0.24 | 0.11 | 0.07 | 0.10 |
| Berkley | 0.11 | 0.20 | 0.09 | 0.13 | 0.08 | 0.05 | 0.10 |
| Cars_inria | 0.05 | 0.09 | 0.07 | 0.10 | 0.09 | 0.02 | 0.08 |
| Caltech 101 | 0.08 | 0.16 | 0.09 | 0.15 | 0.10 | 0.03 | 0.09 |
| Caltech 256 | 0.07 | 0.15 | 0.11 | 0.15 | 0.13 | 0.02 | 0.08 |
| Pascal 2007 | 0.07 | 0.13 | 0.14 | 0.11 | 0.16 | 0.02 | 0.10 |
| Pascal 2008 | 0.07 | 0.13 | 0.14 | 0.11 | 0.15 | 0.02 | 0.11 |
| Pascal 2009 | 0.07 | 0.13 | 0.15 | 0.11 | 0.19 | 0.02 | 0.11 |
| Pascal 2010 | 0.07 | 0.14 | 0.12 | 0.11 | 0.14 | 0.02 | 0.13 |
| **Mean** | **0.09** | **0.17** | **0.13** | **0.15** | **0.15** | **0.04** | **0.12** |

### 2.7.5 **Recommendations**

From the discussions and numerical results in the previous sections, it is evident that RDP(mod), Masood(mod), and Carmona(mod) have distinct advantages over the original RDP, Masood, and Carmona methods respectively. Their applicability to curves of various types, non-digital curves, as well and semi-digitized data plots is also shown. Further, their applicability to various image datasets is also shown in section 2.7.4. In terms of the precision metric, reliability metric, maximum deviation, as well as compression ratio, the modified methods perform better than the corresponding original methods. It is also seen that PRO(1.0) performs similar to the modified methods. A natural question to be answered then is that which method should be preferred in practical applications.

The selection of an appropriate method depends upon the requirement of the application. If the local fitting of curves is needed and the details of small curvature changes are important, then Masood(mod) and PRO($\varepsilon_0$), $\varepsilon_0 \in (0.4, 0.7)$ are more suitable than other methods. However, PRO($\varepsilon_0$) is preferred over Masood(mod) in terms of the computational complexity. Further, if even finer fitting of smaller features is desired, PRO($\varepsilon_0$), $\varepsilon_0 \in (0.2, 0.4)$ can serve the purpose.

For general purposes, RDP(mod) and Carmona(mod) are more desirable than Masood(mod). Further PRO($\varepsilon_0$), $\varepsilon_0 \in (0.8, 1.0)$ also serves the purpose. From the point of computational complexity, RDP(mod) and PRO(mod) win over Carmona(mod).

## 2.8 Conclusion

This chapter discussed the theoretical aspects of the problem of polygonal approximation of digital curves. Specifically the nature of local and global fits of the approximate polygon with the curve is discussed. Simple precision and reliability metrics are proposed and used for proving that the least squares fitting of line (edges of the approximate polygon) encounters a conflict in optimizing the local and global qualities of fit together.

A PA method, PRO, that uses precision and reliability measures in the optimization goals is proposed. PRO can be easily tailored by changing the tolerance value. The fit is very close for small tolerance values ($\varepsilon_0 \sim 0.2$) and the closeness of the fit decreases as the value of $\varepsilon_0$ becomes close to 1. It is shown through extensive comparison that PRO(0.2) chooses the dominant points such that each and every change in the curvature is retained though the number of points needed is large. On the other hand, PRO(1.0) avoids the problems of under-fitting as well as over-fitting and provides a good balance for all the performance parameters that indicate local nature of fit, global nature of fit, dimensionality reduction, as well as the computation time.

In addition, a non-parametric framework for PA methods is proposed. The approach is based upon theoretical bound of the deviation of the pixels obtained by the digitization of a line segment. The approach is to use this bound in a PA method as either the optimization goal or the termination condition or both. It is shown that this approach can be incorporated in various types of PA methods easily to make them independent of control parameter and related heuristics. This is illustrated by modifying three different PA methods (RDP [12, 13], Masood [46], and Carmona [35]). The results show that as compared to the use of control parameters in the original versions of these methods, the modifications of the methods using the non-parametric approach provide a more balanced performance and good approximation of the digital curves. It is also shown that the modified versions of the PA detection methods can still retain their original natural characteristics. Though the approach is applied for three methods only in this thesis, the approach can be suitably applied in most dominant point detection methods to make them free of heuristics.

PA methods discussed in this chapter are also applied to non-digital, semi-digital, and noisy digital curves. As a final note, the results of these PA methods for 10 practical computer vision datasets are shown. Such a study is important for the research community that uses dominant point detection as one of the fundamental preprocessing steps in several applications. The contents of Chapter 2 have been reported in [93, 131-135].

# Chapter 3 : Tangent estimation of digital curves

## 3.1 Background

This chapter proposes a novel tangent estimation method that is very simple and has a firm analytical foundation. Since the proposed method has definite continuous curve and digital curve upper bounds, it is called as the Definite Error Bounded (DEB) TE method. It is proven that in a continuous conic, the slope computed by DEB closely matches the slope of the actual tangent. The proof of the analytical error bound for conic shapes is presented and numerical examples are shown. For digital curves (conics or non-conics), this work establishes the numerical error bound for the proposed tangent estimation method using the derivation in section 2.4, which is also derived using rigorous mathematical analysis.

Although a part of the total error bound is derived only for conic curves, DEB is not restricted to the case of perfectly digitized conics only. The proposed tangent estimation method can be applied to any digital or continuous curve which may be noisy or noise-free, though an explicit analytical error bound may be difficult to derive.

Several salient features of the DEB are:

1. Only two points at a certain distance from the point of interest should be identified for estimating the tangent at the point of interest.
2. The computational complexity is very low.
3. It gives good multigrid and isotropicity performance.
4. It performs well for conic and non-conic curves and provides superior performance than 71 other methods.

Section 3.2 presents an example in image processing which highlights the importance of reducing the error in TE. Section 3.3 proposes the DEB algorithm, its pseudocode and computational complexity analysis. Section 3.4 presents the total error bound of DEB. Section 3.5 presents numerical examples to illustrate the error bound. Section 3.6 presents the comparison of DEB with other TE methods for various conic and non-conic curves and the chapter is concluded in section 3.5.

The highlights and contributions of this chapter are summarized here. The novel DEB algorithm is a very simple and computation efficient tangent estimator. It is shown that despite the simplicity of the method, the method has a definite upper bounded error for conics. This method does not suffer from the tangent estimation singularity experienced by other methods [31, 69]. The method uses a simple control parameter and a rule of thumb for choosing the control parameter is also provided. It is notable that the rule of thumb also uses the upper bound of the error in tangent estimation and thus it is less empirical than most other tangent estimation methods. Extensive numerical experiments validate the superiority of the proposed tangent estimator over almost all the existent tangent estimators. Further, the applicability of the proposed tangent estimator for non-conic curves with convex and concave curvatures is also exhibited.

## 3.2 Example of importance of tangent estimation

In this section, using the geometric method proposed by Yuen [28] (which has found vast application in ellipse detection methods [65, 68, 70, 75, 94, 96, 111]), the impact of error in TE on the performance of this method is demonstrated. TE is an important step in the Yuen's three point method for finding the centers of the ellipses.

### 3.2.1 Yuen's three point method [28] for finding the centers of the ellipses

Yuen [28] proposed a method for ellipse detection in which the 5-dimensional parameter space of ellipse detection problem is split into two sets:

1. The first set of two-dimensional parameter space comprising the coordinates of the centers of ellipses – a geometric method that requires three points to find the center of the ellipse is used for the first set.

2. The second set of three-dimensional parameter space, comprising the remaining parameters of the ellipses – the same three points used in the first set are used in a least squares framework to determine the parameters of the second set.

Randomized HT is then used for various such computations to arrive at an acceptable solution.

The geometric method used for the first set of parameters – the coordinates of the center of the ellipse – is presented here. Consider three points $P_1(x_1, y_1)$, $P_2(x_2, y_2)$, and $P_3(x_3, y_3)$, as shown in Figure 3.2-1. The tangents to the ellipse at these three points are marked as $t_1$, $t_2$, and $t_3$ respectively. Let the slopes of the tangents at these points be given by $\tan\theta_1$, $\tan\theta_2$, and $\tan\theta_3$. The intersection point of lines $t_1$ and $t_2$ is denoted as $P_{\tan,12}$ and that of $t_2$ and $t_3$ as $P_{\tan,23}$. Further, the midpoint of the line segment joining $P_1$ and $P_2$ is denoted as $P_{\mathrm{mid},12}$ and the midpoint of the line segment joining $P_2$ and $P_3$ as $P_{\mathrm{mid},23}$. Now, a line $l_{12}$ that passes through $P_{\mathrm{mid},12}$ and $P_{\tan,12}$, and a line $l_{23}$ that passes through $P_{\mathrm{mid},23}$ and $P_{\tan,23}$ are computed. The center of the ellipse is given by the intersection point of the lines $l_{12}$ and $l_{23}$.

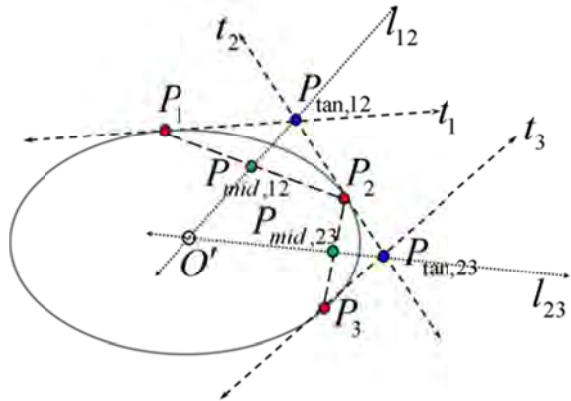

**Figure 3.2-1: Graphical illustration of the Yuen's method.**

### 3.2.2 Error metric and tangent tests

The actual coordinates of the ellipse's center $O'(x', y')$ using Yuen's method in terms of $P_1(x_1, y_1)$, $P_2(x_2, y_2)$, $P_3(x_3, y_3)$, $\theta_1$, $\theta_2$, and $\theta_3$ are computed using eqns. (3-1) and (3-2):

$$x' = -\frac{A_{12}C_{23} - A_{23}C_{12}}{B_{12}C_{23} - B_{23}C_{12}}, \tag{3-1}$$

$$y' = -\frac{A_{12}B_{23} - A_{23}B_{12}}{B_{12}C_{23} - B_{23}C_{12}}, \tag{3-2}$$

where:

$$A_{12} = y_2^2 - y_1^2 - \left(x_2 y_2 - x_1 y_1\right)\left(\tan\theta_1 + \tan\theta_2\right) + \left(x_2^2 - x_1^2\right)\tan\theta_1\tan\theta_2 \tag{3-3}$$

$$A_{23} = y_3^2 - y_2^2 - \left(x_3 y_3 - x_2 y_2\right)\left(\tan\theta_2 + \tan\theta_3\right) + \left(x_3^2 - x_2^2\right)\tan\theta_2\tan\theta_3 \tag{3-4}$$

$$B_{12} = \left(y_2 - y_1\right)\left(\tan\theta_1 + \tan\theta_2\right) - 2\left(x_2 - x_1\right)\tan\theta_1\tan\theta_2 \tag{3-5}$$

$$B_{23} = \left(y_3 - y_2\right)\left(\tan\theta_2 + \tan\theta_3\right) - 2\left(x_3 - x_2\right)\tan\theta_2\tan\theta_3 \tag{3-6}$$

$$C_{12} = 2\left(y_2 - y_1\right) - \left(x_2 - x_1\right)\left(\tan\theta_1 + \tan\theta_2\right) \tag{3-7}$$

$$C_{23} = 2\left(y_3 - y_2\right) - \left(x_3 - x_2\right)\left(\tan\theta_2 + \tan\theta_3\right) \tag{3-8}$$

If the center of the actual ellipse is at $O(x_0, y_0)$, then the net distance between the computed and actual centers can be used as the error metric in eqn. (3-9):

$$r_{\text{err}} = \sqrt{\left(x_0 - x'\right)^2 + \left(y_0 - y'\right)^2} \tag{3-9}$$

For the simplicity of further analysis, and without the loss of generalization, an actual ellipse centered at the origin (i.e. $x_0 = y_0 = 0$) is considered. Further, it is assumed that the ellipse is oriented along the $x-$axis and has $a$ and $b$ as the lengths of semi-major and semi-minor axes. Since the computations become singular when any of the points are on the major axis or minor axis, these points shall be avoided, as they may skew the analysis without yielding any significant insight.

For convenience, the points $P_i\left(x_i, y_i\right)$, $i = 1$ to $3$ are defined using their angular position $\alpha_i$ as in eqn. (3-10).

$$x_i = a\cos\alpha_i; \quad y_i = b\sin\alpha_i \tag{3-10}$$

where $\alpha_i = \left\{-\dfrac{5\pi}{12}, \dfrac{\pi}{4}, \dfrac{11\pi}{12}\right\}$ are chosen for optimal coverage of the curvature of the ellipse [94]. Further, the semi-major axis $b = a\sqrt{1 - e^2}$, where eccentricity $e = 0.8$ is used for illustration (i.e. $b = 0.6a$). The actual tangents can be calculated using eqn. (3-11):

$$\theta_i = \tan^{-1}\left(-\frac{b}{a}\cot\alpha_i\right) \qquad (3\text{-}11)$$

The above choices are almost the perfect and most representative choices for the method. Thus, using these default choices and varying one parameter at a time, it can be safely assumed that the effect of only the varied parameter is being observed.

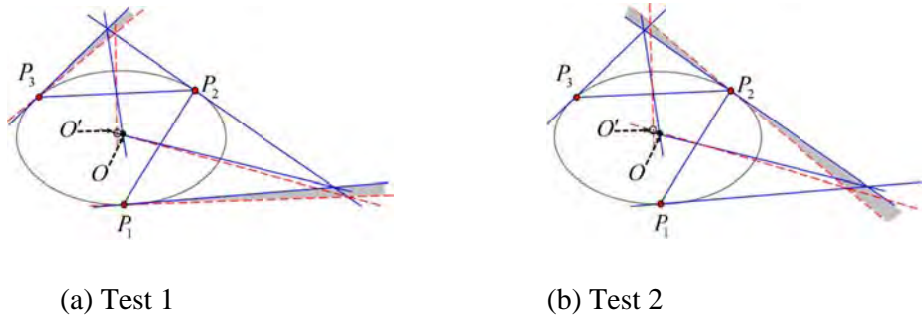

(a) Test 1          (b) Test 2

**Figure 3.2-2: Illustration of the two tests in section 3.2.2.**

Two tests of the sensitivity of the geometric method to the error in tangent computation are considered (illustrated in Figure 3.2-2, where dashed lines show erroneous lines and the unfilled small circles show the erroneous center computed due to the error in tangents):

Test 1: The relative error $r_{err}/a$ for various values of error $\delta\theta$ in TE at points $P_1$ and $P_3$, assuming that there is an equal amount of error in the tangents $\theta_1$ and $\theta_3$ (Figure 3.2-2(a)) as in eqn. (3-12):

$$\theta_i = \tan^{-1}\left(-\frac{b}{a}\cot\alpha_i\right) + \delta\theta; \quad i = 1,3 \qquad (3\text{-}12)$$

Test 2: The relative error $r_{err}/a$ for various values of error $\delta\theta$ in TE at point $P_2$ (Figure 3.2-2(b)) as in eqn. (3-13):

$$\theta_2 = \tan^{-1}\left(-\frac{b}{a}\cot\alpha_2\right) + \delta\theta. \qquad (3\text{-}13)$$

The range $\delta\theta \in [-\pi/12, \pi/12]$ (i.e. $[-15°, 15°]$ is considered.

### 3.2.2.1 Tests 1 and 2 in the absence of digitization

In the absence of digitization, eqn. (3-10) is used for computing the coordinates. The results of tests 1 and 2 are presented in Figure 3.2-3. It is seen that the error in tangents leads to significant non-negligible error for both the tests. Specifically, the error in test 2 is higher than in test 1. This is because the error in the tangent at the second point $P_2$ affects the method more as it is used in all the computations.

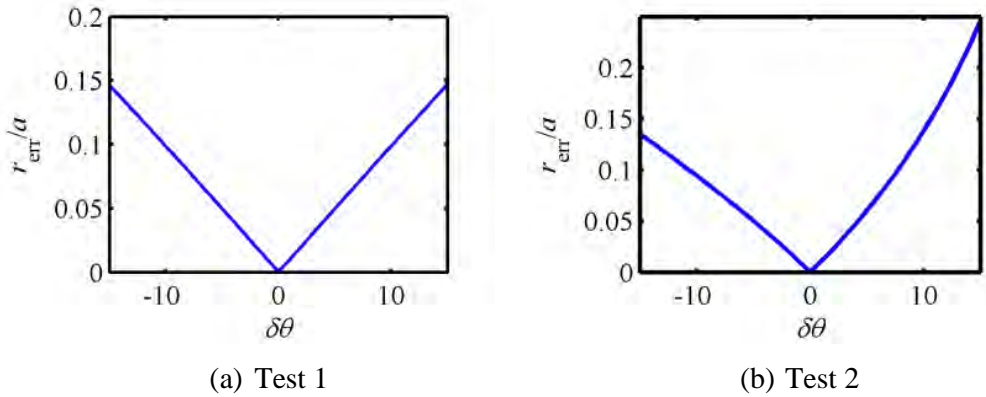

(a) Test 1          (b) Test 2

**Figure 3.2-3: Absence of digitization: Results of the tests 1 and 2.**

### 3.2.2.2 Tests 1 and 2 in the presence of digitization

In the presence of digitization, eqn. (3-14) below is used instead of eqn. (3-10) to compute the coordinates of the points:

$$x_i = \text{round}(a\cos\alpha_i); \quad y_i = \text{round}(b\sin\alpha_i), \qquad (3\text{-}14)$$

where the function $\text{round}(\bullet)$ denotes the rounding to the nearest integer. For this case, a variation of the major axis of the ellipse $a \in [10, 100]$ is considered. In this case, the maximum and average relative error for a test for each value of $a$ is presented.

The results of the two tests in the presence of digitization are shown in Figure 3.2-4. The dotted plot (red) shows the maximum error for the complete range of parameters, while the solid line (blue) shows the average error for the complete range for a particular value of $a$. Both the tests show that the relative error is indeed sensitive to the error in computation of the tangents, but does not vary greatly for ellipses of various sizes. It remains almost flat. The average error for test 2 is higher and almost fixed at about 0.1 (~10%).

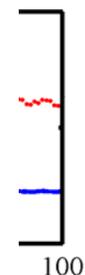

The above tests highlight the importance of reducing the error in the computation of the tangents. In fact it was shown in [25] that the error in TE at point $P_2$ is the single largest contributor of the error in Yuen's three point method. Thus, the need of a good TE is highlighted using the above example.

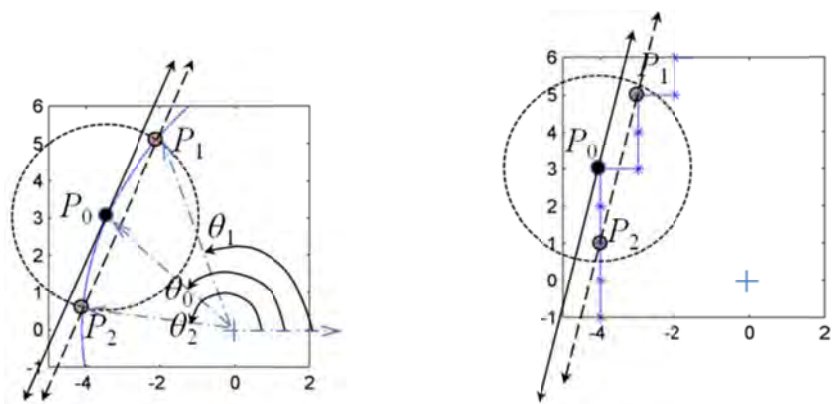

a) Pictorial illustration of the concept and definition of the points

(b) Pictorial illustration for the digitized curve

**Figure 3.2-5: Illustration of the concept for a smooth curve and the corresponding digitized curve.**

## 3.3 Definite error bounded (DEB) tangent estimation

### 3.3.1 Algorithm

Consider a smooth curve shown in Figure 3.2-5(a). Figure 3.2-5(b) is the digital analogue of Figure 3.2-5(a), which shall be used later, and is introduced here for the ease of comparison with the non-digitized case. For this section, Figure 3.2-5(a) is used to introduce the concept and should be referred in the context of the following discussion.

Suppose the tangent at the point $P_0\left(x_0, y_0\right)$ is to be computed, see Figure 3.2-5(a). In practice, since the curve to which $P_0$ belongs is not known, the tangent cannot be computed analytically. Consider a small circle of radius $R$ centered at $P_0$ specified by eqn. (3-15):

$$\left(x - x_0\right)^2 + \left(y - y_0\right)^2 = R^2 \qquad (3\text{-}15)$$

The circle intersects the curve at points $P_1$ and $P_2$, see Figure 3.2-5(a). There are three steps for finding the tangent at $P_0$:

1. find the slope of the line $P_1 P_2$ (denoted by $\tilde{m}$ )

2. find a line with slope $\tilde{m}$ passing through the point $P_0$. The idea is demonstrated in Figure 3.2-5(a). The slope $\tilde{m}$ of the line $P_1 P_2$ is given by eqn. (3-16):

$$\tilde{m} = \left(y_2 - y_1\right)\big/\left(x_2 - x_1\right) \qquad (3\text{-}16)$$

3. find the intercept $c$ : $c = y_0 - \tilde{m}x_0$ .

The equation of the line is then given by eqn. (3-17):

$$y = \tilde{m}x + c \qquad (3\text{-}17)$$

The pseudocode for the proposed tangent estimator is presented in Figure 3.3-1.

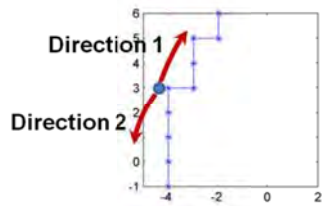

```
begin
p = 0;//finding P₁
    do {  p=p+1;
          current pixel=pᵗʰ pixel from
          P₀ in direction 1;
    }while
```
$$\left( \left| \text{distance of current pixel from } P_0 - R \right| < 1/\sqrt{2} \right)$$

```
    P₁=current pixel;

    p = 0;//finding P₂
    do {  p=p+1;
            current pixel=pᵗʰ pixel from
    P₀ in direction 2;
    }while
```
$$\left( \left| \text{distance of current pixel from } P_0 - R \right| < 1/\sqrt{2} \right)$$

```
    P₂=current pixel;

    m̃'=slope of the line P₁P₂ ;
```
$$\phi = \tan^{-1} \tilde{m}'$$
```
    return( m̃',φ )
end
```

/*This is the pseudocode for computing the tangent at a particular point $P_0$ on a digital curve. The inputs are the pixels of the digital curve, the point $P_0$, and the radius $R$. The output is the estimated slope $\tilde{m}'$ of the tangent and the angle $\phi$ that the tangent makes with the x-axis*/

**Figure 3.3-1: Pseudocode for computing the tangent**

### 3.3.2 Computation complexity

It is evident that the pseudocode in Figure 3.3-1 requires a maximum of $2 \times \text{ceil}(R)$ executions of the do-while loops, where $\text{ceil}(R)$ is the smallest integer larger than or equal to $R$. Considering the additional two steps of computing the slope and the polar angle, the computational complexity of the algorithms is $2 \times \text{ceil}(R) + 2$ computations. Since $R$ is a constant decided using eqn. (3-32) for an application, the time required for computing the tangent is a constant (time taken for $2 \times \text{ceil}(R) + 2$ computations). So the time complexity of the proposed tangent estimator is $O(1)$.

As evident, there are no shape fitting or optimization steps (as needed in most contemporary estimation methods). Thus, the computation complexity of the proposed algorithm is many magnitudes smaller than the other tangent estimation methods. In our knowledge, among other methods, implicit parabolic fitting [63, 64] has the lowest computational complexity, which is of the order $O(Q)$, where $Q$ is the parameters that determines the local region in the vicinity of the point of interest. Clearly, the proposed method has a very small computational complexity.

## 3.4 Error bound of the proposed tangent estimator for conic curves

This section presents the error bound of the DEB tangent estimator. First, the error bound of the proposed method for continuous conics is presented in section 3.4.1. For convenience, this error is called the analytical error. Second, the total error bound including the effect of digitization is presented in section 3.4.2. The error bounds are used to choose the value of the control parameter $R$ of the proposed method in section 3.4.3.

### 3.4.1 **Analytical error bound**

Consider a general conic equation [112] specified by eqn. (3-18):

$$x^2 + y^2 = e^2 \left( x + a \right)^2 \tag{3-18}$$

The above equation describes a conic of eccentricity $e$ with one focus at the origin, the second focus (if any) along the $x-$axis, and the directrix given by the equation $x = -a$, where $a$ is the distance between the first focus and the directrix (generally called the focal parameter).

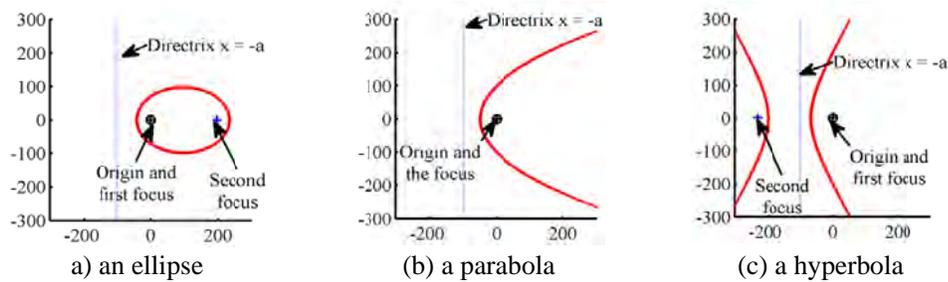

| a) an ellipse | (b) a parabola | (c) a hyperbola |

**Figure 3.4-1: Illustration of the conics, the directrix, and the foci.**

Figure 3.4-1 shows samples of conics represented by the above equation. For clarity the foci and the directrix are also shown. The above equation can be simplified by using the polar parametric representation of the coordinates in eqns. (3-19) and (3-20).

$$x = r\cos\theta, \; y = r\sin\theta \tag{3-19}$$

$$r(1 - e\cos\theta) = ae \tag{3-20}$$

The equation of the slope of the tangent at $P_0$ can be computed analytically using eqns. (3-18)and (3-20) as in eqn. (3-21).

$$m_0 = \frac{dy}{dx}\bigg|_{P_0} = e\csc\theta_0 - \cot\theta_0 \tag{3-21}$$

The points $P_0\left(r_0, \theta_0\right)$, $P_1\left(r_1, \theta_1\right)$, and $P_2\left(r_2, \theta_2\right)$ lie on the conic defined by eqn. (3-20), while the points $P_1$ and $P_2$ are also on the circle defined in eqn. (3-15). For convenience, $\theta_i = \theta_0 + \Delta\theta_i, i = 1$ to $2$ is substituted, where $\Delta\theta_i, i = 1$ to $2$ are the two solutions of the simultaneous equations (3-20) and (3-15). Equations (3-20) and (3-15) are solved simultaneously for $\Delta\theta_i, i = 1$ to $2$ in order to find the points $P_1$ and $P_2$ (details in Appendix B) and the solutions for $\Delta\theta_i, i = 1$ to $2$ are given in eqn. (3-22):

$$\Delta\theta_1 = D\left(dD - 1\right)\sum_{n=0}^{\infty}\left(dD\right)^{2n} \quad ; \quad \Delta\theta_2 = D\left(dD + 1\right)\sum_{n=0}^{\infty}\left(dD\right)^{2n} \tag{3-22}$$

where

$$D = \frac{\left(1 - e\cos\theta_0\right)^2 \left(R/ae\right)}{\sqrt{\left(e\sin\theta_0\right)^2 + \left(1 - e\cos\theta_0\right)^2}} \tag{3-23}$$

$$d = \frac{e\sin\theta_0}{\left(1 - e\cos\theta_0\right)} \tag{3-24}$$

Further, the slope $\tilde{m}$ of the estimated tangent given by (3-16) can be approximated as in eqn. (3-25)(see details in Appendix C):

$$\tilde{m} = m_0 - 0.5ed\, D^3 \csc\theta_0 + O\left(D^4\right) \tag{3-25}$$

For future reference, the maximum value of $D$ is noted in eqn. (3-26).

$$D_{\max} = \max(D) = \left(1 + \frac{1}{e}\right)\left(R/a\right) \tag{3-26}$$

It can be shown that $\max\left(D\right)$ occurs at $\theta_0 = \pi$ (i.e. $\sin\theta_0 = 0$ and $\cos\theta_0 = -1$). The derivation is presented in Appendix D. Thus using eqn. (3-25), $\tilde{m}$ converges to $m_0$, subject to the condition that $D_{max} \ll 1$. In eqn. (3-25), additional attention should be paid to two special cases: $\theta_0 \in \{0, \pi\}$, where $\csc \theta_0$ is singular. However, noting that $d \csc \theta_0$ is not singular, there is no extra singularity other than the singularity of the actual slope $m_0$. The angular error in the computation of the slope is given by eqn. (3-27).

$$\partial \phi = \left| \tan^{-1}(m_0) - \tan^{-1}(\tilde{m}) \right| = \left| \tan^{-1}\left( \frac{m_0 - \tilde{m}}{1 + m_0 \tilde{m}} \right) \right|$$

$$\approx \tan^{-1} \left| \frac{0.5 e d \, D^3 \csc \theta_0}{1 + m_0^2} \right| \approx \left| \frac{0.5 e d \, D^3 \csc \theta_0}{1 + m_0^2} \right| \tag{3-27}$$

Specifically, $\partial \phi = 0$ for circles ($e = 0$). Further the error in the computation of the tangent is bounded by $\left| \dfrac{0.5 e d \, D^3 \csc \theta_0}{1 + m_0^2} \right|$ and can be considered of order $O(D^3)$.

### 3.4.2 Total error bound including the effect of digitization

The effect of digitization on the DEB can be computed using the effect of digitization on the computation of slope of line passing through the points $P_1$ and $P_2$. The derivation of the error in computation of such a line has already been presented in section 2.4.1, though in the context of PA of digital curves [135]. The error bound in the computation of slope given by eqn. (2-36) is reproduced here in a simplified form for the ease of reference in eqn. (3-28) below.

$$\partial \tilde{\phi}_{max} = \max \left( \frac{1}{s^3} \left| \sin \tilde{\phi} \pm \cos \tilde{\phi} \right| \left| s^2 - s \left( \pm \cos \tilde{\phi} \pm \sin \tilde{\phi} \right) + \left( \pm \cos \tilde{\phi} \pm \sin \tilde{\phi} \right)^2 \right| \right) + O\left( s^{-2} \right) \tag{3-28}$$

where $s$ is given by (2-30) and $\tilde{\phi}$ corresponds to the slope $\tilde{m}$ of the line $P_1 P_2$ used in eqn. (3-27). Specifically, $\tilde{\phi}$ is given by eqn. (3-29).

$$\tilde{\phi} = \tan^{-1}\left( \tilde{m} \right) \tag{3-29}$$

The result in eqn. (3-28) proves that the error in the computation of the tangent converges even in the presence of digitization. The error due to digitization for various values of $s$ and $\tilde{\phi}$ is plotted in Figure 3.4-2. It can be seen clearly that small values of $s$ result in significant error while larger values of $s$ significantly reduce the error.

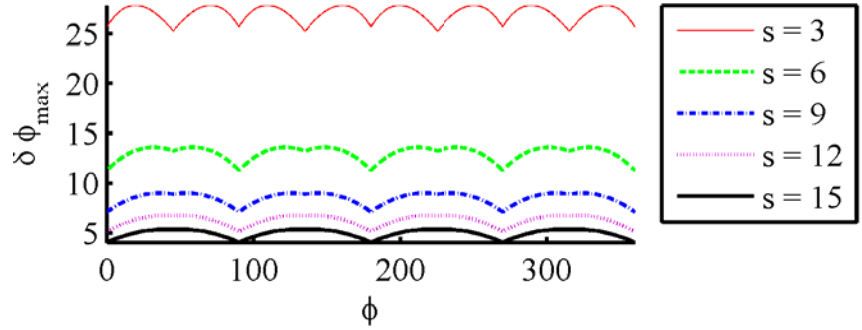

**Figure 3.4-2: Error $\partial\tilde{\phi}_{max}$ for various values of $s$.**

The total error in the computation of the tangent (including the analytical and digital effects) is given by eqn. (3-30):

$$\partial\phi_{max}^{tot} = \partial\phi + \partial\tilde{\phi}_{max} \qquad (3\text{-}30)$$

### 3.4.3 Control parameter R and multigrid performance

For the validity of the above analysis, it is required that $D_{max} \ll 1$. Accordingly, $R$ can be chosen using eqn. (3-31):

$$R = \frac{D_{tol}\,ae}{1+e} \qquad (3\text{-}31)$$

where $D_{tol}$ is chosen to be very small, $D_{tol} \ll 1$.

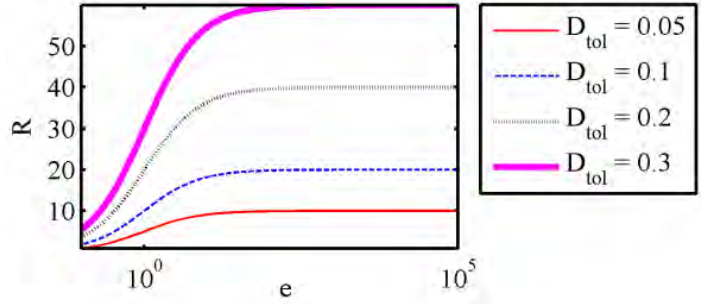

**Figure 3.4-3: Radii computed using (3-31) for different values of eccentricity and $D_{tol}$ where $a = 100$.**

The parameter $R$ for a given eccentricity and selected values of $D_{tol}$ is determined using eqn. (3-31) and is plotted in Figure 3.4-3 for $a = 100$, $e \in \left\{ 10^{-1}, 2 \times 10^{-1}, 3 \times 10^{-1}, \ldots, 10^{5} \right\}$ (corresponding to 10000 conics). The maximum value of $R$ (corresponding to $D_{tol} = 0.2$) is 60. However, the value of $R$ is a few pixels for small ellipses with low eccentricity.

Since the values or estimates of $a$ and $e$ are not generally available *apriori* in most practical scenarios, $R$ can be chosen using eqn. (3-32):

$$R \le D_{tol} \rho_{\min} \qquad (3\text{-}32)$$

where $\rho_{\min}$ is the radius of the smallest circle for which the tangent estimator has to be applied. It is important to consider the total error bound and the effect $R$ in the absence and presence of digitization. In the absence of digitization, the error bound is given by eqn. (3-27). Upon substitution of eqn. (3-31) in eqn. (3-27), it is seen that the error bound $\partial \phi$ is proportional to $R^3$. This implies that the smaller the value of $R$, the lesser is $\partial \phi$. In the case of digitization, the error bound is given by eqn. (3-28) is a decreasing function of $s$, which in turn is related to $R$. The value of $s$ is larger for larger values of $R$. This is illustrated using the family of conics considered above, for which the values of the minimum values of $s$ are plotted in Figure 3.4-4. It is seen that higher values of $R$ result in higher values of $s$ and consequently lower values of error due to digitization given by eqn. (3-28).

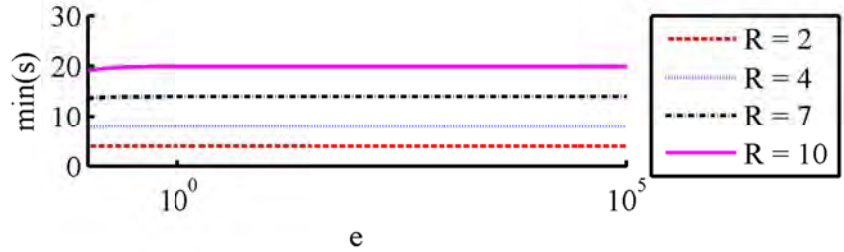

**Figure 3.4-4: Plot of min(s) for various values of $R$.**

In general, the error due to digitization $\partial\tilde{\phi}_{\max}$ decreases with increasing value of $R$, while the analytical error $\max\left(\partial\phi;\ \forall\theta_0\right)$ increases with $R$. Thus, in the case of digitization, it is preferable to use as large values of $R$ as possible, while satisfying eqn. (3-32). In the event of this conflicting influence of the parameter $R$, eqn. (3-32) serves as an important rule of thumb and the values closer to the upper limit given by eqn. (3-32) can be chosen. Our observation is that $D_{tol} \leq 0.5$ is sufficient for the analytical error $\max\left(\partial\phi;\ \forall\theta_0\right)$ to be reasonably small.

For the example of the family of conics considered above, the length of the semi-minor axis of the smallest ellipse is 20.1. Assuming that the smallest circle in the family has radius $\rho_{\min} = 20$ pixels, a suitable value of $R$ for $D_{tol} = 0.5$ is computed using eqn. (3-32) as $R = 10$.

Now the value of $R$ in terms of the multigrid parameter $h$ is considered. In multigrid analysis, the parameter $h$ determines the grid step size of an image. In other words, $h^{-1}$ is the total number of pixels in the image. In the analysis presented in section 3.4, the default value of $h$ is $h = 1$ owing to the digitization model given by eqn. (2-25). However, for a general case, the suitable value of $R$ can be given by modifying eqn. (3-32) as in eqn. (3-33):

$$R \leq D_{tol}\rho_{\min}h \qquad (3-33)$$

where, $\rho_{\min}$ is the radius of the smallest circle in pixels. Further, using eqns. (3-27) and (3-33), DEB is multigrid convergent of the order $\mathrm{O}(h^{-3})$.

**Rule of thumb:** In most images, it is reasonable to consider that the smallest circle may be of radius $\rho_{\min} = 5$ or 6 pixels, which implies that $R = 2.5$ or 3 may be used for estimating tangent. Nevertheless, if the estimated value of $\rho_{\min}$ is higher, it is recommended to choose the largest possible value of $R$ satisfying eqn. (3-32).

## 3.5 Numerical examples to illustrate the error bound

In this section, the analytical error bound, digital error bound, and the total error bound are studied for various families of conics and they are compared against the actual error in tangent estimation. It is shown that indeed the total error bound is the upper bound for a wide range of conics.

### 3.5.1.1 Family of conics

A family of conics given by $a = 200$, $e \in \left\{ 10^{-1}, 2 \times 10^{-1}, 3 \times 10^{-1}, \ldots, 10^{5} \right\}$ (i.e. 10000 conics of different eccentricities) is considered. This family encompasses ellipses of very low eccentricity to hyperbolae of very high eccentricity. This family was also used in Figure 3.4-3 and Figure 3.4-4. The effect of the value of $R$ on the analytical error bound ($\partial \phi$) can be seen in Figure 3.5-1, where the values of $\max \left( \partial \phi; \; \forall \theta_0 \right)$ using four fixed values of $R$ are plotted.

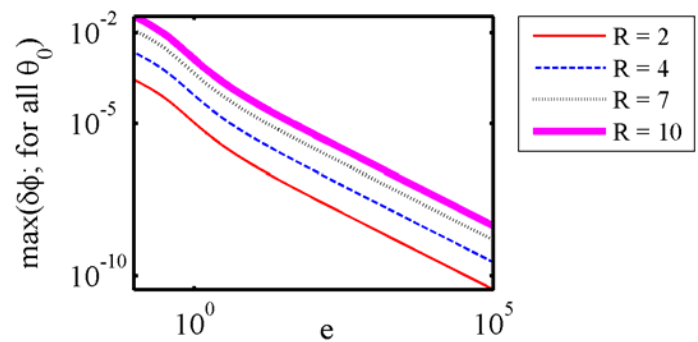

**Figure 3.5-1: Analytical error bounds for conics -** $\max \left( \partial \phi; \; \forall \theta_0 \right)$ **for various values of** $R$ **.**

Evidently, for a given value of $R$, the conics with lower eccentricity demonstrate maximum value of error $\partial \phi$. This validates the applicability of eqn. (3-32) for most practical purposes. Further, using the recommended value $D_{tol} = 0.5$, the suitable value of $R$ is computed using (3-32) as $R = 10$. It is seen in Figure 3.5-1 that the maximum analytical error for the value of $R = 10$ is $0.035°$.

Now, the digital error bound for the above considered family of conics is studied. For a given value of $R$, the two points $P_1, P_2$ and their corresponding digital pixels $P_1', P_2'$ are computed. These are used to compute $\tilde{m}$ and $\tilde{m}'$, using eqn. (3-16) and the slope of $P_1' P_2'$ respectively. Subsequently, the actual error $\max\left(\partial\tilde{\phi}\right)_{\forall \theta_0}$ due to digitization is computed for a family of conics and compared against $\partial\tilde{\phi}_{\max}$. For the family of conics considered above, the results are plotted in Figure 3.5-2. Figure 3.5-2(a) plots the digital error bound $\partial\tilde{\phi}_{\max}$ and the actual error due to digitization. It is noted that the actual error due to digitization is always less than $\partial\tilde{\phi}_{\max}$. Thus, it is verified that $\partial\tilde{\phi}_{\max}$ is indeed the upper bound of the error due to digitization.

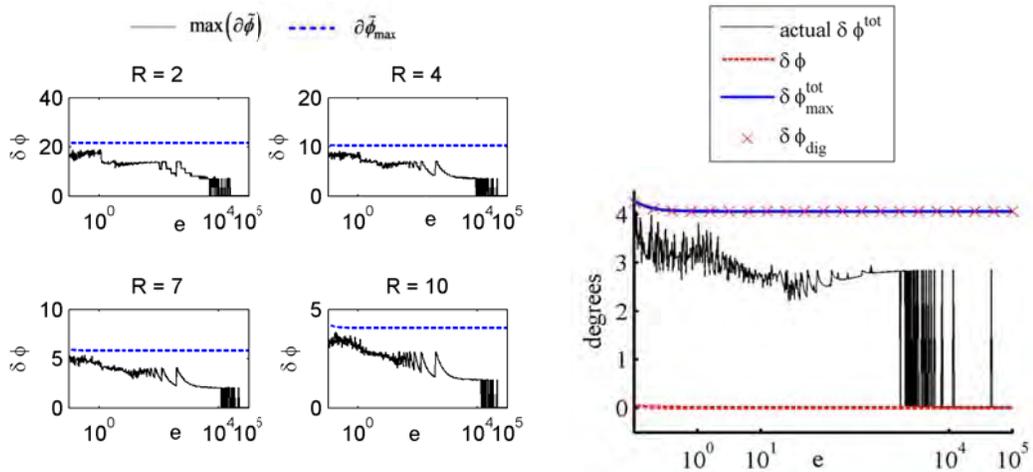

(a) Plots of $\partial\tilde{\phi}_{\max}$ and $\max(\partial\tilde{\phi})$ (in degrees) for various values of $R$.

(b) Plot of actual error, analytical error bound, digital error bound, and total bound for $R = 10$

**Figure 3.5-2: Error in the computation of the tangents due to digitization for section 3.5.1.1.**

Finally, the total error in tangent estimation by the proposed method is considered. The actual total error is computed as $\partial\phi^{\text{tot}} = \left|\tan^{-1} m_0 - \tan^{-1}\tilde{m}'\right|$ and is used to find the value of $\max\left(\partial\phi^{\text{tot}}\right)$. The values of $\max\left(\partial\phi^{\text{tot}}\right)$ for $R = 10$ are plotted in Figure 3.5-2(b). The total error bound $\partial\phi^{\text{tot}}_{\max}$ computed using eqn. (3-30), the analytical error bound $\max\left(\partial\phi; \forall \theta_0\right)$, and the digital error bound $\partial\tilde{\phi}_{\max}$ are also plotted in Figure 3.5-2(b). It is noted that $\partial\tilde{\phi}_{\max}$ is very close to $\partial\phi^{\text{tot}}_{\max}$, due to which the plot of $\partial\tilde{\phi}_{\max}$ is hardly visible in Figure 3.5-2(b). This means that the error due to digitization is the main contributor. Further, the actual maximum error in the computation $\max\left(\partial\phi^{\text{tot}}\right)$ of the ellipses is always less than $\partial\phi_{\max}^{\text{tot}}$.

### 3.5.1.2 Family of parabolae

Now, consider a family of parabolas (i.e. $e = 1$) with $a \in [30, \ 500]$. Analytical error bound for each value of $\theta_0$ for a given value of $R$ is computed using eqn. (3-27). The computed maximum angular errors are plotted in Figure 3.5-3. The maximum error in tangent estimation $\max\left(\partial\phi; \ \forall\theta_0\right)$, corresponding to $R = 10$, is $0.3913°$.

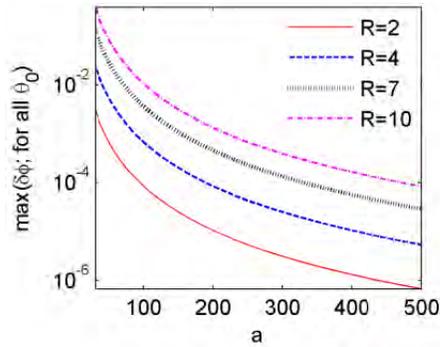

**Figure 3.5-3: Analytical error bounds for family of parabola.**

( $\max\left(\partial\phi; \ \forall\theta_0\right)$ **for various values of** $R$ ).

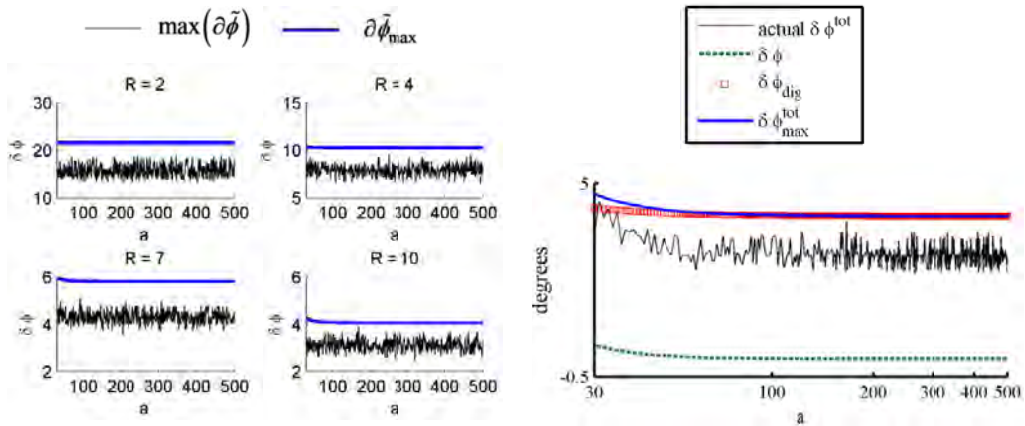

(a) Plots of $\partial\tilde{\phi}_{\max}$ and $\max(\partial\tilde{\phi})$ (in degrees) for various values of $R$.

(b) Plot of actual error, analytical error bound, digital error bound, and total bound for $R = 10$

**Figure 3.5-4: Error in the computation of the tangents due to digitization for section 3.5.1.2.**

The actual error $\max\left(\partial\tilde{\phi}\right)_{\forall\theta_0}$ due to digitization is computed for a family of parabolae and compared against $\partial\tilde{\phi}_{\max}$ in Figure 3.5-4(a). It is noted that the actual error due to digitization is always less than $\partial\tilde{\phi}_{\max}$. The actual total error in the computation of tangent $\max\left(\partial\phi^{\text{tot}}\right)$ is also smaller than the total error bound $\partial\phi^{\text{tot}}_{\max}$, as shown in Figure 3.5-4(b). It is highlighted that for parabola with small values of $a$, the analytical error bound $\max\left(\partial\phi;\ \forall\theta_0\right)$ is non-negligible.

### 3.5.1.3 Family of circles

It was discussed in section 3.4.1, just after eqn. (3-27), that the analytical error $\partial\phi = 0$ for circles. However, the error due to digitization $\partial\tilde{\phi}_{\max}$ is non-zero for circles. Thus, in this section, a family of circles is considered. The family contains circles with radii $\rho \in 10^z; z \in \{1.3, 1.31, 1.32, \ldots, 5\}$, corresponding to 371 circles among which the smallest circle is of radius 19.95 and the largest circle of radius $10^5$. Thus, this family contains very small circles as well as very large circles.

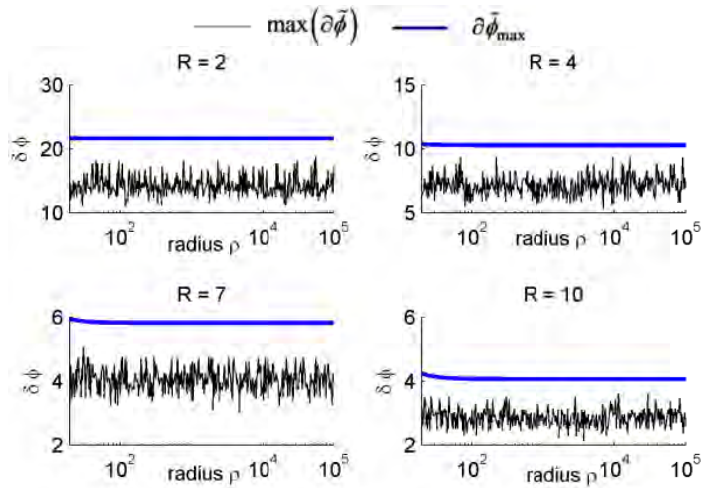

**Figure 3.5-5: Error in the computation of the tangents due to digitization for the family of circles.**

(plots of $\partial\tilde{\phi}_{\max}$ and $\max\left(\partial\tilde{\phi}\right)$ (in degrees) for various values of $R$)

The results for this family are plotted in Figure 3.5-5. It is noted again that the actual errors $\max\left(\partial\tilde{\phi}\right)_{\forall\theta_0}$ are always less than the error bound $\partial\tilde{\phi}_{\max}$. Also, the error does not change much with the size of ellipses. This is because $\min(s)$, which is the main contributor in $\partial \tilde{\phi}_{max}$, is strongly related to the value of $R$, and not $\rho$ (until $\rho$ is very small).

## 3.6 Comparison of DEB with other tangent estimators

### 3.6.1 Algorithms used for comparison

The summary of the existing tangent estimators was provided in [20]. Based on the study performed in [20], the performance of the proposed Definite Error Bounded (DEB) tangent estimator is compared with the following tangent estimators (the codes of all of which have been developed by the author in Matlab 2010):

1. Linear regression – order 1 to order 5 (LR1-LR5): This involves fitting an equation of order $N$ on the coordinates of $2Q+1$ pixels in the neighborhood of the point of interest.

2. Explicit parabola fitting (EPF): This involves fitting a parabolic equation on the coordinates of $2Q+1$ pixels in the neighborhood of the point of interest.

3. Implicit parabola fitting (IPF) [63, 64]: This is very similar to EPF with one difference that the coordinates are translated to a new coordinate system such that the point of interest is the new origin. The analytical solutions of the parabolic equation can then be directly computed.

4. Independent coordinate IPF (ICIPF) [63]: In this method, the coordinates of $2Q+1$ pixels in the neighborhood of the point of interest are represented as two independent parabolic functions of a fictitious parameter, say $l$, where the parabolic equations are determined using IPF. The derivatives for both $x$ and $y$ coordinates are computed with respect to the parameter $l$, which are subsequently used for computing the tangent.

5. Gaussian derivative (GD) [27, 44]: In this method, three variables, viz. (1) $x$ coordinates of $2Q+1$ pixels in the neighborhood of the point of interest, (2) y coordinates of $2Q+1$ pixels in the neighborhood of the point of interest, and (3) the pixel number $q = -Q$ to $Q$, are considered. Two functions in the space of $q$ are defined, which are the convolution of $x(q)$ and $y(q)$ variables with the Gaussian derivative function of $q$. The tangent is then defined as the ratio of the convolved $y$ function to the convolved $x$ function. It is essentially similar to linear Gaussian filter [172] and is based on Gaussian smoothing of curves.

6. Median method by Matas (Matas) [66]: In this method, the angles of the slopes of the lines connecting $2Q$ pixels in the neighborhood of the point of interest (excluding the point of interest) to the point of interest are computed. The median value of these angles is used as the estimated tangent's angle.

7. λMSG and λMST [21, 22]: In all the above methods, the parameter $Q$ is the control parameter and often determines the performance of the method. A parameter independent method based on maximal segments was proposed in [21, 22]. In this method, a pencil of maximal segments is found for the point of interest. A weighted sum of the slopes of the segments in this pencil is then taken as the estimate of the tangent, where the weights are computed using a chosen function. For convenience, the method is called λMSG if Gaussian curve is used to determine the weights and λMST if triangular curve is used to determine the weights. It was noticed that λMST generates huge errors in the computation of tangents for certain situations because it forces the weights of the segments at the extreme ends of the pencil to zero. Thus, a modified tangent function was used, in which the tangent function is elevated by 0.4, such that the floor of the tangent function is at 0.4.

8. Hybrid methods [20]: It has already been mentioned that methods 1-6 are dependent upon the parameter $Q$. In order to make them parameter free, the theory of maximal segments was used and many hybrid ways of determining the parameter $Q$ adaptively and independently for each point of interest were proposed in [20]. Six hybrid ways of determining the parameter $Q$ were proposed and are referred to as 10, 11, 12, 2, 3, 4 in [20]. The details are avoided for the sake of brevity and unnecessary diversion. The method used for computing the tangents is used as prefix. For example, EPF(01) implies that EPF was used at the core of tangent estimation and the hybrid way 01 was used for determining the parameter $Q$.

Summarizing the methods used for comparison, 10 parameter dependent methods (LR1-LR5, EPF, IPF, ICIPF, GD, Matas), 60 hybrid methods (6 hybrid ways for each of the 10 parameter dependent methods), and 2 parameter independent methods (λMSG and λMST) were used for comparison against the proposed method (DEB). Since it is difficult to represent the comparison of all the 72 methods together in a figure, the most representative methods are selectively presented in all the figures, beginning from Figure 3.6-1. By saying most representative methods, it is meant that a few methods that give the best performance are chosen. Also, if a hybrid method [20] gives better performance than the original method and other hybrid methods, it is preferred over all other variants of the same method. For example, if EPF(3) performs better than EPF, EPF(10,11,2,4,5), it is chosen among all of them as the representative of EPF.

### 3.6.2 Setup for comparison

For comparison, only closed curves are used. The curves are in a square digital image space of $300 \times 300$ pixels. For generating digital curves, the curve is first generated using the continuous function of the curve where the center of the curve is randomly chosen within two pixel region of the center of the image space. This continuous curve is then digitized. For each geometry, 100 such random digital curves are generated.

For each of the 100 curves for a geometry, the data of $\partial \phi$ is computed for every pixel on the closed curve, such that almost all angles (with very small angular difference between them) in the range $[0, 360]$ are considered. Since there are 100 curves (very slightly different from each other due to randomly chosen centers within two pixel region), one-to-one correlation between the pixels and angles is not present. So, for each curve, $\partial \phi$ is interpolated over the range $[0, 360]$ with uniform interval of $0.5$ degrees. Thus, for each geometry, one value of $\partial \phi$ for angles $\theta = \{0, 0.5, 1, \cdots, 360\}$ degrees is obtained. As relevant, average and maximum values are computed over $\theta = \{0, 0.5, 1, \cdots, 360\}$.

Three experiments are performed. In the first experiment, circular geometry of radius 100 is considered. In the second experiment, ellipses of various eccentricities are considered. In the third experiment, non-conic shapes with inflexion points are considered. The details of each experiment and the results are discussed in subsequent sub-sections.

### 3.6.3 **Experiment with circular geometry**

In this experiment, 100 circles of radius 100 are generated, and the coordinates of the centers are randomly chosen from the range $[149,151]$, where the point $(150,150)$ is the center of the digital image space of size $300 \times 300$ pixels. The actual envelope is shown in the Figure 3.6-1(a) and the 100 randomly chosen centers are shown in Figure 3.6-1(b). In Figure 3.6-1(c), the average value of the error in tangent estimation $\partial \phi$ is presented as a function of the control parameter for DEB and the parameter dependent methods (1-6) of the list in section 3.6.1.

First, the performance of the proposed method (DEB) is discussed. It is seen that as the value of $R$ increases, the average error monotonically reduces. Since the analytical error in tangent estimation is zero for the circles, the digitization error is the only contributor of the error. As discussed in section 3.4.3, the value of $s$ increases with the values of $R$, and as a consequence, the error in tangent estimation decreases. This explains the monotonic decrease in the $\partial \phi$ as the value of $R$ increases.

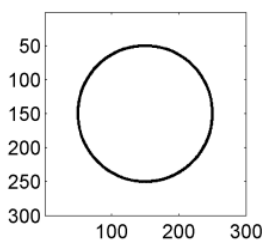

(a) The envelope of the circles used in section 3.6.3

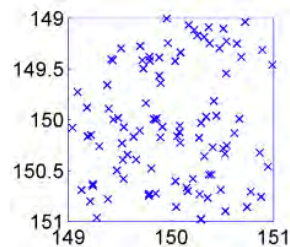

(b) The box shows the region from which the centers are randomly chosen and the cross 'x' show the 100 randomly selected centers

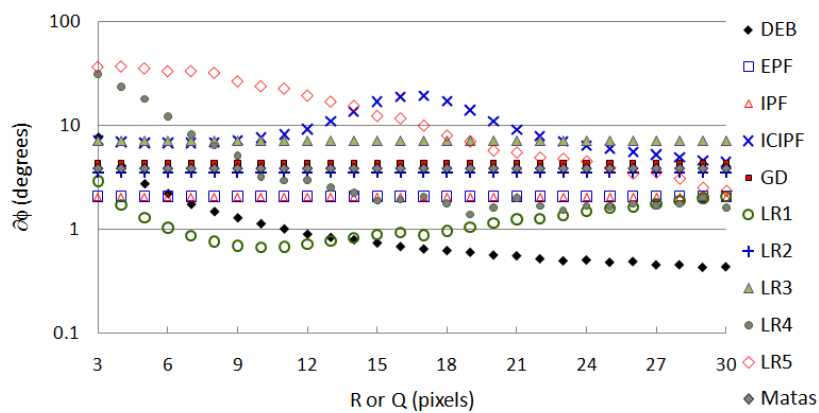

(c) Average error in estimation of tangents for 100 digital circles for various parameter dependent methods. The average error for the complete family circles is plotted as a function of the control parameter (R for the proposed method and Q for methods 1-6 of section 3.6.1).

**Figure 3.6-1: The experiment with circles of radius 100 and randomly chosen centers (section 3.6.3)**

Now, the performance of DEB against other parameter dependent methods is compared. It is clearly evident that for $R \geq 14$, DEB performs better than any other parameter dependent method. For smaller values of $R$, LR1 performs better than DEB for $R$ and $Q < 14$. This can be explained using the fact that the circles have large radius in comparison to the edge segments considered using $2Q+1$ pixels in the neighborhood of any pixel of interest. Thus, the first order linear regression also performs well in this case. Further, it is noted that for $R$ and $Q \leq 6$, EPF and IPF also perform better than DEB[2].

---

[2] This can be explained as follows. First make note of the fact that the error of EPF and IPF is almost invariant of the value of $Q$. This is because in EPF and IPF, though least squares fitting of parabola may result in large errors as

Figure 3.6-2 summarizes the average error and maximum error of 48 various algorithms. Here, the average and maximum error for most algorithms and their hybrids are provided (a total of 48 methods). Hybrids of LR2-LR5 are avoided for brevity. For all the parameter dependent methods, the value of $R$ or $Q$ for which the average error in Figure 3.6-1(c) is minimum[3] is considered. The results clearly demonstrate that DEB has lowest value of average and maximum errors. In terms of average error, GD(3) is the closest competitor, though it performs poorly for maximum error. In terms of maximum error, EPF(3) and IPF(3) and GD(2) are the closest competitors.

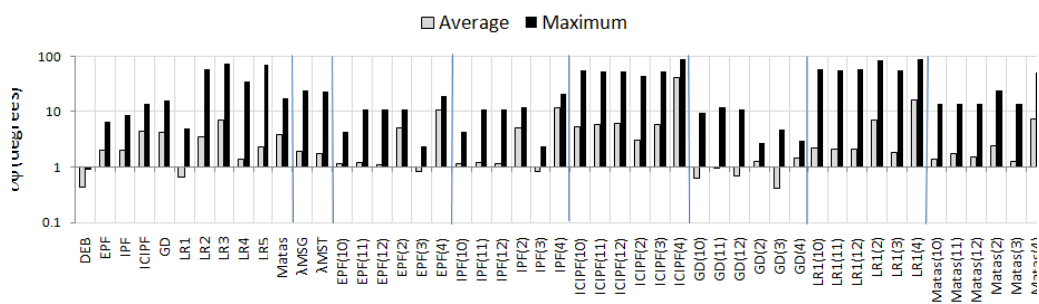

**Figure 3.6-2: Summary of results for the experiment tangent estimation for digital circles.**

### 3.6.4 Experiment with elliptic geometry

In this experiment, 10000 ellipses are generated and divided into $g = 1$ to $100$ groups, each group $g$ containing $n = 1$ to $100$ digital ellipses. The length of the semi-major axis of all the ellipses is fixed, i.e. $A = 100$ (note that this is different from the focal parameter $a$). The eccentricity of the ellipses within one group is fixed. i.e. $e_{g,n} = 0.01(g-1); \forall n$. However, within one group, the 100 ellipses have 100 different randomly chosen centers, where the centers are randomly chosen from the range $[149, 151]$ and the point $(150,150)$ is the center of the digital image space of size $300 \times 300$ pixels.

---

the value of $Q$ increases, the fit has the least residue in the close proximity of the point of interest and thus, the computed tangent has a reasonably small value of error.

[3] The hybrid versions of LR2-LR5 are skipped because of their non-relevance for this curve, since the circle is a quadratic curve. While LR2, EPF, and IPF, all require quadratic curve fitting, EPF and IPF clearly perform better than LR2 and are thus sufficient in representing second order curve fitting. Thus, LR2 has been omitted.

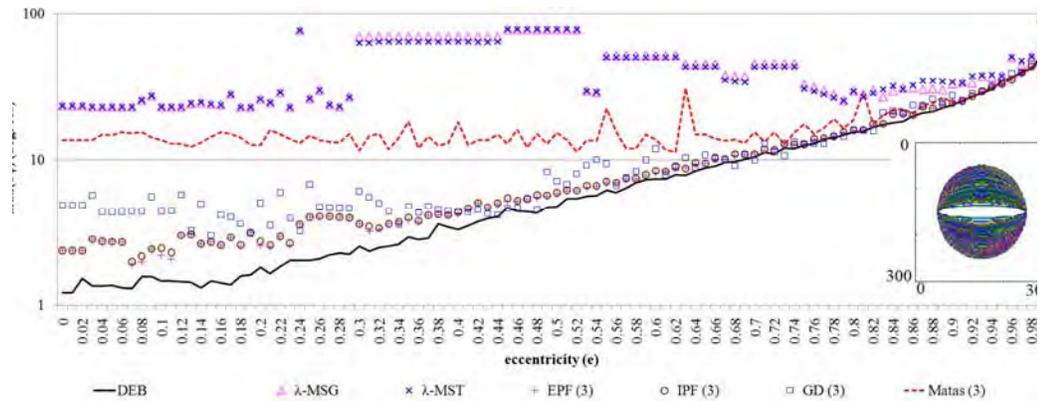

**Figure 3.6-3: Maximum error in TE using various algorithms.**

The envelope of the digital ellipses for all the groups is illustrated in the inset of Figure 3.6-3. In Figure 3.6-3, the error in TE is plotted against eccentricities of the ellipses for the experiment in section 3.6.4. The envelope of the ellipses is shown in the inset. As the eccentricity increases, the effect of digitization is expected to be severe for the left and right portions of the curves, as major curvature changes occur over a small digital portion of the image. On the other hand, for the bottom and top portions, the curvature hardly changes over a long portion of the curve for ellipses with high eccentricity and the error in computation of the tangent is expected to be low. See the most elliptic ellipse in the inset of Figure 3.6-3, corresponding to $e = 0.99$ in context of the preceding discussion. Thus, for elliptic curves, the average error (averaged over all the angles) in the computation of the tangents is not a good measure of the quality of tangent computation. So, the maximum error over all the angles is used as the most representative parameter of the error in tangent computation. The maximum error for each ellipse in a group is computed and then averaged over the group. Thus, a maximum error for one value of eccentricity is obtained.

The maximum errors in tangent computation using various algorithms are plotted against the eccentricity of the ellipses in Figure 3.6-3. First it is noted that the proposed method (DEB, $R = 10$) performs better than all the methods for almost all the values of eccentricity. For low values of eccentricity, in general EPF(3) and IPF(3) are the next best competitors. However, for high eccentricities, GD(3) is also a good competitor. It is also noted that λMSG and λMST give reasonable error in tangent computation for low eccentricities $e < 0.3$. On the other hand, their errors become very high for high eccentricity ellipses, bordering close to 90 degrees for many values of eccentricities. In our opinion, this is because in the current forms as reported in [20], both λMSG and λMST are incapable of handling cases with such eccentricity. It is a reasonable expectation that a more sophisticated design of the shape function used in λMSG and λMST, this effect can be mitigated to a certain extent. However, this is beyond the scope of the present work.

### 3.6.5 Experiment with non-conic shapes containing inflexion points

In this experiment, a family of non-conic curves given by eqn. (3-34) is considered.

$$r = \frac{R_{out}\left(1 - b\sin\left(n\theta\right)\right)}{1 + b} \quad ; \quad x = r\cos\theta + x_0; \quad y = r\sin\theta + y_0, \qquad (3\text{-}34)$$

where $R_{out}$ is the radius of the smallest circle (centered at origin) encompassing the entire geometry and the value of $b$ determines the largest inner circle (centered at the origin) that touches the inflexion points, $R_{in}/R_{out} = \left(1 - b\right)/(1 + b)$. Also, the value of $n$ determines the actual shape and also represents the number of inflexion points. Here, $R_{out} = 100$ and $b = 0.5$ are used. The example curves for $n = 1$ to $7$ are shown in Figure 3.6-4. Hundred such digital curves are generated for each value of $n$, $n = 1$ to $7$ with the center coordinates $x_0$ and $y_0$ chosen randomly from the range $\left[149, 151\right]$.

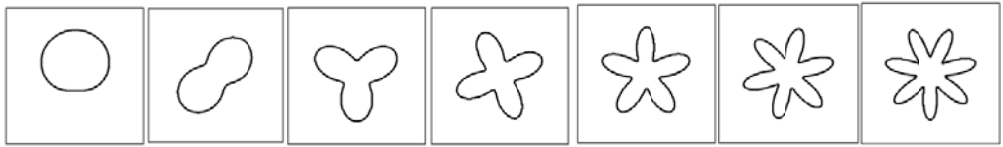

**Figure 3.6-4: Geometry given by (3-34) for the values** $n = 1$ to $7$ **.**

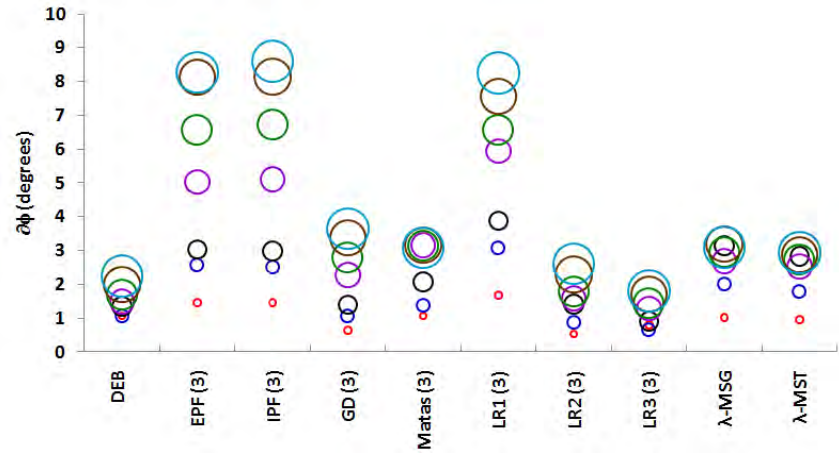

**Figure 3.6-5: Average values of errors for various algorithms for non-conic digital curves.**

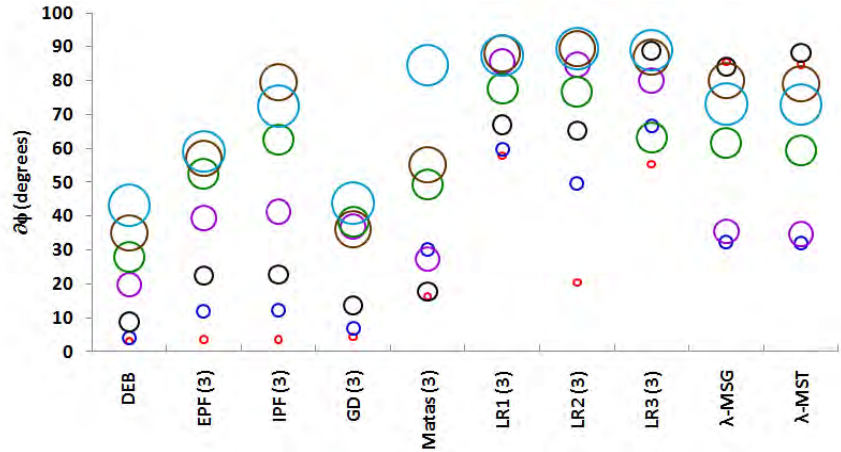

**Figure 3.6-6: Maximum values of errors for various algorithms for non-conic digital curves.**

For various algorithms, the mean and maximum are presented in Figure 3.6-5 and Figure 3.6-6, respectively. The size of the circles indicate the value of $n$. Smallest circle corresponds to $n = 1$, while the largest circle corresponds to $n = 7$ for each algorithm. $R = 10$ has been used for the proposed method, DEB. In terms of the mean error, LR3(3) gives the best performance. This is because the complicated nature of curve can be better represented locally using a higher order curve than the second order curve[4]. Nevertheless, the performance of DEB comes really close to the performance of LR3(3). On the other hand, DEB clearly outperforms all the algorithms in terms of the maximum error, demonstrating consistent good performance for non-conic curves as well.

## 3.7 Conclusion

A simple definite error bound (DEB) TE for digital curves is proposed. Explicit error bounds have been analytically derived for all kinds of conics. The upper bounds have been derived for both continuous and digitized conic curves. Extensive numerical experiments in section 3.5 confirm the error bounds as being the upper bounds. In addition, since the performance and upper bounds of the algorithm are verified and shown to be small for very small as well as very large geometries (based on the value of focal parameter $a$), the results inherently also demonstrate the multigrid performance of the proposed algorithm (DEB). Owing to the simplicity of the proposed method, the algorithm is computationally inexpensive. The performance of DEB is compared against 72 contemporary methods of tangent estimation for various geometries, which include a large circle, ellipses of various eccentricities, and non-conic curves with inflexion points. Using the example of circle as the standard, it is shown that the performance of DEB is superior to all the methods. Since 100 digital circles and all possible angles are considered, the results also demonstrate the isotropicity property as discussed in [20]. DEB also gives the lowest maximum error in tangent computation for ellipses of almost all values of eccentricities. Finally, DEB demonstrates reasonable performance and the lowest value of maximum error in tangent computation for non-conic shapes with inflexion points as well. In conclusion, the proposed tangent estimator has been extensively studied, in terms of error bounds, multigrid performance, isotropicity, and statistical averages and maximum values of error in tangent computation for conic (continuous as well as digitized) curves. In addition, its utility as a generic tangent estimator is also clearly demonstrated, though the mathematical definite error bound may be difficult to compute for generic curves. The content of this chapter have been reported in [31, 69].

---

[4] On the other hand, curves of order 4, 5 or higher are too complicated for local fitting and the value of $Q$ often results into underfitting of the high order curves in the local regions. For other families of curves, this effect may vary.

# Chapter 4 : Least squares fitting of ellipses

## 4.1 Background

The least squares methods currently in use are based on the fundamental work by Rosin [82, 83, 86-88] and Fitzgibbon [80]. They both employ the algebraic equation of general conics to define the minimization problem and additional numeric constraints are introduced in order to restrict the solutions to the elliptic curves. In other works [86-89], Rosin developed and tested several error metrics for quantifying the quality of fit. Fitzgibbon [80] solves the constrained minimization problem using generalized eigenvalue decomposition.

For the applications that require either ellipse fitting or ellipse detection, Fitzgibbon's method is very popular. The reason behind its popularity is the fact that it is non-iterative, fast, linear time algorithm with robust performance against noisy data. Although Fitzgibbon's method [80] is quite elegant and has several merits, some issues with Fitzgibbon's method [80] have been recognized [32, 173-175]. For example, Maini [173] recognized that Fitzgibbon's method [80] has the problem of numerical instability, ill-posedness of the scatter matrix, and the singularity of the constraint matrix. Further, if the data points are exactly on the ellipse, all generalized eigenvalues are zero and do not result in any solution at all. Maini [173] suggests a modification of Fitzgibbon's method [80] which include the translation and scaling of the data points, followed by an iterative optimization where if Fitzgibbon's algorithm suffers from the zero eigenvalue problem, some random noise is added to the data points. Maini therefore improves Fitzgibbon's method of fitting ellipses for non-noisy as well as noisy set of elliptic data points.

Liu [32] on the other hand aims at using Fitzgibbon's method as an initial guess and applies the gradient descent on one point at a time. As a consequence, the ellipse taken as the initial guess gets optimized for the data points in the sequence they occur. Thus, the final result of Liu is biased towards the last few data points in the sequence. Here, the reference is only for the least squares and gradient descent based elliptic parameter extraction done by Liu in [32] and not to other pre-processing and post-processing steps discussed therein.

Another approach taken by Harker [174] and Halir [175] (Harker extend the work of Halir), where the scatter matrix was partitioned into two smaller interdependent matrices and the constraint matrix was changed to a smaller non-singular constraint matrix. Harker and Halir addressed the low eccentricity bias in Fitzgibbon's method and proposed a bias correction technique for fitting ellipses of high eccentricity. Harker does improve the method of Fitzgibbon. However, due to the bias correction in which Harker use a linear combination of an elliptic and a hyperbolic solution, its selectivity for elliptic shapes is poor.

Although Maini [173] mentioned cursorily about the instability of Fitzgibbon's method and ill-conditioned nature of the scatter matrix $\mathbf{S}$ (defined later in eqn. (4-6)), detailed discussion about this topic has not been presented in the literature. Further, though Maini suggests a translation of the data points to the center of the bounding box of the data points. A more rigorous numerical analysis of the scatter matrix and a more suitable translation of the data points are proposed in section 4.2. The modified Fitzgibbon's method is called numerically stable algebraic fitting (NSAF) method.

All the methods discussed above employ algebraic equation of the ellipse and the algebraic distance of the points on the ellipse as the cost function. As opposed to these, the Ahn's method [90] uses the geometric distances of the data points from the ellipse as the central quantity for fitting the ellipse. Ahn's method uses two nested iterative loops. The outer loop considers the data of pixels as a whole and uses gradient descent to optimize the estimated geometric parameters (namely, coordinates of the center, angle made by the major axis with $x$- axis, and the lengths of semi-major and semi-minor axes). The inner loop is executed for each pixel and the estimation of a point on the ellipse that is nearest to the considered pixel is optimized iteratively. Ahn's method uses geometric parameters as the driving factors of the algorithm in both the loops. Hence, it is not incorrect to say that the distance of the pixels from the ellipse forms the central concept of the method. Ahn's method also employs a bias correction technique which considers a linear combination of an elliptic and a hyperbolic solution that is further iteratively optimized. Ahn's algorithm is expected to perform better than algebraic minimization methods due to the use of geometric distance as the main criterion. However, there are two issues with Ahn's method. Both the issues are related to the optimization process in Ahn's algorithm. First, Ahn's method is computationally expensive owing to the iterative optimization for each pixel within the outer loop. Second, the problem of local minima is present because non-linear optimizations are involved in both inner and outer loops. It is seen in the numerical simulations that Ahn's algorithm has tendency to get stuck in the local minimum during optimization in both the inner and outer loops.

The concept of the geometric distance of a data point from an ellipse is presented in section 4.3. A new least squares ellipse fitting (ElliFit) method that is based on geometric distance minimization and does not require constrained optimization is proposed in section 4.4. As opposed to other ellipse fitting methods that invariably use constrained optimization, unconstrained optimization is sufficient for ElliFit because it uses a set of unconventional variables. These are related to the actual parameters of ellipses in a non-linear manner. Thus, the constraints are directly incorporated in the definition of the new variables. Due to this, ElliFit does not need additional constraints or non-linear optimization and still demonstrates high selectivity for elliptic curves. Rather than designing a least squares formulation using a general quadratic equation and satisfying certain constraints, the geometric model of ellipse can be used as the basic model and the distance of the pixels from the fitted ellipse as the criteria for designing the least squares formulation.

The outline and contributions of this chapter are presented here. The conventionally used algebraic fitting of ellipses is presented in section 4.2. Specifically, the popular method of Fitzgibbon [80] is considered in this section and the problem of numerical instability of this method is discussed. Based on the discussion, a simple solution for dealing with this numerical instability is proposed in section 4.2.2.

The concept of geometric distance as a minimization goal in least squares fitting is discussed in section 4.3. The geometric equation of a general ellipse, its simplified form, and the distance of a data point from the ellipse to be fit is presented here. These expressions are useful for the theory presented in section 4.4.

Section 4.4 is the highlight and novelty of this chapter in which the novel ElliFit method is presented. A modification of the minimization function (the geometric distance) is presented in section 4.4.1. The mathematical model of the method is presented in section 4.4.2. The non-linear operator, its injective mapping, and inversion are presented in section 4.4.3. The numerical stabilization of the linear operator is discussed in section 4.4.4. Finally the computational complexity of the proposed method is presented in section 4.4.5. Several numerical simulations and comparison with other least squares methods are presented in section 4.5. The proposed method performs better than the other least squares method for variety of cases like digital incomplete curves, noisy clusters of data points, as well as images with several incomplete ellipses. The proposed method is also tested for non-elliptic conics and non-conics as well and it is seen that the proposed method has lower false positives as compared to other least squares based methods. The chapter is concluded in section 4.6.

## 4.2 Algebraic distance minimization and Fitzgibbon's method

### 4.2.1 Introduction to the method

A set of data points $\left\{ P_i\left(x_i, y_i\right); i = 1 \text{ to } N \right\}$ on which an ellipse has to be fitted are considered. Considering a general quadratic equation given in eqn. (4-1):

$$ax^2 + bxy + cy^2 + dx + fy + g = 0 \tag{4-1}$$

In the least squares approach, the aim is to find the coefficients $(a, b, c, d, f, g)$ such that the residue in eqn. (4-2) is minimized and the constraint in eqn. (4-3) is satisfied.

$$r = \sum_{i=1}^{N} \left( ax_i^2 + bx_i y_i + cy_i^2 + dx_i + fy_i + g \right)^2 \tag{4-2}$$

$$b^2 - 4ac < 0 \tag{4-3}$$

The constraint in eqn. (4-3) should be satisfied in order to restrict the solutions to ellipses. Further in general, $\left| ax_i^2 + bx_i y_i + cy_i^2 + dx_i + fy_i + g \right|$ represents the algebraic distance of a data point $P_i\left(x_i, y_i\right)$ from the conic in eqn. (4-1). Thus, the methods in which the residue term to be minimized is similar to the term in eqn. (4-2) are referred to as the algebraic distance minimization methods.

As discussed above, Fitzgibbon [80] is such a method and it is elaborated here. In order to solve this problem the matrices and vectors in eqns. (4-4) - (4-7) are defined.

$$\overline{A} = \begin{bmatrix} a & b & c & d & f & g \end{bmatrix}^T \tag{4-4}$$

$$\mathbf{D} = \begin{bmatrix} \vdots & \vdots & \vdots & \vdots & \vdots & \vdots \\ x_i^2 & x_i y_i & y_i^2 & x_i & y_i & 1 \\ \vdots & \vdots & \vdots & \vdots & \vdots & \vdots \end{bmatrix} \tag{4-5}$$

$$\mathbf{S} = \mathbf{D}^T \mathbf{D} \tag{4-6}$$

$$\mathbf{C} = \begin{bmatrix} 0 & 0 & 2 & 0 & 0 & 0 \\ 0 & -1 & 0 & 0 & 0 & 0 \\ 2 & 0 & 0 & 0 & 0 & 0 \\ 0 & 0 & 0 & 0 & 0 & 0 \\ 0 & 0 & 0 & 0 & 0 & 0 \\ 0 & 0 & 0 & 0 & 0 & 0 \end{bmatrix} \tag{4-7}$$

For convenience, $\overline{A}$, $\mathbf{D}$, $\mathbf{S}$, and $\mathbf{C}$ are referred to as the coefficient vector, data matrix, scatter matrix, and the constraint matrix respectively. Eqns. (4-2) and (4-3) can then be rewritten as in eqns. (4-8) and (4-9), respectively.

$$r = \overline{A}^T \mathbf{D}^T \mathbf{D} \overline{A} = \overline{A}^T \mathbf{S} \overline{A} \tag{4-8}$$

$$\overline{A}^T \mathbf{C} \overline{A} > 0 \tag{4-9}$$

Eqns. (4-8) and (4-9) can then be cast as a set of simultaneous equations in (4-10).

$$\mathbf{S} \overline{A} = \lambda \mathbf{C} \overline{A} \quad ; \quad \overline{A}^T \mathbf{C} \overline{A} = 1 \tag{4-10}$$

Since the constraint matrix $\mathbf{C}$ has only one positive eigenvalue, the generalized eigenvalues $\overline{u}_j$ and eigenvectors $\lambda_j$ ( $j = 1 \text{ to } 6$ ) for the pair of matrices $\mathbf{S}$ and $\mathbf{C}$ are computed and the solution of $\overline{A}$ is given as in eqn. (4-11).

$$\overline{A} = \overline{u}_j (\lambda_j > 0) \tag{4-11}$$

### 4.2.2 Numerically stable algebraic fitting (NSAF)

It is noted that all the components of the matrix $\mathbf{S}$ are $N$ times the raw statistical moments $m_{p,q}(x_i, y_i)$ of the bivariates $(x_i, y_i)$ given by eqn. (1). The Bauer Skeel condition number of the scatter matrix $\mathbf{S}$ defined in eqn. (4-6) has the strongest contribution from either $\sum_{i=1}^{N} x_i^4$, $\sum_{i=1}^{N} y_i^4$, and $\sum_{i=1}^{N} x_i^2 y_i^2$. The statistical moments are minimum when the moments are taken along the mean $(\tilde{x}, \tilde{y})$ of the bivariates $(x_i', y_i')$ given in eqn. (4-12).

$$\tilde{x} = \frac{1}{N} \sum_{\forall i} x_i; \quad \tilde{y} = \frac{1}{N} \sum_{\forall i} y_i \tag{4-12}$$

Since Fitzgibbon's method is affine invariant, translation of the coordinate system to a new origin with coordinates $(\tilde{x}, \tilde{y})$ can be done and is effective in reducing the condition number of the scatter matrix. When Fitzgibbon's method is used with the above mentioned coordinates translation, the least squares ellipse fitting technique shall be called the numerically stable algebraic fitting technique (NSAF) [176].

## 4.3 Geometric distance minimization

Section 4.2 uses the minimization of the algebraic distance of the data points from the algebraic equation of ellipse (in eqn. (4-1)). Beginning from this section, the use of the geometric distance of a data point from the ellipse as the minimization goal is considered. This section derives an expression for the geometric distance of a data point from an ellipse. Consider the geometric equation of a generic ellipse with semi-major axis length $a$, semi-minor axis length $b$, angle of orientation (angle made by the major axis of the ellipse with the $x$ axis) $\theta_c$, and center $C(x_c, y_c)$. This geometric equation is presented in eqn. (4-13).

$$\frac{\left(\left(x - x_c\right)\cos\theta_c - \left(y - y_c\right)\sin\theta_c\right)^2}{a^2} + \frac{\left(\left(x - x_c\right)\sin\theta_c + \left(y - y_c\right)\cos\theta_c\right)^2}{b^2} = 1 \tag{4-13}$$

where $a$, $b$, $\theta_c$, $x_c$, and $y_c$ satisfy the following constraints:

$$
\begin{aligned}
&\text{C1:} \quad a, b \in \mathbb{R}^+ \\
&\text{C2:} \quad b \leq a \\
&\text{C3:} \quad \theta_c \in [0, \pi) \\
&\text{C4:} \quad x_c, y_c \in \mathbb{R}
\end{aligned}
\tag{4-14}
$$

Eqn. (4-13) can be simplified as eqns. (4-15) and (4-16).

$$\alpha\left(x-x_c\right)^2 + \beta\left(y-y_c\right)^2 + \gamma\left(x-x_c\right)\left(y-y_c\right) = a^2 b^2 \qquad (4\text{-}15)$$

$$\alpha = \left(a^2 \sin^2 \theta_c + b^2 \cos^2 \theta_c\right) = 0.5\left(\left(a^2 + b^2\right) - \left(a^2 - b^2\right)\cos 2\theta_c\right)$$

$$\beta = \left(a^2 \cos^2 \theta_c + b^2 \sin^2 \theta_c\right) = 0.5\left(\left(a^2 + b^2\right) + \left(a^2 - b^2\right)\cos 2\theta_c\right) \qquad (4\text{-}16)$$

$$\gamma = \left(a^2 - b^2\right)\sin 2\theta_c$$

The slope of the tangent at the point $P(x, y)$ is given by eqn. (4-17).

$$\frac{dy}{dx} = -\frac{2\alpha\left(x-x_c\right) + \gamma\left(y-y_c\right)}{2\beta\left(y-y_c\right) + \gamma\left(x-x_c\right)} \qquad (4\text{-}17)$$

and consequently, the equation of the tangent at a point $P_i(x_i, y_i)$ on the ellipse is given by eqns. (4-18) - (4-20).

$$y = m_i x + c_i \qquad (4\text{-}18)$$

$$m_i = \frac{dy}{dx}\bigg|_{P_i} = -\frac{2\alpha\left(x_i - x_c\right) + \gamma\left(y_i - y_c\right)}{2\beta\left(y_i - y_c\right) + \gamma\left(x_i - x_c\right)} \qquad (4\text{-}19)$$

$$c_i = \frac{2\beta\left(y_i - y_c\right)y_i + 2\alpha\left(x_i - x_c\right)x_i - \gamma\left(y_i x_c + x_i y_c - 2x_i y_i\right)}{2\beta\left(y_i - y_c\right) + \gamma\left(x_i - x_c\right)} \qquad (4\text{-}20)$$

Suppose, there is a data point $P_i'\left(x_i', y_i'\right)$; $x_i', y_i' \in \mathbb{Z}$, with its nearest point on the ellipse being $P_i(x_i, y_i)$. The actual distance between the ellipse and the pixel $P_i'\left(x_i', y_i'\right)$ is then equal to the distance of $P_i'\left(x_i', y_i'\right)$ from the tangent passing through $P_i(x_i, y_i)$ and is given by eqn. (4-21).

$$d_i = \left|\frac{y_i' - m_i x_i' - c_i}{\sqrt{1 + m_i^2}}\right| \qquad (4\text{-}21)$$

## 4.4 Geometry based least squares fitting of ellipses - ElliFit

### 4.4.1 Modification of the minimization function

For a sequence of pixels $\left\{ P_i'\left( x_i', y_i' \right); i = 1 \text{ to } N \right\}$, the intention is to find the parameters $a, b, \theta_c, x_c,$ and $y_c$, such that the square of the above distance is minimized. Here, $\left| \bullet \right|$ denotes the absolute value in the case of scalars and Euclidean norm in the case of vectors. Mathematically, this minimization is described by eqn. (4-22)

$$\min \left[ \left( d_i \right)^2 = \frac{\left( y_i' - m_i x_i' - c_i \right)^2}{1 + m_i^2} \right] : \text{ subject to C1-C4} \qquad (4\text{-}22)$$

where C1-C4 are defined in eqn. (4-14). For the minimization problem above, the minima occurs when $\partial \left( d_i \right)^2 / \partial m_i = 0$ and $\partial^2 \left( d_i \right)^2 / \partial m_i^2 > 0$. The expressions of $\partial \left( d_i \right)^2 / \partial m_i$ and $\partial^2 \left( d_i \right)^2 / \partial m_i^2$ are as shown in eqns. (4-23) and (4-24).

$$\frac{\partial \left( d_i \right)^2}{\partial m_i} = -2 \frac{\left( y_i' - m_i x_i' - c_i \right) \left( m_i y_i' + x_i' - m_i c_i \right)}{\left( 1 + m_i^2 \right)^2} \qquad (4\text{-}23)$$

$$\frac{\partial^2 \left( d_i \right)^2}{\partial m_i^2} = \frac{2}{\left( 1 + m_i^2 \right)^3} \left[ \left( m_i y_i' + x_i' - m_i c_i \right)^2 + \left( y_i' - m_i x_i' - c_i \right) \left( 2m_i^2 y_i' - 2m_i^2 c_i + 3mx_i' - y_i' + c_i \right) \right]$$

$$(4\text{-}24)$$

From the above, it is seen that $\left( y_i' - m_i x_i' + c_i \right) = 0$ is the appropriate solution for minimizing $\left( d_i \right)^2$. However, since the point $P_i(x_i, y_i)$ may not be actually on the ellipse, $\left( y_i' - m_i x_i' + c_i \right) = 0$ cannot be satisfied. Nevertheless, $\left| y_i' - m_i x_i' - c_i \right|$ can be minimized such that it is as close to zero as possible. Thus, in effect, solving the minimization problem of eqn. (4-22) can be reformulated as determining the parameters $a, b, \theta_c, x_c,$ and $y_c$, such that $r_i$ in eqn. (4-25)

$$r_i = \left| y_i' - m_i x_i' - c_i \right| \qquad (4\text{-}25)$$

is minimized subject to constraints C1-C4 defined in eqn. (4-14). Assuming that the pixels $\left\{ P_i'\left(x_i', y_i'\right); i = 1 \text{ to } N \right\}$ were obtained by digitizing the points on the ellipses, that is only digitization noise is present, the point $P_i(x_i, y_i)$ on ellipse nearest to the pixel $P_i'\left(x_i', y_i'\right)$ satisfies the constraint defined by eqn. (4-26).

$$\left| x_i' - x_i \right| \le 0.5; \ \left| y_i' - y_i \right| \le 0.5 \tag{4-26}$$

For the above condition, it can be proven that $m_i'/m_i \to 1$ and $c_i'/c_i \to 1$, where $m_i'$ and $c_i'$ are obtained by substitution of $P_i'\left(x_i', y_i'\right)$ in the place of $P_i(x_i, y_i)$ in eqns. (4-19) and (4-20). Thus, $r_i'/r_i \to 1$, where:

$$r_i' = \left| y_i' - m_i' x_i' - c_i' \right|. \tag{4-27}$$

Thus, minimizing the residue $r_i'$ described by eqn. (4-27) will in fact indirectly minimize the square of the geometric distance defined by eqn. (4-21).

### 4.4.2 Mathematical model of ElliFit

Eqns. (4-28) - (4-31) present the metric spaces and maps used for ElliFit.

$$\overline{A} = \begin{bmatrix} a & b & \theta_c & x_c & y_c \end{bmatrix}^{\mathrm{T}}, \text{ subject to C1-C4} \tag{4-28}$$

$$\overline{\Phi} = \begin{bmatrix} \phi_1 & \phi_2 & \phi_3 & \phi_4 & \phi_5 \end{bmatrix}^{\mathrm{T}}, \ \overline{\Phi} \in \mathbb{R}^5 \tag{4-29}$$

$$\overline{Y} = \begin{bmatrix} y_1' & y_2' & \dots & y_N' \end{bmatrix}^{\mathrm{T}}, \ \overline{Y} \in \mathbb{Z}^N \tag{4-30}$$

$$\mathbf{K} : \overline{A} \to \overline{\Phi}; \ \mathbf{X} : \overline{\Phi} \to \overline{Y} \tag{4-31}$$

The upright and non-bold letter 'T' in the superscript denotes the matrix and vector transpose of the matrices and vectors. The vector $\overline{A}$ is a five dimensional vector that contains the parameters of the fitted ellipse. The vector $\overline{Y}$ contains the $y$ coordinates of the pixels. A new five-dimensional vector $\overline{\Phi}$ is defined containing new real valued variables $\phi_1$ to $\phi_5$, which acts as an intermediate stage defined for splitting the non-linear map $\mathbf{M} : \overline{A} \to \overline{Y}$. The non-linear map $\mathbf{M}$ is split into two maps $\mathbf{K}$ and $\mathbf{X}$, as defined in eqn. (4-31). The variables $\phi_1$ to $\phi_5$ and the map $\mathbf{X}$ are explicitly designed, such that the mapping $\mathbf{X} : \overline{\Phi} \to \overline{Y}$ is linear and can be written as in eqn. (4-32).

$$\overline{Y} = \mathbf{X} \cdot \overline{\Phi} \qquad (4\text{-}32)$$

Since $\mathbf{M} : \overline{A} \to \overline{Y}$ is a non-linear map and $\mathbf{X} : \overline{\Phi} \to \overline{Y}$ is designed to be a linear map, hence the map $\mathbf{K} : \overline{A} \to \overline{\Phi}$ is non-linear.

In the subsequent sections, the properties of the operators $\mathbf{K}$ and $\mathbf{X}$ will be outlined and it will be shown that $\mathbf{K}$ is non-linear but is a one-to-one mapping due to the constraints of $\overline{\Phi}$ and $\overline{A}$. This implies that the parameters $a, b, \theta_c, x_c$, and $y_c$ can be obtained uniquely from $\phi_1$ to $\phi_5$, if $\overline{\Phi}$ can be determined by employing eqn. (4-32) to minimize the residue $r_i'$ given by eqn. (4-27).

For designing the variables $\phi_1$ to $\phi_5$ and the map $\mathbf{X}$, the following residual distance is defined in eqn. (4-33).

$$\sum_{i=1}^{N} \left( r_i'^2 \right) = \left\| \overline{Y} - \mathbf{X} \cdot \overline{\Phi} \right\|^2 \qquad (4\text{-}33)$$

where $\left\| \bullet \right\|$ denotes the Frobenius norm for matrices and Euclidean norm for vectors. The advantage in using the above definition is that the Moore Penrose pseudoinverse can be used to determine the unique optimal solution of $\overline{\Phi}$ in order to minimize the sum of squares of the residues $r_i'$, $i = 1$ to $N$ as shown in eqn. (4-34)

$$\overline{\Phi} = \left( \mathbf{X}^{\mathrm{T}} \cdot \mathbf{X} \right)^{-1} \cdot \mathbf{X}^{\mathrm{T}} \cdot \overline{Y} \qquad (4\text{-}34)$$

In order to obtain a form similar to eqn. (4-33), the variables $\phi_1$ to $\phi_5$ are defined in eqns. (4-35) - (4-39) and the map $\mathbf{X}$ is defined in eqn. (4-40).

$$\phi_1 = \alpha / \beta \qquad (4\text{-}35)$$

$$\phi_2 = \gamma / \beta \qquad (4\text{-}36)$$

$$\phi_3 = \frac{2\alpha x_c + \gamma y_c}{\beta} = 2\phi_1 x_c + \phi_2 y_c \qquad (4\text{-}37)$$

$$\phi_4 = \frac{2\beta y_c + \gamma x_c}{\beta} = 2y_c + \phi_2 x_c \tag{4-38}$$

$$\phi_5 = \frac{\alpha x_c^2 + \beta y_c^2 + \gamma x_c y_c - a^2 b^2}{\beta} = \phi_1 x_c^2 + y_c^2 + \phi_2 x_c y_c - a^2 b^2 / \beta \tag{4-39}$$

$$\mathbf{X} = \begin{bmatrix} \vdots & \vdots & \vdots & \vdots & \vdots \\ -x_i'^2 / y_i' & -x_i' & x_i' / y_i' & 1 & -y_i'^{-1} \\ \vdots & \vdots & \vdots & \vdots & \vdots \end{bmatrix} \tag{4-40}$$

For computing the actual parameters in $\overline{A}$, the following inverse map from the parameters ($\phi_1$ to $\phi_5$) to the parameters $a, b, \theta_c, x_c$, and $y_c$ as shown in eqns. (4-41) - (4-45) may be used.

$$a = 2\sqrt{\frac{\phi_2 \phi_3 \phi_4 - \phi_4^2 \phi_1 - \phi_3^2 - \phi_5\left(\phi_2^2 - 4\phi_1\right)}{\left(\phi_2^2 - 4\phi_1\right)\left(\left(1 + \phi_1\right) - \sqrt{\left(1 - \phi_1\right)^2 + \phi_2^2}\right)}} \tag{4-41}$$

$$b = 2\sqrt{\frac{\phi_2 \phi_3 \phi_4 - \phi_4^2 \phi_1 - \phi_3^2 - \phi_5\left(\phi_2^2 - 4\phi_1\right)}{\left(\phi_2^2 - 4\phi_1\right)\left(\sqrt{\left(1 - \phi_1\right)^2 + \phi_2^2} + \left(1 + \phi_1\right)\right)}} \tag{4-42}$$

$$\theta_c = -0.5 \tan^{-1}\left(-\phi_2 / \left(1 - \phi_1\right)\right) \tag{4-43}$$

$$x_c = \left(\phi_2 \phi_4 - 2\phi_3\right) / \left(\phi_2^2 - 4\phi_1\right) \tag{4-44}$$

$$y_c = \left(\phi_2 \phi_3 - 2\phi_1 \phi_4\right) / \left(\phi_2^2 - 4\phi_1\right) \tag{4-45}$$

It is notable that the inverse tangent function used in eqn. (4-43) is a four quadrant inverse tangent function.

### 4.4.3 Uniqueness of non-linear operator

In this section, it is shown that the map $\mathbf{K} : \overline{A} \rightarrow \overline{\Phi}$ is injective. This means that for a vector $\overline{A}$ defined by eqn. (4-28), there is one and only one vector $\overline{\Phi}$ as computed by the map $\mathbf{K} : \overline{A} \rightarrow \overline{\Phi}$, and for a vector $\overline{\Phi}$ given by eqn. (4-29) (subject to the conditions of existence of solution), there is one and only one vector $\overline{A}$ satisfying the constraints C1-C4 as described by eqn. (4-14).

### *4.4.3.1 Forward mapping* $\mathbf{K}: \bar{A} \rightarrow \bar{\Phi}$

Due to the constraint C1 in eqn. (4-14), $a^2$ and $b^2$ are one-to-one functions of $a$ and $b$. Owing to this, $\phi_1$ and $\phi_2$, both are one-to-one functions of $a$ and $b$. Although $\phi_1$ and $\phi_2$ are not individually injective functions of $\theta_c$ when $\theta_c$ is subject to constraint C3, however as a pair of functions $\{\phi_1, \phi_2\}$, the pair has a one-to-one relationship with $\theta_c$, as constrained by C3. This is similar to the fact that for a general angle $\theta \in [0, 2\pi)$, $\sin\theta$ and $\cos\theta$ are individually many-to-one, but a vector defined as $[\sin\theta \quad \cos\theta]^T$ is a unique vector for any $\theta \in [0, 2\pi)$.

According to eqns. (4-37) and (4-38), since $\phi_3$ and $\phi_4$ are simply linear combinations of $x_c$ and $y_c$, they are also one-to-one functions of $x_c$ and $y_c$. Further, $\phi_3$ and $\phi_4$ are linear combinations of $\phi_1$ and $\phi_2$, which have a one-to-one relationship with $a$, $b$, and $\theta_c$ as a pair. Thus, as a pair, $\phi_3$ and $\phi_4$ also have a one-to-one relationship with $a$, $b$, and $\theta_c$. Following the same logic, although $\phi_5$ is a many-to-one function of $\theta_c$, $x_c$ and $y_c$, the vector $\bar{\Phi}$ itself is a one-to-one map of the vector $\bar{A}$. This means that corresponding to a vector $\bar{A}$ in the five-dimensional space, constrained by C1-C4 as described by eqn. (4-14), there is one and only one $\bar{\Phi}$ in the five-dimensional real space.

### *4.4.3.2 Inverse mapping* $\tilde{\mathbf{K}}: \bar{\Phi} \rightarrow \bar{A}$

Now considering the inverse mapping, i.e. the mapping from $\bar{\Phi}$ to $\bar{A}$, $\tilde{\mathbf{K}}: \bar{\Phi} \rightarrow \bar{A}$, as specified by eqns. (4-41)-(4-45). Since the four quadrant inverse tangent is used in eqn. (4-43), $\theta_c$ is a one-to-one function of $\phi_1$ and $\phi_2$.

It can be shown that for a given $\bar{\Phi}$, a solution $\bar{A}$ exists if and only if the following conditions of existence are satisfied as expressed by eqn. (4-46).

$$
\begin{aligned}
&\text{E1: } \phi_1 > 0 \\
&\text{E2: } \begin{cases} \text{Either} \quad \phi_5 > 0 \text{ AND } \phi_2\phi_3\phi_4 > \left(\phi_4{}^2\phi_1 + \phi_3{}^2\right) \\ \text{Or} \quad \phi_5 < 0 \text{ AND } \phi_2\phi_3\phi_4 < \left(\phi_4{}^2\phi_1 + \phi_3{}^2\right) \end{cases}
\end{aligned}
\quad (4\text{-}46)
$$

Now, it is shown that for a set of values of $\phi_1$ to $\phi_4$, the pair of $\left( x_c, y_c \right)$ is a unique pair. The expression $\left( \phi_2^2 - 4\phi_1 \right)$ can be written as eqn. (4-47).

$$\left( \phi_2^2 - 4\phi_1 \right) = \left( \left( 1-\phi_1 \right) \sec 2\theta_c \right)^2 - \left( 1+\phi_1 \right)^2 \tag{4-47}$$

Thus, given the condition E1 of existence of solution and the solution $\theta_c$, the denominator $\left( \phi_2^2 - 4\phi_1 \right)$ in eqns. (4-44) and (4-45) is a one-to-one function of $\phi_1$ and $\theta_c$. Now, considering the numerators, $x_c$ is a many-to-one function of $\phi_2$ and $\phi_4$ (since $\phi_2\phi_4 = \left( -\phi_2 \right)\left( -\phi_4 \right)$). Similarly, $y_c$ is a many-to-one function of $\phi_2$ and $\phi_3$ (it is noted that due to the condition E1, $y_c$ is a one-to-one function of $\phi_1$ and $\phi_4$). However, as a pair, $\left( x_c, y_c \right)$ together form one-to-one functions of $\phi_1$ to $\phi_4$.

Now, the uniqueness of $a$ and $b$ is considered. Owing to the constraint C1 in eqn. (4-14), it is sufficient to prove the uniqueness of $a^2$ and $b^2$ for a given set of values of $\phi_1$ to $\phi_5$ that satisfy the existence conditions E1 and E2 in eqn. (4-46). Using the eqns. (4-41)-(4-45) and (4-47), $a^2$ and $b^2$ can be written as in eqns. (4-48) and (4-49) .

$$a^2 = 2 \frac{\phi_3 x_c + \phi_4 y_c - \phi_5}{\left( \left( 1-\phi_1 \right) \sec 2\theta_c \right) - \left( 1+\phi_1 \right)} \tag{4-48}$$

$$b^2 = 2 \frac{\phi_3 x_c + \phi_4 y_c - \phi_5}{\left( \left( 1-\phi_1 \right) \sec 2\theta_c \right) + \left( 1+\phi_1 \right)} \tag{4-49}$$

Using the arguments presented just after eqn. (4-47), the denominators in eqns. (4-48) and (4-49) are one-to-one functions of $\phi_1$ and $\theta_c$. Since the pair $\left( x_c, y_c \right)$ is one-to-one functions of $\phi_1$ to $\phi_4$, and the numerator in eqns. (4-48) and (4-49) is simply a linear combination of $x_c$ and $y_c$, hence $a^2$ and $b^2$ are also one-to-one functions of $\phi_1$ to $\phi_5$ for a given set of $x_c$, $y_c$, and $\theta_c$. Thus, the vector $\overline{A}$ is a one-to-one function of the vector $\overline{\Phi}$, subject to the existence conditions of E1 and E2, see eqn. (4-46).

Thus, in the framework of constraints C1-C4 and E1-E2, the mapping $\mathbf{K} : \bar{A} \rightarrow \bar{\Phi}$ and the inverse mapping $\tilde{\mathbf{K}} : \bar{\Phi} \rightarrow \bar{A}$ is injective. If any of the existence conditions E1-E2 are not satisfied while inverse mapping, either $a$ or $b$ or both will have complex values. This is an easy and direct way to filter off non-elliptic curves. This is the basis of the high selectivity demonstrated by ElliFit.

### 4.4.4 Numerical stability of linear operator

In this section, considering the mathematical properties of the linear map $\mathbf{X} : \bar{\Phi} \rightarrow \bar{Y}$. Specifically, addressing the properties concerning the computation of the Moore Penrose pseudoinverse of $\mathbf{X}$. The Moore Penrose pseudoinverse $\tilde{\mathbf{X}} : \bar{Y} \rightarrow \bar{\Phi}$ of $\mathbf{X} : \bar{\Phi} \rightarrow \bar{Y}$ is given by eqn. (4-50).

$$\tilde{\mathbf{X}} = \left( \mathbf{X}^{\mathrm{T}} \cdot \mathbf{X} \right)^{-1} \cdot \mathbf{X}^{\mathrm{T}} \tag{4-50}$$

For convenience, $\mathbf{G}$ is defined as in eqn. (4-51).

$$\mathbf{G} = \left( \mathbf{X}^{\mathrm{T}} \cdot \mathbf{X} \right) \tag{4-51}$$

Thus, using (4-40), $\mathbf{G}$ can be written as eqn. (4-52).

$$\mathbf{G} = \begin{bmatrix} g_1 & h_1 & j_1 & h_2 & g_3 \\ h_1 & g_2 & h_2 & j_2 & h_3 \\ j_1 & h_2 & g_3 & h_3 & j_3 \\ h_2 & j_2 & h_3 & g_4 & h_4 \\ g_3 & h_3 & j_3 & h_4 & g_5 \end{bmatrix} \tag{4-52}$$

where:

$$g_1 = \sum_{i=1}^{N} \frac{x_i'^4}{y_i'^2}; \; g_2 = \sum_{i=1}^{N} x_i'^2; \; g_3 = \sum_{i=1}^{N} \frac{x_i'^2}{y_i'^2}; \; g_4 = \sum_{i=1}^{N} 1; \; g_5 = \sum_{i=1}^{N} \frac{1}{y_i'^2};$$

$$h_1 = \sum_{i=1}^{N} \frac{x_i'^3}{y_i'}; \; h_2 = \sum_{i=1}^{N} \frac{-x_i'^2}{y_i'}; \; h_3 = \sum_{i=1}^{N} \frac{x_i'}{y_i'}; \; h_4 = \sum_{i=1}^{N} \frac{-1}{y_i'}; \tag{4-53}$$

$$j_1 = \sum_{i=1}^{N} \frac{-x_i'^3}{y_i'^2}; \; j_2 = \sum_{i=1}^{N} -x_i'; \; j_3 = \sum_{i=1}^{N} \frac{-x_i'}{y_i'^2}$$

The numerical stability of the matrix $\mathbf{G}$ is integral to the accuracy and stability of the solution $\overline{\Phi}$ and subsequently $\overline{A}$. The condition number of the matrix $\mathbf{G}$ is an indication of the numerical stability of its inversion. Thus, the condition number of the matrix $\mathbf{G}$ is studied.

It is to be noted that the matrix $\mathbf{G}$ is a five-dimensional symmetric matrix containing only 12 distinct entries. Further, the perturbation in $\mathbf{G}$ may occur only through the perturbations in the pixels $\left\{ P_i'\left( x_i', y_i' \right); i = 1 \text{ to } N \right\}$, which may be the result of digitization or distortion or other types of noise. Since the perturbations occur only through the pixels, the perturbations in $\mathbf{G}$ are also symmetric. Thus, following [177], the Bauer Skeel condition number (infinity norm condition number) of the matrix $\mathbf{G}$ is within a finite well defined ratio of the actual condition number of the matrix $\mathbf{G}$, and can be considered as a direct indicator of the condition number. This theory allows the study of the condition number as related to the component with the maximum absolute value in the matrix $\mathbf{G}$. It is evident from eqn. (4-53) that $\mathbf{G}$ has the largest condition number when one of the pixels has zero $y$ coordinate, that is, $\exists i \in 1 \text{ to } N : y_i' = 0$. When one of the pixels has zero $y$ coordinate, that is, $\exists i \in 1 \text{ to } N : y_i' = 0$, the determinant of $\mathbf{G}$ becomes infinite, leading to a singular $\mathbf{G}^{-1}$.

However, this numerical instability can be easily avoided by modifying $\mathbf{X}$ and $\overline{Y}$ as follows in eqns. (4-54) and (4-55).

$$\mathbf{X} = \begin{bmatrix} \vdots & \vdots & \vdots & \vdots & \vdots \\ -x_i'^2 & -x_i' y_i' & x_i' & y_i' & -1 \\ \vdots & \vdots & \vdots & \vdots & \vdots \end{bmatrix} \tag{4-54}$$

$$\overline{Y} = \begin{bmatrix} y_1'^2 & y_2'^2 & \dots & y_N'^2 \end{bmatrix}^{\mathrm{T}}, \ \overline{Y} \in \mathbb{Z}^N \tag{4-55}$$

As a result of this modification, instead of minimizing the residues $r_i'$ through the cost function given in eqn. (4-33). The following cost function defined in eqn. (4-56) is used.

$$\sum_{i=1}^{N} \left( y_i' r_i' \right)^2 = \left\| \overline{Y} - \mathbf{X} \cdot \overline{\Phi} \right\|^2 \tag{4-56}$$

where $\mathbf{X}$ and $\overline{Y}$ are given by eqns. (4-54) and (4-55) respectively. Although the modification improves the numerical stability of the inverse of matrix $\mathbf{G}$ as defined in eqn. (4-51), it has an implication on the minimum number of pixels required for fitting the ellipse. From eqn. (4-56), if a pixel $P_i'$ has zero $y$ coordinate, then $\left(y_i' r_i'\right) = 0$, which implies that even though the residue $r_i'$ may be non-zero, its contribution to the cost function is zero. Since any ellipse may intersect the $x$ axis at a maximum of two points, using $N \geq 7$ ensures that at least five points (same as the number of unknowns) contributes directly to the cost function.

Following the modifications given in eqns. (4-54) and (4-55), The Bauer condition number of the matrix $\mathbf{G}$ defined in eqn. (4-51) has the strongest contribution from either $\sum_{i=1}^{N} x_i'^4$ or $\sum_{i=1}^{N} x_i'^2 y_i'^2$. It is noted that due to the modifications in eqns. (4-54) and (4-55), all the components of the matrix $\mathbf{G}$ are $N$ times the statistical moments $\psi_{(m,n)}\left(x_i', y_i'\right)$ of the bivariates $\left(x_i', y_i'\right)$ along $(0, 0)$, where the $(n, m)$th moment around a point $(\tilde{x}, \tilde{y})$ is given by eqn. (4-57).

$$\psi_{(m,n)}\left(x_i', y_i'\right) = \frac{1}{N} \sum_{\forall i} \left(x_i' - \tilde{x}\right)^n \left(y_i' - \tilde{y}\right)^m \tag{4-57}$$

The statistical moments are minimum when $(\tilde{x}, \tilde{y})$ are the averages (the first order moments) of the bivariates $\left(x_i', y_i'\right)$, this is given in eqn. (4-58).

$$\tilde{x} = \frac{1}{N} \sum_{\forall i} x_i'; \quad \tilde{y} = \frac{1}{N} \sum_{\forall i} y_i' \tag{4-58}$$

Thus, if the pixel space is translated at a new origin given by eqn. (4-58), the condition number of the matrix $\mathbf{G}$ can be further reduced.

### 4.4.5 Algorithm and computational complexity

ElliFit is presented in algorithmic form in Figure 4.4-1.

Step 1: Compute $\tilde{x}_i = x_i' - \bar{x}, \tilde{y}_i = y_i' - \bar{y}$ where $(\bar{x}, \bar{y})$ are given by eqn. (4-58).

Step 2: Form the matrix $\mathbf{X}$ and the vector $\bar{Y}$ as given in eqns. (4-59) and (4-60). We highlight that eqns. (4-59) and (4-60) are similar to eqns. (4-54) and (4-55) and $x_i'$ and $y_i'$ in eqns. (4-54) and (4-55) are replaced by $\tilde{x}_i$ and $\tilde{y}_i$ in eqns. (4-59) and (4-60).

$$\mathbf{X} = \begin{bmatrix} \vdots & \vdots & \vdots & \vdots & \vdots \\ -\tilde{x}_i^2 & -\tilde{x}_i \tilde{y}_i & \tilde{x}_i & \tilde{y}_i & -1 \\ \vdots & \vdots & \vdots & \vdots & \vdots \end{bmatrix} \quad (4\text{-}59)$$

$$\bar{Y} = \begin{bmatrix} \tilde{y}_1^2 & \tilde{y}_2^2 & \ldots & \tilde{y}_N^2 \end{bmatrix}^T, \bar{Y} \in \mathbb{Z}^N \quad (4\text{-}60)$$

Step 3: Compute $\bar{\Phi}$ using eqn. (4-34).

Step 4: Compute $a, b, \theta_c, \tilde{x}_c,$ and $\tilde{y}_c$ using eqns. (4-41) - (4-45), where $x_c$ and $y_c$ in eqns. (4-44) and (4-45) are replaced by $\tilde{x}_c$ and $\tilde{y}_c$.

Step 5: Compute $x_c = \tilde{x}_c + \tilde{x}$ and $y_c = \tilde{y}_c + \tilde{y}$.

**Figure 4.4-1: Algorithm of ElliFit.**

The computational complexity of ElliFit is determined as follows:

1. The computational complexity of Step 1 is $O(2N)$.

2. Assuming that the elements of matrix $\mathbf{G}$ are computed directly, the complexity of forming $\mathbf{G}$ is $O(12N)$.

3. The computational complexity of computing $\mathbf{G}^{-1}$ is $O(5^3)$.

4. Computation complexity of determining $\bar{\Phi}$ is $O(30N)$.

5. Computational complexity of determining $\bar{A}$ is $O(5)$.

The computational complexity is determined mainly by the complexity of computing $\bar{\Phi}$: $O(30N)$. Thus, ElliFit has a computational complexity of $O(N)$ only. Hence it is highly efficient technique to fit an ellipse.

## 4.5 Comparison with other methods

The performance of the proposed method (ElliFit) is compared with various methods based on the least squares fitting of ellipses. The methods used for comparison are Fitzgibbon [80], Chaudhuri [84], Harker [174], Ahn [90], Maini [173], and Liu [32]. Fitzgibbon [80], Chaudhuri [84], and Harker [174] are non-iterative ellipse fitting methods. Chaudhuri [84] is basically an ellipse fitting algorithm for any cluster of data. Ahn [90], Maini [173], and Liu [32] are iterative methods. Ahn [90] is the only method based on the geometry of the ellipse. Maini [173] and Liu [32] are based on Fitzgibbon [80] and aimed at improving the performance of ellipse fitting for special cases.

### 4.5.1 Digital incomplete or complete elliptic curves

Let consider a family of elliptic curves given by $b \in [20,150]$, $a \in [b,150]$, $\alpha \in [0,\pi)$, $x_c = 150$, and $y_c = 150$. In the range $[\theta_1, \theta_1 + \Delta\theta]$, where $\theta_1$ is randomly selected, a digital curve corresponding to the analytical ellipse is generated. For various values of $\Delta\theta$, 1000 such random curves are generated, corresponding to randomly chosen values of $a, b, \alpha$, and $\theta_1$. For these curves, the ellipses are fitted using ElliFit and the six methods used for comparison. For the fitted ellipses, ellipses that satisfy the following conditions are retained for further considerations:

   i.   The semi-major axis of a fitted ellipse should be less than 200.

   ii.   The ratio of the semi-minor to the semi-major axis (aspect ratio) should be more than 0.1 (that is the eccentricity of the ellipse should be less than 0.995).

The following performance parameters are plotted for each of the methods for the various values of $\Delta\theta$:

1. The mean of the error metric E13 as proposed by Rosin in [86].

2. The mean of the error metric E14 as proposed by Rosin in [88].

3. The mean of the distances $d = \sum_{i=1}^{N} d_i \bigg/ N$ of the pixels from the fitted ellipse.

4. Total number of detected ellipses $E_{total}$.

5. Recall: Recall $= E_{O \geq 0.8}/1000$, where $E_{O \geq 0.8}$ is the number of detected ellipses that have overlap ratios $O$ (given by Jaccard index [178]) more than or equal to 0.8 with the corresponding actual ellipses.

6. Precision: Precision $= E_{O \geq 0.8}/E_{total}$.

The first three are error metrics and the lower values of these parameters are indicator of better fitting. On the other hand, the remaining three are related to the ellipse detection characteristics. The results are plotted in Figure 4.5-1 (error metrics) and Figure 4.5-2 (ellipse detection characteristics). For a better representation of the data in Figure 4.5-1, the parameters E13, E14, and $d$ with respect to the data for ElliFit is plotted. This means that instead of plotting E13, E14, and $d$, the plots use $(E13-E13_P)$, $(E14-E14_P)$, and $(d-d_P)$, where the subscript P denotes the proposed ElliFit method. The values of $E13_P$, $E14_P$, and $d_P$ are listed in the sub-figures (a,b,c) of Figure 4.5-1 respectively. The results in Figure 4.5-1 show that ElliFit has the lowest values of the error metrics E13, E14, and $d$ for all the values of $\Delta\theta$, while Ahn [90] and Harker [174] are the closest competitors.

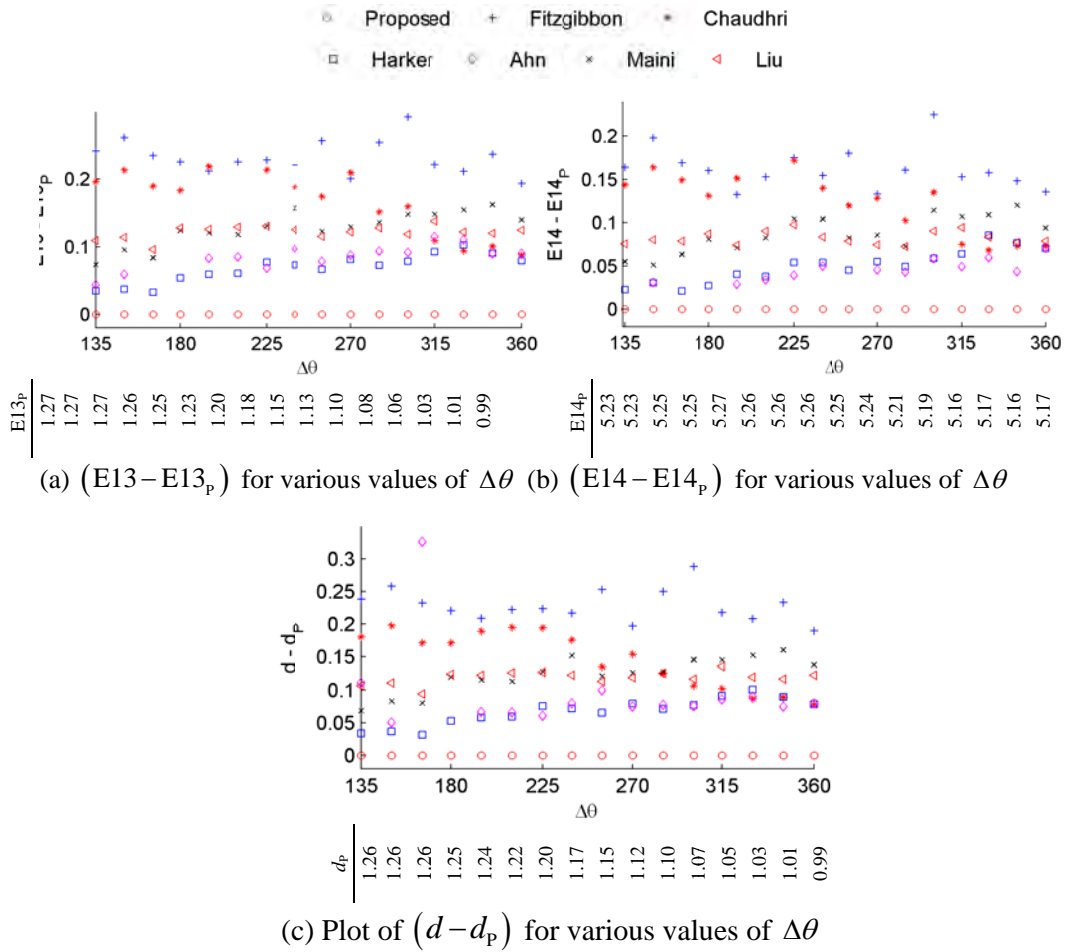

(a) $(E13-E13_P)$ for various values of $\Delta\theta$  (b) $(E14-E14_P)$ for various values of $\Delta\theta$

(c) Plot of $(d-d_P)$ for various values of $\Delta\theta$

**Figure 4.5-1: Plot of error metrics for Experiment 4.5.1.**

Figure 4.5-2 shows the total detected ellipses (Figure 4.5-2(a)), recall (Figure 4.5-2(b)), and precision (Figure 4.5-2(c)) for various values of $\Delta\theta$. It is seen that ElliFit has successfully detected all the ellipses for all values of $\Delta\theta$ except for $\Delta\theta = 135°$, for which ElliFit detected 982 ellipses out of 1000. While Chaudhuri [84], Maini [173], and Liu [32] detected close to 1000 ellipses for each value of $\Delta\theta$, their recall and precision values are very poor. This is understandable because while Chaudhuri [84] is an ellipse fitting algorithm that fits ellipses on any given cluster of points, it is not selective for elliptic shapes.

On the other hand, Liu [32] and Maini [173] are marginal improvements of Fitzgibbon [80] which in most cases perform very similar to Fitzgibbon. Although Harker [174] detected close to 1000 ellipses for each value of $\Delta\theta$, it has poor recall and precision as compared to ElliFit and Ahn [90]. Although Ahn [90] has good precision (very close to ElliFit and the maximum value 1), it performed poorer than ElliFit for recall and significantly poorer than ElliFit in terms of total detections. The total number of detected ellipses is low for Ahn [90] because it uses two nested non-linear iterative optimization processes and it is easy for Ahn [90] to either misconverge to a local minimum or to become non-convergent. On the other hand, the use of non-linear iterative optimization based on geometric principles helped Ahn [90] to be very precise if it converges to the global minimum.

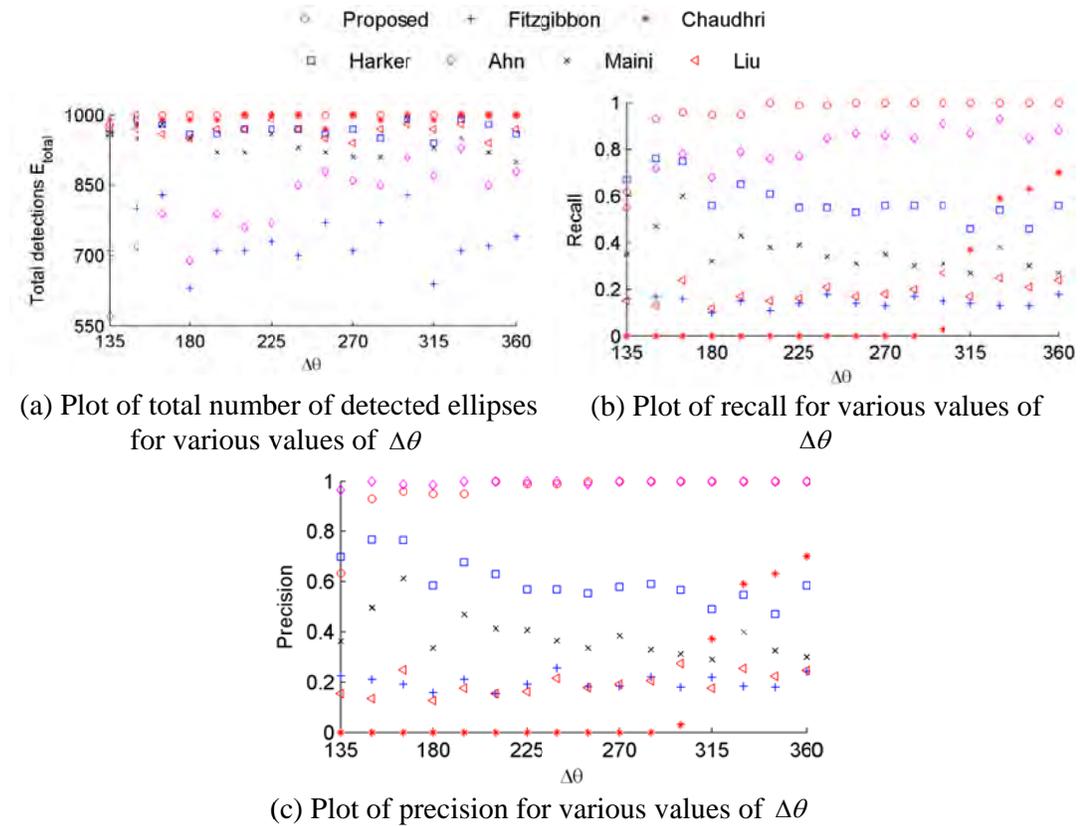

(a) Plot of total number of detected ellipses for various values of $\Delta\theta$

(b) Plot of recall for various values of $\Delta\theta$

(c) Plot of precision for various values of $\Delta\theta$

**Figure 4.5-2: Ellipse detection characteristics for Experiment 4.5.1.**

For illustration, an example of a digitized ellipse with partial data corresponding to $\Delta\theta = 140°$ is presented in Figure 4.5-3. The results of various algorithms are shown. The actual ellipse is shown using thin (black) curve while the fitted ellipses are shown in thick (red) curves. It is noted that Ahn [90] does not detect the ellipse due to misconvergence, Fitzgibbon [80] and Liu [32] does not generate an ellipse that satisfies the conditions (i and ii) in section 4.5.1, Maini [173] fits an ellipse that is quite different from the actual ellipse, Harker [174] generates an ellipse similar (but not exactly matching) to the actual ellipse, Chaudhri [84] fitted an inaccurate ellipse, and only ElliFit managed to successfully detect the ellipse that is very close to the actual ellipse from the partial digital ellipse data given as its input.

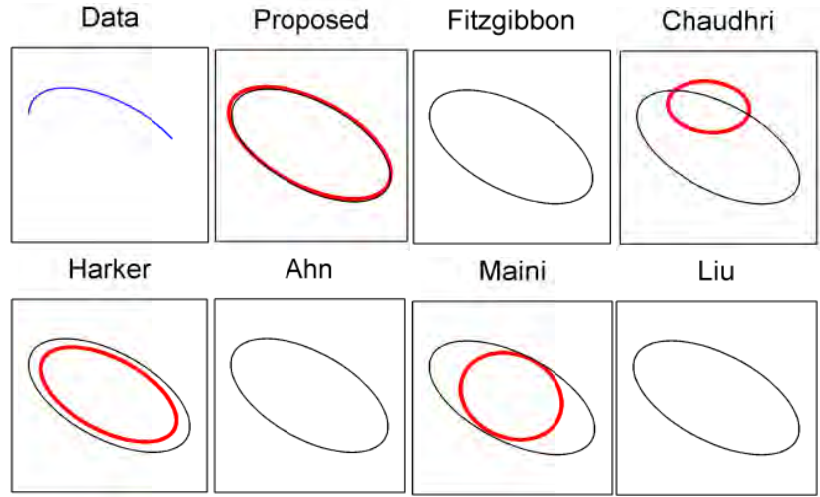

**Figure 4.5-3: An example of digital incomplete elliptic curve (Experiment 4.5.1).**

The results demonstrate good applicability of ElliFit for detecting elliptic shapes from digital images. Thus, ElliFit should be effective for applications that require detection of elliptic shapes from digital images. Since digital images are ubiquitous in today's world and several natural and man-made structures are elliptic, a wide range of applications may benefit from ElliFit.

### 4.5.2 Noisy cluster of points around an ellipse

Considering a family of ellipses given by $b \in [20,150]$, $a \in [b,150]$, $\alpha \in [0,\pi)$, $x_c = 150$, and $y_c = 150$. First, points on the ellipse corresponding to various values of $\theta$ at an interval of 10 degrees are computed. Let the set of points be denoted as $\{P_\theta(x_\theta, y_\theta)\}$. Subsequently, zero mean Gaussian noise is added to the value of the coordinates, such that the standard deviations of the noise for the $x$ and $y$ coordinates are $\sigma_x = \kappa \max|x - x_c|/100$ and $\sigma_y = \kappa \max|y - y_c|/100$ respectively, where $\kappa$ is the noise percentage. The same error metrics as in section 4.5.1 are computed for the various values of $\kappa$. The results are plotted in Figure 4.5-4 (error metrics) and Figure 4.5-5 (ellipse detection characteristics). Harker [174] used a maximum noise percentage of 3% for elliptic data points. Here, the results for up to 20% noise are shown.

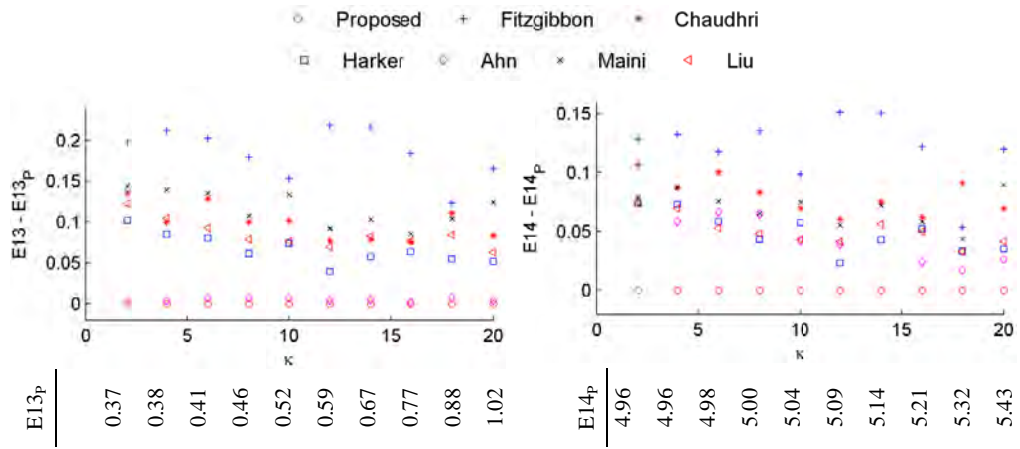

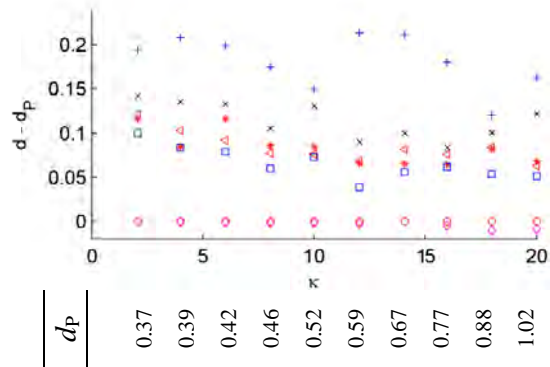

(a) Plot of (E13−E13$_P$) for various values of noise percent $\kappa$.

(b) Plot of (E14−E14$_P$) for various values of noise percent $\kappa$.

(c) Plot of (d−d$_P$) for various values of noise percent $\kappa$.

**Figure 4.5-4: Plot of error metrics for Experiment 4.5.2.**

From Figure 4.5-4, it is evident that Ahn [90] and ElliFit have the lowest values of E13 and $d$. For high noise levels ($\kappa > 14$) Ahn [90] performs slightly better than ElliFit for the parameter $d$. However, ElliFit clearly outperformed against all the other methods with respect to the parameter E14. For the current experiment, Chaudhri [84] demonstrated the third best performance. This is as expected because Chaudhri [84] benefitted in this experiment by the uniform availability of data points across the complete curvature of the ellipse.

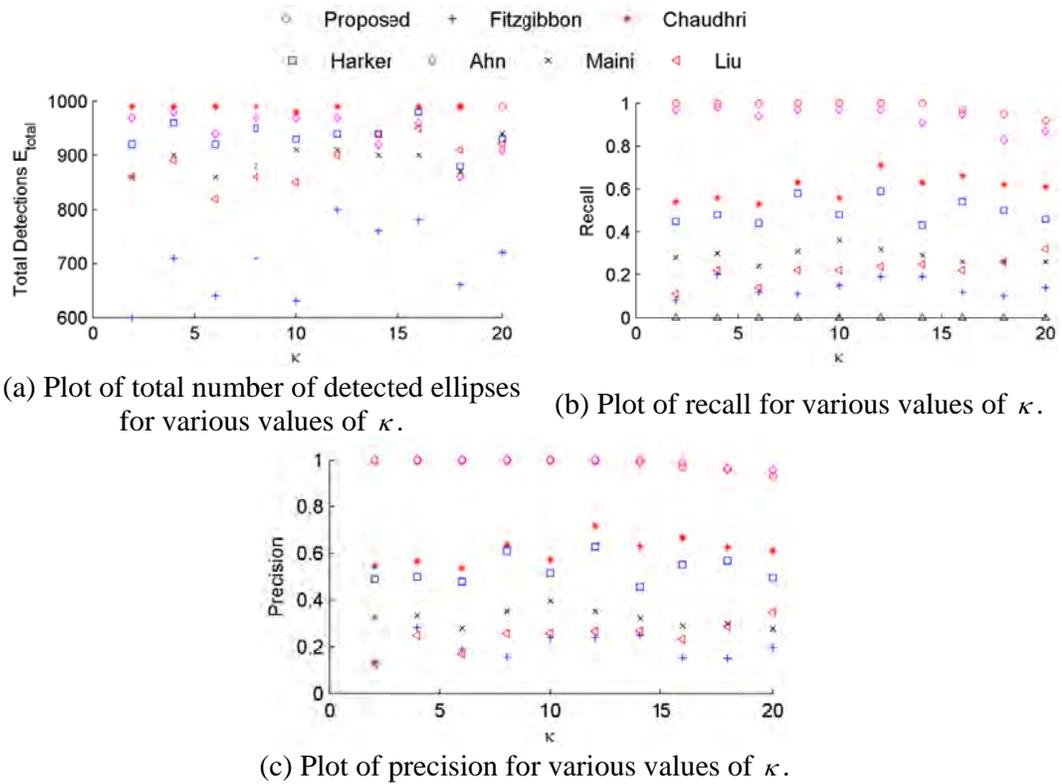

(a) Plot of total number of detected ellipses for various values of $\kappa$.

(b) Plot of recall for various values of $\kappa$.

(c) Plot of precision for various values of $\kappa$.

**Figure 4.5-5: Ellipse detection characteristics for Experiment 4.5.2**

The ellipse detection characteristics are shown in Figure 4.5-5. It is seen that ElliFit detected the maximum number of ellipses and the values of the recall and precision for ElliFit are close to 1. In comparison, Ahn [90] has slightly poorer recall and better or same precision as ElliFit.

An example is illustrated in Figure 4.5-6. The results of various algorithms are shown. The actual ellipse is shown using thin (black) curve while the fitted ellipses are shown in thick (red) curves. The data points are quite deviated from the actual ellipse as seen in Figure 4.5-6. It is seen that ElliFit and Ahn [90] have detected ellipses that are quite similar and have good overlap with the actual ellipse. Chaudhuri [84] also detected an ellipse that has good overlap with the actual ellipse, but the detected ellipse has lower eccentricity as compared to the actual ellipse. Harker [174] and Maini [173] detected similar but smaller ellipses than the actual ellipse. Fitzgibbon [80] detects a similar but larger ellipse and Liu [32] misconverges to an inaccurate ellipse with different orientation.

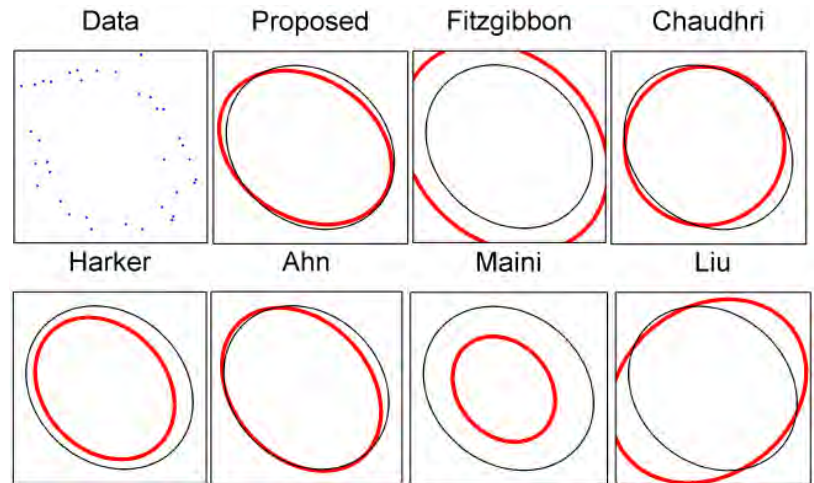

**Figure 4.5-6: An example of noisy cluster of points ($\kappa = 20$) around an ellipse (Experiment 4.5.2).**

Another challenging situation is considered where the data is not available around the complete circumference of the ellipse. It is available for only 270° angular sector. The error metrics and ellipse detection characteristics are plotted in Figure 4.5-7 and Figure 4.5-8 respectively. In terms of E13 and $d$, it is seen that the performance of ElliFit and Ahn [90] are the best among all the methods. However, ElliFit outperformed all other methods for E14. It is noted that for the noise percent $\kappa = 18, 20$, Ahn [90] misconverges for a few ellipses, resulting in very high value of $(d - d_{\mathrm{p}})$ (close to 50) and $(\mathrm{E}14 - \mathrm{E}14_{\mathrm{p}})$ (close to 25).

For the ellipse detection characteristics shown in Figure 4.5-8, ElliFit detected the maximum number of ellipses. Though Ahn [90] detected a lesser number of ellipses, its recall and precision is very close to ElliFit. For higher values of noise, if Ahn [90] detects ellipses, they are detected with greater precision and recall as compared to ElliFit. This is as expected because ElliFit employs a single step of linear least squares retrieval of parameters, Ahn [90] recoursed to several iterations in which it gets the chance to correct and improve on the fitting.

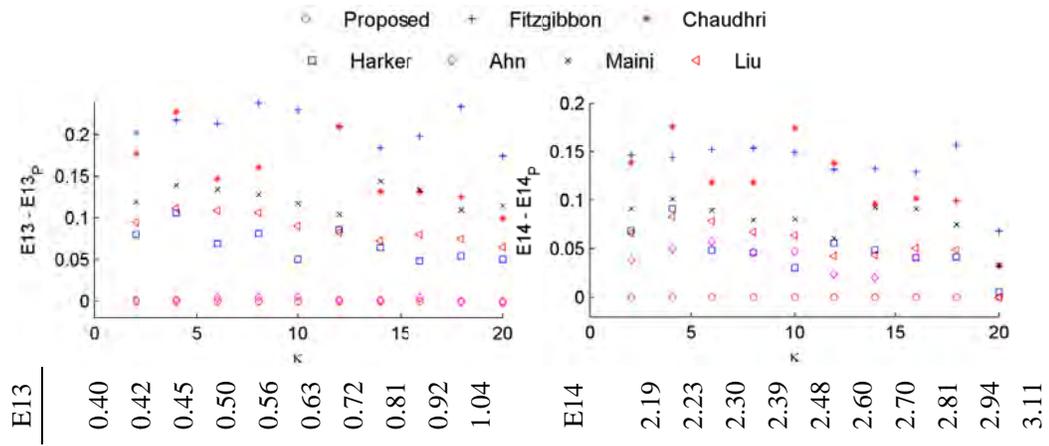

| E13 | 0.40 | 0.42 | 0.45 | 0.50 | 0.56 | 0.63 | 0.72 | 0.81 | 0.92 | 1.04 |
|---|---|---|---|---|---|---|---|---|---|---|

| E14 | 2.19 | 2.23 | 2.30 | 2.39 | 2.48 | 2.60 | 2.70 | 2.81 | 2.94 | 3.11 |
|---|---|---|---|---|---|---|---|---|---|---|

(a) Plot of $(E13 - E13_P)$ for various values of noise percent $\kappa$.

(b) Plot of $(E14 - E14_P)$ for various values of noise percent $\kappa$.

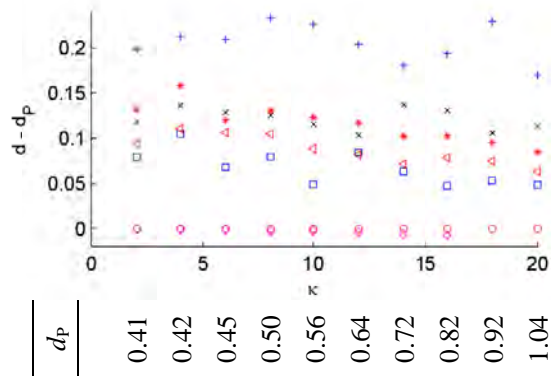

| $d_P$ | 0.41 | 0.42 | 0.45 | 0.50 | 0.56 | 0.64 | 0.72 | 0.82 | 0.92 | 1.04 |
|---|---|---|---|---|---|---|---|---|---|---|

(c) Plot of $(d - d_P)$ for various values of noise percent $\kappa$.

**Figure 4.5-7: Plot of error metrics for Experiment 4.5.2, 270°.**

An example with 270° angular sector data and $\kappa = 20$ is illustrated in Figure 4.5-9. The result of various algorithms are shown. The actual ellipse is shown using thin (black) curve while the fitted ellipses are shown in thick (red) curves. It is noted that the ellipse fitted by ElliFit is closest to the actual ellipse, followed by Ahn [90]. Fitzgibbon [80] fitted to a very small ellipse, Chaudhri [84], Harker [174], Maini [173], and Liu [32] detected the ellipse with greater overlap. Furthermore, these are not close to the actual ellipse. Noise is often present in most practical scenarios. The effect is more prominently seen in astronomical and medical data where the images often have clusters of points that are close to elliptic shapes but with high variance. Further, such applications are quite critical to the accurate detection of elliptic patterns. Robust performance of ElliFit makes it a good candidate for such applications involving significant amount of noise.

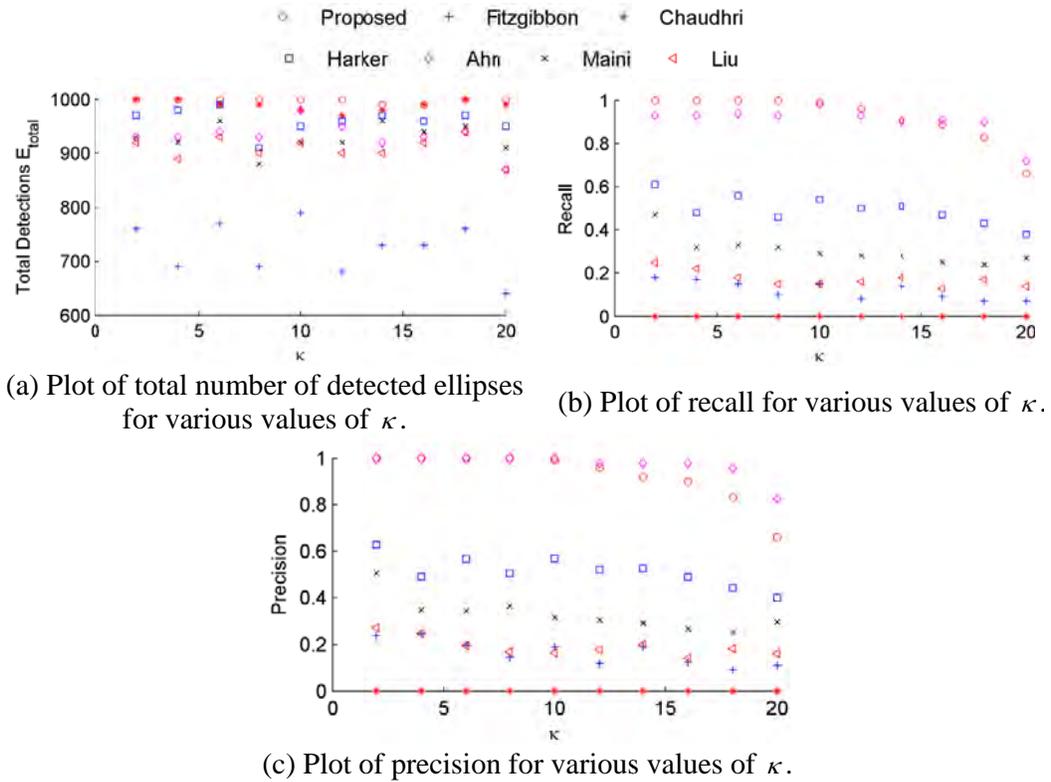

(a) Plot of total number of detected ellipses for various values of $\kappa$.

(b) Plot of recall for various values of $\kappa$.

(c) Plot of precision for various values of $\kappa$.

**Figure 4.5-8: Ellipse detection characteristics for Experiment 4.5.2, 270°.**

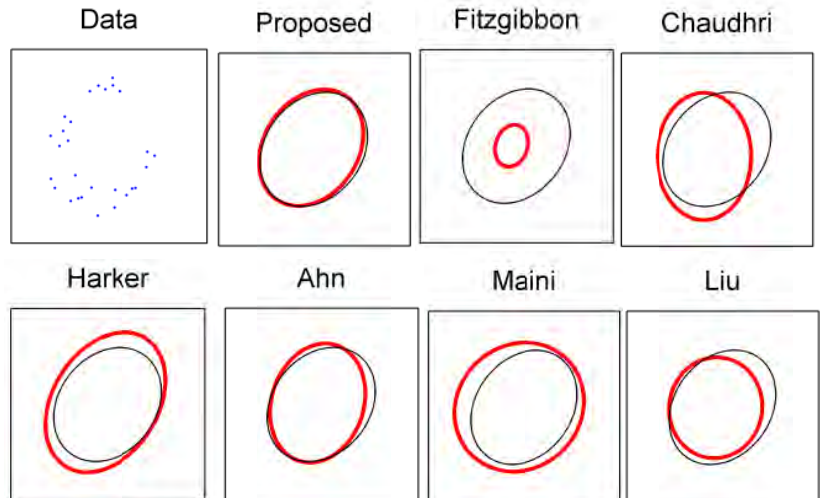

**Figure 4.5-9: An example of noisy cluster of points around an ellipse.**

### 4.5.3 Multiple incomplete ellipses within an image

In this section, a more challenging scenario is considered in which an image may contain several digital ellipses and these may or may not be complete. To generate such images, the setup described as follows is used. To generate the images, consider an image size of $300 \times 300$ pixels and randomly generate $N_E \in \{4, 8, 12, 16, 20, 24\}$ ellipses within the region of the image. The parameters of the ellipses are generated randomly as follows: the center points of the ellipses are arbitrarily located within the image, the lengths of semi-major and semi-minor axes are randomly assigned values in the range $\left[ 10, 300/\sqrt{2} \right]$, and the orientations of the ellipses are also chosen randomly.

The only constraint is that each ellipse must be completely contained in the image and overlap with at least one ellipse. For each value of $N_E$, 100 images containing edge curves of the incomplete ellipses are generated. From the edge map of the image, continuous edge curves with smooth edges (removal of sharp changes in curvature and inflexion points) are obtained. Each edge curve is given as one input to the ellipse detection algorithm. Thus, the number of edge curves may be larger than the number of ellipses in the image and some of the edge curves may be very small. Some example images and the edge curves used to test the ellipse detection are shown in Figure 4.5-10. It is highlighted that the curves have been derived using the techniques in section 5.3.

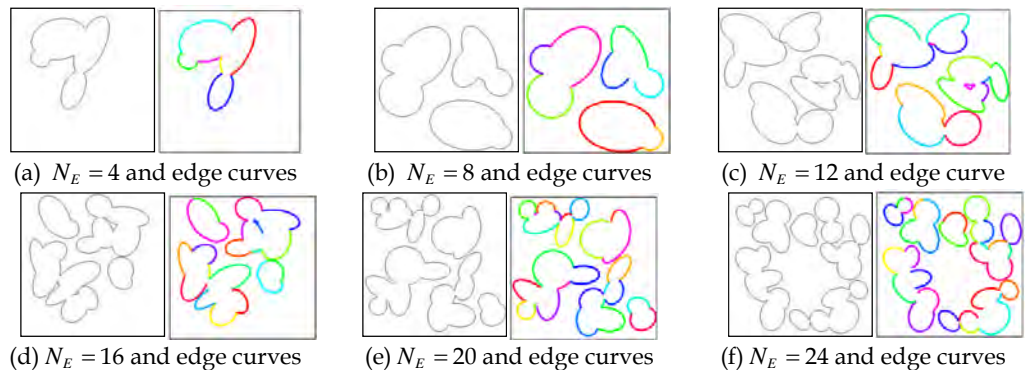

(a) $N_E = 4$ and edge curves    (b) $N_E = 8$ and edge curves    (c) $N_E = 12$ and edge curve

(d) $N_E = 16$ and edge curves    (e) $N_E = 20$ and edge curves    (f) $N_E = 24$ and edge curves

**Figure 4.5-10: Examples of images with multiple incomplete digital ellipses.**

Since there are several ellipses and edge curves in each image, the standard deviation of the error metrics is also considered. Thus, for this experiment, the mean values of the error metrics E13, E14, and $d$ (Figure 4.5-11 (a-c)), as well as their standard deviations Figure 4.5-11(e-f) are plotted.

It is evident that ElliFit outperformed the rest of the methods in terms of E13 and $d$, while it performed similar to or better than most methods in terms of E14. It is notable that Ahn [90] has significantly high mean values for all three error metrics. This is because the misconvergence for one of the edge curves in the image results in very high mean value of the error metrics although the error metrics may have small value for the other edge curves. This is also evident from the standard deviation data plotted in Figure 4.5-11(d-f), where it is seen than Ahn [90] has very high value for standard deviation as well. Similar effect happens with less frequency for Harker [174] as well, occasionally resulting in high mean and standard deviations (for example, see the data of Harker [174] for $N_E$ =12 and $N_E$ =24 in Figure 4.5-11(b,e)).

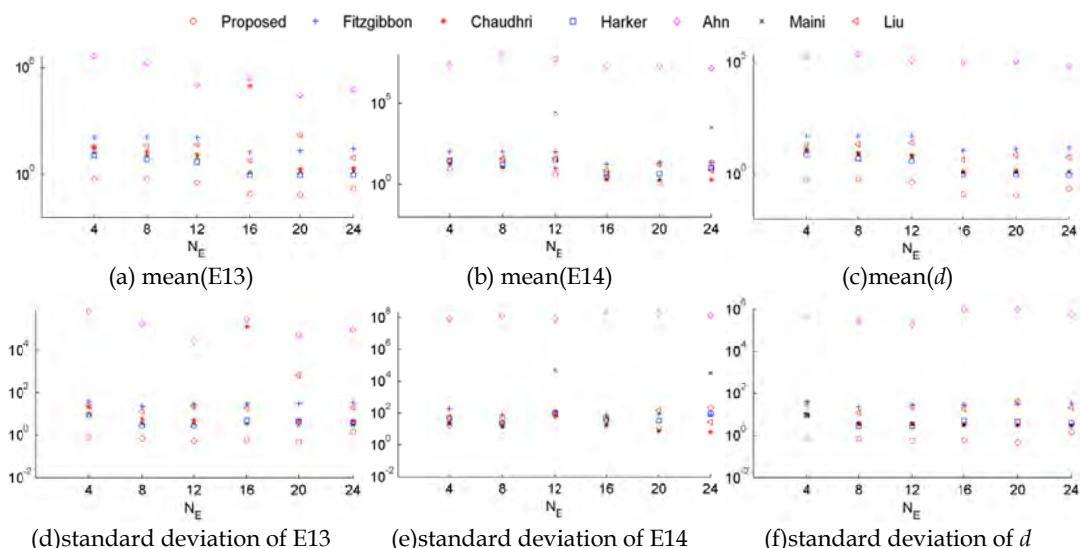

<div align="center">
(a) mean(E13)      (b) mean(E14)      (c)mean($d$)

(d)standard deviation of E13    (e)standard deviation of E14    (f)standard deviation of $d$
</div>

**Figure 4.5-11: Error metrics (mean and standard deviations) for the experiment in section 4.5.3.**

The ellipse detection characteristics are plotted in Figure 4.5-12. In Figure 4.5-12(a), the black line denotes the actual number of ellipses corresponding to different values of $N_E$. The number of ellipses detected by ElliFit is always close to the actual number of ellipses. The recall of ElliFit is the highest among all the methods and is close to one for all values of $N_E$. However, the precision of ElliFit is slightly poorer than the Ahn [90]. This is because the total number of ellipses detected by ElliFit is slightly higher than the actual number of ellipses for all values of $N_E$. Ahn [90] has a slightly poor recall ratio since the number of ellipses detected by Ahn [90] is less than the number of actual ellipses. However, Ahn detected the ellipses with slightly better precision than ElliFit. After ElliFit and Ahn, the next best performance is demonstrated by Harker [174]. Figure 4.5-13 provides some examples to illustrate the ellipse detection results in actual images.

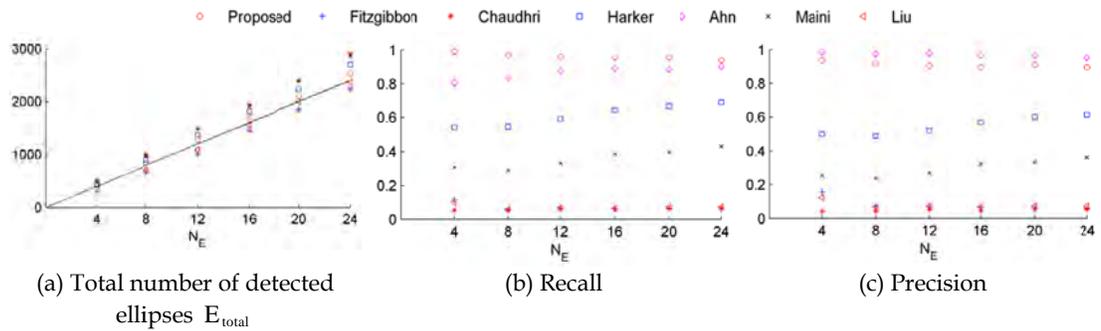

(a) Total number of detected ellipses $E_{total}$

(b) Recall

(c) Precision

**Figure 4.5-12: Ellipse detection characteristics for the experiment in section 4.5.3.**

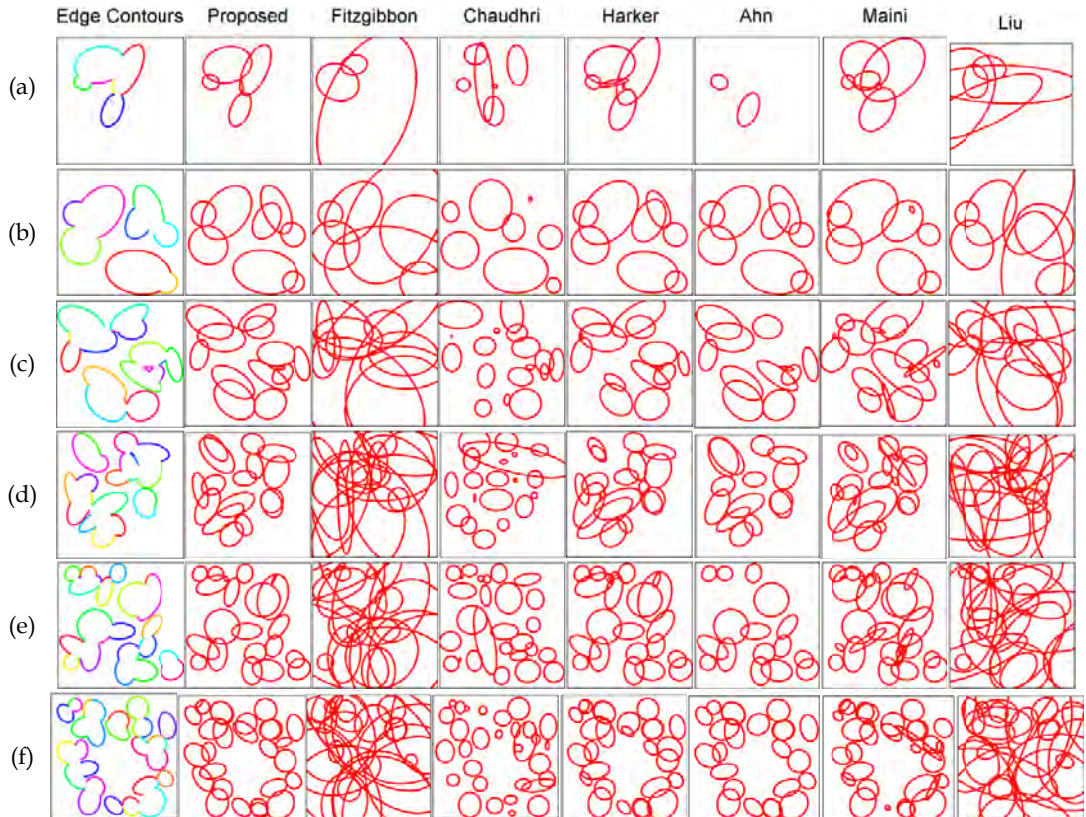

**Figure 4.5-13: Multiple ellipses in an image region (section 4.5.3).**

The scenarios of multiple occluded elliptic shapes is often encountered in biological cell segmentation problems [179] and object detection problems [4]. Both Bai [179] and Chia [4] use Fitzgibbon's method [80] as the ellipse fitting approach. The results in this sub-section demonstrate that such applications will benefit substantially by using ElliFit.

### 4.5.4 Non-elliptic conics – analytical and digital

In this section, non-elliptic conics are considered and it is inspected if the ellipse fitting methods detect ellipses in non-elliptic data. It is desirable in several applications that the ellipses are fitted only for elliptic data and the non-elliptic data is not fitted falsely to an ellipse. This characteristic of selectivity of the ellipse fitting methods to elliptic data is studied in this experiment.

Consider the mathematical model of conics given in eqn. (4-61).

$$x = \frac{l \cos \theta}{1 - e \cos \theta} + x_0 ; y = \frac{l \sin \theta}{1 - e \cos \theta} + y_0 \qquad (4\text{-}61)$$

where $l$ is the semi-latus rectum of the conic and $e$ is the eccentricity of the conic. The following family of conics given by $x_0 = 150$, $y_0 = 150$, $l \in [20,150]$, $e \in [1,2]$ (i.e., parabolae and hyperbolae, but no ellipses) are considered in this analysis. A conic is randomly chosen from this family and is used to generate the corresponding curves using $\theta \in [180° - \Delta\theta, 180° + \Delta\theta]$ in the image region $300 \times 300$ pixels. 1000 such images for each value of $\Delta\theta$ are generated, where $\Delta\theta$ is chosen from 90° to 180° at an interval of 10°. These digital curves are tested by the ellipse fitting methods.

For this experiment, only the number of ellipses detected for each value of $\Delta\theta$. Ideally no non-elliptic conic should be detected as elliptic by the ellipse fitting methods. The total numbers of detected ellipses for various values of $\Delta\theta$ are plotted in Figure 4.5-14(a). It is noted that ElliFit detected no curves as elliptic, while Ahn [90] detected only a few ellipses, and the remaining methods performed poorly by detecting numerous ellipses with false positive rate ranging from ~60% to 100% (about 600 to 1000 ellipses out of 1000, as shown in Figure 4.5-14(a)). This can be attributed to the fact that ElliFit and Ahn, both use geometric models for the least squares fitting of ellipses, hence would have higher selectivity towards the elliptic curves.

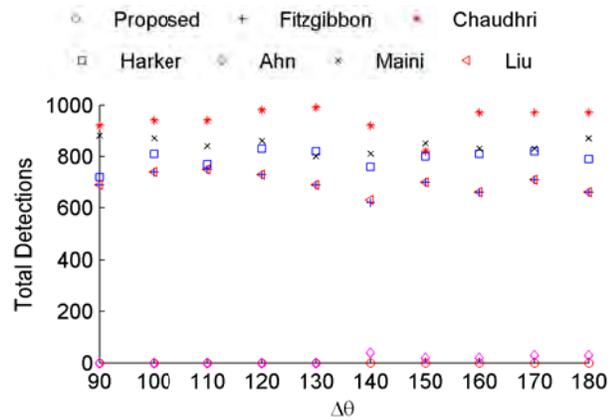

(a) total number of detected ellipses

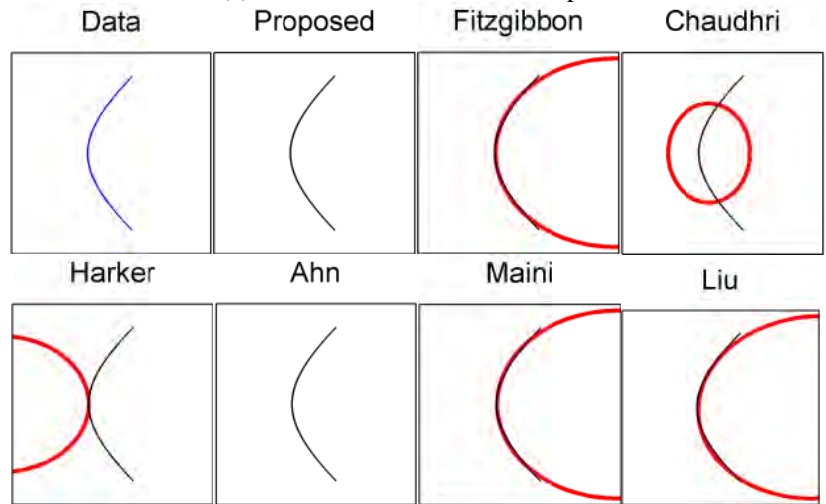

(b) an example and detected ellipses

**Figure 4.5-14: Performance of ellipse detection methods for analytical non-elliptic conics of section 4.5.4.**

The digital curves for the same family of conics are considered. The experimental setup is the same as based on eqn. (4-61) except that the curves are now digitized. The total number of detected ellipses for the digital non-elliptic conics is plotted in Figure 4.5-15(a). ElliFit and Ahn [90] are the only methods detecting no ellipses for all the curves, while the others detected numerous ellipses with a false positive rate ranging from ~65% to 100%. This is in agreement with the discussion presented in the previous paragraph.

Two examples, one for the analytical non-elliptic conic and the other for the digital non-elliptic conics, are presented in Figure 4.5-14(b) and Figure 4.5-15(b) respectively. The curves in black thin line denote the actual conics. Ahn and ElliFit do not detect the ellipses while the remaining methods fitted ellipses on the non-elliptic analytical and digital curves.

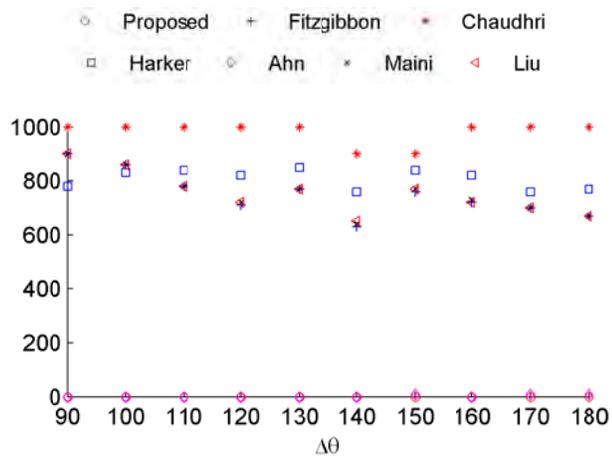

(a) total number of detected ellipses

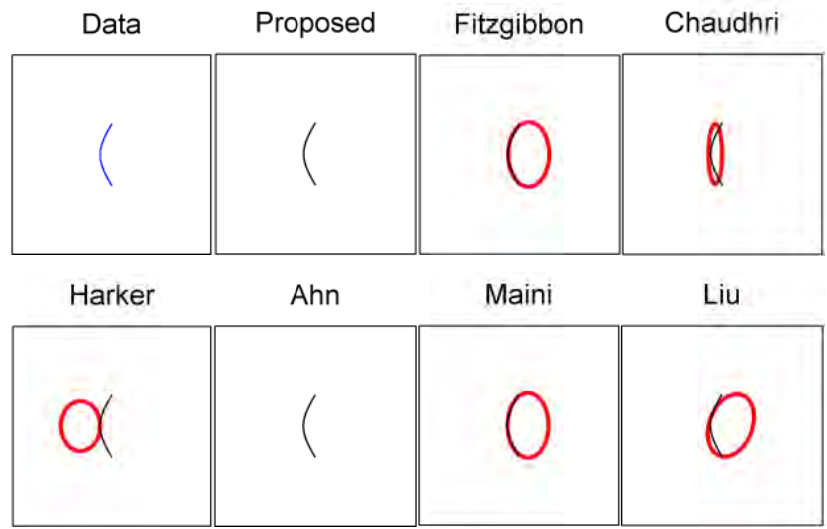

(b) an example and detected ellipses

**Figure 4.5-15: Performance of ellipse detection methods for digital non-elliptic conics of section 4.5.4.**

### 4.5.5 Non-elliptic noisy conics

In this subsection, the case of non-elliptic noisy conics is considered. For the mathematical model of conics in eqn. (4-61) and the family of non-elliptic conics as described in section 4.5.4, the data points on the conics for $\theta \in \left[ 180° - \Delta\theta, 180° + \Delta\theta \right]$ are generated. The data points are restricted in the image region of $300 \times 300$ pixels. Let the set of points be denoted as $\left\{ P_\theta \left( x_\theta, y_\theta \right) \right\}$. Zero mean Gaussian noise is added to the value of coordinates, such that the standard deviations of the noise for $x$ and $y$ coordinates are $\sigma_x = \kappa \max\left|x - x_c\right|/100$ and $\sigma_y = \kappa \max\left|y - y_c\right|/100$ respectively, where $\kappa \in [1,30]; \kappa \in \mathbb{N}$ is the noise percentage. 1000 such images for each value of $\kappa$ and $\Delta\theta$ were generated, where $\Delta\theta$ is chosen from $90°$ to $180°$ at an interval of $10°$. Thus, for each value of $\kappa$, there are 10,000 images.

The total number of detected ellipses for each value of $\kappa$ is plotted in Figure 4.5-16. Even in the case of very high noise $\kappa = 30$, ElliFit has a low false positive rate of 3.61%, that is only 361 ellipses were detected out of 10,000 images from the conic family.

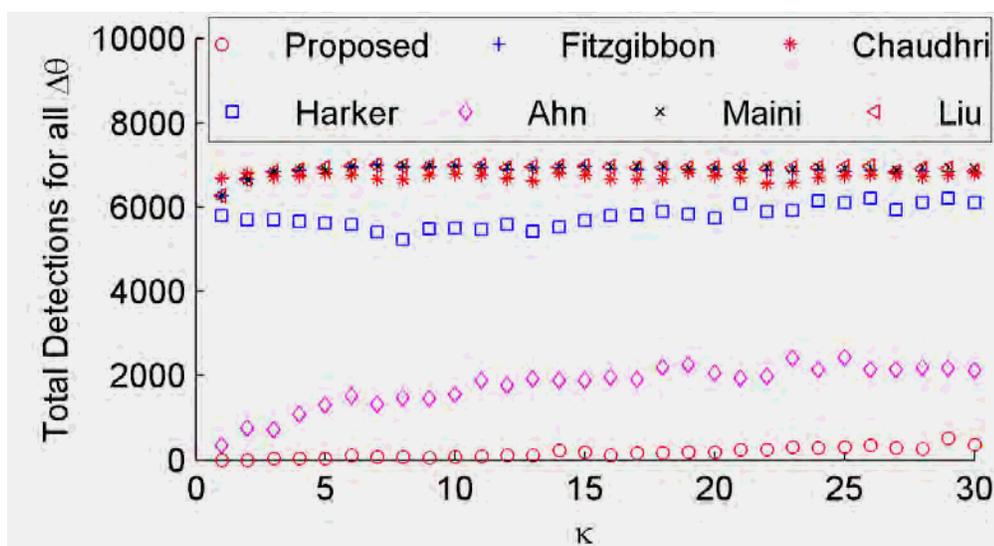

**Figure 4.5-16: Performance of ellipse detection methods for non-elliptic noisy conics of section 4.5.5.**

ElliFit has the least number of detected ellipses for every value of $\kappa$. Ahn [90] performed the second best, though the number of ellipses detected by Ahn rises rapidly as the amount of noise increases. In order to study the impact of the availability of curvature, the total number of detected ellipses for three specific values of $\Delta\theta$, i.e., $\Delta\theta = 90°,135°,180°$ in Figure 4.5-17(a,b,c) are analyzed respectively.

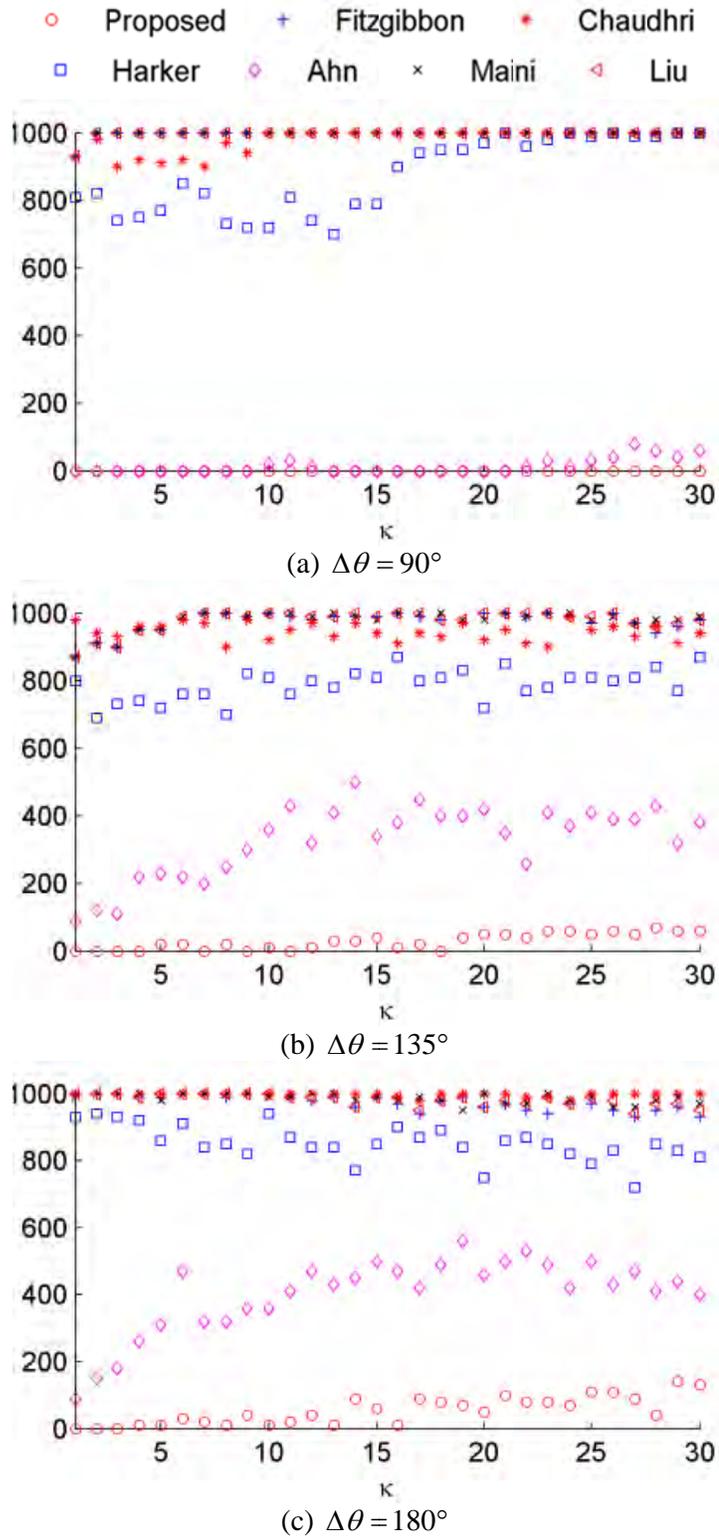

(a) $\Delta\theta = 90°$

(b) $\Delta\theta = 135°$

(c) $\Delta\theta = 180°$

**Figure 4.5-17: Total number of ellipses detected $E^{total}$ for noisy non-elliptic conics of section 4.5.5.**

For $\Delta\theta = 90°$, ElliFit did not detect any ellipse while Ahn [90] detected a few ellipses for high level of noise ($\kappa \geq 25$). Even though the numbers of ellipses detected by ElliFit increase when the value of $\Delta\theta$ is increased, the numbers are significantly less as compared against other methods. Some examples for illustration of the detected ellipses by various methods are shown in Figure 4.5-18. The curves in black thin line denote the actual conics. In Figure 4.5-18(a) and Figure 4.5-18(c), only Ahn [90] and proposed method did not detect the conic as ellipse. Figure 4.5-18(b) is more challenging because the clusters of data points are quite similar to an ellipse. In this case, even Ahn [90] fitted an ellipse and ElliFit is the only method that did not detect the non-elliptic conic as ellipse.

The results in sub-sections 4.5.4 and 4.5.5 demonstrate that ElliFit shows good selectivity of elliptic shapes and low false positive rates for non-elliptic shapes. This is a greatly desirable property for medical and robotic applications that require ellipse detection with low false positive rates and decision making is critically dependent on the correct detection of elliptic shapes.

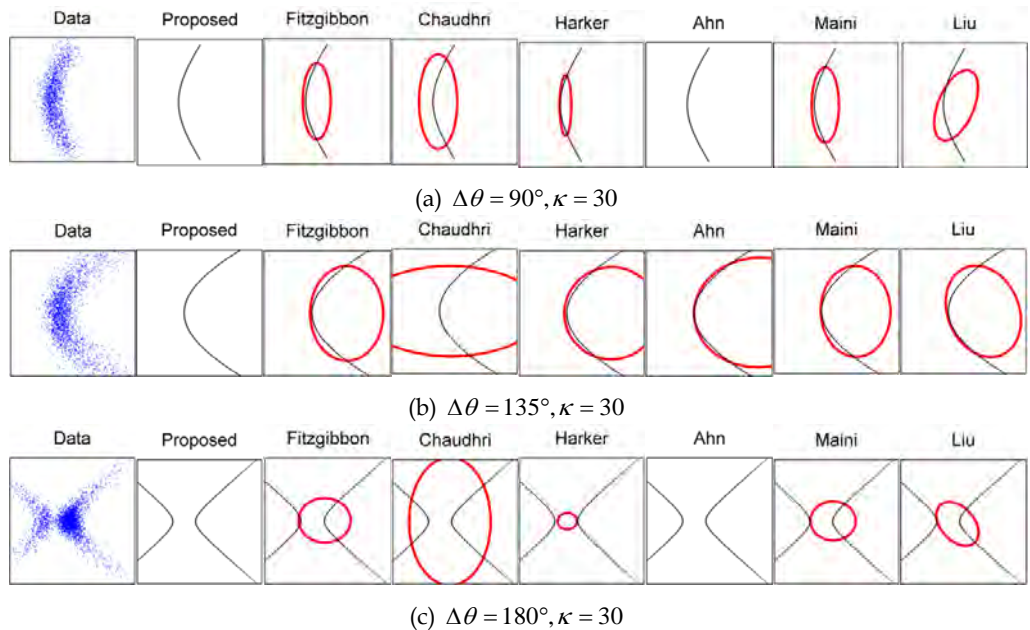

(a) $\Delta\theta = 90°, \kappa = 30$

(b) $\Delta\theta = 135°, \kappa = 30$

(c) $\Delta\theta = 180°, \kappa = 30$

**Figure 4.5-18: Examples of noisy non-elliptic (section 4.5.5).**

## 4.6 Conclusion

For the algebraic fitting of ellipses using Fitzgibbon's method [80], it was identified by Maini [173] that translation of coordinates may reduce the condition number of the scatter matrix. However, the exact reason for this was not discussed in detail and a seemingly heuristic translation was chosen. In this chapter, the exact reason is explained using the theory of statistical moments.

Also, a better translation has been suggested based upon the theory of statistical moments. However, the focus of the chapter is the geometric equation based ElliFit method which uses the concept of minimization of the geometric distance of the data points from the ellipse. ElliFit splits the mathematical problem of ellipse fitting into two operators such that the overall algorithm is non-iterative, does not involve constrained optimization, and is numerically stable. Since the model is based upon the geometric distance and not the algebraic equation of ellipse, the selectivity of the method for a set of elliptic data points is higher than most contemporary methods even if the data points are quite noisy (up to 20% Gaussian noise for positive test data and up to 30% Gaussian noise for the negative test data). ElliFit shows superior performance in terms of several performance parameters like E13 (proposed by Rosin in [86]) and E14 (proposed by Rosin in [88]), mean distance of the set of data points from the fitted ellipse, total number of detections, recall, and precision of fitted ellipses. Empirically, it is noted that only Ahn [90] is the closest technique, which is expected since Ahn [90] also uses geometric distance as the principle concept. However, since Ahn is an iterative non-linear optimization method, it suffers from the problem of local minima and high time complexity. Thus, ElliFit has a significant advantage over Ahn [90]. The contents of this chapter have been reported in [25, 137-139].

# Chapter 5 : Ellipse detection method

## 5.1 Background

Researchers have started using edge contour following methods, in which the connectivity of the edge pixels in the form of edge contours and the continuity of edge contours were used in addition to the mathematical model of ellipse. Though the idea is old [81, 92], the effective use of the idea is fairly recent [33, 34, 68, 97-99]. Evidently, use of these new tools improved the applicability of the ellipse detection methods for simple real images, typically containing one or two ellipses in foreground. These are currently the benchmark in the ellipse detection methods.

The edge contour following techniques group the edge contours based on continuity [33, 34, 68, 97-99]. Considering one edge contour at a time, the edge contour is followed to its ends and other edge contours in the proximity of the edge contour with reasonable angular continuity with the edge contour are found. Such edge contours are then merged with the current edge contour and the newly formed edge contour after merging is followed for finding other edge contours continuous to it. Thus, effectively the parameters of continuity are used as additional constraints to the ellipse detection scheme (which may be based on least squares fitting or Hough transform or random consensus). The use of continuity as the additional constraint introduces three main problems. First, edge contour following needs recursive algorithms. Second, the edge contours that belong to a common ellipse, but are far apart cannot be grouped together based on the continuity. Third, these methods are dependent on many control parameters. For example, the continuity is typically tested using the proximity of the two edge contours and the angular deviation between them [34]. For reliable results, these parameters have to be sufficiently large so that various possible edge contours may be grouped together. However, setting large value of the parameters usually allows many more false positives and deteriorates the performance. Thus, the advantage of using continuity as a constraint is limited by the choice of several control parameters.

An ellipse detection (ED) method that uses the information of edge curvature and convexity in relation to the other edge contours is proposed in this chapter. The method uses edge curvature and convexity as the constraints for the ellipse detection method, instead of the conventionally used continuity constraint [34, 68, 98]. So the method is called the edge curvature and convexity (ECC) based ellipse detector.

A general introduction to the ECC method is presented in section 5.2. The three major portions of the method are presented in sections 5.3, 5.4, and 5.5 respectively. Numerical evaluation of the method and comparison with other methods is presented in section 5.6. The chapter is concluded in section 5.7.

## 5.2 Introduction and novelty of the ECC ellipse detector

The ECC ellipse detector considers a search region for every edge contour that contains other edge contours eligible for grouping with the current edge contour. The edge contours inside the search region of an edge contour and satisfying the associated convexity are considered as the only eligible edge contours for grouping with the considered edge contour. The quality of grouping is further improved by using a two-dimensional Hough transform in an intermediate step, in which a new 'relationship score' is used instead of the conventional histogram count. The new relationship score is used for ranking the edge contours in a group and identifying the lower ranked candidates in a group, which may be subsequently removed from the group if required. The new constraints and the new relationship score improve the grouping of edge contours and consequently the overall performance. Additional thrust in the performance comes from the non-heuristic saliency criteria that are effective in quantifying the goodness of the detected elliptic hypotheses (EH) and finally selecting good EH.

Several salient and distinguishing features of ECC ellipse detector are:

1. Constraints based on the curvature and associated convexity of the edge contours, different from the conventionally used continuity constraint.

2. Relationship score that quantifies the strength of the relationship between a group and its edge contours is used for ranking the edge contours in a group.

3. For extracting edge contours with smooth curvature, a simple PA based inflexion point detection scheme is proposed.

4. The search region used in the proposed method is different from continuity constraint used in general edge contour following methods. Since the search region does not use proximity or angular continuity, it is able to consider distant edge contours as the grouping candidates as well.

5. Simple expression for defining and computing the associated convexity (earlier defined in complicated ways [68, 94]) is proposed.

6. HT is not used for generating EH. It is used for forming the groups of edge contours, ranking the edge contours in a group, and generating evaluation criteria for the least squares method. The use of geometry based criterion in addition to the value of residue for the least squares fitting increases the effectiveness of least squares fitting and reduces the chances of detecting outliers.

7. A saliency scheme that uses three saliency scores for evaluating the goodness of the elliptic hypotheses is proposed. Most contemporary methods do use some form of saliency or distinctiveness criteria [74, 97, 98, 106-109]. The difference in the current method is that three simple and effective saliency criteria are used, which are combined together such that the selection of the salient elliptic hypotheses is non-heuristic and no threshold or control parameters need to be specified for the selection of salient elliptic hypotheses.

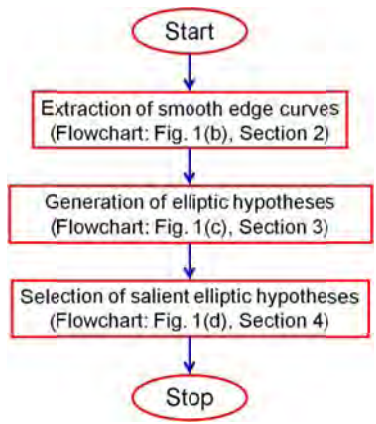

(a) Flowchart of the proposed method

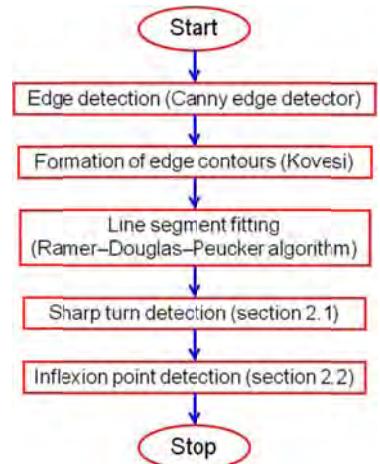

(b) Flowchart for the extraction of smooth edge curves

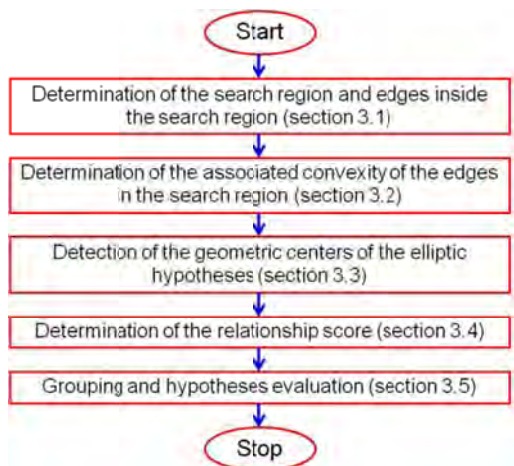

(c) Flowchart for the generation of elliptic hypotheses

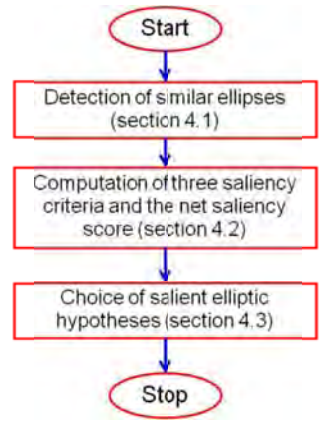

(d) Flowchart for the selection of salient elliptic hypotheses

**Figure 5.2-1: Block diagram of the proposed method.**

The overall framework of the proposed method can be concisely presented as follows. The flowchart of the proposed method is shown in Figure 5.2-1(a). It shows three major blocks: extraction of smooth edge curves, generation of elliptic hypotheses, and selection of salient elliptic hypotheses. In the first block (Figure 5.2-1(b) and section 5.3), the edge contours with smooth curvatures (smoothness as defined in section 5.3) are extracted. This step is motivated by the fact that the curvature of the ellipse does not change sharply. The extraction of smooth edge curves is essential so that the properties of curvature and associated convexity can be used. In the second block (Figure 5.2-1 (c) and section 5.4), considering one edge contour at a time, the edge contours inside the search region of the edge contour and satisfying the associated convexity are found. Using the selected edge contours, the centers of the elliptic hypotheses are computed and Hough Transform (HT) is applied using a method similar to [28, 70]. However, instead of using the conventional general histogram count, a novel relationship score is used here. This step of finding the centers and applying HT is not used for finding the actual parameters of the elliptic hypotheses. Instead, it is used for forming the groups of edge contours that potentially belong to the same ellipse, ranking the edge contours within a group, and for providing a criterion that is used for judging the elliptic hypotheses generated in the next step. After forming the groups using HT, least squares method [80] is used to evaluate the group and generate the elliptic hypotheses. The result of least squares method is evaluated using the residue in the least squares formulation as well as the center generated using the HT. If the criteria are not satisfied, poorly ranking edge contours are dropped from the group in order to improve the quality of the group. In the third block (Figure 5.2-1(d) and section 5.5), three novel saliency criteria for selecting salient EH are proposed. The three criteria are combined and selection is performed such that no heuristic selection of thresholds needs to be done.

## 5.3 Pre-processing – obtaining edge contours of smooth curvature

It is well known that the curvature of any elliptic shape changes continuously and smoothly. So, edge contours with smooth curvature are extracted in the first step. The term smooth curvature is defined here as follows. A portion of an edge contour which does not have a sudden change in curvature, either in terms of the amount of change or the direction of change, is called here as a smooth portion of edge contour. It should be noted that no smoothing operation of any kind has been applied. The edge curves are extracted from the existing data itself. The flowchart of the process of extracting the smooth edge curves is shown in Figure 5.2-1(b).

First, the input image is converted to gray scale. It is preferable in the case of real image to perform histogram equalization on the gray image for improving the contrast and enhancing the boundaries. Next, the edge pixels are extracted from the gray image using the canny edge detector [172]. The control parameters for Canny edge detector used in this chapter are as follows: low hysteresis threshold $T_L = 0.1$ , high hysteresis threshold $T_H = 0.2$, and standard deviation for the Gaussian filter $\sigma = 1$ . This choice of control parameters works satisfactorily for most of the images. This edge map is then used to derive non-branched edge contours. For this purpose, Kovesi's codes [113] of junction removal and extracting connected edge contours have been used. Edge contours less than 5 pixels long are excluded from further consideration since they may be the effect of noise or do not contribute in ellipse detection process due to the lack of curvature.

After extracting the edge contours, approximate polygon of each edge curvature is obtained using RDP-mod (proposed in section 2.5.1.2) PA method. After fitting the approximate polygons on the edge contours, the edges of the polygons are used for extracting smooth edge curves. In order to avoid confusion between the edges of the approximate polygon and the edge contour, the edges of the approximate polygon shall be referred to as line segments. Beginning from one end of the edge contour, the points where the curvature becomes irregular are sought. The edge contour is broken at the points (vertices of the approximate polygons) where the curvature becomes irregular, such that every new edge contour formed out of this process is a smooth edge contour. In this context, the regularity is defined in two aspects – the amount of change of curvature (sharp turns), and the change of direction of curvature (presence of inflexion points). In the following two subsections, the methods for detecting sharp turns and inflexion points are proposed.

### 5.3.1 Sharp turns detection

Consider an edge contour $e$, for which approximate polygon (AP) has been computed. Let the sequence of line segments that represent the edge contour $e$ be $\{l_1, l_2, \ldots, l_N\}$. Let the angles between all the pairs of consecutive line segments be denoted as $\{\theta_1, \theta_2, \ldots, \theta_{N-1}\}$, where $\theta_i \in [-\pi, \pi]$ is the anticlockwise angle from $l_{i+1}$ to $l_i$. An example is shown in Figure 5.3-1(a).

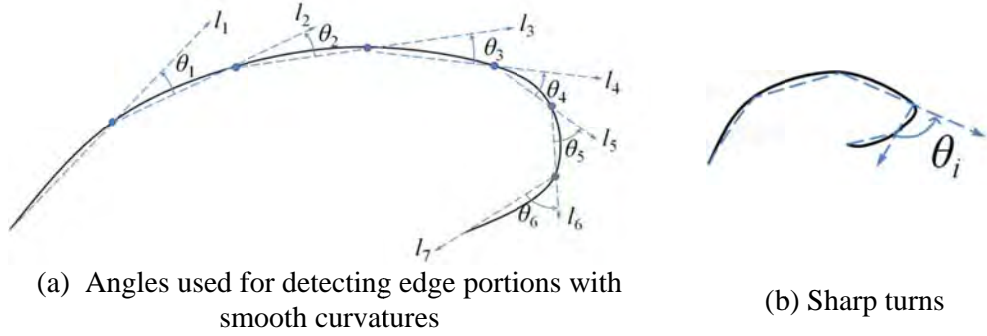

(a) Angles used for detecting edge portions with
smooth curvatures

(b) Sharp turns

**Figure 5.3-1: Calculation of angles for detecting edge portions with
smooth curvatures and sharp turns.**

The value of angle $\theta_i$ is an indicator of the change in the curvature of the edge contour. If $\theta_i$ is small, then it implies that the change in the curvature of the edge contour is small. On the other hand, if $\theta_i$ is large, it implies that the change in the curvature of the edge contour is large.

In the sequence of the angles, $\{\theta_1, \theta_2, \ldots, \theta_{N-1}\}$, if any angle $\theta_i$ is large, i.e., equal to or greater than a chosen threshold $\theta_0$ (say $\pi/2$, empirically determined), then the change in the curvature of the edge contour at such points $P_{i+1}$ (the intersection point of line segments $l_i$ and $l_{i+1}$) is considered to be large (or sharp), and the edge contour is split at $P_{i+1}$ to ensure smooth curvature. An example is shown in Figure 5.3-1(b).

### 5.3.2 Inflexion points detection

Considering the sequence of angles defined in section 5.3.1, $\{\theta_1, \theta_2, \ldots, \theta_{N-1}\}$, the change of the sign of the angles (negative or positive) represents the change in direction of the curvature. Thus, a Boolean sequence $\{b_1, b_2, \ldots, b_{N-1}\}$ is created, where $b_i$ is given as follows:

$$b_i = \begin{cases} 1 & |\theta_i + \theta_1| < |\theta_i| + |\theta_1| \\ 0 & \text{otherwise} \end{cases} \tag{5-1}$$

The value of $b_i$ is '0' if the signs of $\theta_i$ and $\theta_1$ are the same and $b_i$ is '1' if the signs of $\theta_i$ and $\theta_1$ are different. This Boolean sequence can be used to identify the inflexion points and decide the exact places where the edge contour should be split. It is worth noticing that there are three possibilities of the occurrence of inflexion points:

1. $b_i = 1$ AND $b_{i-1} = b_{i+1} = 0$ (in words, one '1' Boolean element between two '0' Boolean elements).

2. $b_i = 1, b_{i+1} = 1$ AND $b_{i-1} = b_{i+2} = 0$ (in words, two '1' Boolean elements between two '0' Boolean elements).

3. $b_{i-1} = 0$ AND $b_i = b_{i+1} = b_{i+2} = 1$ (in words, more than two '1' Boolean elements after one '0' Boolean element).

The possibilities, the point of splitting, and the graphical representations of the possibilities are presented in Table 5.3-1.

**Table 5.3-1: Inflexion points: various possibilities**

| Sl. No. | Sequence | The point of split | Graphical illustration (split at the highlighted points) |
|---|---|---|---|
| 1 | $b_i = 1$ AND $b_{i-1} = b_{i+1} = 0$ | $P_{i+1}$ | $P_{i+1}$ ... 0 ① 0 ... $b_i$ |
| 2 | $b_i = 1, b_{i+1} = 1$ AND $b_{i-1} = b_{i+2} =$ | $P_{i+1}, P_{i+2}$ | $P_{i+1}$ $P_{i+2}$ ... 0 ① ① 0 ... $b_i$ $b_{i+1}$ |
| 3 | $b_{i-1} = 0$ AND $b_i = b_{i+1} = b_{i+2} = 1$ | $P_{i+1}$ | $P_{i+1}$ ... 0 ① 1 1 ... $b_i$ |

Based on the absence or presence of sharp turns and inflexion points, there may be none or many points at which an edge contour needs to be split in order to obtain smooth contours. If there are $N'$ such points on an edge contour, the edge contour can be split at these points to form ($N'+1$) smaller edge contours of smooth curvature. Here, it is highlighted that other algorithms may be used for removing inflexion points [114]. However, the propose algorithm deals with more cases of inflexion points in comparison to [114].

## 5.4 Grouping and elliptic hypotheses evaluation

After extracting smooth edge curves in section 5.3, the properties of the smooth curvature are used to form groups of edge contours that are effective for generating EH. The flow chart of generation of EH is shown in Figure 5.2-1(c). The first four steps are used for finding the edge contours that are suitable for grouping. The actual grouping and elliptic hypotheses generation is done in the last step. The details are presented in the following subsections.

### 5.4.1 **Search region and edge contours inside it**

Given an edge contour of smooth curvature, a search region can be found such that the edge contours outside the search region of an edge contour can be safely excluded from grouping with the considered edge contour. Such exclusion is possible by defining a suitable search region such that the edge contours inside the search region are the only edge contours that may be grouped with the considered edge contour for forming ellipses. Such a search region is shown in Figure 5.4-1(a) . The shaded region $R_2$ is the search region and the edge contours that do not lie completely inside the search region will not be grouped with $e_1$ .

#### *5.4.1.1 Finding the search region*

The search region $R$ is defined as follows. For a given edge contour $e$ , let the tangents to the edge contour at its end points $P_1$ and $P_2$ be denoted by $l_1$ and $l_2$ , and the line segment connecting the end points $P_1$ and $P_2$ be denoted by $l_3$ . The tangents to the edge contours are computed using DEB (proposed in Chapter 3, $R = 4$ has been used). The two tangents $l_1$ and $l_2$ and the line $l_3$ divide the image into two regions, $R_1$ and $R_2$ (notations used only for nomenclature), as shown in Figure 5.4-1. The shaded region shows the search region for the edge contour $e_1$ . Based on the search region, it can be concluded that the edge contours $e_2$ , $e_3$ , and $e_5$ cannot be grouped with $e_1$ for generating elliptic hypotheses. The search region can be found using the lines $l_1$ , $l_2$ , and $l_3$ and the midpoint of the edge contour $P_{mid}$ shown in subfigure (b). The search region $R$ is the region that does not contain $P_{mid}$ , where $P_{mid}$ is the middle pixel of the edge contour $e$. See Figure 5.4-1(b) for illustration. Mathematically, the search region $R$ is defined as follows:

$$R = \begin{cases} R_2 & P_{mid} \notin R_2 \\ R_1 & \text{otherwise} \end{cases} \tag{5-2}$$

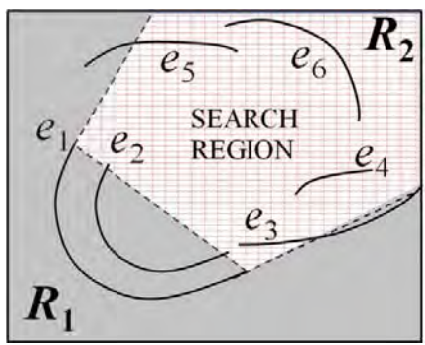
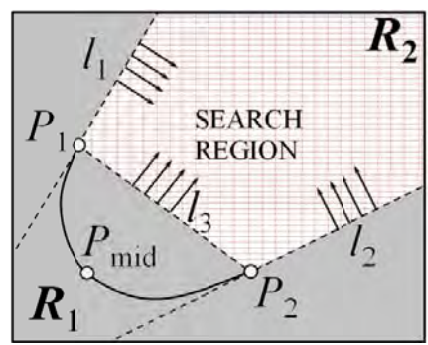

(a) Illustration of the search region    (b) Definition of the search region

**Figure 5.4-1: Illustration and detection of the search region.**

In Figure 5.4-1, $R_1$ and $R_2$ are just symbolic representation of two regions. If the nomenclature is changed in Figure 5.4-1(b) then $P_{mid}$ will not be in $R_1$. Thus, according to eqn. (5-2), $R_1$ will be the search region.

### 5.4.1.2 Finding the edge contours within the search region of an edge contour

For a given edge contour $e$, after finding its search region $R$, the edge contours within the search region can be found as follows. An edge contour $e_i \in R$ if all the three criteria S1-S3 below are satisfied:

- Search region criterion 1 (S1): $e_i$ and $P_{mid}$ are on the same side of $l_1$,

- Search region criterion 2 (S2): $e_i$ and $P_{mid}$ are on the same side of $l_2$, and

- Search region criterion 3 (S3): $e_i$ and $P_{mid}$ are on the opposite sides of $l_3$.

### 5.4.2 Associated convexity of the edge contours inside the search region

The associated convexity of a pair of edge contours can be studied in order to further exclude grouping of the edge contours that are not suitable (for grouping). Figure 5.4-2(a-e) shows five scenarios of the associated convexity between two edge contours. It is evident that the scenario presented in Figure 5.4-2(e) is the only scenario that should be considered for optimal grouping. A simple method is proposed below that can identify if the two edge contours have their associated convexity as shown in Figure 5.4-2(e).

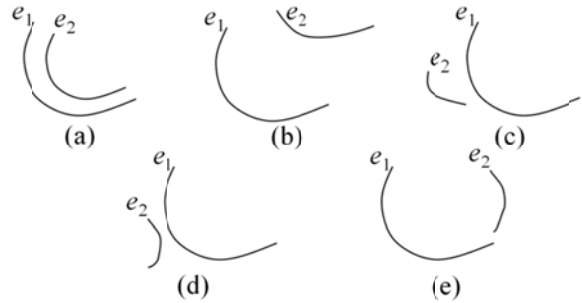

**Figure 5.4-2: Possible associated convexities between two edge contours.**

Consider the line segments $l_1$ and $l_2$ formed by joining the end points of $e_1$ and $e_2$, respectively, as shown in Figure 5.4-3. Let $P_1$ and $P_2$ be the midpoints of the line segments $l_1$ and $l_2$. Let $l_3$ be a line passing through $P_1$ and $P_2$, such that it intersects the edge contours $e_1$ and $e_2$ at $P_1'$ and $P_2'$ respectively. This is illustrated in Figure 5.4-3.

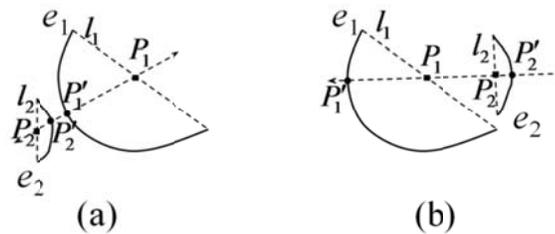

**Figure 5.4-3: Illustration of the concept of associated convexity.**

The pair of edge contours $e_1$ and $e_2$ are suitable for grouping if and only if:

$$P_1'P_2' \approx P_1P_1' + P_1P_2 + P_2P_2' \,. \tag{5-3}$$

The approximation is attributed to the fact that $P_1'$ and $P_2'$ have to be the edge pixels nearest to the line $l_3$, and may not be exactly on $l_3$ due to digitization of the edge pixels.

### 5.4.3 **Detection of the geometric centers of EH**

Yuen's three point method presented in section 3.2.1 is used for retrieving the centers [28]. The proof of its geometrical concept is provided in [75]. There are two geometric exceptions to Yuen's method. The first case is that the tangents $t_1$ and $t_2$, or $t_2$ and $t_3$ are parallel to each other. The second case is that the line segments $l_{12}$ and $l_{23}$ are parallel to each other (i.e., the pixels in the set are collinear, though they may belong to a curved edge contour). The second case is unlikely to appear if the edge contours are of smooth curvature. In both these cases, the center of the ellipses cannot be found. In order to generate reliable estimate of the ellipse's center, many sets of points have to be generated for each edge contour.

Here, readers are referred to [94] for studying the effect of selection of points. Their work suggests that every pair of points in a set (of three points) should satisfy these conditions such that chances of occurrence of the above mentioned problems are reduced: (1) proper convexity, (2) proper distance between the points, and (3) reasonable angle between the tangents at these points. From this perspective, it is reasonable and simple to split an edge contour into three sub-edge contours and choose points sequentially from each sub-edge contour. However, if the edge contour is not long enough, using this method will not ensure sufficiently high number of sets. Thus, it is reasonable to split an edge contour into three sub-edge contours and choose points randomly from each sub-edge contour to form the desired number of sets.

The digitization in the images results into the computation of centers to be inaccurate [25]. In effect, the centers computed above cannot be used directly because the centers calculated from various sets may be close but not exactly the same. So, for improving the robustness, the parametric space of centers is quantized into bins. Let the input image of size $M \times N$ pixels be divided into $B_m$ and $B_n$ equally sized bins along the rows and columns respectively (total number of bins is $B = B_m B_n$). Thus, each bin is of size $m \times n$ pixels, where $m = M/B_m$, $n = N/B_n$.

### 5.4.4 **Determination of the relationship score**

Given an edge contour $e$, $S$ sets of three pixels are generated. For each set, a center can be computed using the geometric concept in 5.4.3. As discussed in section 5.4.3, all the sets may not generate valid centers. Let the number of sets that generated a valid center be $S_e$. Ideally, all $S_e$ centers should fall in the same bin (which should coincide with the bin containing the center of the actual ellipse). In practice, all the computed centers will not fall in the same bin.

A score is assigned to each bin-edge contour pair, which shall be referred to as the relationship score. The relationship score is an indicator of the trust that can be put upon their relationship. All the edge contours in a single bin are considered as a group (potentially belonging to a common ellipse). However, they are ranked on the basis of their score within a group.

A simple relationship score is presented in eqn. (5-4).

$$\tilde{r}_e^b = S_e^b \qquad (5\text{-}4)$$

where $S_e^b \leq S_e$ is the number of sets of the edge contour $e$ that voted for the bin $b$. The count of the votes for bin $b$, commonly referred to as the general histogram count, is therefore $\sum_{\forall e} \tilde{r}_e^b$. Below, the novel relationship score $r_e^b$ (which can be considered as an enhancement of $\tilde{r}_e^b$) is proposed in eqn. (5-5).

$$r_e^b = S_e^b \, r_1 \, r_2 \qquad (5\text{-}5)$$

where, $r_1$ is a function of $S_e^b / S_e$ and $r_2$ is a function of $S_e / S$. For avoiding confusion, it is reiterated that $\tilde{r}_e^b$ of (5-4) is used for general histogram count, and $r_e^b$ of eqn. (5-5) is the proposed relationship score.

The ratio $S_e^b / S_e \in [0,1]$ is an indicator of the relative weight of the bin $b$ as compared to other bins that were computed for the same edge contour. If $S_e^b / S_e$ is high, the bin is better ranked than the rest of the bins, indicating that the relation between the edge contour and the bin $b$ is stronger, and thus should be given more priority. On the other hand if $S_e^b / S_e$ is less, the bin might have been computed by a chance combination of the randomly selected pixels, and should not be given significant importance.

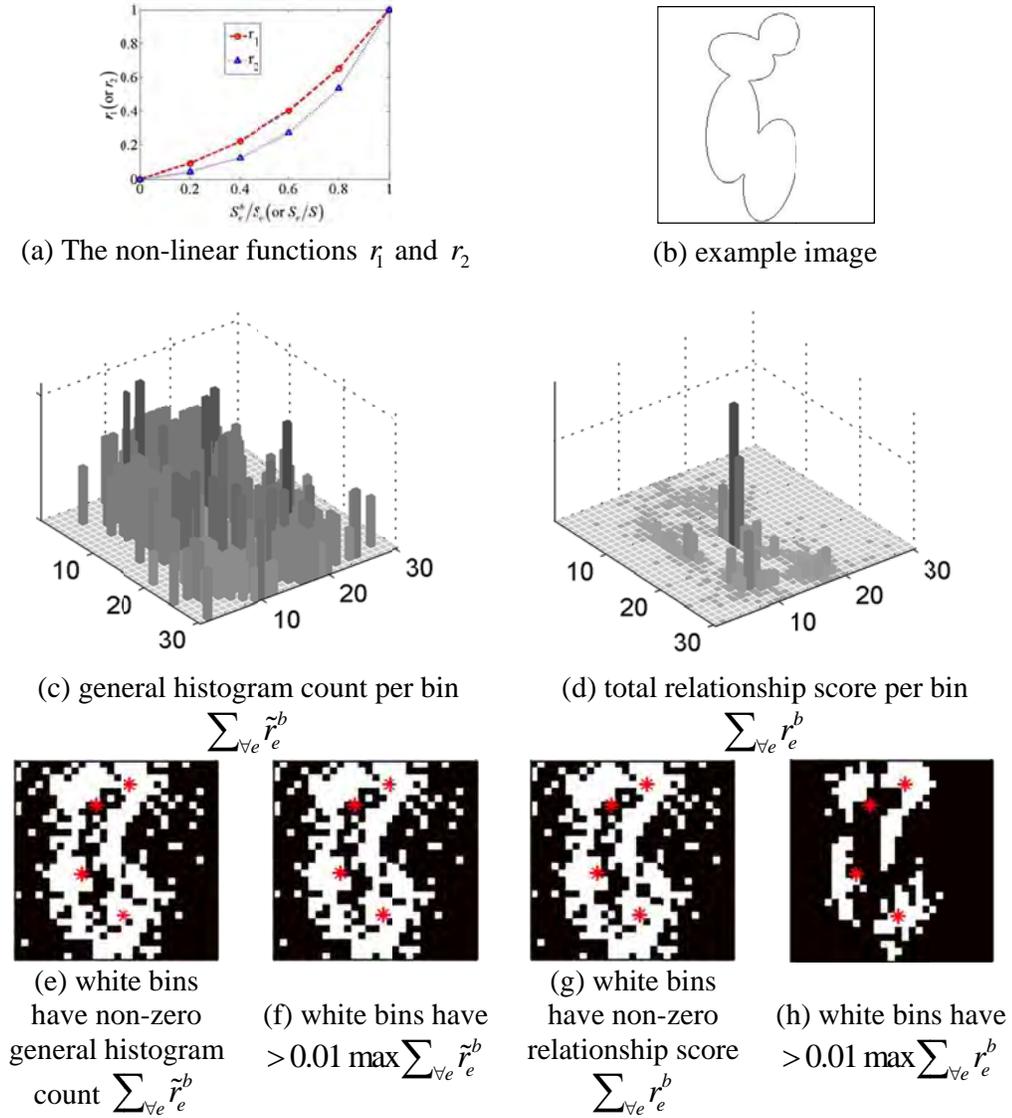

(a) The non-linear functions $r_1$ and $r_2$      (b) example image

(c) general histogram count per bin      (d) total relationship score per bin

(e) white bins have non-zero general histogram count $\sum_{\forall e} \tilde{r}_e^b$    (f) white bins have $> 0.01 \max \sum_{\forall e} \tilde{r}_e^b$    (g) white bins have non-zero relationship score $\sum_{\forall e} r_e^b$    (h) white bins have $> 0.01 \max \sum_{\forall e} r_e^b$

**Figure 5.4-4: The proposed relationship score and its effect.**

**Asterisks in (e-h) denote the actual centers of the ellipses in (b).**

A non-linear $r_1$ as shown in Figure 5.4-4(a) is preferred to dampen $r_e^b$ for low $S_e^b / S_e$. The ratio $r_1$ is computed as in eqn. (5-6).

$$r_1 = \left( \frac{S_e^b}{S_e} \right) \exp\left( \frac{S_e^b}{S_e} - 1 \right),$$
(5-6)

where $\exp(\bullet)$ denotes the exponential function. Though other functions might be chosen to achieve similar effect, this is not the scope of the current work to compare with other types of functions. Here, it suffices to say that the above function emulates well the desired effect.

As discussed before, out of the total $S$ sets generated for an edge contour, all may not result into valid bins. If there are only a few valid sets $S_e$ in comparison to $S$, it may mean that the edge contour is a poor elliptic candidate (and thus an outlier). On the other hand, if the ratio $S_e/S \in [0,1]$ is high, then it is indicative of the edge contour being a good elliptic arc. Similar to $r_1$, $r_2$ is defined as in eqn. (5-7).

$$r_2 = \left( \frac{S_e}{S} \right) \exp \left( 2 \left( \frac{S_e}{S} - 1 \right) \right), \qquad (5\text{-}7)$$

where $r_2$ has stronger dampening effect than $r_1$. It should also be noted that while $r_1$ is indicative of relative importance of a bin (among various bins computed for an edge contour), $r_2$ is indicative of the relative trust of an edge contour (in comparison to other edge contours).

In order to illustrate the effect of using the relationship score as compared to the general histogram typically used in HT, a simple image of size $300 \times 300$ pixels shown in Figure 5.4-4(b) is considered. $m = n = 10$ pixels is used for forming the bins. The total histogram count per bin $\sum_{\forall e} \tilde{r}_e^b$ and total relationship score per bin $\sum_{\forall e} r_e^b$ are plotted in Figure 5.4-4(c,d) respectively. The bins which received non-zero (white) histogram count and relationship score are plotted in Figure 5.4-4(e,g) respectively. Figure 5.4-4(f,h) show bins with values $> 0.01 \max \sum_{\forall e} \tilde{r}_e^b$ and $> 0.01 \max \sum_{\forall e} r_e^b$ respectively. In Figure 5.4-4(e-h), the actual centers of the ellipses are shown using asterisks (*).

It is seen that Figure 5.4-4(e-g) are exactly the same. The similarity of Figure 5.4-4(e,g) can be understood from the fact that if there is a non-zero vote in a bin, both $\sum_{\forall e} \tilde{r}_e^b$ and $\sum_{\forall e} r_e^b$ are non-zero. On the other hand, if a bin has zero votes, both $\sum_{\forall e} \tilde{r}_e^b$ and $\sum_{\forall e} r_e^b$ are zero. Figure 5.4-4(f) is same as Figure 5.4-4(e) because no bins have been filtered away when 1% filtering is used on $\sum_{\forall e} \tilde{r}_e^b$. On the other hand, with 1% filtering on $\sum_{\forall e} r_e^b$, many bins are filtered away, as shown in Figure 5.4-4(h). It is clearly visible that the proposed relationship score is more selective than the general histogram count even when only 1% filtering is used for $\sum_{\forall e} \tilde{r}_e^b$ and $\sum_{\forall e} r_e^b$ to eliminate irrelevant center hypotheses.

Thus, the specific advantages of the proposed relationship score over the conventional histogram count used in Hough transform are listed as follows:

1. The relationship score is more selective than the conventional histogram count. Since this step is used for generating initial guess for grouping, the selective nature of relationship score is expected to generate fewer but better groups. This increases the reliability of this step as initial guess and reduces the computational complexity of the complete scheme.

2. While the general histogram count (or $\tilde{r}_e^b$) is insensitive to the information in $S_e^b / S_e$ and $S_e / S$ regarding the edge contour, the relationship score $r_e^b$ uses this information to generate a fairer score for ranking the edge contours within a group.

### 5.4.5 Grouping and hypotheses evaluation

The following method is used for grouping the edge contours that possibly belong to a common ellipse and finding the remaining parameters of the elliptic hypothesis. Here, grouping does not mean physical merging or connecting of the edge contours. In the current context, grouping means collecting the edge contours that may possibly belong to same ellipse as one set.

After assigning the relationship score $r_e^b$, all the edge contours having a common bin $b$ may initially be considered as a group. The various edge contours in a group are ranked based on their bin-edge contour relationship scores $r_e^b$. The edge pixels of the edge contours in a group are appended in the descending order of their relationship scores $r_e^b$ and NSAF method of ellipse fitting proposed in section 4.2.2 is used on the group to find all the parameters of the ellipse. Now, the quality of this group is evaluated based on the two criteria listed below:

- Criterion 1 (C1): Error of least squares fitting $\leq \varepsilon_{ls}$, a chosen threshold error value.

- Criterion 2 (C2): The center bin $b$ of the group is inside the detected elliptic hypothesis. If the quadratic equation of the ellipse is given by $f(x, y) = 0$, then a bin $b$ is inside the ellipse if the center of the bin $(x_b, y_b)$ is such that $f(x_b, y_b) < 0$.

If both C1 and C2 are satisfied, then the parameters of the ellipses computed using the least squares fitting are given as output. If anyone of the two criterions is not fulfilled, the weakest edge contour (with the lowest relationship score $r_e^b$) is removed from the group and the above process is repeated till either the above criteria are satisfied or the group becomes empty.

## 5.5 Saliency and elliptic hypotheses selection

The above elliptic hypotheses generation scheme may generate multiple elliptic hypotheses corresponding to an actual ellipse because various groups may correspond to a common ellipse. The elliptic hypotheses selection is performed in two stages. In the first stage, the hypotheses that are similar to each other are found out and only one representative elliptic hypothesis is kept among them (section 5.5.1). This increases the chances of one elliptic object being represented by a single elliptic hypothesis. In the second stage, the elliptic hypotheses that remain after the clustering are evaluated for their saliency. Three kinds of saliency criteria and methods to combine them are proposed in sections 5.5.2, 5.5.3, and 5.5.4.

### 5.5.1 Detection of similar ellipses

Let the quadratic equation describing an ellipse be given by eqn. (5-8).

$$f(x, y) = 0 \tag{5-8}$$

A pixel $P(x', y')$ in the image belongs to the elliptic region (inside or on the boundary of the ellipse) if $f(x', y') \leq 0$. Using this concept, the image pixels outside the ellipse can be labeled Boolean '0', while the rest can be labeled Boolean '1'. Let the two dimensional matrix that stores these Boolean variables for various pixels be denoted as $I$. In order to find if two ellipses $E_1$ and $E_2$ are similar, the Boolean matrices $I_1$ and $I_2$ are formed using the quadratic functions $f_1(x, y)$ and $f_2(x, y)$ respectively. The following similarity measure is then defined as:

$$D = 1 - \frac{\text{count}(XOR(I_1, I_2))}{\text{count}(OR(I_1, I_2))} \tag{5-9}$$

where count$(A)$ gives the number of Boolean '1' elements in the matrix $A$. Here, $XOR(I_1, I_2)$ gives Boolean values '1' at the pixels which belong to only one of the ellipses (and not both), i.e. the non-overlapping region of the ellipses. $OR(I_1, I_2)$ gives the total region jointly overlapped by the two regions. Thus $D$ gives the ratio of overlap of the two ellipses. The above overlap ratio is an adaptation of the Jaccard index used in set theory [178]. The concept is illustrated graphically in Figure 5.5-1.

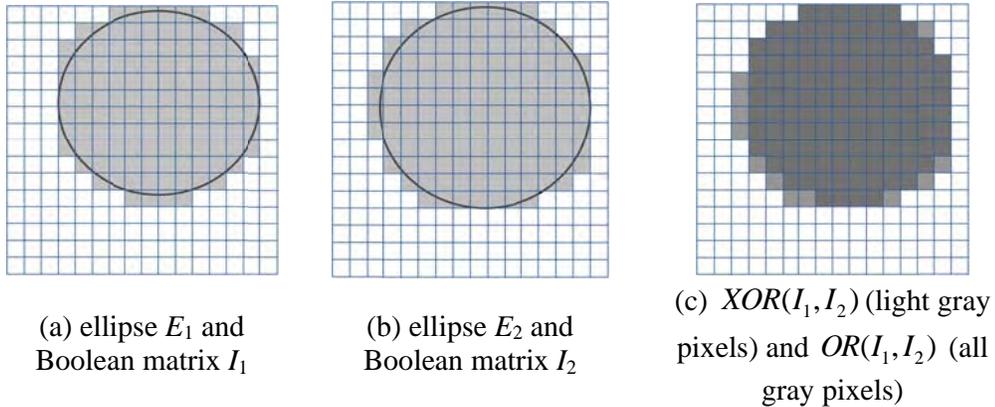

(a) ellipse $E_1$ and Boolean matrix $I_1$

(b) ellipse $E_2$ and Boolean matrix $I_2$

(c) $XOR(I_1, I_2)$ (light gray pixels) and $OR(I_1, I_2)$ (all gray pixels)

**Figure 5.5-1: Illustration of the concept of overlap measure.**

For a given ellipse, all the ellipses that have overlap ratio $D > D_0$; $D_0 = 0.9$ are clustered together. Among the ellipses in a cluster, the choice of representative candidate should depend upon the reliability of the ellipses in the cluster. One way of determining the reliability is to choose the ellipse that was formed by maximum amount of data. Thus, the angular circumference ratio (introduced in section 5.5.2.1) has been used for this purpose.

### 5.5.2 Computation of three saliency criteria

The following three criteria are used for computing the saliency of the elliptic hypotheses:

1. Angular circumference ratio

2. Alignment ratio

3. Angular continuity ratio

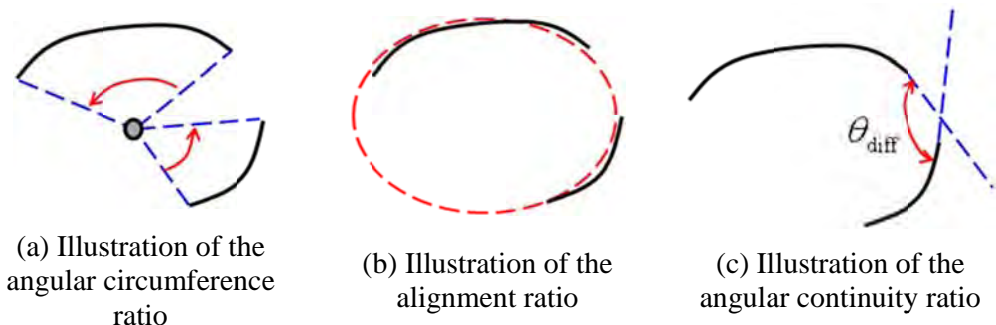

(a) Illustration of the angular circumference ratio

(b) Illustration of the alignment ratio

(c) Illustration of the angular continuity ratio

**Figure 5.5-2: Illustration of the three saliency criteria.**

### *5.5.2.1 Angular circumference ratio*

Instead of using the pixel count feature [106], the angle subtended by an edge contour on the center of the ellipse is used as the measurement of circumference (as illustrated in Figure 5.5-2(a)). Such scheme is less sensitive to the quantization problem in the pixel count feature. Suppose an ellipse $E$ was fitted to a group $G$, then the angular circumference ratio $c(E,G)$ is defined as below:

$$c(E,G) = \frac{\sum_{\forall e \in G} \alpha(E,e)}{2\pi} \tag{5-10}$$

where $\alpha(E,e)$ is the angle subtended by the ends of the edge contour $e$ at the center of the ellipse $E$. A higher value of $c(E,G)$ implies a larger support of $E$ on $G$.

### *5.5.2.2 Alignment ratio*

This saliency criterion considers the distribution of the pixels around the elliptic hypothesis as shown in Figure 5.5-2(b). This idea was first proposed by [108], though in the context of straight lines primarily. In the context of elliptic hypotheses, the following method is proposed.

Consider the pixels $\{P\}$ in the edge contour that generated an EH. The Euclidean distance, $d$, of these pixels from the EH are computed. The lesser the Euclidean distance, the more reliable is an edge pixel for generating the elliptic hypothesis. After applying a threshold ( $d_0 = 2$ ) on the Euclidean distance, the number of pixels that are reliable for the current hypothesis are counted and normalized with respect to the total number of pixels, $N_G$, in the edge contours that generated the EH. The alignment ratio $a(E, G)$ is computed in (5-12) below using eqn. (5-11).

$$s(E, P) = \begin{cases} 1 & if \ d < d_0 \\ 0 & otherwise \end{cases} \tag{5-11}$$

$$a(E, G) = \frac{\sum_{i=1}^{N_G} s(E, P_i)}{N_G} \tag{5-12}$$

The higher the value of $a(E, G)$, the better is the fit between ellipse $E$ and group $G$.

### 5.5.2.3 Angular continuity ratio

Another criterion for choosing salient hypotheses is based on the angular continuity of the edge contours that generated a hypothesis. Consider two edge contours as shown in Figure 5.5-2(c). The angle between the two intersecting tangents made at the two nearest end points of $e_1$ and $e_2$, $\theta_{\text{diff}}$, is the angle that determines the continuity between the two edge contours. It can have a maximum value $\pi$. Thus, the ratio of the angle, $\theta_{\text{diff}}$, and $\pi$ is an indicator of the continuity between the two edge contours. Similar idea was proposed in [98]. The angular continuity ratio is defined as:

$$\phi(E, G) = \begin{cases} 1 & \text{if } N = 1 \\ \dfrac{1}{N-1} \sum_{i=1}^{N-1} \dfrac{\theta_{\text{diff}}\left(e_i, e_{i+1}\right)}{\pi} & \text{if } N > 1 \end{cases} \tag{5-13}$$

where $N$ is the number of edge contours in the group $G$. It is worth noting that if an elliptic object has large $\phi(E, G)$, the reliability of such elliptic hypotheses is better than an elliptic object that appears in the form of far apart contours.

### 5.5.3 Determination of the net saliency score

All the three criteria discussed in section 5.5.2 are representative of the quality of an elliptic hypothesis in the context of the image. While all of them have their respective strengths, all of them have their own deficiencies. Combining them to form a single saliency measure serves as supplementing their individual capabilities. Various schemes of combining them and the impact of these schemes are discussed in this section.

### 5.5.3.1 Multiplicative combination

The first way of combining them is to multiply the three criteria as in eqn. (5-14). The net saliency of an ellipse $E$ detected using a group $G$ is the product of $a(E,G)$, $c(E,G)$, and $\phi(E,G)$:

$$\sigma_{\mathrm{mul}}(E,G) = a(E,G)\,c(E,G)\,\phi(E,G)\,. \tag{5-14}$$

The net saliency as defined above takes into account each of the three criteria. Due to its multiplicative nature, if any of the criteria is very poor, the net saliency of the hypothesis becomes very low, even though it might be salient in terms of the other two criteria. In effect, it is similar to Boolean AND function. Thus, on one hand it rewards the hypotheses that are salient in terms of all three criteria, and on the other hand it is very strict on the rest.

### 5.5.3.2 Additive combination

The problem with multiplicative combination motivates us to consider the additive combination of the three criteria. Given the fact that the range of all the three criteria is $[0,1]$, the value 1 being the most desirable, the net saliency may be considered as the average of the three criteria as follows:

$$\sigma_{\mathrm{add}}(E,G) = \frac{a(E,G) + c(E,G) + \phi(E,G)}{3}\,. \tag{5-15}$$

While the multiplicative combination is too sensitive to the poor performance of any criterion, the additive combination is less sensitive to the poor performance of a single criterion. This means that though one of the criteria may have low value, if the other two criteria have sufficiently high values, the net saliency in additive combination will still be sufficient to have the hypothesis selected.

### 5.5.3.3 Choosing a threshold

Whatever be the saliency measure, higher value of saliency indicates that the detected ellipse is better than the ellipses with lower saliency. Based on this value, the best ellipses can be determined in various ways. The most prevalent and straight forward manner is to choose all the ellipses with saliency higher than a threshold. The performance of the ellipse detection method greatly depends upon the choice of the threshold. However, there is no hard and fast rule that ensures the suitability of a chosen threshold in most scenarios. The value that seemingly works very well for one image(s) may completely fail in another image(s). So, the use of the statistical mean of the net saliency for all the elliptic hypotheses as the threshold is more reasonable. Once the saliency measure is computed for each elliptic hypothesis, the average of all the saliency measures can be computed and used as the threshold. An alternate scheme is proposed and discussed in more detail in section 5.5.4. Our observation is that it works well for most images.

### 5.5.3.4 Examples

In order to further elucidate the discussion above, two example images are considered and various saliency criteria, the multiplicative and additive saliency scores are studied. The effect of using a threshold for selecting the hypotheses is also studied. The examples are presented in Figure 5.5-3 and Figure 5.5-4, respectively. In both the examples, the smooth edge curves are extracted according to the procedure in section 5.3 (see Figure 5.5-3(c) and Figure 5.5-4(c)). Scheme 3, discussed in section 5.6.3.3, is used for generating EH. After generating EH and removing similar ellipses, the three saliency criteria discussed in section 5.5.2 are computed for each EH. The multiplicative net saliency score (5-14) and additive net saliency score (5-15) are also computed for each EH. For reference, the elliptic hypotheses chosen by the proposed elliptic hypotheses selection scheme (section 5.5.4) are shown in Figure 5.5-3(d) and Figure 5.5-4(d).

In subfigures (e)-(g) of Figure 5.5-3 and Figure 5.5-4, the results of the three saliency scores (section 5.5.2.1 - 5.5.2.3) are plotted. In subfigures (h)-(i) of Figure 5.5-3 and Figure 5.5-4, the results of the multiplicative and additive net saliency scores are plotted. In these subfigures, the darkness of an elliptic hypothesis is proportional to the value of the score. This means that darker ellipses have higher value of the saliency score as compared to the lighter ellipses.

Now, a threshold of 0.8 is applied on each of the saliency scores. This means that for a particular saliency score, the ellipses that have a score of greater than or equal to 0.8 times the maximum score (for all elliptic hypotheses) are chosen. In subfigures (j)-(l) of Figure 5.5-3 and Figure 5.5-4, the elliptic hypotheses chosen by applying the threshold 0.8 on the three saliency scores (section 5.5.2.1 - 5.5.2.3) are plotted. In subfigures (m)-(n) of Figure 5.5-3 and Figure 5.5-4, the elliptic hypotheses chosen by applying the threshold 0.8 on the multiplicative and additive net saliency scores are plotted. The chosen elliptic hypotheses are shown in the dark, while the remaining elliptic hypotheses are shown in light color.

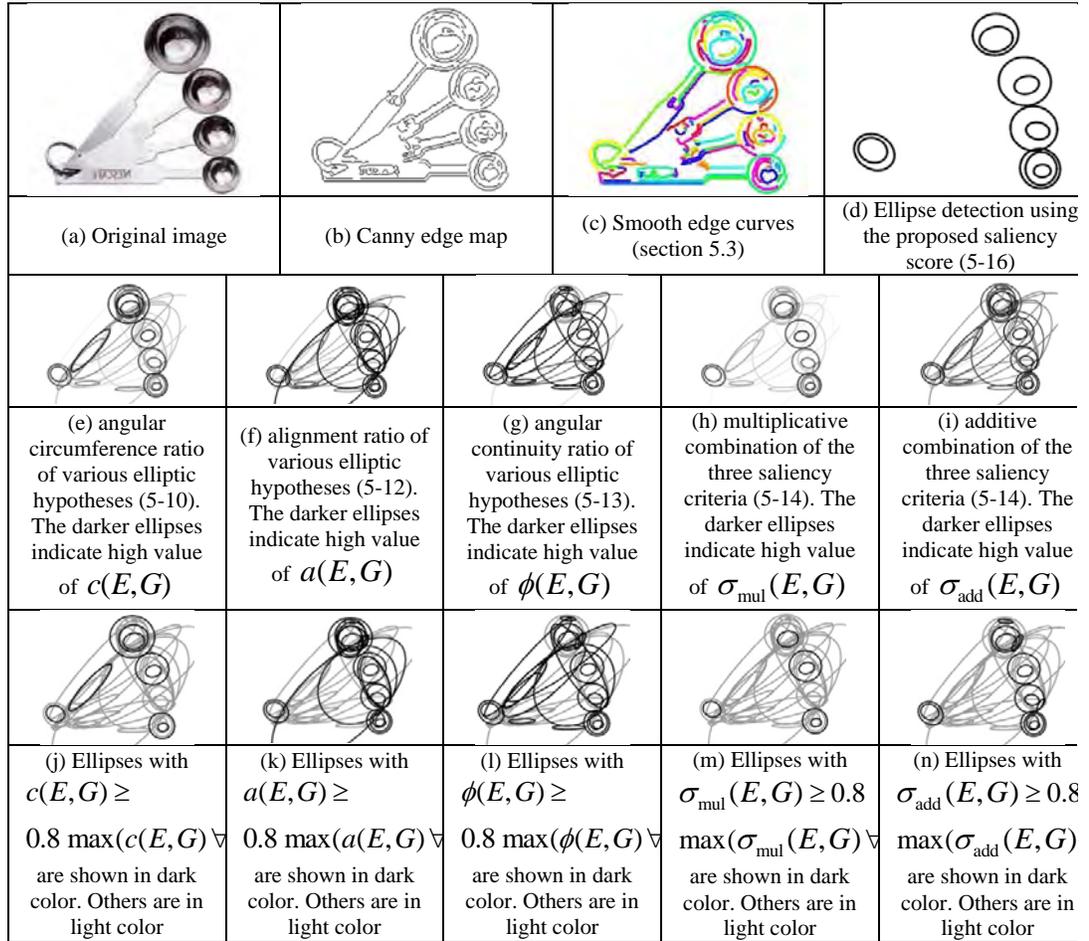

| (a) Original image | (b) Canny edge map | (c) Smooth edge curves (section 5.3) | (d) Ellipse detection using the proposed saliency score (5-16) | |
|---|---|---|---|---|
| (e) angular circumference ratio of various elliptic hypotheses (5-10). The darker ellipses indicate high value of $c(E,G)$ | (f) alignment ratio of various elliptic hypotheses (5-12). The darker ellipses indicate high value of $a(E,G)$ | (g) angular continuity ratio of various elliptic hypotheses (5-13). The darker ellipses indicate high value of $\phi(E,G)$ | (h) multiplicative combination of the three saliency criteria (5-14). The darker ellipses indicate high value of $\sigma_{mul}(E,G)$ | (i) additive combination of the three saliency criteria (5-14). The darker ellipses indicate high value of $\sigma_{add}(E,G)$ |
| (j) Ellipses with $c(E,G) \geq$ $0.8 \max(c(E,G)\forall$ are shown in dark color. Others are in light color | (k) Ellipses with $a(E,G) \geq$ $0.8 \max(a(E,G)\forall$ are shown in dark color. Others are in light color | (l) Ellipses with $\phi(E,G) \geq$ $0.8 \max(\phi(E,G)\forall$ are shown in dark color. Others are in light color | (m) Ellipses with $\sigma_{mul}(E,G) \geq 0.8$ $\max(\sigma_{mul}(E,G)\forall$ are shown in dark color. Others are in light color | (n) Ellipses with $\sigma_{add}(E,G) \geq 0.8$ $\max(\sigma_{add}(E,G)\forall$ are shown in dark color. Others are in light color |

**Figure 5.5-3: Illustration of the saliency criteria for example 1.**

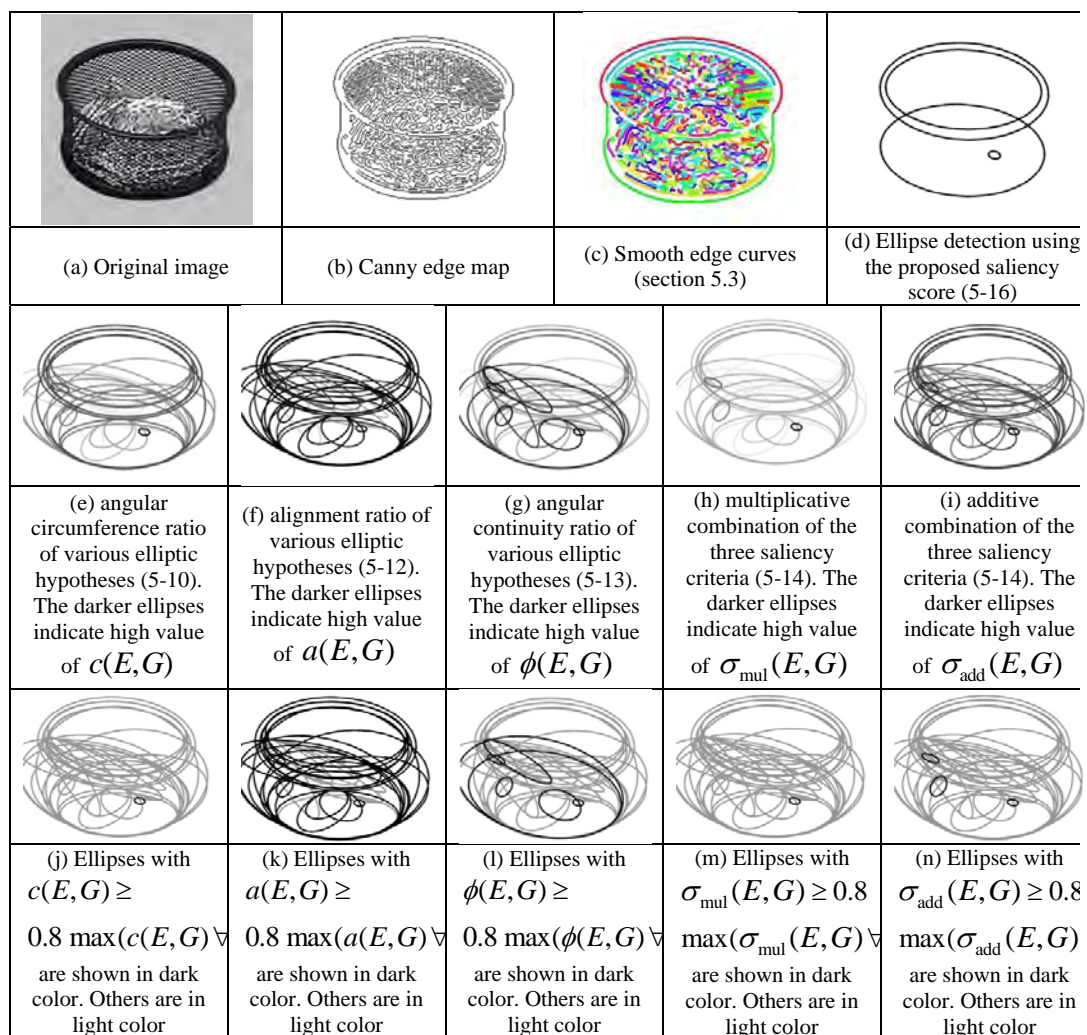

**Figure 5.5-4: Illustration of the saliency criteria for example 2.**

The first observation is that in both the examples, none of the saliency criteria performs well in choosing the elliptic hypotheses. For example 1, the circumference ratio (Figure 5.5-3(e,j)) performs better than the remaining saliency criteria (Figure 5.5-3(f,g,k,l)). However, the circumference ratio performs very poor for the second example (Figure 5.5-4(e,j)). In both the examples, while alignment ratio (Figure 5.5-4(f,k) and Figure 5.5-4(f,k)) is able to select the true positive elliptic hypotheses, it is not able to reject the false positive elliptic hypotheses effectively. The angular continuity ratio performs better for example 1 ratio (Figure 5.5-3(g,l)) than example 2 (Figure 5.5-4(g,l)). Thus, use of a single saliency criterion is not reliable for diverse images.

Second observation is that the multiplicative net saliency score (Figure 5.5-3(h,m) and Figure 5.5-4(h,m)) is more selective than the additive net saliency score (Figure 5.5-3(i,n) and Figure 5.5-4(i,n)). Only the elliptic hypotheses that have high scores in all the three saliency criteria stand a chance to get selected using the multiplicative net saliency score (Figure 5.5-3(h,m) and Figure 5.5-4(h,m)). As already discussed before, the saliency criteria may behave differently in different images. Thus, only very good elliptic hypotheses are selected by the multiplicative combination. As a consequence, multiplicative net saliency score will give good precision but significantly poor recall.

Third observation is that the additive net saliency score performs reasonably well for the first example (Figure 5.5-3(i,n)), but significantly poor for the second example (Figure 5.5-4(i,n)). The main reason of the poor performance of the additive net saliency score in the second example (Figure 5.5-4(i,n)) is the threshold value of 0.8. For the image in the second example, only three elliptic hypotheses are selected (Figure 5.5-4(i,n)), and all of them are false positives. The relevant (true positive) elliptic hypotheses have lower values than the chosen threshold in (Figure 5.5-4(i,n)) and thus are not selected. A choice of threshold value of 0.7 would be more appropriate for example 2, while it would be very lenient for example 1. Thus, using a strict threshold for all the images is not suitable.

### 5.5.4 The proposed scheme for hypotheses selection

As shown using the examples in section 5.5.3.4, neither the multiplicative combination, nor the additive combination is suitable for elliptic hypotheses selection. Further, the suitable threshold for selecting the elliptic hypotheses varies from image to image. In order to make the selection of the elliptic hypotheses non-heuristic, the decision of selecting the elliptic hypothesis $E$ is made using the expression below:

$$\text{AND}\begin{cases} a(E,G) \geq avg\left\{a(E,G)\right\}, \\ c(E,G) \geq avg\left\{c(E,G)\right\}, \\ \phi(E,G) \geq avg\left\{\phi(E,G)\right\}, \\ \sigma_{add}(E,G) \geq avg\left\{\sigma_{add}(E,G)\right\} \end{cases} \tag{5-16}$$

Here, $avg\left\{a(E,G)\right\}$ is the average value of the alignment ratios calculated for all the elliptic hypotheses remaining after the similar ellipses identification. The same applies for the other expressions in (5-16). The use of the average values of the saliency scores as the threshold for selection makes the selection procedure non-heuristic and free of human intervention. This is unlike other methods where thresholds are chosen based on heuristics and their efficacy depends largely on the considered application and dataset. This selection method chooses the appropriate threshold values in eqn. (5-16) for each image independently and is therefore not dependent on the type of dataset.

Further, the use of AND operation ensures that the elliptic hypotheses that are good in every respect are selected finally. This selection method assures that the selected hypotheses perform better than the average elliptic hypotheses (for a particular image) in every criterion and have overall good saliency. Due to the non-heuristic thresholds and the AND operation, the method is able to generate good results for a large variety of images as demonstrated by the numerical results in the following section.

## 5.6 Numerical evaluation

This section presents various numerical results to test the proposed ellipse detection method. Based on section 5.4, three different schemes of ellipse detection are considered in section 5.6.3. These schemes are tested using the synthetic dataset. Based on the results, scheme 3 is considered for the remaining numerical results. The performance of scheme 3 is compared against four other ellipse detection methods in section 5.6.4.

### 5.6.1 Synthetic dataset of overlapping and occluded ellipses

The proposed method is tested under various scenarios such as occluded ellipses and overlapping ellipses using synthetic images. To generate the synthetic images, an image size of $300 \times 300$ is considered and $\alpha \in \{4, 8, 12, 16, 20, 24\}$ ellipses generated randomly within the region of image. The parameters of the ellipses are generated randomly: center points of the ellipses are arbitrarily located within the image, lengths of semi-major and semi-minor axes are assigned values randomly from the range $\left[10, 300/\sqrt{2}\right]$, and the orientations of the ellipses are also chosen randomly. The only constraint applied is that each ellipse must be completely contained in the image and overlap with at least one ellipse.

For each value of $\alpha$, 100 images containing occluded ellipses and 100 other images containing overlapping ellipses are generated. In the occluded ellipses, the edges of the overlapped regions are not available, while in the overlapping images all the edge contours of the ellipses are available. Thus, in total there are 600 images with occluded ellipses and 600 images with overlapping ellipses. Examples of synthetic images with occluded and overlapping ellipses are presented in Figure 5.6-1.

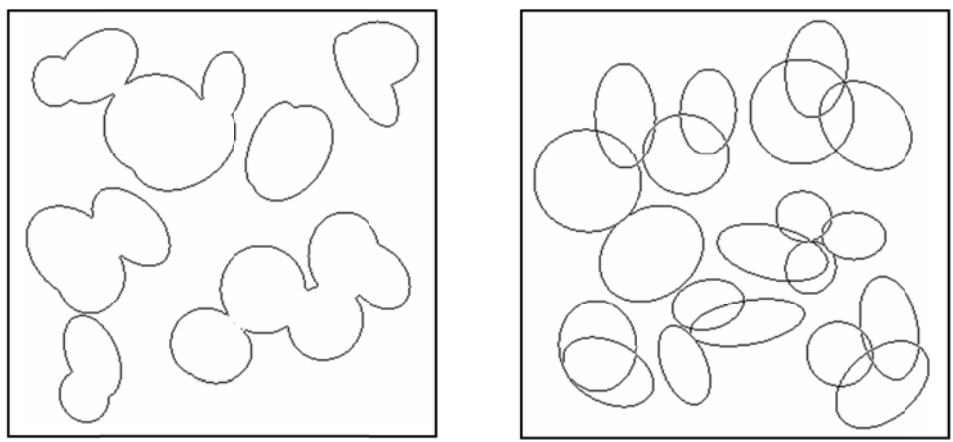

(a)  Example of image with occluded ellipses

(b)  Example of image with overlapping ellipses

**Figure 5.6-1: Examples of images with occluded and overlapping ellipses.**

### 5.6.2 Comparison metrics and the experimental setup

The following metrics are used for evaluating the performance of the proposed ellipse detection method:

$$\text{Precision} = \frac{\text{number of true positive elliptic hypotheses}}{\text{total number of elliptic hypotheses}} \tag{5-17}$$

$$\text{Recall} = \frac{\text{number of true positive elliptic hypotheses}}{\text{number of actual ellipses}} \tag{5-18}$$

$$\text{F-measure} = \frac{2 \times \text{Precision} \times \text{Recall}}{\text{Precision} + \text{Recall}} \tag{5-19}$$

where the true positive elliptic hypotheses are the hypotheses that have high overlap with the ellipses in the ground truth. For synthetic dataset, an overlap ratio of 0.95 is used for determining true positive hypotheses. For the synthetic dataset (occluded ellipses), the performance metrics are computed for each of the 100 images corresponding to a value of $\alpha$, and the mean for 100 images is used as the performance metric for that value of $\alpha$. Similar procedure is used for the synthetic images containing the occluded ellipses.

Square bins of size $25 \times 25$ pixels based on the guidelines in [180] have been used, and the threshold for C1 is chosen as $\varepsilon_{ls} = 0.01$. The number of sets used for finding the centers of EH (section 5.4.3) using an edge contour is $S = 200$.

### 5.6.3 Various schemes of ellipse detection and their performance

#### 5.6.3.1 Scheme 1

In the generation of elliptic hypotheses, some steps may be dropped to form simplified schemes for ellipse detection method. An example of a simplified scheme is presented in Figure 5.6-2(a). In this scheme, the search region and the associated convexity are not used. Further, instead of the relationship score $\tilde{r}_e^b$ in eqn. (5-5), just the histogram count is used for ranking the edges falling into a bin. This scheme is referred to as scheme 1.

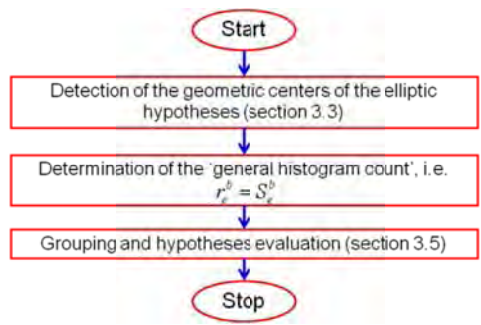
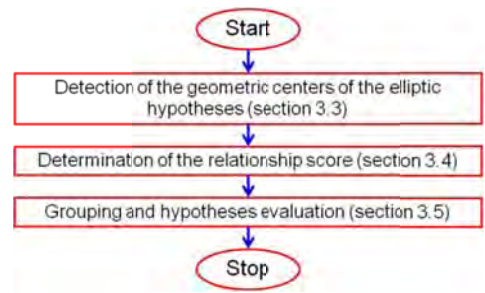

(a) Flowchart of the generation of elliptic hypotheses for scheme 1.

Search region and associated convexity are not used. Further, general histogram count using $r_e^b = S_e^b$, instead of eqn. (5-5), is used.

(b) Flowchart of the generation of elliptic hypotheses for scheme 2.

Search region and associated convexity are not used.

**Figure 5.6-2: Flowcharts of the elliptic hypotheses generation step for schemes 1 and 2.**

#### 5.6.3.2 Scheme 2

Scheme 2 is similar to scheme 1, with only one difference. In scheme 2, the relationship score proposed in eqn. (5-5) is used. The comparison of the results of scheme 1 and 2 shall illustrate the advantage of using the proposed relationship score.

### 5.6.3.3 Scheme 3 (the proposed scheme)

Scheme 3 is the complete elliptic hypotheses generation scheme as shown in Figure 5.2-1(c) and proposed in section 5.4. The comparison of scheme 2 and scheme 3 illustrates the impact of using the search region and associated convexity for filtering away the edges that are not suitable for grouping with an edge.

### 5.6.3.4 Comparison of the performance of the three schemes

The results of all the three schemes for the synthetic dataset are presented in Figure 5.6-3 and Figure 5.6-4. Figure 5.6-3 shows the results for the images with the occluded ellipses, while Figure 5.6-4 shows the results for the overlapping ellipses.

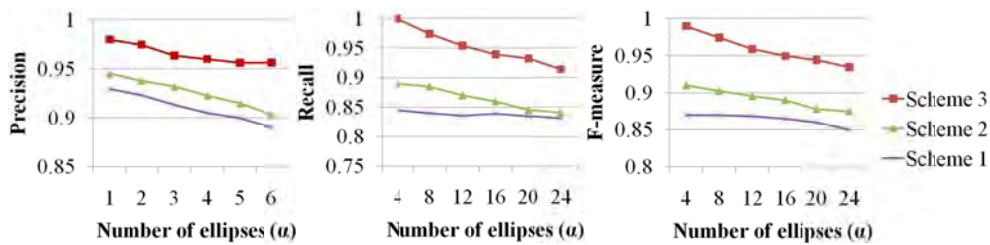

**Figure 5.6-3: Comparison of the schemes 1-3 for images with occluded ellipses.**

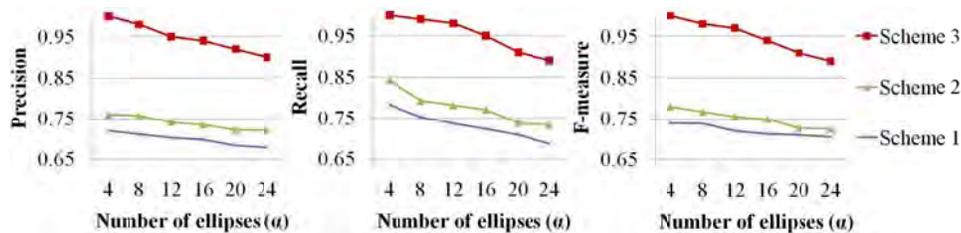

**Figure 5.6-4: Comparison of the schemes 1-3 for images with overlapping ellipses.**

Comparing the performance of schemes 1 and 2, it is evident that the use of the relationship score in scheme 2 improves the performance of the ellipse detection scheme. This is because scheme 2 is more selective than scheme 1. Thus, the number of elliptic hypotheses generated in scheme 2 is lesser than scheme 1 in general, while the number of true positive elliptic hypotheses generated in scheme 2 is more than scheme 1 in general.

Similarly, comparing the performance of schemes 2 and 3, it is evident that scheme 3 performs significantly better than scheme 2. Again, the reason is the higher selectivity of the scheme 3 that occurs due to the selection of better grouping candidates using the search region and associated convexity.

### 5.6.4 Performance comparison with other methods

The performance of the proposed ECC method of ellipse detection (scheme 3) is compared against one standard ellipse detection method and four recently proposed ellipse detection methods. The standard method considered is the randomized Hough transform (RHT) [70]. The recently proposed methods considered here are the methods proposed by Mai [34], Kim [68], Bai [179], and Liu [32]. For Bai [179] and Liu [32], the values of control parameters used in their original articles did not provide good results. Thus, the control parameters were varied and the best combination of control parameters were obtained to get the best results for synthetic images (Bai [179]: $\text{disTh} = 0.03$, $\text{dMinTh} = 10$, $\text{dTh} = 3.0$ and Liu [32]: $\eta = 1/N$, $e_{th} = 0.05$, $\alpha = 0.1$, other control parameters retain the default value of the respective algorithm). It is highlighted that these methods are very sensitive to the choice of control parameters. For instance, using $\text{disTh} = 0.009$ as mentioned in Table 2 of Bai [179], results in less than 5% precision for synthetic datasets and almost no detections for real dataset. Similarly, using $\alpha = 0.5$, as suggested in [32] results in almost zero precision for synthetic and real datasets.

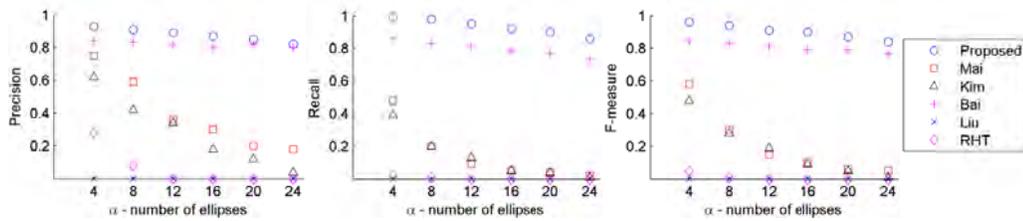

**Figure 5.6-5: Comparison of the proposed method with other methods for images with occluded ellipses.**

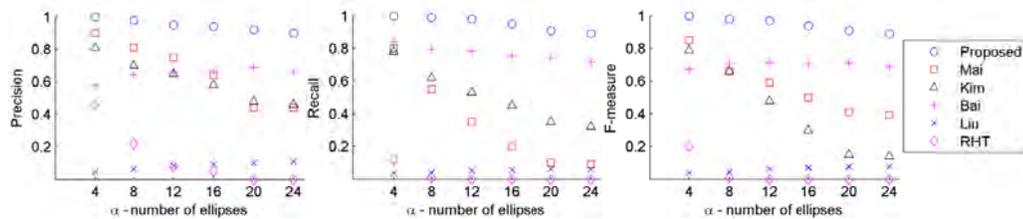

**Figure 5.6-6: Comparison of the proposed method with other methods for images with overlapping ellipses.**

The results for synthetic dataset are presented in Figure 5.6-5 (occluded ellipses) and Figure 5.6-6 (overlapping ellipses). It is clearly evident that the proposed method outperforms the existing methods in terms of precision, recall, as well as F-measure. Further, it is close to the best values of precision, recall, and F-measure (the best value is 1 for all these measures). Lastly, even with substantial increase in the number of ellipses in an image, the performance does not deteriorate significantly. The proposed method took an average time of 5.63 seconds for the synthetic dataset.

It is noted that Liu [32] has zero detections for the case of occluded ellipses. This is because Liu [32] does not deal with the inflexion points, as also highlighted in [32]. It is also noted that Bai [179] gives the closest and most consistent performance with respect to the proposed method for the case of occluded ellipses. This is consistent with the nature of dataset considered in Bai [179]. However, it performs poorer for the case of overlapping ellipses as compared to the occluded ellipses. Mai [34] and Kim [68] show deteriorating performance as the number of ellipses increases.

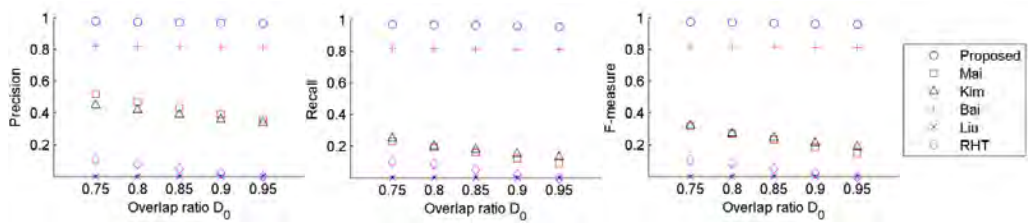

**Figure 5.6-7: Performance of various methods for different values of overlap ratio.**

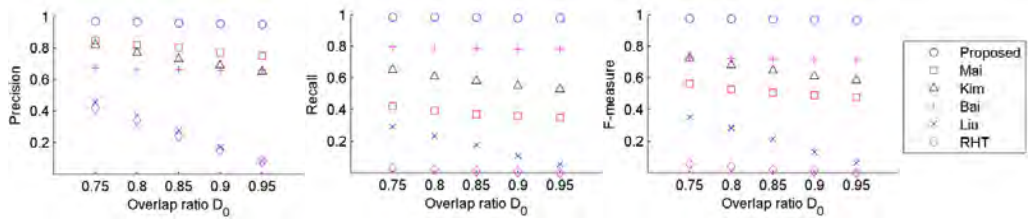

**Figure 5.6-8: Performance of various methods for different values of overlap ratio.**

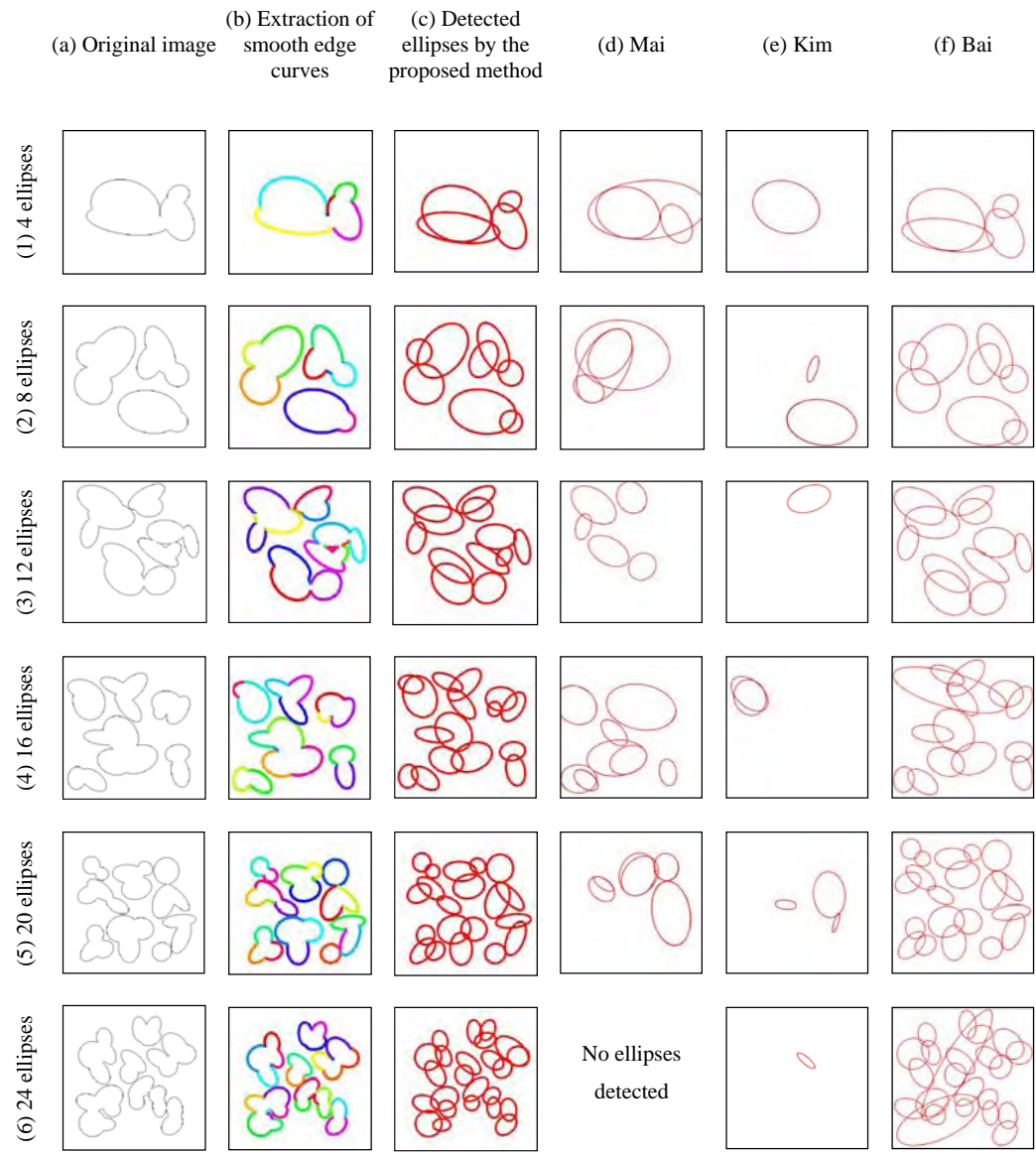

**Figure 5.6-9: Examples of synthetic images with occluded ellipses.**

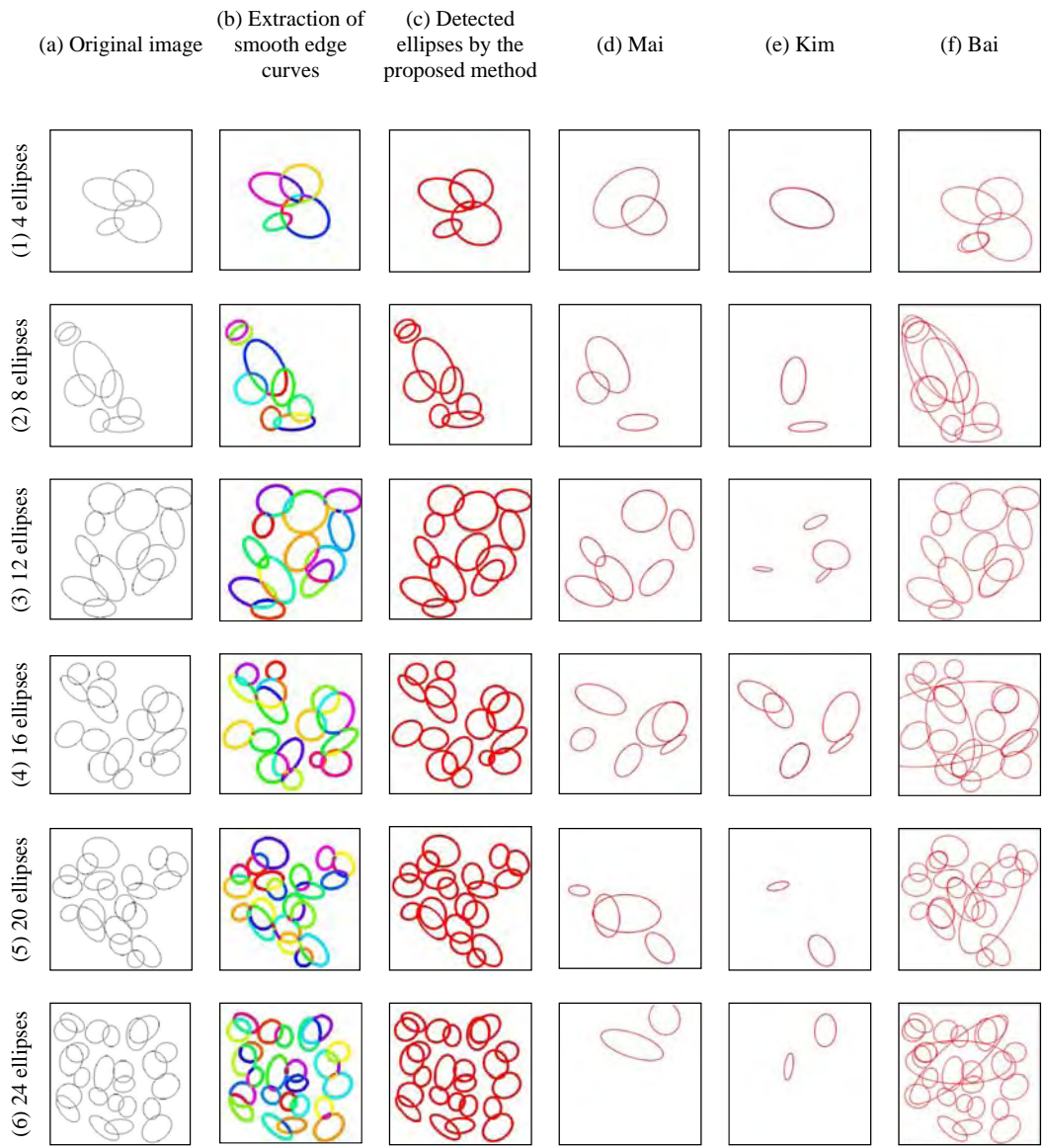

**Figure 5.6-10: Examples of synthetic images with overlapping ellipses.**

As mentioned before, an overlap ratio of 0.95 between the elliptic hypotheses and the ground truth (true parameters of the ellipses) has been used for determining the true positive hypotheses. In order to show that the poor performance of the other methods is not due to the high overlap ratio, the performance of the proposed method and the other methods is further tested using different values of the overlap ratio used for determining the true positive hypotheses. For this purpose, images with $\alpha = 12$ are used. The comparison results of the proposed method and the other methods are shown in Figure 5.6-7 and Figure 5.6-8. It is evident that the performance of the proposed method does not depend significantly on the overlap ratio used for finding the true positive elliptic hypotheses. Some example images and the detected ellipses with occluded and overlapping ellipses are presented in Figure 5.6-9 and Figure 5.6-10, respectively.

## 5.7 Conclusion

A method for ellipse detection based on edge curvature and convexity is proposed. The ECC method performs significantly better than the existing methods for complicated synthetic and real images. The demonstrated performance is close to the best achievable performance for the synthetic images. The primary reason of the good performance of ECC is the enhanced selectivity of the method while grouping the edges for detecting ellipses. The selectivity is due to the use of smooth curvature (in the form of search region and associated convexity), the novel relationship score, and the robust non-heuristic saliency criteria. The contents of this chapter have been reported in [45, 95, 96, 139].

# Chapter 6 : Image processing applications

This chapter presents three image processing applications of the algorithms proposed in Chapter 2 – Chapter 5. The algorithms proposed in these chapters have several interesting and practically useful applications. Some interesting applications are surface grain analysis of materials, medical diagnostics (like malarial and sickle cell counting), analysis of geological and astronomical pictures, assisted robotics for applications like sorting ores, vegetables, and cans, etc. in automatic mineral processing, grocery, and recycling units respectively, bowling arena management by robot, etc. Advanced applications may include using the elliptic shapes for object detection [181], face detection [182], biometric iris based systems [183, 184], shape representation of complex objects [185], etc.

Here, three specific applications are presented. The first application involves ellipse detection from real images. It relates to most of the robotics applications presented above and is presented in section 6.1. The second application is the segmentation and sorting of the biological cell organelles in the images obtained by a biological microscope and is presented in section 6.2. The third application is that of object detection using *polygonal and elliptical features* only and a hierarchical object template. This application is presented in section 6.3.

## 6.1 Ellipse detection in real images

Ellipse detection in real images has several applications in the field of robotics. Assisted robotics of interest are applications like sorting ores, vegetables, cans, etc. in automatic mineral processing unit, grocery shops and warehouses, and recycling units respectively. Other robotic applications include managing bowling arenas, garage inventory, or being golf caddies. For such dedicated applications, the method can be customized for better and faster performance. All of these applications require that elliptic objects present in the images are detected. Thus, the ellipse detection method proposed in Chapter 5 (scheme 3) is of particular interest here. It should be noted that the algorithms and schemes in Chapter 5 use algorithms from Chapter 2 to Chapter 4.

In order to show the applicability for the above mentioned robotic applications, the Caltech-256 database [1] is used as the dataset of real images in which several classes correspond to elliptic objects. These images present greater challenges than the synthetic images used in Chapter 4 and Chapter 5 because the edge contours of elliptical shaped objects are corrupted by complex and varied backgrounds, illumination variations, partial occlusions, image noise, shadows and spectral reflections. In the Caltech 256 dataset, 400 real images were randomly chosen from 48 categories (each category corresponding to an elliptic object class). Only one condition was used for selection. At least 5 (among 20 volunteers) should identify a minimum of one recurring ellipse as the ground truth. The details of the procedure to generate the ground truth are presented in Appendix E. Among the 400 real images used for generating the results, the minimum number of ellipses in an image was 1 and the maximum number of ellipses was 60. An overlap ratio of 0.8 between the detected ellipse and the ground truth is used for determining the true positive hypotheses in the real images dataset.

Examples of images with the detected ellipses using several ellipse detection methods are presented in Figure 6.1-1. It is seen that the proposed method is effective in detecting elliptic shapes in real images containing elliptic object categories including balls, blimps, music systems, calculator and phone buttons, optical disks, coins, joysticks, wheels, tires, etc. The overall performance parameters of the various ellipse detection methods for these 400 images are presented in Table 6.1-1. In Figure 6.1-1 and Table 6.1-1, the proposed method is the scheme 3 of section 5.6.3.

**Table 6.1-1. Performance metrics for various hybrid ellipse detection methods.**

|  | Proposed | Mai | Kim | Bai | Liu | RHT |
|---|---|---|---|---|---|---|
| **Average Precision** | **0.8748** | 0.2862 | 0.1831 | 0.2248 | 0.0716 | 0.0213 |
| **Average Recall** | **0.7162** | 0.1632 | 0.1493 | 0.2955 | 0.1403 | 0.0157 |
| **Average F-measure** | **0.7548** | 0.1831 | 0.1591 | 0.1685 | 0.0808 | 0.0186 |
| **Average time taken (seconds)** | 38.68 | 11.41 | 60.87 | 9.10 | 7.78 | 810.41 |

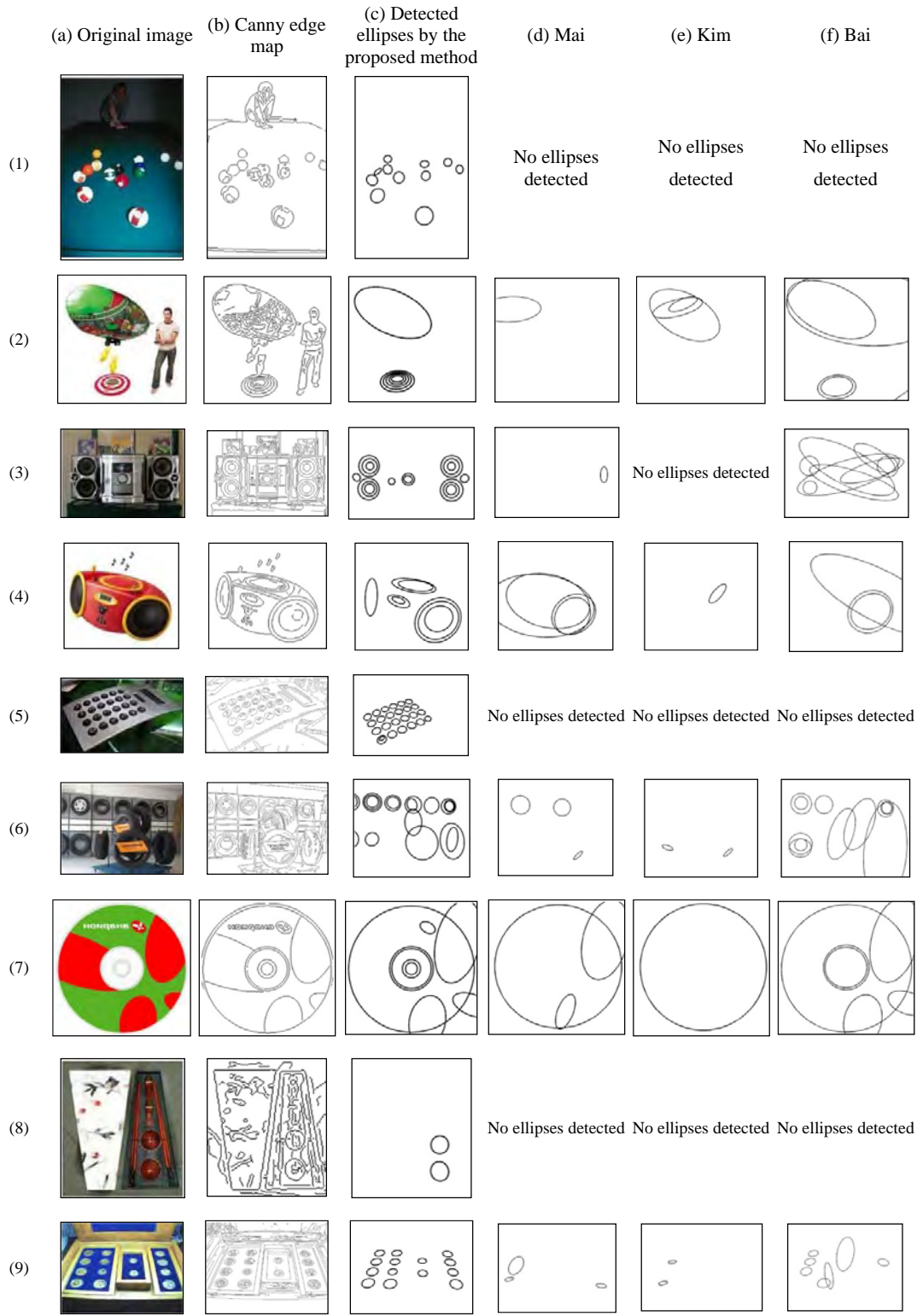

**Figure 6.1-1 : Examples of real images: proposed method (scheme 3 of section 5.6.3) vs. other methods.**

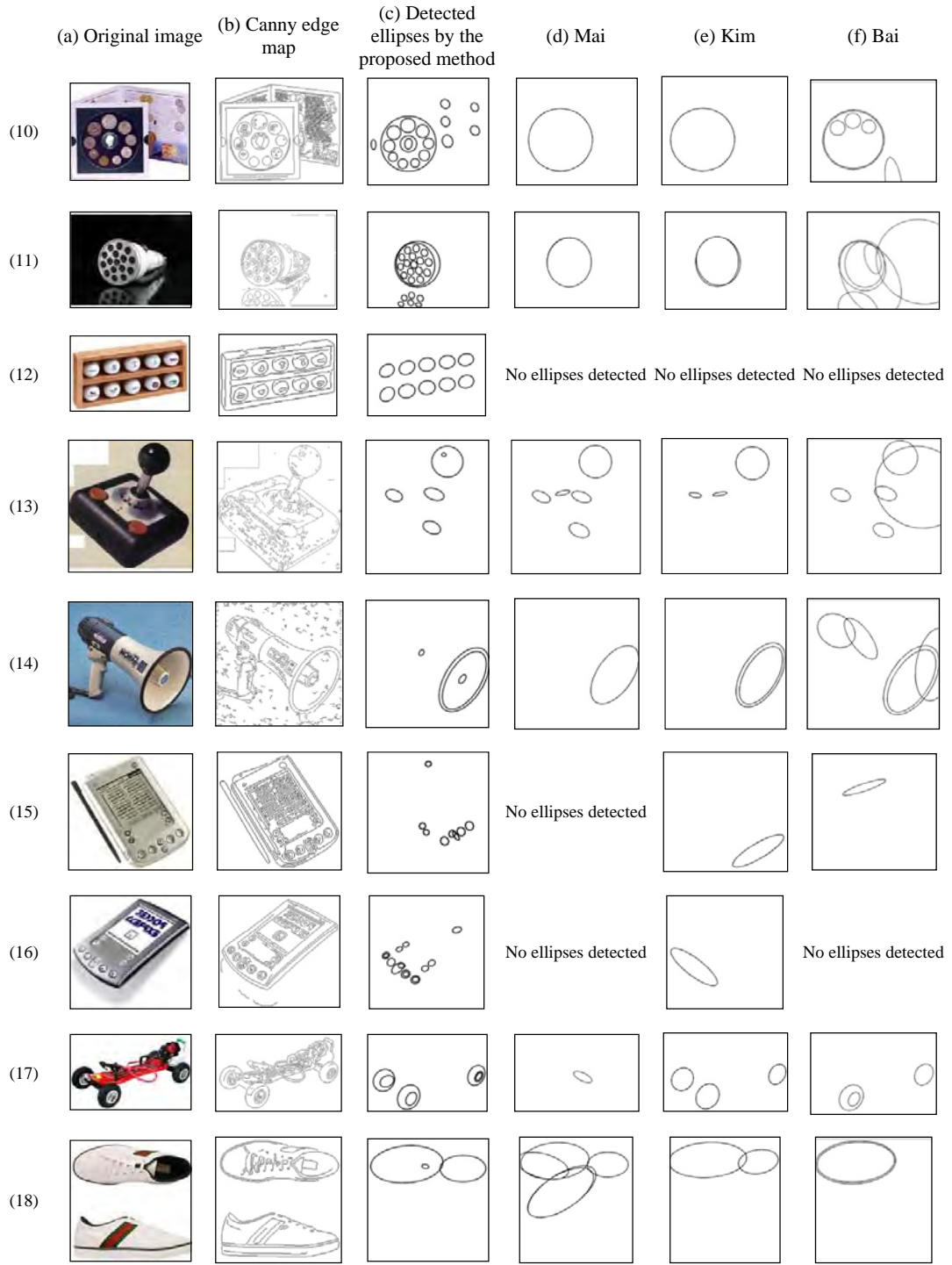

**Figure 6.1-1 : Examples of real images: proposed method (scheme 3 of section 5.6.3) vs. other methods. (contd.)**

## 6.2 Segmentation of sub-cellular organelles

Cell and sub-cellular segmentation in biomedical images is helpful in diagnosis and cell biology research. Often manual segmentation and classification is slow since one image may contain numerous cells or sub-cellular structures. Software can be used for this purpose but the accuracy is often low and is often unable to filter away the artifacts on its own. Here, a segmentation algorithm for cells or sub-cellular structures that can be modeled as elliptic is proposed. Examples of such datasets can be found in [114, 186-188]. The dataset considered in this paper is a dataset of images of mixed cell organelle types (mitochondria and lysosomes) [188]. Fluorescence confocal microscope is used to generate the images. Preprocessing of the images (in the materials and methods section) leads to cleaner images in which the cells appear in the foreground (for example as shown in Figure 6.2-3(a)). These images are used as input images for the proposed algorithm. It is highlighted that in several of the images, no sub-cellular structures are seen. Thus, images that contain sub-cellular features are selected manually. There are a total of 444 such images, each of size 1349 x 1030 pixels.

Using the algorithms proposed in Chapter 2 – Chapter 5, an algorithm for a given medical dataset is proposed here. The algorithm is tested on a dataset of images of two types of sub-cellular structures [188] and the algorithm shows a good performance. Further it is quite fast and easily parallelizable. Thus, with some code optimization, it can be made real time.

The proposed algorithm is presented in section 6.2.1. The results are presented in section 6.2.2.

### 6.2.1 Proposed algorithm

The proposed algorithm employs three simple blocks, viz., pre-processing, ellipse fitting, and ellipse selection. The flowchart of the proposed algorithm is shown in Figure 6.2-1.

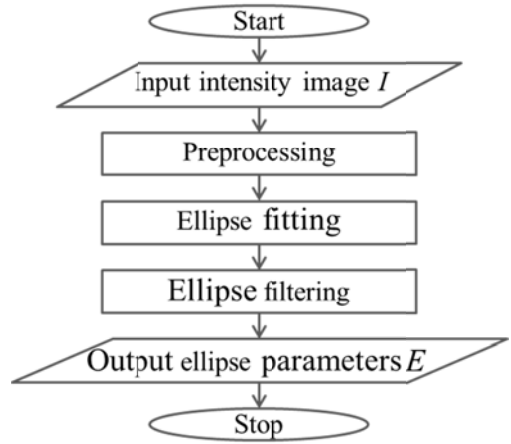

**Figure 6.2-1: Flowchart of the proposed method.**

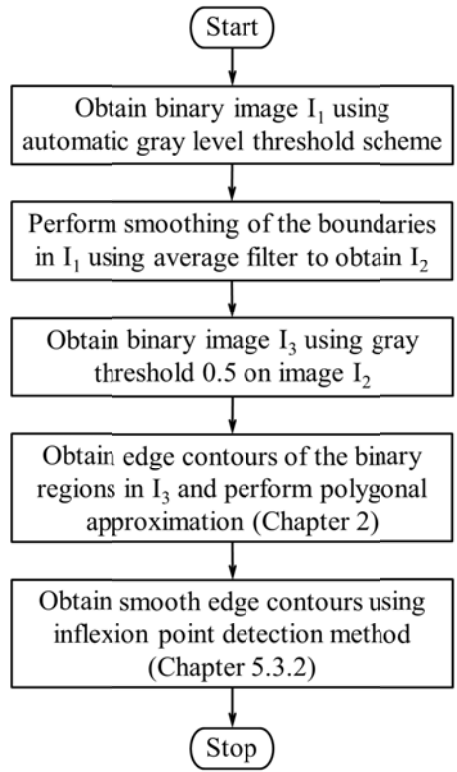

**Figure 6.2-2: Flowchart for preprocessing block**

### 6.2.1.1 Preprocessing block

It is assumed that the input is the intensity image $I$ (gray scale 0-255). The flowchart of this block is presented in Figure 6.2-2. One of the critical aspects is the choice of the threshold level for obtaining the binary images. The images in the biomedical dataset obtained using microscopy can suffer from several issues. Examples include low contrast, bleaching (background illumination), noise, scattering from irrelevant organelles, etc. However, assuming that the same instrument and measurement setup is used to generate the images in a particular dataset, a suitable threshold level $t_1$ can be determined *apriori* for binarizing the image (step 1 in Figure 6.2-2), binary image is being referred to as $I_1$.

In order to deal with extremely low contrast features, it is preferable to use a low value of $t_1$. Consider for example the image in Figure 6.2-3(a). The highlighted circle shows a region that contains a cell but is invisible due to extremely low contrast. As a consequence, this cell is present in the binarized image when $t_1 = 5$ is used (Figure 6.2-3 (b)) but absent when $t_1 = 15$ is used (Figure 6.2-3 (c)). It is also notable that artifacts due to noise appear when low threshold is used.

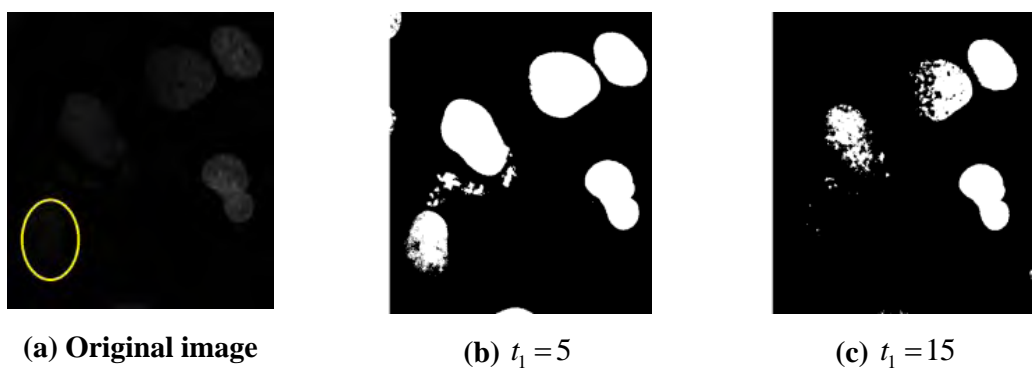

(a) Original image      (b) $t_1 = 5$      (c) $t_1 = 15$

**Figure 6.2-3: Example of the effect of threshold value $t_1$ on the binary image $I_1$**

Thus, a suitable selection of the threshold is quite important. A statistical scheme is proposed here for choosing the value of threshold $t_1$. For determining the suitable value of $t_1$, the histogram of each image in the dataset is generated for gray levels 0 – 255. Let the histogram count for a gray level $g$, $g = 1$ to 256 for an image $I$ be denoted as $h(g, I)$. Cumulative histogram is generated for each gray level and image is computed as:

$$\overline{C}(g, I) = \sum_{g'=1}^{g} h(g', I) \qquad (6\text{-}1)$$

The normalized cumulative histogram is then computed as follows:

$$C(g,I) = \frac{\bar{C}(g,I)}{\max\left(\bar{C}(g,I)\big|\forall g\right)} \tag{2}$$

The values of the normalized cumulative histogram for a given gray level $g$ are averaged for all the images and plotted in Figure 6.2-4(a). It is seen that the images have low intensity since only lower gray levels (till $g = 100$) have contribution in the images. A zoom-in of Figure 6.2-4(a) is provided in Figure 6.2-4(b) where only $g = 0$ to 31 is considered). It is seen that $g = 4$ is sufficient for more than 80% of the cumulative histogram of the images. Thus, $t_1 = 4$ is chosen as the threshold for binarizing the images (step 1 of Figure 6.2-2).

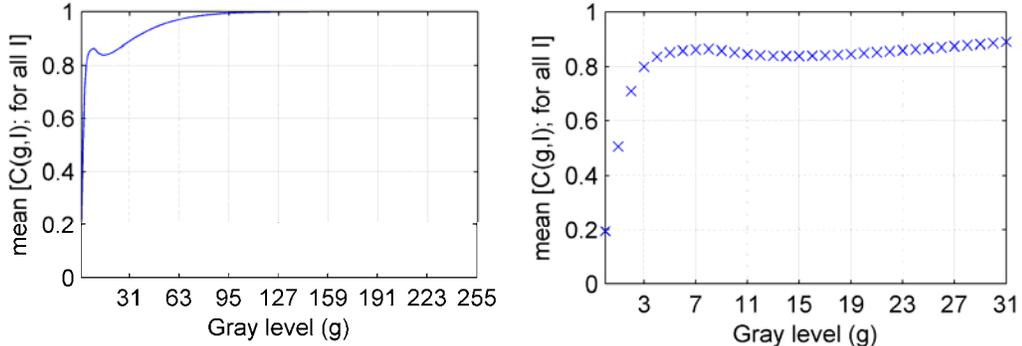

(a) Normalized cumulative histogram count averaged over all the selected images.

(b) Zoom in of (a) above for gray levels $0 - 32$.

**Figure 6.2-4: Normalized cumulative histogram averaged over all the images for choosing the threshold $t_1$.**

Yet, as shown in Figure 6.2-3, the boundaries of the cells may not be well defined due the binarization. Thus, mean filter of size 5 is applied on the binary image $I_1$ to get a gray scale image $I_2$ with smoother features (step 2 in Figure 6.2-2). The image $I_2$ is binarized again using a threshold value of 0.5, since average filter is applied on binary image (step 3 in Figure 6.2-2) The final binary image used for further processing is denoted as $I_3$.

The edge contours of the white regions in $I_3$ are extracted (step 4 in Figure 6.2-2) and are denoted by index $e = 1$ to $\hat{e}$. For each edge $e$, PA of the edge is derived using

RDP-mod proposed in section 2.5.1.2 (step 4 in Figure 6.2-2). Other PA methods may be used, but as discussed in section 2.7.5, RDP-mod is preferred. The PA of the edge contour is consequently used to remove the inflexion points and obtain smooth edges $e' = 1$ to $\tilde{e}'$ using the algorithm for removing inflexion points proposed in section 5.3.2 (step 5 in Figure 6.2-2).

### 6.2.1.2 Ellipse fitting block

This block calls a least squares based ellipse fitting method for each edge $e' = 1$ to $\tilde{e}'$. If the method generates a valid ellipse, the geometric parameters of the ellipse (length of semi-major axis $a$, length of semi-minor axis $b$, x-coordinate of the center $x_0$, y-coordinate of the center $y_0$, and the angle made by semi-major axis with the $x$ axis $\alpha$ of the fitted ellipse) is appended in the array containing the parameters of ellipses $E$.

The choice of the ellipse fitting method has a significant impact on the overall performance of the method. Hough transform based methods have several problems like a huge number of samples are required to obtain robust results, five-dimensional parameter space of ellipses is difficult to deal with computationally, and the whole process can be significantly time consuming. So, least squares based methods for fitting ellipses can be used. In our numerical results, it shall be shown that the geometry based least squares method proposed in section 4.4 performs better than some of the other least squares method.

### 6.2.1.3 Ellipse filtering block

After fitting ellipses on each edge contour, available apriori information about the dataset can be used to filter or remove some unreasonable ellipses. The filtering criteria depend upon the dataset and *apriori* information known about it. For example the imaging resolution and the CCD grid size can be used to determine the size range of the cells and thus the bounds on the lengths of semi-major axis $a$ and semi-minor axis $b$ may be generated. Also, the biological information about the cells can be used to generate an estimate of the maximum ratio of the semi-major and semi-minor axes $a/b$. In the proposed algorithm, the following filtering criteria have been used:

$$a/b \leq t_2 \tag{6-3}$$

$$t_3 \leq b \leq t_4 \tag{6-4}$$

where $t_2$, $t_3$, and $t_4$ are the thresholds determined based on the *apriori* knowledge of the cells and the imaging system. The ellipses satisfying eqns. (6-3) and (6-4) are retained. It is known that the size of the lysosomes and mitochondria varies in the range $0.1\,\mu m$ to $1\,\mu m$. Thus, the value of the threshold $t_2$ is chosen as the ratio of the maximum to the minimum size of these sub-cellular structures, i.e., $t_2 = 10$. Further, based on the general size of the sub-cellular structures in the dataset, the threshold values of $t_3$ and $t_4$ are chosen as 10 and 100 respectively.

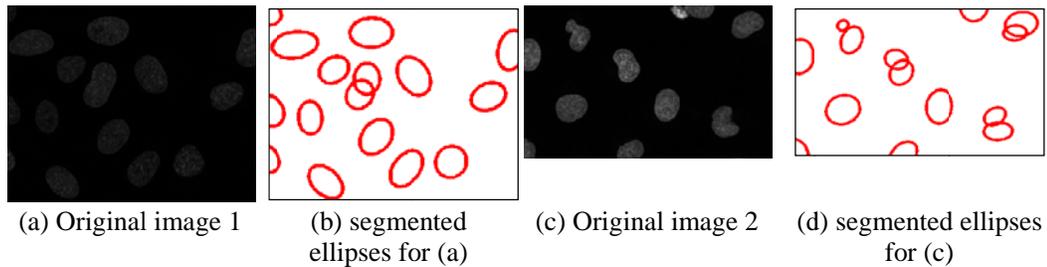

(a) Original image 1    (b) segmented ellipses for (a)    (c) Original image 2    (d) segmented ellipses for (c)

**Figure 6.2-5: Examples of images and ellipses segmented by the proposed method.**

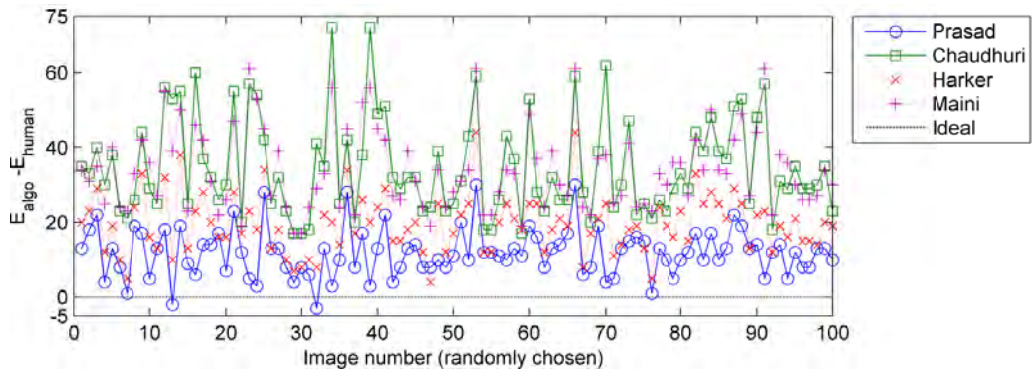

**Figure 6.2-6: Comparison of the performance for various least squares methods.**

### 6.2.2 Numerical Results

Two examples of the result of the proposed algorithm are shown in Figure 6.2-5. It is clearly seen that the proposed algorithm can segment the sub-cellular structures very well for low contrast as well as high contrast images, even when some of the sub-cellular structures may be occluded by other structures. The number of ellipses found using the proposed algorithm is denoted using $E_{\text{algo}}$ and the number of ellipses found by a human (occluded or otherwise) is denoted by $E_{\text{human}}$. The quantity $E_{\text{algo}} - E_{\text{human}}$ is plotted for 100 randomly chosen images (since it is difficult to collectively present and compare the results for all 444 images) in Figure 6.2-6. As highlighted in section 2.2, the method chosen for least squares ellipse fitting has a significant impact on the performance of the algorithm. Thus, four methods (Prasad [138] – used in the proposed algorithm and presented in section 4.4, Chaudhuri [84], Harker [174], and Maini [173]) are considered and their performances are compared in Figure 6.2-6. It is seen that Prasad gives the best performance, thus showing to be the best choice among the four least squares ellipse fitting methods.

**Table 6.2-1: Statistics of computation time without parallel processing**

|  | Average time (s) | Standard Deviation |
|---|---|---|
| Preprocessing | 0.52 | 0.43 |
| Ellipse fitting (Prasad) | 7.86 | 2.04 |
| Ellipse fitting (Chaudhuri) | 9.42 | 2.93 |
| Ellipse fitting (Harker) | 11.27 | 3.38 |
| Ellipse fitting (Maini) | 9.46 | 3.01 |
| Ellipse filtering | 0.02 | 0.02 |

**Table 6.2-2: Statistics of computation time with parallel processing (parallelization of step 2 of Figure 6.2-1 is performed using 8 parallel cores)**

|  | Average time (s) | Standard Deviation |
|---|---|---|
| Preprocessing | 0.52 | 0.43 |
| Ellipse fitting (Prasad) | 0.99 | 0.30 |
| Ellipse fitting (Chaudhuri) | 1.21 | 0.38 |
| Ellipse fitting (Harker) | 1.42 | 0.43 |
| Ellipse fitting (Maini) | 1.23 | 0.39 |
| Ellipse filtering | 0.02 | 0.02 |

The time comparison of the four methods is presented in Table 6.2-1 (without any parallelization). It is seen that among the four least squares ellipse fitting methods, Prasad takes the least computation time as well. It is also noted that the preprocessing and ellipse filtering steps take very little time. In fact, the computation time for each image can be easily reduced below 1 second by parallelizing the ellipse fitting block. This is illustrated in Table 6.2-2, where it is shown that the proposed method with Prasad's unconstrained least squares method takes less than 1 second for the most time consuming portion when 8 cores are used. The time performance can be further improved by increasing the number of cores and more effective programming. Thus, the proposed algorithm is capable of providing computation time less than the image acquisition time of a typical fluorescence microscope.

## 6.3 Object detection

The object detection problem is an advanced image processing application which finds potential application in many real life scenarios. The theory and implementation of object detection problem is an amalgamation of various techniques and theories in different fields of computer engineering, electrical engineering, and mathematics. For example, various feature types in images (textures, colors, edges, shapes, etc.), kernel and matrix theory, topology and graph theory, theory of probability, machine learning, feature matching, compact representation, parallel computing, etc.

The vast literature and research work in the problem of object detection inhibit a detailed discussion of these aspects here. A concise summary is presented in Appendix F. Here, the approach taken by us to illustrate the utility of the geometric primitives in the object detection problem is briefly presented.

The main distinguishing characteristics of the proposed objected detection method are as follows:

1. Only the geometric primitives (polygon and ellipse) have been used as the object features. This is quite rare in the object detection community.
2. A hierarchical object representation template (called hierarchical code) with both generative and discriminative qualities is used.

For convenience, the object detection method is called the Geometry based Hierarchical code Object Detection (GHOD) method. When the edge contour are used as the object features in the usual manner, the edge contours are used as they are without any smoothing, noise removal, data compression, or any other form of treatment. Due to this, if the edge features representing an object are learnt and tested on a true positive test image, the result of matching the edge feature and the detection may vary even if two different edge detection algorithms are used for generating the edge map of the test image. In order to alleviate this effect, instead of using the usual edge contours as features, here the polygonal approximation of the edge contours are used as the features. In addition to the removal of noise and the compression of the data, it also aids in designing more robust matching schemes for the approximate polygons. In addition to the approximate polygonal features, elliptic features are also used as geometric features.

A hierarchical model is used for representing the object. The details of the hierarchical model are presented in section 6.3.1. This model is a novel proposition and is quite different from the commonly used object representation models like bag of words, dictionary, codebooks, random Markov fields, etc. It has both generative and discriminative capabilities and can use any general feature type including but not limited to edge features, patch features, color features, texture features, geometric features, and spectral features.

The results of four object detection methods (including GHOD) for 6 classes are presented in section 6.3.3. It is seen that GHOD can indeed provide better results than the other three methods. The result of GHOD is also shown for 10 more object classes from a different dataset.

### 6.3.1 Hierarchical object representation

Instead of using the regular object class representation schemes like codebook, dictionary, bag of words, Markov random fields, etc., a new hierarchical code is used for representing each object class. The hierarchical code has several hierarchy levels in which the first level (top level) has the most generic (generative) features. The generic features have more likelihood ratios than the other features, implying that these are the most commonly appearing features in the true positive image. The features in the next level are less generic and more class-specific (discriminative) than the ones at the top. There is no restriction on the type of feature. The node may represent edge feature, contour feature, shape feature, patch feature, texture feature, color feature, kernel feature or any other kind of feature. Thus, the architecture of the object class is quite flexible. Further, the architecture also allows for scalability since a few or many levels may exist and there is no restriction on the number of children a node may contain.

An example of such a code is shown in Figure 6.3-1. Two training images of cars are considered on which the edge features, geometric features and texture based region features are shown. All of them have been heuristically generated only for the sake of illustration of the concept. The features that represent the object class car have been heuristically marked using three kinds of arrows and three different notations. The numbers 1-5 marked using large black arrows show the highest level in the hierarchical (most generic features). The alphabets a-i marked using medium brown arrows show next level of hierarchical code (these features are more discriminative than upper level features). The Greek alphabets α-γ marked using small arrows show the lowest level in the hierarchical code. The text boxes near features 1 and 5 show examples of the details of using the features 1 and 5. Such info about the expected variability and type of matching should be stored with each node.

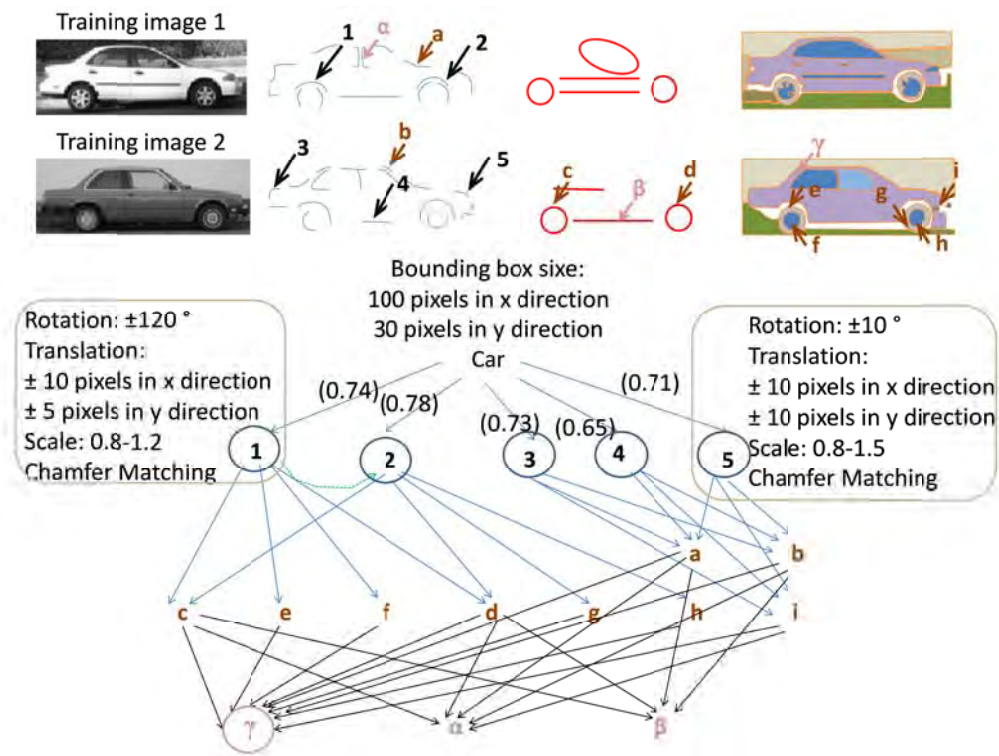

**Figure 6.3-1: An example of the proposed hierarchical code.**

Each connection is given a weight that is equal to the likelihood of presence of a feature given the upper node if the connection is present (i.e., conditional likelihood).

The hierarchical code as discussed above has several merits. First, beginning with a feature on top level, each traversal path represents the object model independent of the other paths. Thus, for a test image, even if matching along all the paths except one (the one with very high conditional likelihoods) indicate that the object may not be present, the object still stands a chance to be detected. However, it is unlikely that one of the paths in the code has very high conditional likelihoods since it is unlikely that an object can be represented very well by only a certain features. Thus, for a test image, if several paths indicate the presence of the object, the chances of correct detection increase significantly. Second, this model is expected to provide good discriminative performance as well because the lower level features may have their likelihoods very low, but conditional to the presence of generic features, their likelihood increases, which implies that the simultaneous presence of generic and discriminative features is required to infer the class. Third, appending or updating such hierarchical code is quite easy. Thus, adaptive learning techniques, learning while testing, and scalable training schemes (with no fixed training data size) may be used.

The architecture of the node, parent-to-child topology, and child to parent topology are next discussed. Each node in the hierarchical code can be uniquely identified as $\text{node}(l,k)$ or simply $(l,k)$ where $l$ is the level in the hierarchical code in which the node is present and $k$ is the index of the node in the level $l$. For the ease of reference, $(l,k)$ is referred to as the nodal address of the node. Each node has the following attributes:

1. Parent id (array) – this attribute stores the nodal address of the parents of a given node. Every node except the nodes in level 1 has at least one parent id. The level of each parent id is essentially $l-1$. It is denoted using $Parent(l,k) = \left\{ \left(l-1,k_p'\right); p = 1 \text{ to } N_{(l,k)} \right\}$ where $k_p'$ is the index of the $p$ th parent node and $N_{(l,k)}$ is the number of parents of the node $(l,k)$.

2. Feature details – this attribute has all the details related to the feature that the node represents. Its sub-attributes are feature type, feature, matching scheme (or algorithm), and the likelihood of the presence of the feature. These are represented as $Type(l,k)$, $Feature(l,k)$, $MatchSch(l,k)$, and $P(l,k)$. The likelihood of the presence of the feature is considered as a feature property since the likelihood of the presence of the feature is independent of the nodal address of the node representing the feature. It is required that the maximum value of the match between two features is 1, which indicates an exact match and the minimum value is zero, which indicates no match.

3. Matching parameters – this attribute stores the control parameters and thresholds used for the particular node when the nodal feature is matched to the features in an image. Currently, for a given matching algorithm, the control parameters are kept constant throughout the learning time and through all the nodes and features. However, this attribute is created so that in a later stage the adaptive learning of these parameters can be incorporated. This attribute is represented as $MatchPar(l,k)$.

4. Conditional likelihood of the node (array) – this attribute stores the likelihood of the presence of the feature subject to the condition that the parent node is present. This attribute is represented by $C(l,k) = \left\{ C\left(l, k, k'_p\right); k'_p = 1 \text{ to } N_{(l,k)} \right\}$ and it is an array of size $N_{(l,k)}$ (i.e. the number of parents of the node). For the nodes in level 1, the conditional likelihood value is the likelihood of the feature itself, i.e., $C(1,k) = P(1,k)$.

The nodes and child-to-parent topology (where child node identifies its parents nodes) for the code in Figure 6.3-1 is shown in Figure 6.3-2 as an example.

| | k = 1 | k = 2 | k = 3 | ... | |
|---|---|---|---|---|---|
| l = 1 | Node(1,1)<br>Parent(1,1)=NULL<br>Feature(1,1)=1<br>P(1,1)=0.74<br>MatchSch = 'Chamfer'<br>C(1,1)=P(1,1) | Node(1,2)<br>Parent(1,2)=NULL<br>Feature(1,2)=2<br>P(1,2)=0.78<br>C(1,2)=P(1,2) | Node(1,3)<br>Parent(1,3)=NULL<br>Feature(1,3)=3<br>P(1,3)=0.73<br>C(1,3)=P(1,3) | Node(1,4)<br>Parent(1,4)=NULL<br>Feature(1,4)=4<br>P(1,4)=0.65<br>C(1,4)=P(1,4) | Node(1,5)<br>Parent(1,5)=NULL<br>Feature(1,5)=5<br>P(1,5)=0.71<br>C(1,5)=P(1,5) |
| l = 2 | Node(2,1)<br>Parent(2,1)={(1,3),(1,4),(1,5)}<br>Feature(2,1)=α<br>P(2,1)=0.32<br>C(2,1)={0.65,0.51,0.71} | Node(2,2)<br>Parent(2,2)={(1,3),(1,4),(1,5)}<br>Feature(2,2)=ɮ<br>P(2,2)=0.32<br>C(2,2)={0.51,0.73,0.67} | Node(2,3)<br>Parent(2,3)={(1,1),(1,2)}<br>Feature(2,3)=c<br>P(2,3)=0.32<br>C(2,3)={0.71,0.66} | ... | Node(2,9)<br>Parent(2,9)={(1,3),(1,4),(1,5)}<br>Feature(2,9)=i<br>P(2,9)=0.21<br>C(2,N₂)={0.57,062,0.51} |
| l = 3 | Node(3,1)<br>Parent(3,1)={(2,1),(2,2),(2,3),(2,4),(2,5),(2,6),(2,7),(2,8),(2,9)}<br>Feature(3,1)=γ<br>P(3,1)=0.48<br>C(3,1)={0.91,0.54,0.62,0.71,0.78,0.56,0.67,0.69,0.76} | Node(3,2)<br>Parent(3,2)={(2,1),(2,2),(2,3),(2,4),(2,9)}<br>Feature(3,2)=α<br>P(3,2)=0.48<br>C(3,2)={0.84,0.63,0.92,0.76,0.55} | Node(3,1)<br>Parent(3,1)=γ={(2,1),(2,2),(2,3),(2,4)}<br>Feature(3,1)=β<br>P(3,1)=0.48<br>C(3,1)={0.91,0.67,0.59} | | |

**Figure 6.3-2: Sample nodes and child-to-parent topology for code in Figure 6.3-1.**

The above child-to-parent topology calls for more computation efficient learning of the tree as it uses lesser number of computations per node. If the code learning for each node in a level is done in parallel CPUs, then such topology benefits from faster computation.

However, the matching of features with the nodes in a hierarchical code is done in a top-down manner. So a parent-to-child topology is useful as well. This purpose can be easily met by a child look-up table. The look-up table for the code in Figure 6.3-1 is shown in Figure 6.3-3 as an example. General mathematical notation used to represent the children of a node is $Child(l,k) = \left\{ \left( l+1, \tilde{k}_c \right); c = 1 \text{ to } M_{(l,k)} \right\}$ where $c$ denotes the $c$ th child of the parent and $M_{(l,k)}$ is the number of children of the node $(l,k)$.

| Parent's nodal address | Child's nodal address |
|---|---|
| (1,1) | {(2,3),(2,4),(2,5),(2,6)} |
| (1,2) | {(2,3),(2,4),(2,7),(2,8)} |
| (1,3) | {(2,1),(2,2),(2,9)} |
| (1,4) | {(2,1),(2,2),(2,9)} |
| (1,5) | {(2,1),(2,2),(2,9)} |
| (2,1) | {(3,1),(3,2),(3,3)} |
| (2,2) | {(3,1),(3,2),(3,3)} |
| (2,3) | {(3,1),(3,2),(3,3)} |
| (2,4) | {(3,1),(3,2),(3,3)} |
| (2,5) | {(3,1)} |
| (2,6) | {(3,1)} |
| (2,7) | {(3,1)} |
| (2,8) | {(3,1)} |
| (2,9) | {(3,1),(3,2)} |
| (3,1) | NULL |
| (3,2) | NULL |
| (3,3) | NULL |

**Figure 6.3-3: Look up table for parent-to-child topology.**

The matching begins at the top and progress down the hierarchy. Thus, effectively, the matching begins with a high optimism to find an object, and become more and more decisive as the lower nodes in the path are traversed. Such matching is done for every traversal path in the code. This provides multiple paths in the object detection and recognition chain and improves the robustness of overall algorithm. The match score of the feature in a node $(l,k)$ with an input image is represented by $m_{self}(l,k)$. The match score of a node $(l,k)$ due to its children alone is denoted as $m_{child}(l,k)$. The net match score which is a combination of $m_{self}(l,k)$ and $m_{child}(l,k)$ is denoted as $m_{node}(l,k)$

Now, the scheme for calculating the net match between a hierarchical code and an image is presented. If the node $(l,k)$ has children $\left\{ \left( l+1, \tilde{k}_c \right); \tilde{k}_c = 1 \text{ to } M_{(l,k)} \right\}$, the match score due to children is computed using eqn. (6-5):

$$m_{child}(l,k) = E\left( C\left( l, \tilde{k}_c \right) m_{node}\left( l, \tilde{k}_c \right); \tilde{k}_c = 1 \text{ to } M_{(l,k)} \right) \qquad (6\text{-}5)$$

However, if the node has no child, $m_{child}(l,k)$ is assigned a value zero.

The net match score at the node $(l,k)$ is given by eqn. (6-6):

$$m_{node}(l,k) = \begin{cases} m_{self}(l,k) & \text{if } m_{child}(l,k) = 0 \\ \dfrac{m_{self}(l,k) + m_{child}(l,k)}{2} & \text{otherwise} \end{cases} \qquad (6\text{-}6)$$

In this way, the match score for the complete code can be computed in a bottom-up manner (beginning at the lowest level feature nodes), though the matching is actually done in a top-down manner. The match score of the complete code is compared against a threshold in order to make an inference.

### 6.3.2 Algorithmic setup

Geometric features – polygonal approximation of edge curves and ellipses detected in the images – are used as the object features. Since the polygons are described by vertices (as belonging to continuous domain), they can be scaled as desired for matching purposes. For convenience, each polygon feature to be compared is scaled and rotated such that the maximum dimension of the polygon is 90 pixels long and the polygon is centered at the point (50,50). Here the term maximum dimension is defined as the maximum distance between any pair of vertices of the polygon. Chamfer distance is used for matching the polygons (rotation threshold of 5 degrees). In addition, if the collection of all the features (learnt into the hierarchical code with their relative position intact) are restricted to be present in a bounding square box of size 200 pixels, then the relative positions of each polygon is stored as an extra information for the polygons. If two polygonal features have low Chamfer distance between them (threshold 0.25), the match score is considered as 'probably true' (temporary $m_{self} = 0.5$).

After matching all the features in the hierarchical code, the relative positions of the features with temporary $m_{self} = 0.5$ are translated so that the features in level 1 (the most likely features) have an offset of a maximum 10 pixels in either direction. The features are then rematched. The actual matching score $m_{self}$ is computed using eqn. (6-7):

$$m_{self} = \frac{m_{ch} + m_{cen}}{2} \qquad (6\text{-}7)$$

where $m_{ch}$ and $m_{cen}$ are the match scores due to Chamfer distance $d_{ch}$ of the polygons and maximum horizontal or vertical distance $d_{cen}$ between the centers. The expressions of $m_{ch}$ and $m_{cen}$ are given in eqns. (6-8) and (6-9) respectively:

$$m_{ch} = \frac{1 - d_{ch}}{0.75} \qquad (6\text{-}8)$$

$$m_{cen} = \frac{1 - 0.025 d_{cen}}{0.75} \qquad (6\text{-}9)$$

Similarly, for ellipses, first the centers of ellipses are translated to a center $(0,0)$ and the semi-major and semi-minor axes $a$ and $b$ are scaled such that the semi-major axes of the ellipses to be matched are both 100 pixels. The similarity ratio $D$ presented in section 5.5.1 is used to match the elliptic features. Similar to the polygons, elliptic features with $D > 0.75$ are assigned the match score as 'probably true' (temporary $m_{self} = 0.5$).

The procedure of translating the centers and rematching the features (as done for polygons) is then performed and the actual match score is calculated using eqn. (6-10).

$$m_{self} = \frac{m_{sim} + m_{cen}}{2} \qquad (6\text{-}10)$$

where $m_{sim} = D/0.75$ and $m_{cen}$ is computed using eqn. (6-10).

The likelihoods and conditional likelihoods are learnt using the Bayesian learning approach. For each class, the initial learning of features is done using 10 positive training images (with bounding box specified) and 5 negative training images. The features with likelihood of more than 0.7 are assigned to the level 1 of the hierarchical codes. The hierarchical code is then built with the condition that a new feature shall be added as child to parent node if the condition likelihood of the child for the parent is at least 0.4. After constructing the hierarchical code with 10 positive and 5 negative training images, the features unused in the hierarchical code are discarded. When a new training image comes, the likelihoods and conditional likelihoods of all the nodes are adjusted. After the re-training (essentially similar to validation) is done for a set of 15 new images (2:1 ratio for positive and negative images), the hierarchical code is flattened and reconstructed using the same conditions as mentioned above. Before allowing for testing, condition that the training and re-training has been done on at least 30 positive training and validation images is imposed.

### 6.3.3 Numerical results

The first set of results are presented for 6 classes of objects in the Graz-17 dataset [2]. For these classes, the results of three object detections based on edge curves and geometric features [2-4] are compared with GHOD. The results and comparison are presented in Table 6.3-1. It is seen that GHOD indeed performs better than the three methods for most of the classes considered.

**Table 6.3-1. Performance comparison with recent methods based on edge or shape primitives**

| Class | GHOD (proposed) | | | Opelt [2] | Shotton [3] | Chia [4] | |
|---|---|---|---|---|---|---|---|
| No. of test images↓ | | RP-AUC | RP-EER | RP-EER | RP-AUC | RP-AUC | RP-EER |
| **Motorbike** | 100 | 0.9999 | 0.7% | 4.4 | 1.0000 | 0.9996 | 1.0% |
| **Car-rear** | 100 | 0.9935 | 1.1% | 2.3 | 0.9912 | 0.9797 | 4.0 |
| **Car-2/3-rear** | 14 | 0.9152 | 8.3% | 12.5 | 0.6925 | 0.6843 | 35.7% |
| **Bike-side** | 50 | 0.9504 | 5.7% | 28.0 | 0.6957 | 0.8299 | 18.8% |
| **Mug** | 20 | 0.9734 | 4.3% | 6.7 | 0.9035 | 0.9619 | 5.0% |
| **Cup** | 20 | 0.9196 | 10.1% | 18.8 | 0.9158 | 0.8964 | 15.0% |

**Table 6.3-2: Performance of GHOD for various classes of Caltech-256 [1] dataset**

| Class | RP-AUC | RP-EER |
|-------|--------|--------|
| Car-tire | 0.8352 | 4% |
| Mountain-bike | 0.8546 | 7.10% |
| Bowling-ball | 0.7568 | 15% |
| Video-projector | 0.9854 | 2.10% |
| Boom-box | 0.6584 | 21.40% |
| Calculator | 0.9241 | 6.70% |
| Soda-can | 0.8142 | 18.30% |
| Coffee-mug | 0.8912 | 3.60% |
| Car-side | 0.9598 | 2.30% |
| Motorbikes | 0.8953 | 9.20% |

Here, it should be highlighted that GHOD has not yet been tested on articulate object classes like cow, horse, etc., and the classes considered till now indeed have some important geometric consistencies. For example, cars and bikes do have tires or elliptic groves for them. Similarly, in general (though not always), images of mugs and cups may have ellipses to denote their open faces.

Other classes with possible strong geometric features have been considered in Table 6.3-2, which lists the performance of GHOD for 10 object classes in Caltech 256 dataset [1]. It is seen that GHOD gives good detection results for these categories as well.

However, it is recommended that GHOD should be tested for more variety of classes, especially articulate classes. Some parametric changes might be needed for this purpose. Further, the usual Chamfer distance based matching of polygons may not be suitable for articulate objects. The feature types should also be increased. Texture features and kernel based features are the next types of features that the author wants to include in GHOD. It is also intended that the semi-supervised learning, learning while testing (adaptive learning), and machine learning of the control parameters be included in future. Thus, the current version of GHOD is a work in progress and GHOD is expected to work better and for more variety of classes in future. Nevertheless, the work presented here conclusively supports the utility of geometric primitive features for object detection.

# Chapter 7 : Conclusion and future work

## 7.1 Conclusions

The focus of the entire body of work presented in this thesis is on the geometric primitives present in the images. Three important geometric primitives, polygons, tangents, and ellipses are discussed. For each of these primitives, some fundamental and important results have been derived.

For the polygonal approximation (PA) of digital curves, two main points have been addressed in Chapter 2:

1. While approximating the digital curves using polygons, it is difficult to optimize both the local and global qualities of fit. None of the known performance metrics are capable of providing a balance of the local and global qualities of fit.

2. The known optimization goals are based on heuristics but it can be shown that there is a definite upper bound of the error due to PA and thus the heuristics are not needed. In fact, if heuristics are used, the PA methods may over-fit and under-fit on certain sections of the edge curves.

For the first point above, simple metrics – precision for local quality of fit and reliability for global quality of fit – are proposed. Explicit derivations show that the local and global qualities of fit are always at conflict when least squares fitting is used for obtaining the polygonal approximation. It is also seen that the existing metrics focus on either the local quality of fit of the global quality of fit. Thus, if the usual performance metrics are used in the optimization goals of the PA methods, it is expected that the PA methods will also focus on only one of these aspects. Precision and reliability optimization (PRO) method proposed in the thesis allows for simultaneous optimization of the local and global quality of fit and can be easily tweaked for the desired performance.

For the second point, an explicit derivation shows that the error due to digitization can be analytically derived and it is a convergent upper bounded term. This upper bound depends upon the length of the line segment (edge of the PA) and its orientation. Thus, using a fixed heuristically chosen threshold in a PA method is unsuitable. This is because if the threshold is more than the upper bound, the fit for that edge is poor and if the threshold is less than the upper bound, unnecessary computation resources are wasted in optimizing the edge though the error cannot be reduced further due to digitization. Based on the upper bound, a non-parametric framework which can be used in most PA method is proposed. Three distinct methods have been adapted into this framework (RDP-mod, Masood-mod, and Carmona-mod) and show good performance in the non-parametric framework. Other PA methods are also compared in the context of the precision, reliability, and the upper bound of the error due to digitization.

In summary of the performance of all the PA methods proposed in Chapter 2, several practical aspects like performance of the proposed PA methods for scaled digital curves, noisy digital curves, non-digital curves, as well as for images in huge datasets are considered. The results show that the proposed PA methods perform quite well for all these practical aspects. This corroborates with the two fundamental but often neglected points mentioned above.

For the problem of tangent estimation (TE) in digital curves, addressed in Chapter 3, a simple and extremely cost effective method DEB for estimating the tangents of the digital curves is proposed. Using an analytical derivation, it is shown that DEB is quite simple, the error in TE by the method has a definite upper bound for the conic curves. Though in principle, it can be shown that the method has a definite upper bound for curves that can be represented by polynomial bases, the derivation might be quite tedious and thus has not been presented here. Nevertheless, DEB is tested extensively for digital and continuous conics as well as several digital non-conic curves and DEB performs quite well. In fact, the comparison of the error of DEB with 72 other TE methods show that despite the simplicity, DEB in general outperforms all the methods.

In Chapter 4, the popularly used method of Fitzgibbon [80] is improved such that the modified method NSAF is numerically stable. However, the more important contribution of Chapter 4 is the unconstrained, non-iterative, numerically stable, computation efficient least squares fitting method ElliFit, which is based on minimization of the geometric distance (instead of conventionally used algebraic distance) and can be easily implemented. Several important and fundamental mathematical concepts are employed to design ElliFit and derive the salient features exhibited by ElliFit. Numerical results show that the method performs better than other least squares based fitting methods for a variety of challenging experiments.

Chapter 5 proposed a hybrid ellipse detection method ECC. In the development of ECC, several interesting contributions have been proposed. The focus has been on designing simple computation efficient ways of solving some problems of discrete geometry. One example is the method of detection of inflexion points in digital curves. Another example is the convexity analysis of the curves in the form of detecting the search region for edge contours (section 5.4.1). The concept of associated convexity which gives the convexity relationship between two edge curves is also worth mentioning. Similarly, a simple overlap measure designed using Jaccard index solves the problem of finding the area overlap between two ellipses (or their digital counter parts).

Another set of contributions of ECC is the saliency measures and a non-heuristic parameter-independent scheme of selecting good elliptic hypotheses for an image. Three simple saliency criteria, which can be easily computed, are used to quantify the different aspects of the quality of ellipse fit on a group of edges. The selection scheme chooses the threshold parameters from the images themselves and do not require user-specified heuristic inputs. Examples show that the non-parametric selection scheme proposed in eqn. (5-16) performs better than other schemes.

Numerical experiments on 1200 synthetic images show that ECC performs much better than several hybrid ellipse detection methods and gives recall, precision, and F-measure close to 100%. In fact, ECC can be implemented in three different schemes. For more time critical applications, simpler scheme 1 can be used which uses lesser steps while grouping. However, for better accuracy and performance, scheme 3 which incorporates all the steps is the best scheme.

The work in Chapter 2 - Chapter 5 serves to highlight that a careful analysis of the discrete geometry problem – in  continuous domain as well as the effect of digitization on the geometry – can help the researchers to design simple, cost-effective, typically error-bounded methods for extracting and manipulating geometric primitives in digital images. This is consistent with the basic theme and the main aim of this thesis.

The practical usage of the work in Chapter 2 - Chapter 5 is shown in Chapter 6 which presents three practical image processing applications. For the first application – detection of ellipses in real images for robotics applications – the ECC method is shown to perform better than other methods. However, the time taken by it is more than two methods among the methods used for comparison.

The Caltech 256 dataset is a sufficiently complex and varied dataset with real images which emulate real life scenario sufficiently. Thus, the good performance of ECC indicates that the method can be widely used for practical applications. Some interesting application may be surface grain analysis of materials, medical diagnostics (like malarial and sickle cell counting), analysis of geological and astronomical pictures, assisted robotics for applications like sorting ores, vegetables, and cans, etc. in automatic mineral processing, grocery, and recycling units. For such dedicated applications, the method can be customized for better and faster performance. Further, parallelization schemes can be incorporated for faster processing.

For the second application – detecting cell organelles in microscopy images – ElliFit is shown to perform very well. The computation time is reduced to below 1 second using parallel computation over 8 cores. This time is less than the typical acquisition time of the fluorescence microscopes. Thus, this method can be used for online cell counting. The diagnosis and analysis of the biological tests can be sped up considerably and made less dependent on human expertise using such methods.

For the third application – object detection using geometric primitives only – it is seen that the geometric primitives are indeed sufficient to provide reasonable accuracy in object detection as compared to other methods. The accuracy demonstrated is indeed insufficient for real life object detection application. But it is well-known in the object detection research community that no single type of feature is sufficient for providing good object detection performance. Thus, the performance of the proposed object detection method should be improved by including more feature types. Texture and kernel features should provide good and complimentary variation in the feature types. Thanks to the flexible hierarchical object template which can use any type of features, such integration of geometric features, texture features, and kernel features can be done easily.

## 7.2 Specific contributions

In order to summarize the major and significant contributions of the thesis, the algorithms and important concepts presented in this thesis are listed here.

### 7.2.1 Algorithms proposed

For the problem of polygonal approximation of digital curves, four algorithms have been proposed. Modified Ramer-Douglas-Peucker (RDP(mod)) method for polygonal approximation of digital curves – proposed in section 2.5.1 – presents a parameter independent modification of Ramer-Douglas-Peucker method. Precision and reliability optimization (PRO) method for polygonal approximation of digital curves – proposed in section 2.3 – optimizes both precision and reliability metrics for PA and provides the capability to tweak the algorithm for the desired performance of fit (very fine or very crude fit). Modified Carmona-Poyato's (Carmona(mod)) method for polygonal approximation of digital curves – proposed in section 2.5.3 – presents a parameter independent modification of Carmona-Poyato's method. Modified Masood's (Masood(mod)) method for polygonal approximation of digital curves – proposed in section 2.5.2 – presents a parameter independent modification of Masood's method.

For the problem of tangent estimation of digital curves, Definite Error Bounded (DEB) method – proposed in section 3.3 – is a simple and computation efficient method for estimating tangents of digital curves which performs better than most other TE methods for a variety of experiments.

For the problem of ellipse fitting on edge curves, two algorithms have been proposed. Numerically stable algebraic fitting (NSAF) method – proposed in section 4.2 – is a simple modification of Fitzgibbon's method of least squares ellipse fitting using algebraic distance. Ellipse Fitting (ElliFit) method – proposed in section 4.4.5 – an unconstrained, non-iterative, numerically stable, and computation efficient method for least squares fitting of ellipses using geometric distance.

Several algorithms have been proposed for dealing with several sub-problems in the problem of ellipse detection in images using edge curves. However, two algorithms among them are the most notable ones. Method for detecting inflexion points in digital curves – proposed in section 5.3.2 – is a low computational complexity algorithm for detecting the inflexion points in digital edge curves. Edge curvature and convexity (ECC) ellipse detection method – proposed in section 5.2 – is a hybrid ellipse detection method for ellipse detection from discontinuous digital curves.

### 7.2.2 Important derivations and concepts with fundamental insights

The following important derivations and concepts which provided fundamental insights into geometric primitives for digital curves were presented in the thesis.

Precision and reliability metrics and the conflict in the global and local quality of fit for least squares framework was presented in section 2.2. Derivation of the error due to digitization of a continuous line segment and the non-parametric framework for PA were presented in section 2.4.

Derivation of the analytical expression of the error of DEB for continuous conics and digital conics was presented in section 3.4.

Use of the Bauer Skeel condition number and statistical moments to quantify and reduce the numerical instability of the least squares ellipse fitting method was demonstrated in sections 4.2.2 and 4.4.4. The modification of the minimization function (geometric distance) in order to enable non-iterative linear least squares formulation was presented in section 4.4.1. The definition of intermediate variables $\phi_1$ to $\phi_5$ in ElliFit such that the ellipse fitting problem can be split into two operators – one linear well-posed operator and another non-linear but injective and easily invertible operator – was presented in section 4.4.2.

Concepts of search region and associated convexity used in ECC used for grouping edges were presented in sections 5.4.1 and 5.4.2. Relationship score for ranking the edges in a group was presented in section 5.4.4. Saliency criteria and non-heuristic parameter-independent selection scheme used in ECC was presented in section 5.5.

Selection of control parameter $t_1$ for enhancing the contrast and binarizing the microscopy images of bio cell organelles was presented in section 6.2.1.1. Hierarchical object template used in object detection method was presented in section 6.3.1.

## 7.3 Future work

The work presented in this thesis has a lot of future potential. For convenience, the potential future directions are classified into two types – 1) theoretical fundamentals and algorithms design; 2) Practical applications. In each of these categories, the potential future work directions have been listed in a point wise manner.

### 7.3.1 Theoretical fundamentals and algorithms design

**Identifying the shape distortion model without knowing optical model:** It is well known that line gets distorted to a hyperbolic curve in a point camera or fish eye lens model. Other forms of distortion are also present due to optics or perspective. In some cases, the distortion can be corrected by knowing the optics model. However, it is of practical importance if the distortion model can be predicted without knowing the optics model. The easiest cues for such distortion modeling are the intersecting digital hyperboles which can be correlated to the corners and edges of rectangular objects. This can be used for distortion model prediction.

**Three dimensional polygonal surface fitting:** All the geometric primitives considered in this thesis are in the two-dimensional digital domain. It is of theoretical as well as practical interest to extend the work of non-parametric polygonal approximation to three-dimensional spaces. This is of practical interest for new and advanced applications like fast graphics and motion rendering in 3D video and gaming technology and virtual display environments. It might also be of significant importance for generating robust computational meshes.

**Object super-classes:** The hierarchical template of GHOD can be used for demonstrating super-classes and re-using features (across classes) by performing merge and comparison of object templates of two classes. Another possibility with the hierarchical template is the clustering of object classes based on the generic features [189]. The idea is that super-classes of the object classes may be identified based on the generic features. Accordingly, a newly learnt object class can also be identified as potentially belonging to a super-class, and already learnt features may be reused. For example, the learnt codes may have the potential of grouping horses, cows, dogs, etc. as animals; cups, vases, kettles, etc. as pots; bicycles, cars, carts, etc. as vehicles; and so on.

**Spectral matching of polygons:** It is discussed at the end of section 6.3.3 that Chamfer distance based matching of polygons may not be quite suitable for articulate objects like animals and machines. The authors are currently working on spectral matching techniques for this problem [190].

**Non-digitization noises:** most of the work presented in this thesis consider only digitization noise. However, the impact of other forms of noises should also be incorporated in the derivations of the upper bounds presented in sections 2.4 and 3.4. It is expected that some form of signal to noise ratio or noise variance parameter will appear in the expression of these error bounds.

**Error bound of DEB for polynomial curves:** In general any continuous curve can be represented in terms of polygonal bases. Thus, deriving a general error bound for polygonal curves of any order shall establish the validity of DEB for any digital curve. This will be an important future work. **Parameter-independent DEB:** Using a combination of RDP-opt and a carefully designed function for determining the value of control parameter $R$ of DEB, it must be possible to make DEB control parameter independent. This may increase the computation cost of DEB by a small factor but will help in making the algorithms non-heuristic.

**Least squares fitting of conics using geometric distance:** The extension of least squares ellipse fitting method for other conics can also be done. However, it will be interesting to see if least squares fitting of polynomial curves based on geometric distance can be done with greater efficacy than the algebraic form of fitting. **Fitting of three dimensional geometric surfaces** like ellipsoids, hyperboloids, paraboloids, catenaries, etc.: This are of importance in civil engineering, computer simulation of modern architectural structures, etc.

### 7.3.2 Practical applications

**Fluid and field velocities computation using a combination of PA and TE methods:** Given time stamped images or video of the fluid profiles (pressure maps) or intensity maps, or other kind of field distributions, a combination of PA (for edge contours corresponding to different gray levels) and TE, the normal outward velocity at various location can be computed. These normal outward velocities can then be used for predicting the pressure or field flow or to predict the approximate dynamic model. This can be especially important for oceanic current analysis in oceanology, wind current analysis in climatology, grain growth analysis in material technology, and other such scientific applications.

**Stereo imaging using PA:** For two stereo-shifted images, when PA methods are applied, the key points of the objects in the two images can be identified and correlated to each other. Based on the relative displacement between these correlated pairs of points and the sensor arrangement, the points can be back-projected to obtain the depth data in an accurate and fast manner. **Fast object tracking and rendering for sports and ballistic videos and games:** This shall require a combination of PA and fast polygon matching algorithms for designing computation efficient and very fast performance.

**Fast pupil or iris tracking:** A combination of suitable binarization scheme, smoothing operations, PA, and least squares method can be used for real time tracking of pupil or iris of eye. **Shape based cell classification in bio-medical images:** A combination of ellipse fitting, polygon matching, and sorting based on shapes and sizes can be used for automatic cell sorting, classification, cell counting, and other diagnostic functions.

**Non-linear optics and camera calibration:** Given some reference shapes, profiles, and illuminations, the images captured in image sensors can be processed using conic shape fitting. Such fitting can give insight into either numerical correction (using algorithms) or optical aberration correction (using non-linear optics) required for correcting the optical aberration effects.

The above mentioned are some of the potential future directions of the work in this thesis. It can be predicted that there are several other possible avenues for this work. The author intends to work on at least a few of the above.

## 7.4 One sentence summary

In summary, it can be seen that if the discrete geometry is studied carefully, simple and effective solutions for various fundamental problems can be devised and used in several practical applications with significant impact.

# Appendix

## A. Proof of eqn. (2-8)

**Theorem.**

If $\mathbf{X} = \left[ \begin{bmatrix} x_1 & x_2 & \cdots & x_M \end{bmatrix}^T \quad \begin{bmatrix} y_1 & y_2 & \cdots & y_M \end{bmatrix}^T \right]$ and $\mathbf{B} = \mathbf{X} \left( \mathbf{X}^T \mathbf{X} \right)^{-1} \mathbf{X}^T = \left[ b_{i,j} \right]$,

then $\sum_{i=1}^{M} \sum_{j=1}^{M} b_{i,j} \leq M$

**Proof.**

The main concept used in the proof is that of raw moments, central moments, and correlation coefficient used in bivariate statistics. Accordingly, $m_{p,q}$ and $\mu_{p,q}$ represent the raw moment and central moment respectively of $x^p y^q$ :

$$m_{p,q} = \frac{1}{M} \sum_{i=1}^{M} x^p y^q \tag{1}$$

$$\mu_{p,q} = \frac{1}{M} \sum_{i=1}^{M} \left( x - m_{1,0} \right)^p \left( y - m_{0,1} \right)^q \tag{2}$$

$\left( \mathbf{X}^T \mathbf{X} \right)$ can be written in terms of the raw moments as follows:

$$\mathbf{X}^T \mathbf{X} = \begin{bmatrix} \sum_{i=1}^{M} x_i^2 & \sum_{i=1}^{M} x_i y_i \\ \sum_{i=1}^{M} x_i y_i & \sum_{i=1}^{M} y_i^2 \end{bmatrix} = M \begin{bmatrix} m_{2,0} & m_{1,1} \\ m_{1,1} & m_{0,2} \end{bmatrix} \tag{3}$$

Thus, $\left( \mathbf{X}^T \mathbf{X} \right)^{-1}$ can be written as:

$$\left( \mathbf{X}^T \mathbf{X} \right)^{-1} = d \begin{bmatrix} m_{0,2} & -m_{1,1} \\ -m_{1,1} & m_{2,0} \end{bmatrix} \tag{4}$$

where $d = M^{-1} \left( m_{2,0} m_{0,2} - m_{1,1}^2 \right)^{-1}$ is defined for convenience.

Given that $\mathbf{B} = \mathbf{X}\left(\mathbf{X}^{\mathrm{T}}\mathbf{X}\right)^{-1}\mathbf{X}^{\mathrm{T}}$, $\mathbf{B}$ can be expanded as follows:

$$\mathbf{B} = \left[\, b_{i,j} \,\right] \tag{5}$$

where:

$$b_{i,j} = d\left(m_{0,2}x_i x_j - m_{1,1}\left(x_i y_j + x_j y_i\right) + m_{2,0}y_i y_j\right); \quad i, j = 1 \text{ to } M \tag{6}$$

Thus, $\displaystyle\sum_{i=1}^{M}\sum_{j=1}^{M} b_{i,j}$ is written as:

$$\sum_{i=1}^{M}\sum_{j=1}^{M} b_{i,j} = d\left(m_{0,2}\sum_{i=1}^{M}\sum_{j=1}^{M}\left(x_i x_j\right) - 2m_{1,1}\sum_{i=1}^{M}\sum_{j=1}^{M}\left(x_i y_j\right) + m_{2,0}\sum_{i=1}^{M}\sum_{j=1}^{M}\left(y_i y_j\right)\right) \tag{7}$$

The following identity can be used to simplify eqn. (7):

$$\sum_{i=1}^{M}\sum_{j=1}^{M}\left(a_i b_j\right) = \left(\sum_{i=1}^{M} a_i\right)\left(\sum_{i=1}^{M} b_j\right) \tag{8}$$

Thus, $\displaystyle\sum_{i=1}^{M}\sum_{j=1}^{M}\left(x_i x_j\right) = \left(\sum_{i=1}^{M} x_i\right)^2 = M^2 m_{1,0}^2$, $\displaystyle\sum_{i=1}^{M}\sum_{j=1}^{M}\left(x_i y_j\right) = \left(\sum_{i=1}^{M} x_i\right)\left(\sum_{i=1}^{M} y_i\right) = M^2 m_{1,0} m_{0,1}$,

and $\displaystyle\sum_{i=1}^{M}\sum_{j=1}^{M}\left(y_i y_j\right) = \left(\sum_{i=1}^{M} y_i\right)^2 = M^2 m_{0,1}^2$. Consequently, simplified form of eqn. (7) is

shown in eqn. (9) below:

$$\sum_{i=1}^{M}\sum_{j=1}^{M} b_{i,j} = M^2 d\left(m_{0,2}m_{1,0}^2 - 2m_{1,1}m_{1,0}m_{0,1} + m_{2,0}m_{0,1}^2\right) \tag{9}$$

Now, the following identities (10) - (12) from bivariate statistics can be used in eqn. (9)

$$m_{1,0}^2 = m_{2,0} - \mu_{2,0} \tag{10}$$

$$m_{1,0}m_{0,1} = m_{1,1} - \mu_{1,1} \tag{11}$$

$$m_{0,1}^2 = m_{0,2} - \mu_{0,2} \tag{12}$$

to obtain:

$$\sum_{i=1}^{M}\sum_{j=1}^{M} b_{i,j} = M^2 d \left( m_{0,2} m_{2,0} - m_{0,2}\mu_{2,0} - 2m_{1,1}^2 + 2m_{1,1}\mu_{1,1} + m_{2,0}m_{0,2} - m_{2,0}\mu_{0,2} \right)$$

$$= M^2 d \left( 2d^{-1}M^{-1} - m_{0,2}\mu_{2,0} + 2m_{1,1}\mu_{1,1} - m_{2,0}\mu_{0,2} \right)$$

$$= M \left( 2 - \frac{m_{0,2}\mu_{2,0} - 2m_{1,1}\mu_{1,1} + m_{2,0}\mu_{0,2}}{\left( m_{2,0}m_{0,2} - m_{1,1}^2 \right)} \right) \tag{13}$$

$$= M \left( 2 - \frac{\dfrac{\mu_{2,0}}{m_{2,0}} - 2\dfrac{m_{1,1}^2}{m_{2,0}m_{0,2}}\dfrac{\mu_{1,1}}{m_{1,1}} + \dfrac{\mu_{0,2}}{m_{0,2}}}{\left( 1 - \dfrac{m_{1,1}^2}{m_{2,0}m_{0,2}} \right)} \right)$$

Now, since all the terms in (10) and (12) are positive definite, inequality (14) below is valid:

$$0 < \frac{\mu_{2,0}}{m_{2,0}}, \frac{\mu_{0,2}}{m_{0,2}} < 1 \tag{14}$$

Also, using Cauchy-Schwartz inequality $\left( \sum_{i=1}^{M} a_i b_i \right)^2 \leq \left( \sum_{i=1}^{M} a_i^2 \right)\left( \sum_{i=1}^{M} b_i^2 \right)$, eqn. (15) can be obtained:

$$\frac{m_{1,1}^2}{m_{2,0}m_{0,2}} \leq 1 \tag{15}$$

Now two cases corresponding to no correlation and perfect (positive or negative) correlation between the $x$ and $y$ variables are considered. The correlation is defined using the correlation coefficient:

$$r_{xy} = \frac{\mu_{1,1}}{\sqrt{\mu_{2,0}\mu_{0,2}}} \tag{16}$$

**Case 1: No correlation** $r_{xy=0}$.

For this case, $\mu_{1,1} = 0$ and $\dfrac{m_{1,1}^2}{m_{2,0}m_{0,2}}$ can be written using (10) - (12) as:

$$\frac{m_{1,1}^2}{m_{2,0}m_{0,2}} = \frac{\mu_{1,0}^2 \mu_{0,1}^2}{m_{2,0}m_{0,2}} = \frac{\left( m_{2,0} - \mu_{2,0} \right)\left( m_{0,2} - \mu_{0,2} \right)}{m_{2,0}m_{0,2}} = 1 - \left( \frac{\mu_{2,0}}{m_{2,0}} + \frac{\mu_{0,2}}{m_{0,2}} \right) + \frac{\mu_{2,0}}{m_{2,0}}\frac{\mu_{0,2}}{m_{0,2}} \tag{17}$$

Thus eqn. (13) in case one can be simplified to eqn. (18) below:

$$\sum_{i=1}^{M}\sum_{j=1}^{M}b_{i,j} = M\left(2 - \frac{\dfrac{\mu_{2,0}}{m_{2,0}} + \dfrac{\mu_{0,2}}{m_{0,2}}}{\left(\dfrac{\mu_{2,0}}{m_{2,0}} + \dfrac{\mu_{0,2}}{m_{0,2}}\right) - \dfrac{\mu_{2,0}}{m_{2,0}}\dfrac{\mu_{0,2}}{m_{0,2}}}\right) \tag{18}$$

Using eqn. (14) that the term in the parentheses is less than 1 and hence, $\sum_{i=1}^{M}\sum_{j=1}^{M}b_{i,j} < M$ .

**Case 2: Perfect correlation** $r_{xy} = \pm 1$**.**

For this case, eqn. (11) can be written using eqn. (16) and $r_{xy} = \pm 1$ as:

$$\left(m_{1,0}m_{0,1}\right)^2 = \left(m_{1,1} - \mu_{1,1}\right)^2 = m_{1,1}^2 + \mu_{2,0}\mu_{0,2} - 2m_{1,1}\mu_{1,1} \tag{19}$$

Further, using eqns. (10) and (12), eqn. (19) can be written as:

$$\begin{aligned}
2m_{1,1}\mu_{1,1} &= m_{1,1}^2 + \mu_{2,0}\mu_{0,2} - \left(m_{1,0}m_{0,1}\right)^2 \\
&= m_{1,1}^2 + \mu_{2,0}\mu_{0,2} - \left(m_{2,0} - \mu_{2,0}\right)\left(m_{0,2} - \mu_{0,2}\right) \\
&= m_{1,1}^2 - m_{2,0}m_{0,2} + \mu_{2,0}m_{0,2} + \mu_{0,2}m_{2,0} \\
&= m_{2,0}m_{0,2}\left(\frac{\mu_{2,0}}{m_{2,0}} + \frac{\mu_{0,2}}{m_{0,2}} - 1 + \frac{m_{1,1}^2}{m_{2,0}m_{0,2}}\right)
\end{aligned} \tag{20}$$

Eqn. (13) can be written by substituting eqn. (20) as in eqn. (21) below:

$$\begin{aligned}
\sum_{i=1}^{M}\sum_{j=1}^{M}b_{i,j} &= M\left(2 - \frac{\dfrac{\mu_{2,0}}{m_{2,0}} - 2\dfrac{m_{1,1}^2}{m_{2,0}m_{0,2}}\dfrac{\mu_{1,1}}{m_{1,1}} + \dfrac{\mu_{0,2}}{m_{0,2}}}{\left(1 - \dfrac{m_{1,1}^2}{m_{2,0}m_{0,2}}\right)}\right) \\
&= M
\end{aligned} \tag{21}$$

From the above, it can be deduced that the maximum value of $\sum_{i=1}^{M}\sum_{j=1}^{M}b_{i,j}$ is $M$ and it happens only for perfectly correlated variables $x$ and $y$ .

## B. Solution $\Delta\theta_i, i = 1$ to $2$ for the simultaneous equations (3-15) and (3-20)

For the model of conics given by eqn. (3-18)–(3-20) and the equation of the slope of the tangent given by eqn. (3-21) computed at a point $P_0(r_0, \theta_0)$, it is required to calculate the points $P_1(r_1, \theta_1)$, and $P_2(r_2, \theta_2)$ on the conic as well as the circle given by eqn. (3-15). For convenience, $\theta_i = \theta_0 + \Delta\theta_i$, $i = 1$ and $2$ is used, where $\Delta\theta_i, i = 1$ to $2$ are the two solutions of the simultaneous equations (3-20) and (3-15). Accordingly, eqns. (3-20) and (3-15) are solved to find the points $P_1$ and $P_2$ as follows (while truncating the terms higher than the second order of $\Delta\theta$):

For simplicity of expressions, the assignments in eqn. (22) are used.

$$k = 1 - e\cos\theta; \ k_0 = 1 - e\cos\theta_0 \tag{22}$$

1. Eqn. (23) is obtained by substituting eqns. (3-19) and (3-20) in eqn. (3-15).

$$k^{-2} + k_0^{-2} - 2\cos\Delta\theta\left(k\,k_0\right)^{-1} = \left(R/ae\right)^2 \tag{23}$$

2. Using Taylor series expansion [112] for $\cos\Delta\theta$ given in eqn. (24), eqn. (25) is obtained.

$$\cos\Delta\theta = 1 - \left(\Delta\theta\right)^2\Big/2 + O\left(\left(\Delta\theta\right)^4\right) \tag{24}$$

$$\left(k^{-1} - k_0^{-1}\right)^2 + \left(\Delta\theta\right)^2\left(k\,k_0\right)^{-1} = \left(R/ae\right)^2 \tag{25}$$

3. Simplifying the above, eqn. (26) is obtained.

$$e^2\left(\cos\theta - \cos\theta_0\right)^2\left(k\,k_0\right)^{-2} + \left(\Delta\theta\right)^2\left(k\,k_0\right)^{-1} = \left(R/ae\right)^2 \tag{26}$$

4. Using Taylor series expansion for $\cos\theta$ given in eqn. (27) in the first term (left hand side), eqn. (28) is obtained.

$$\cos\theta = \cos\theta_0 - \sin\theta_0\left(\Delta\theta\right) - \cos\theta_0\left(\Delta\theta\right)^2\Big/2 + O\left(\left(\Delta\theta\right)^3\right) \tag{27}$$

$$\left(\Delta\theta\right)^2\left(e^2\left(\sin\theta_0 + \cos\theta_0\left(\Delta\theta/2\right)\right)^2\left(k\,k_0\right)^{-2} + \left(k\,k_0\right)^{-1}\right) = \left(R/ae\right)^2 \tag{28}$$

*5.* Eqn. (29) is derived by truncating the higher order terms of $\Delta\theta$.

$$\left(\Delta\theta\right)^2\left(e^2\sin^2\theta_0 + k\,k_0\right) = \left(R/ae\right)^2\left(k\,k_0\right)^2 \tag{29}$$

6. Using Taylor series expansion for $\cos\theta$ given in eqn. (27), eqn. (29) can be written as eqn. (30):

$$\left(\Delta\theta\right)^2\left\{e^2\sin^2\theta_0 + k_0\left(k_0 + e\sin\theta_0\left(\Delta\theta\right)\right)\right\} = \left(R/ae\right)^2\left\{k_0\left(k_0 + e\sin\theta_0\left(\Delta\theta\right)\right)\right\}^2 \tag{30}$$

7. Eqn. (31) is obtained by substituting $B = e\sin\theta_0, C = R/ae$.

$$\left(\Delta\theta\right)^2\left(B^2 + k_0^2 + k_0 B\Delta\theta\right) = \left(k_0 C\right)^2\left(k_0 + B\Delta\theta\right)^2 \tag{31}$$

8. Eqn. (32) is obtained from eqn. (31) by truncating the higher order terms of $\Delta\theta$.

$$\left(\Delta\theta\right)^2\left(B^2 + k_0^2\right) = \left(k_0 C\right)^2\left(k_0 + B\Delta\theta\right)^2 \tag{32}$$

9. Eqn. (33) is obtained by simplifying the above equation and solving it:

$$\Delta\theta = k_0^2 C\left(-k_0 BC \pm \sqrt{B^2 + k_0^2}\right)^{-1} \tag{33}$$

10. Eqn. (34) is obtained by re-substituting for $B$ and $C$ in eqn. (33):

$$\Delta\theta = \pm\frac{k_0^2\left(R/ae\right)}{\sqrt{\left(e\sin\theta_0\right)^2 + k_0^2}}\left(1 \mp \frac{\left(e\sin\theta_0\right)k_0\left(R/ae\right)}{\sqrt{\left(e\sin\theta_0\right)^2 + k_0^2}}\right)^{-1} \tag{34}$$

11. It can be proven that $\left|\left(\sin\theta_0\right)k_0 \middle/ \sqrt{\left(e\sin\theta_0\right)^2 + k_0^2}\right| < 1$. Thus, if $\left(R/a\right) < 1$, infinite geometric series expansion can be applied to get a converging series for $\Delta\theta$, as shown in eqn. (35):

$$\Delta\theta = \pm\frac{k_0^2\left(R/ae\right)}{\sqrt{\left(e\sin\theta_0\right)^2 + k_0^2}}\left\{\sum_{n=0}^{\infty}\left(\pm\frac{\left(e\sin\theta_0\right)k_0\left(R/ae\right)}{\sqrt{\left(e\sin\theta_0\right)^2 + k_0^2}}\right)^n\right\}$$
$$= \pm D\left(1 \pm dD + \left(dD\right)^2 \pm \left(dD\right)^3 \pm \cdots\right) \tag{35}$$

where $D$ and $d$ are given by eqns. (3-23) and (3-24) respectively. For convenience, the negative and positive solutions are referred to as $\Delta\theta_1$ and $\Delta\theta_2$ respectively, which is in accordance with Figure 3.2-5(a). Accordingly, the angles $\Delta\theta_1$ and $\Delta\theta_2$ can be written as in eqns. (36) and (37):

$$\Delta\theta_1 = D\left(dD-1\right)\sum_{n=0}^{\infty}\left(dD\right)^{2n} \qquad (36)$$

$$\Delta\theta_2 = D\left(dD+1\right)\sum_{n=0}^{\infty}\left(dD\right)^{2n} \qquad (37)$$

Two special test cases are considered to verify the validity of eqns. (36) and (37).

**Case 1: Circle $e=0$:** For this case, it is known that the focus becomes the center of the circle and the focal parameter $a \to \infty$, the structure is rotationally symmetric, and the radius of the circle $\rho = ae$. Additionally, $\Delta\theta_1 = -\Delta\theta_2, \forall\,\theta_0$. Using (35), $\Delta\theta \approx \pm\left(R/\rho\right)$. Hence, this case is verified.

**Case 2: Symmetry along the $x-$axis, $\theta_0 = 0$ and $\theta_0 = \pi$:** Since the considered conic equation (3-18) is symmetric along the $x-$axis, $\Delta\theta_1 = -\Delta\theta_2, \theta_0 = \left\{0, \pi\right\}$. Using (36), $\Delta\theta\big|_{\theta_0=0} \approx \pm\left(1-e\right)\left(R/ae\right)$, $\Delta\theta\big|_{\theta_0=\pi} \approx \pm\left(1+e\right)\left(R/ae\right)$. Hence, this case is also verified.

## C. Computation of the slope of the tangent

Here, the derivation in Appendix B is used to derive the relationship between the estimated slope $\tilde{m}$ and the analytical slope of the tangent $m_0$ given by eqn. (3-21). Continuing from eqn. (3-16):

$$\tilde{m} = \frac{r_2 \sin \theta_2 - r_1 \sin \theta_1}{r_2 \cos \theta_2 - r_1 \cos \theta_1} = \frac{\sin \theta_2 - \sin \theta_1 - e \sin \left( \theta_2 - \theta_1 \right)}{\cos \theta_2 - \cos \theta_1}$$
$$= \frac{\left( e - \cos \theta_0 \right) \alpha + 2 \sin \theta_0 \beta - 2 e \gamma}{\alpha \sin \theta_0 + 2 \beta \cos \theta_0} \tag{38}$$

$$\alpha = \left( \sin \Delta \theta_1 - \sin \Delta \theta_2 \right) \tag{39}$$

$$\beta = \left( \sin^2 \left( \Delta \theta_1 / 2 \right) - \sin^2 \left( \Delta \theta_2 / 2 \right) \right) \tag{40}$$

$$\gamma = \left( \sin \Delta \theta_1 \sin^2 \left( \Delta \theta_2 / 2 \right) - \sin \Delta \theta_2 \sin^2 \left( \Delta \theta_1 / 2 \right) \right) \tag{41}$$

Now imposing the condition that $D_{max} \ll 1$ (see Appendix C), such that $\Delta \theta_i, i = 1$ to $2$ are very small, $\Delta \theta_i \to 0$, eqn. (38) can be simplified using eqns. (3-21) and (3-22) as follows:

$$\lim_{\Delta \theta_i \to 0} \tilde{m} = \frac{\left( e - \cos \theta_0 \right) + 0.5 \sin \theta_0 \left( \Delta \theta_1 + \Delta \theta_2 \right) + 0.5 e \Delta \theta_1 \Delta \theta_2}{\sin \theta_0 + 0.5 \cos \theta_0 \left( \Delta \theta_1 + \Delta \theta_2 \right)}$$
$$= \left( 1 + \cot \theta_0 D \left( \sum_{n=1}^{\infty} \left( dD \right)^n \right) \right)^{-1} \tag{42}$$
$$\left\{ m_0 + D \left( \sum_{n=1}^{\infty} \left( dD \right)^n \right) + 0.5 e \csc \theta_0 D^2 \left( \left( dD \right)^2 - 1 \right) \left( \sum_{n'=0}^{\infty} \sum_{n=0}^{\infty} \left( dD \right)^{n+n'} \right) \right\}$$

Here, $\lim_{t \to 0} \sin t = t$ has been used. It can be shown that if $D_{max} \ll 1$, then $\left| \cot \theta_0 dD^2 \right| \ll 1$. Thus, eqn. (43) is obtained by applying infinite geometric series expansion [112] and retaining terms up to $O\left( D^3 \right)$.

$$\tilde{m} \approx m_0 - 0.5 e \, d \, D^3 \csc \theta_0 \tag{43}$$

Thus, $\tilde{m}$ converges to $m_0$, subject to the condition that $D_{\max} \ll 1$. In the above expression, additional attention should be paid to two special cases: $\theta_0 \in \{0, \pi\}$, where $\csc \theta_0$ is singular. However, noting that $d \csc \theta_0$ is not singular, there is no extra singularity other than the singularity of the actual slope $m_0$. The angular error in the computation of the slope is given by eqn. (44).

$$
\begin{aligned}
\partial \phi &= \left| \tan^{-1}(m_0) - \tan^{-1}(\tilde{m}) \right| = \left| \tan^{-1}\left( \frac{m_0 - \tilde{m}}{1 + m_0 \tilde{m}} \right) \right| \\
&\approx \tan^{-1} \left| \frac{0.5 e\, d\, D^3 \csc \theta_0}{1 + m_0^2} \right| \approx \left| \frac{0.5 e\, d\, D^3 \csc \theta_0}{1 + m_0^2} \right|
\end{aligned}
\tag{44}
$$

Specifically, for circle, i.e. $e = 0$, thus $\partial \phi = 0$. Further the error in the computation of the tangent is bounded by $\left| \dfrac{0.5 e\, d\, D^3 \csc \theta_0}{1 + m_0^2} \right|$ and can be considered of order $O(D^3)$.

In the above analysis (and in Appendix A), a conic with focus at the origin, and the directrix at $x = -a$ was considered. Since an arbitrarily placed conic can be represented using the general equation for conic (3-18) by applying suitable rotation and translation, the analysis is applicable to all the possible conics.

### D. Maximum value of *D*

Here the maximum value of $D$ is derived. Let $A = \left(1 - e\cos\theta_0\right)^2 \left(\dfrac{R}{ae}\right)$ and $B = \sqrt{\left(e\sin\theta_0\right)^2 + \left(1 - e\cos\theta_0\right)^2}$. Thus using eqn. (3-23), $D = A/B$. For finding the maximum value of $D$, first impose $\partial D/\partial\theta_0 = 0$ is imposed. The expression of $\partial D/\partial\theta_0$ is computed as in eqn. (45).

$$\frac{\partial D}{\partial\theta} = \frac{1}{B}\left(\frac{\partial A}{\partial\theta} - \frac{A}{B}\frac{\partial B}{\partial\theta}\right) \tag{45}$$

where $\dfrac{\partial A}{\partial\theta} = 2\left(1 - e\cos\theta_0\right)\left(e\sin\theta_0\right)\left(\dfrac{R}{ae}\right) = 2A\dfrac{\left(e\sin\theta_0\right)}{\left(1 - e\cos\theta_0\right)}$ and $\dfrac{\partial B}{\partial\theta} = \left(\dfrac{1}{B}\right)\left(e\sin\theta_0\right)$.

Substituting the above in eqn. (45), $\dfrac{\partial D}{\partial\theta} = \dfrac{A\left(e\sin\theta_0\right)\left(2B^2 - \left(1 - e\cos\theta_0\right)\right)}{\left(1 - e\cos\theta_0\right)B^3}$ is obtained.

Thus, $\partial D/\partial\theta_0 = 0$ can be satisfied in three different cases shown in Table D-1.

**Table D-1: The three cases for satisfying $\partial D/\partial\theta_0 = 0$.**

| | |
|---|---|
| Case 1: $A = 0$ and $\left(1 - e\cos\theta_0\right)B^3 \neq 0$. | This is not possible, since for $A = 0$, the condition $\left(1 - e\cos\theta_0\right) = 0$ should be satisfied, which violates the condition $\left(1 - e\cos\theta_0\right)B^3 \neq 0$. |
| Case 2: $\sin\theta_0 = 0$ and $\left(1 - e\cos\theta_0\right)B^3 \neq 0$ | This is possible for $\theta_0 = 0$ or $\pi$. <br> For $\theta_0 = 0$, we get $D = \left(1 - e\right)R/ae$ <br> For $\theta_0 = \pi$: we get $D = \left(1 + e\right)R/ae$ |
| Case 3: $\left(2B^2 - \left(1 - e\cos\theta_0\right)\right) = 0$ and $\left(1 - e\cos\theta_0\right)B^3 \neq 0$ | Investigated below. |

For investigating case 3, considering the condition $\left(2B^2 - \left(1 - e\cos\theta_0\right)\right) = 0$ and substituting the expression of $B$ gives $\theta_0 = \pm\cos^{-1}\left(\left(2e^2 + 1\right)/3e\right)$. Using this expression in $D$ in eqn. (3-23), $D = 2\left(1 - e^2\right)\left(R/ae\right)\left(5 - e^2 - 4e^4\right)^{-1/2}$ is obtained. Evidently, in case 2 as well as case 3, $\theta_0 = \pi$ is the point of maxima for $D$.

# E. Ground truth generation for real images in section 6.1.

**Application for collecting the ground truth**

A Matlab application was developed for generating the ground truth for the problem of ellipse detection in real images. Some snapshots of the application are presented in Figure E-1 to Figure E-3. The volunteer can open the application after providing a user name. For example, in Figure E-1, the user name 'Sachan' appears above the white colored text box.

First, the volunteer loads an image and then clicks the button 'Draw contour', as shown in Figure E-1. A cross-hair cursor then appears on the image area and the volunteer can choose some points on the ellipse as he or she thinks to be existing in the image. The volunteer can close the curve by right clicking. At least six points have to be selected or the application generates an error and asks the volunteer to draw the contour again. After closing the curve (see Figure E-2 for example), the volunteer can press the button 'Verify contour' if he or she is satisfied by his or her input or the button 'Clear contour' if he or she is dissatisfied by it. On pressing the verify button, the application applies NSAF proposed in section 4.2.2 and shows the fitted ellipse in red contour on the image (as shown in Figure E-3). If the volunteer is satisfied by the ellipse, he or she presses the button 'Accept contour'. Otherwise the volunteer presses the button 'Clear contour'.

The application saves the parameters of the ellipse in an array when the 'Accept contour' button is clicked. Finally the data of all the ellipses is saved in a file that contains the user name and the image name when the volunteer clicks the button 'Save XML'.

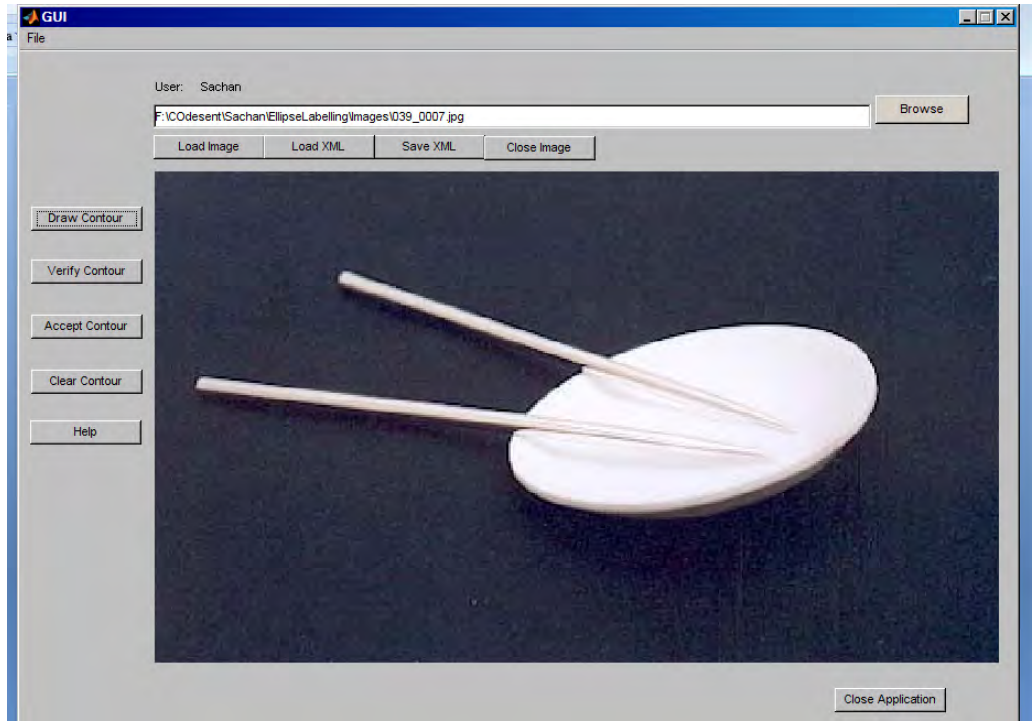

**Figure E-1: The ground truth generation tool with image loaded.**

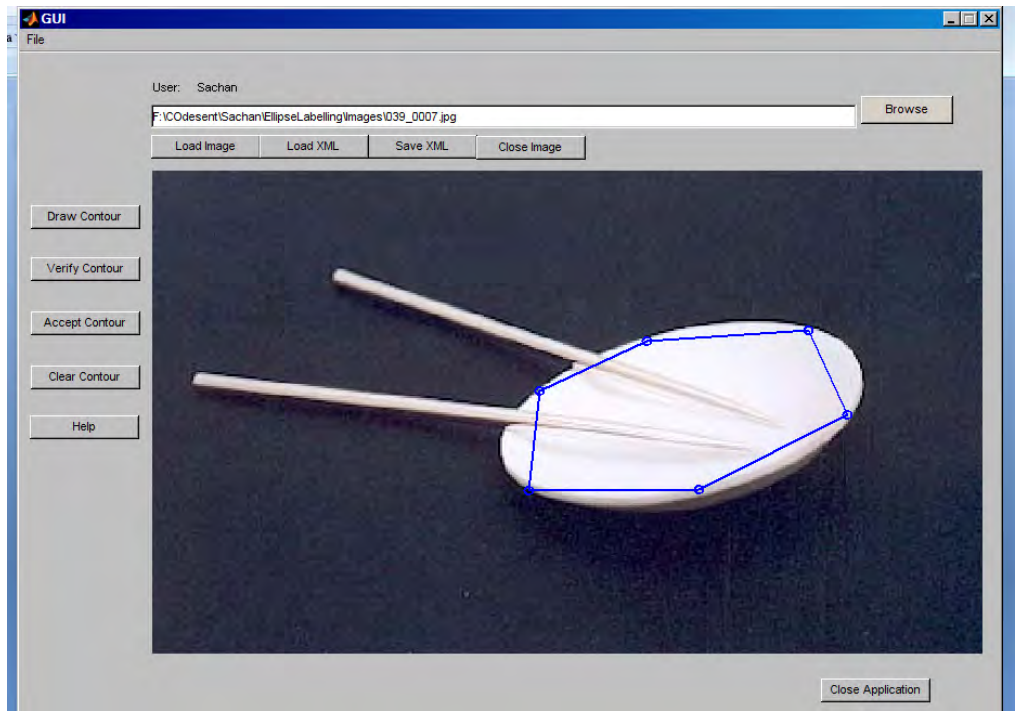

**Figure E-2: The ground truth generation tool with contour drawn.**

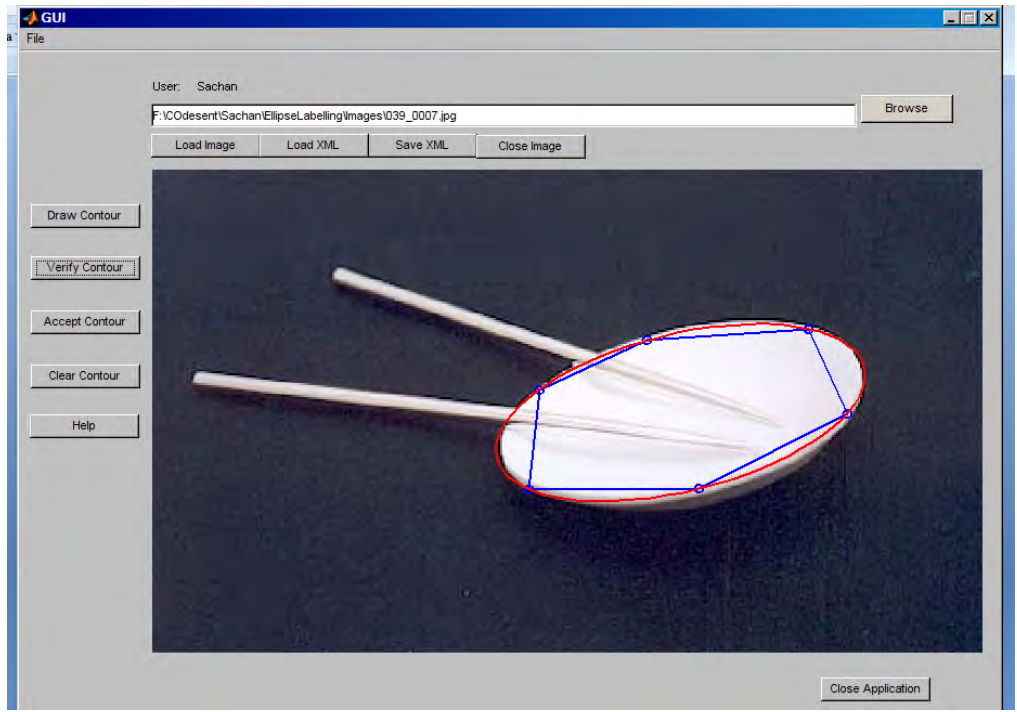

**Figure E-3: The ground truth generation tool with a fitted ellipse.**

**Post processing and collating data from all the volunteers**

The ground truth is been generated for each image by 20 volunteers. For each image, the elliptic clusters voted by more than 10 volunteers are considered as the global ground truth. An elliptic cluster is a cluster in which all the ellipses have an overlap of more than 0.9 with each other. The hypothetic ellipse that is computed using the average parameters of the ellipses within a cluster is used to represent the cluster. Other clusters in which ellipses were contributed by less than 10 volunteers are rejected.

# F. Concise summary of the various aspects of object detection problem [191].

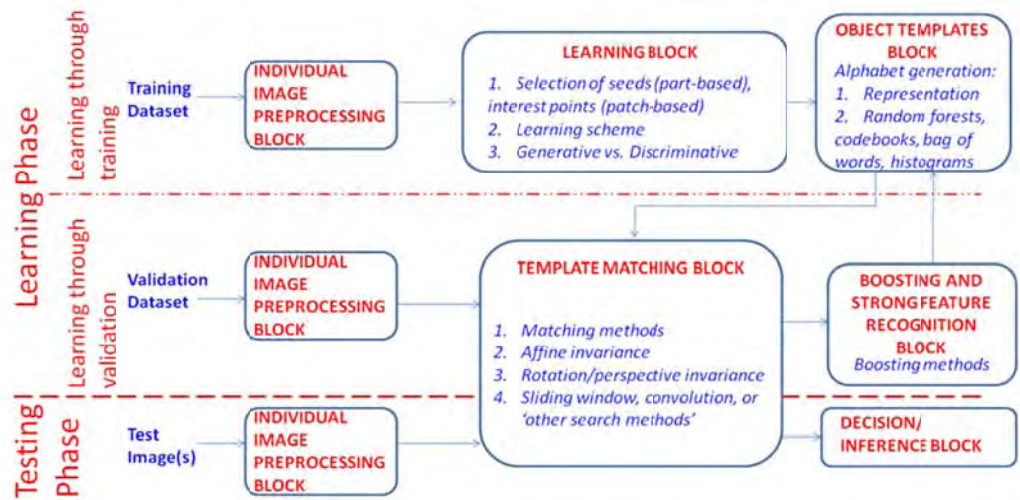

**Figure F-1: Basic block diagram of a typical object detection/recognition system.**

A lot of research is being done in the area of object recognition and detection. In order to facilitate the discussion about the methods and ideas of various bodies of research works, a general block diagram applicable to any object detection and recognition method is presented in Figure F-1. Specific methods proposed by various researchers may vary slightly from this generic block diagram.

Any such algorithm can be divided into two different phases, viz. learning phase and testing phase. In the learning phases, the machine uses a set of images which contains objects belonging to specific pre-determined class(es) in order to learn to identify the objects belonging to those classes. Once the algorithm has been trained for identifying the objects belonging to the specified classes, in the testing phase, the algorithm uses its knowledge to identify the specified class objects from the test image(s).

The researchers have worked upon many specific aspects of the above mentioned system. Some examples include the choice of feature type (edge based or patch based features), the method of generating the features, the method of learning the consistent features of an object class, the specificity of the learning scheme (does it concentrate on inter-class variability or intra-class variability), the representation of the templates, the schemes to find a match between a test or validation image and an object template (even though the size and orientation of an object in the test image may be different from the learnt template), and so on. The following discussion considers one aspect at a time and provides details upon the work done in that aspect.

**Feature types**

There are various feature types used in object detection problems. Examples include edge based features, patch based features, contour based features (similar to PA of the shape of the object or its portion), texture features, geometric features, etc.

***Edge and contour based features***

The methods that use edge-based feature type extract the edge map of the image and identify the features of the object in terms of edges. Some examples include [2, 3, 181, 192-206]. Using edges as features is advantageous over other features due to various reasons. As discussed in [3], they are largely invariant to illumination conditions and variations in objects' colors and textures. They also represent the object boundaries well and represent the data efficiently in the large spatial extent of the images.

In this category, there are two main variations: use of the complete contour (shape or its polygonal approximation) of the object as the feature [192, 194-198, 200, 202] and use of collection of edge contour fragments as the feature of the object [2, 3, 181, 193, 199-205]. Figure F-2 shows an example of complete contour and collection of contours for an image.

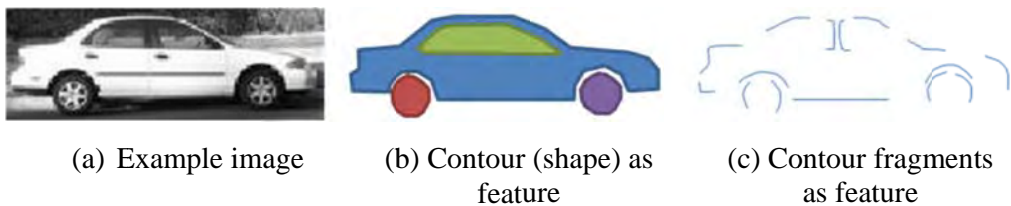

(a) Example image     (b) Contour (shape) as feature     (c) Contour fragments as feature

**Figure F-2: Edge-based feature types for an example image**

In real images, typically incomplete contours are inevitable due to occlusion and noise. Various researchers have tried to solve this problem to some extent [192, 197, 198, 200, 202, 205]. Hamsici [192] identified a set of landmark points (similar to dominant points) from the edges and connected them to obtain a complete shape contour.

Schindler [197] used segmenting approaches [207, 208] to obtain closed contours from the very beginning (he called the areas enclosed by such closed contours as super pixels). Ferrari [202, 205] used a sophisticated edge detection method that provides better edges than contemporary methods for object detection. These edges were then connected across the small gaps between them to form a network of closed contours. Ren [200] used a triangulation to complete the contours of the objects in natural images, which are significantly difficult due to the presence of background clutter. Hidden state shape model was used by Wang [209] in order to detect the contours of articulate and flexible or polymorphic objects.

It is noticeable that all of these methods require additional computation intensive processing and are typically sensitive to the choice of various empirical contour parameters. The other problem involving such feature is that in the test and validation images, the available contours are also incomplete and therefore the degree of match with the complete contour is typically low [197]. Though some measures, like kernel based [192, 210] and histogram based methods [194, 195], can be taken to alleviate this problem, the detection of the severely occluded objects is still very difficult and unguaranteed. Further, such features are less capable of incorporating the pose or viewpoint changes, large intra-class variability, articulate objects (like horses) and flexible or polymorphic objects (like cars) [197, 202, 205]. This can be explained as follows. Since this feature type deals with complete contours, even though the actual impact of these situations is only on some portions of the contour, the complete contour has to be trained.

On the other hand, the contour fragment features are substantially robust to occlusion if the learnt features are good in characterizing the object [2, 3, 194, 201, 202, 205, 211]. They are less demanding in computation as well as memory as the contour completion methods need not be applied and relatively less data needs to be stored for the features. The matching is also expected to be less sensitive to occlusion [3, 212]. Further, special cases like viewpoint changes, large intra-class variability, articulate objects and flexible or polymorphic objects can be handled efficiently by training the fragments (instead of the complete contour) [3, 193, 194, 202, 205, 212]. However, the performance of the methods based on contour fragment features significantly depends upon the learning techniques. While using these features, it is important to derive good feature templates that represent the object categories well (in terms of both inter-class and intra-class variations) [2, 213]. Learning methods like boosting [18, 166, 211, 213-232] become very important for such feature types.

The selection of the contour fragments for characterizing the objects is an important factor and can affect the performance of the object detection and recognition method. While all the contour fragments in an image cannot be chosen for this purpose, it has to be ensured that the most representative edge fragments are indeed present and sufficient local variation is considered for each representative fragment. In order to look for such fragments, Opelt [2] used large number of random seeds that are used to find the candidate fragments and finally derives only two most representative fragments as features. Shotton [3] on the other hand generated up to 100 randomly sized rectangular units in the bounding box of the object to look for the candidate fragments. It is worth noting that the method proposed in [2] becomes computationally very expensive if more than two edge fragments are used as features for an object category. While the method proposed by Shotton [3] is computationally efficient and expected to be more reliable as it used numerous small fragments (as compared to two most representative fragments), it is still limited by the randomness of choosing the rectangular units.

### Geometric shape features

Chia [181] used some geometrical shape support (ellipses and quadrangles) in addition to the fragment features for obtaining more reliable features. Use of geometrical structure, relationship between arcs and lines, and study of structural properties like symmetry, similarity and continuity for object retrieval were proposed in [233]. Though the use of geometrical shape (or structure) for estimating the structure of the object is a good idea, there are two major problems with the methods in [181, 233]. The first problem is that some object categories may not have strong geometrical (elliptic and quadrangle) structure (example horses) and the use of weak geometrical structure may not lead to robust descriptors of such objects. Though [181] demonstrates the applicability for animals, the geometrical structure derived for animals is very generic and applicable to many classes. Thus, the inter-class variance is poor. The classes considered in [181], viz., cars, bikes and four-legged animals (four-legged animals is considered a single class) are very different from each other. Similarly, [233] concentrates on logos and the images considered in [233] have white background, with no natural background clutter and noise. Its performance may degrade significantly in the presence of noise and natural clutter. The second problem is that sometimes occlusion or flexibility of the object may result in complete absence of the components of geometrical structure. For example, if the structural features learnt in [233] are occluded, the probability of detecting the object is very low. Similarly, if the line features learnt in [181], used for forming the quadrangle are absent, the detection capability may reduce significantly.

### *Patch-based features*

The other prevalent feature type is the patch based feature type, which uses appearance as cues. This feature has been in use since more than two decades [234], and edge-based features are relatively new in comparison to it. Moravec [234] looked for local maxima of minimum intensity gradients, which he called corners and selected a patch around these corners. His work was improved by Harris [235], which made the new detector less sensitive to noise, edges, and anisotropic nature of the corners proposed in [234].

In this feature type, there are two main variations:

1) patches of rectangular shapes that contain the characteristic boundaries describing the features of the objects [2, 236-241]. Usually, these features are referred to as the local features.

2) irregular patches in which, each patch is homogeneous in terms of intensity or texture and the change in these features are characterized by the boundary of the patches. These features are commonly called the region-based features.

Figure F-3 shows these features for an example image. Subfigures (b)-(d) show local features while subfigure (e) shows region based features (intensity is used here for extracting the region features). As shown in Figure F-3(b)-(d), the local features may be of various kinds [242, 243].

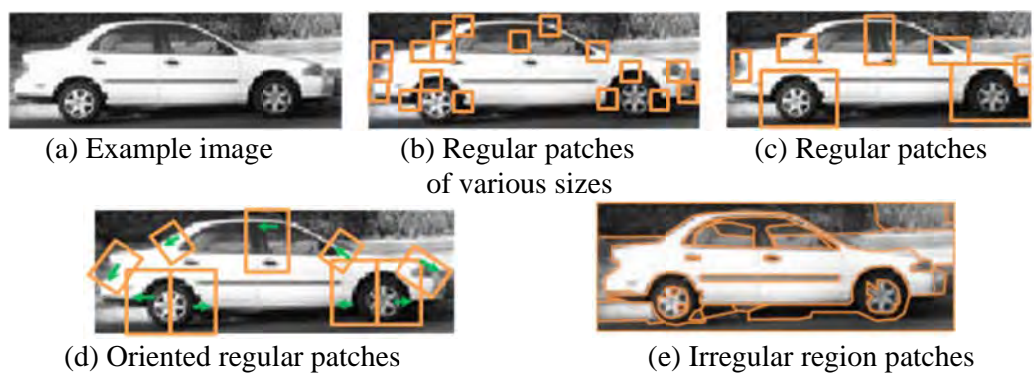

(a) Example image    (b) Regular patches    (c) Regular patches
                         of various sizes

(d) Oriented regular patches        (e) Irregular region patches

**Figure F-3: Patch-based feature types for an example image.**

**Feature types shown in (b)-(d) are called local features, while the feature type shown in (e) is called region-based features.**

A pioneering work was done by Lowe [239], which enabled the use of appropriately oriented variable sized features for describing the object. He proposed a scale invariant feature transformation (SIFT) method. Lowe describes his method of feature extraction in three stages. He first identified potential corners (key points) using difference of Gaussian function, such that these feature points were invariant to scale and rotation. Next, he identified and selected the corners that are most stable and determined their scale (size of rectangular feature). Finally, he computed the local image gradients at the feature points and used them to assign orientations to the patches. The use of oriented features also enhanced the features' robustness to small rotations. With the use of orientation and scale, the features were transformed (rotated along the suitable orientation and scaled to a fixed size) in order to achieve scale and rotational invariance. In order to incorporate the robustness to illumination and pose or perspective changes, the features were additionally described using the Gaussian weighing function along various orientations.

The identification of good corner points (or key-points) is investigated in [239, 244-247]. Lowe [239] studied the stability of the feature points. However, his proposal would apply to his schema of features only. Carneiro [246] and Comer [248] proposed stability measures that could be applied to wide range and varieties of algorithms.

Another major concern is to describe these local features. Though the features can be directly described and stored by saving the pixel data of the local features, such method is naive and inefficient. Researchers have used many efficient methods for describing these local features. These include PCA vectors of the local feature (like PCA-SIFT) [206, 249], Fischer components [250, 251], wavelets and Gabor filters [199], Eigen spaces [252], kernels [192, 206, 210, 253, 254], etc. It is important to note that though these methods use different tools for describing the features, the main mathematical concept behind all of them is the same. The concept is to choose sufficient (and yet not many) linearly independent vectors to represent the data in a compressed and efficient manner [199]. Another advantage of using such methods is that each linearly independent vector describes a certain property of the local feature (depending on the mathematical tool used). For example, a Gabor wavelet effectively describes an oriented stroke in the image region [199]. Yet another advantage of such features is that while matching the features in the test images, properties of linear algebra (like linear dependence, orthogonality, null spaces, rank, etc.) can be used to design efficient matching techniques [199].

The region-based features are inspired by segmentation approaches and are mostly used in algorithms whose goal is to combine localization, segmentation, and/or categorization. While intensity is the most commonly used cue for generating region based features [166, 244, 255], texture [193, 255-258], color [257-259], and minimum energy or entropy [185, 260] have also been used for generating these features. It is notable that conceptually these are similar to the complete contours discussed in edge-based features. Such features are very sensitive to lighting conditions and are generally difficult from the perspective of scale and rotation invariance. However, when edge and region based features are combined efficiently, in order to represent the outer boundary and inner common features of the objects respectively, they can serve as powerful tools [193]. Some good reviews of feature types can also be found in [236, 261, 262].

It has been argued correctly by many researchers that a robust object detection and characterization scheme shall typically require more than one feature types to obtain good performance over large number of classes [2, 193, 202, 203, 205, 229, 263-269].

**Generative model vs. discriminative model**

The relationship (mapping) between the images and the object classes is typically non-linear and non-analytic (no definite mathematical model applicable for all the images and all the object classes is available). Thus, typically this relationship is modeled using probabilistic models [270]. The images are considered as the observable variables, the object classes are considered as the state variables, and the features are considered as intermediate (sometimes hidden) variables. Such modeling has various advantages. First, it provides a generic framework which is useful for both the problems of object detection and recognition (and many other problems in machine vision and outside it). Second, such framework can be useful in evaluating the nature and extent of information available while training, which subsequently helps us to design suitable training strategies.

The probabilistic models for our problems can be generally classified into two categories, viz. discriminative models and generative models [271, 272]. It shall be helpful to develop a basic mathematical framework for understanding and comparing the two models. Let the observable variables (images) be denoted by $\mathbf{x}_i, i = 1$ to $N$, where $N$ is the number of training images. Let the corresponding state variables (class labels) be denoted as $c_i$ and the intermediate variables (features or feature descriptors) be denoted as $\boldsymbol{\theta}_i$. Accordingly, a simplistic graphical representation [272] of the discriminative and generative models is presented in Figure F-4.

(a) Discriminative model      (b) Generative model

**Figure F-4: Graphical illustration of the discriminative and generative models.**

As seen in Figure F-4, the discriminative model uses a map from the images to the class labels, and thus the flow of information is from the observables (images) to the state variables (class labels) [272]. Considering the joint probability $P(c, \boldsymbol{\theta}, \mathbf{x})$, discriminative models expand $P(c, \boldsymbol{\theta}, \mathbf{x})$ as $P(c, \boldsymbol{\theta}, \mathbf{x}) = P\big(c \mid (\boldsymbol{\theta}, \mathbf{x})\big) P\big(\boldsymbol{\theta} \mid \mathbf{x}\big) P\big(\mathbf{x}\big)$. Thus, $P\big(c \mid (\boldsymbol{\theta}, \mathbf{x})\big)$ is the model defining probability [271] and the training goal is:

$$P\big(c \,|\, (\theta, \mathbf{x})\big) = \begin{cases} \alpha & \text{if } \mathbf{x} \text{ contains object of class } c \\ \beta & \text{otherwise} \end{cases} \qquad (46)$$

Ideally, $\alpha = 1$ and $\beta = 0$. Indeed, practically this is almost impossible to achieve, and values between [0,1] are chosen for $\alpha$ and $\beta$.

In contrast, the generative model uses a map from the class labels to the images, and thus the flow of information is from the state variables (class labels) to the observables (images) [272]. Generative models use the expansion of the joint probability $P(c, \boldsymbol{\theta}, \mathbf{x}) = P\big(\mathbf{x} \,|\, (\boldsymbol{\theta}, c)\big) P\big(\boldsymbol{\theta} \,|\, c\big) P\big(c\big)$ $P(c, \boldsymbol{\theta}, \mathbf{x}) = P\big(\mathbf{x} | (\boldsymbol{\theta}, c)\big) P(\boldsymbol{\theta}|c) P(c)$. Thus, $P\big(\mathbf{x}|(\boldsymbol{\theta}, c)\big)$ and $P(c)$ are the model defining probabilities [271] and the training goal is:

$$P\big(\mathbf{x} \,|\, (\boldsymbol{\theta}, c)\big) P\big(c\big) = \begin{cases} \alpha & \text{if } \mathbf{x} \text{ contains object of class } c \\ \beta & \text{otherwise} \end{cases} \qquad (47)$$

Ideally, $\alpha = 1$ and $\beta = 0$. Indeed, practically this is almost impossible to achieve, and values between [0,1] are chosen for $\alpha$ and $\beta$. It is important to note that in unsupervised methods, the prior probability of classes, $P(c)$ is also unknown.

Further mathematical details can be found in [271, 272]. The other popular model is the descriptive model, in which every node is observable and is interconnected to every other node. It is obvious that the applicability of this model to the considered problem is limited. Therefore, we do not discuss this model any further. It shall suffice to make a note that such models are sometimes used in the form of conditional random fields or forests [166, 198, 256].

**Training data size and supervision**

Mathematically, the training data size required for generative model is very large (at least more than the maximum dimension of the observation vector $\mathbf{x}$). On the other hand, discriminative models perform well even if the training dataset is very small (more than a few images for each class type). This is expected because the discriminative models invariably use supervised training dataset (the class label is specifically mentioned for each image). On the other hand, generative models are unsupervised (semi-supervised, at best) [189]. Not only the posterior probability $P\big(\mathbf{x}|(\boldsymbol{\theta}, c)\big)$ is unknown, the prior probability of the classes $P(c)$ is also unknown for the generative models [271]. Another point in this regard is that since generative models do not require supervision and the training dataset can be appended incrementally [167, 203, 271] as vision system encounters more and more scenarios, generative models are an important tool for expanding the knowledge base, learning new classes, and keeping the overall system scalable in its capabilities.

**Learning methods**

Generative models use methods like Bayesian classifiers and Bayesian networks [167, 203, 211, 240], likelihood maximization [167, 273], and expectation maximization [189, 245, 259, 273]. Discriminative models typically use methods like logistic regression, support vector machines [202, 205, 259, 274-278], and k-nearest neighbors [259, 279, 280]. The k-nearest neighbors scheme can also be used for multi-class problems directly, as demonstrated in [279]. Boosting schemes are also examples of methods for learning discriminative models [2], though they are typically applied on already learnt weak features (they shall be discussed later in greater detail). In the schemes where generative and discriminative models are combined [259, 281], there are two main variations: generative models with discriminative learning [245, 266, 271], and discriminative models with generative learning [272]. In the former, typically maximum likelihood or Bayesian approaches are combined with boosting schemes or incremental learning schemes [167, 229, 245, 266, 271], while in the latter, usual discriminative schemes are augmented by 'generate and test' schemes in the feedback loop [272, 282].

**Object templates and their representation**

The learning method has to learn a mapping between the features and the classes. Typically, the features are extracted first, which is followed by either the formation of class models (in generative models) or the most discriminative features for each class (in discriminative models) or random fields of features in which a cluster represents an object class (descriptive models, histogram based schemes, Hough transform based methods, etc.). Based on them, the object templates suitable for each class are learnt and stored for the future use (testing). This section will discuss various forms of object templates used by researchers in computer vision.

While deciding on an object template, following factors have to be considered:

- Is the template the most representative form of the class (in terms of the aimed specificity, flexibility of the object, intra-class variation, etc.)? For example, does it give the required intra-class and inter-class variability features? Does it need to consider some common features among various classes or instances of hierarchical class structure? Does it need to consider various poses and/or perspectives? Does it need to prioritize certain features (or kind of features)?

- Is the model representation an efficient way of storing and using the template? Here, memory and computations are not the only important factors. We need to also consider if the representation enables good decision mechanisms.

The above factors will be the central theme in discussing the specific merits and demerits of the various existing object templates. We begin with the object templates that use the spatial location of the features. Such templates specifically represent the relative position of the features (edge fragments, patches, regions) in the image space. For this, researchers typically represent each feature using a single representative point (called the centroid) and specify a small region in which the location of the centroid may vary in various objects belonging to the same class [2, 3]. All the centroids are then collected together using a graph topology. For example some researchers have used a cyclic or chain topology [197]. This simplistic topology is good to represent only the external continuous boundary of the object. Due to this, it is also used for complete contour representation, where the contour is defined using particular pivot points which are joined to form the contour [197]. Such a topology may fail if the object is occluded at one of the centroid locations, as the link between the chains is not found in such case and the remaining centroids are also not detected as a consequence. Further, if some of the characteristic features are inside the object boundary, deciding the most appropriate connecting link between the centroids of the external and internal boundaries may be an issue and may impact the performance of the overall algorithm. Other topology in use is the constellation topology [167, 283, 284], in which a connected graph is used to link all the centroids. A similar representation is being called multi-parts-tree model in [185], though the essentials are same. However, such topology requires extra computation in order to find an optimal (neither very deep nor very wide) representation. Again, if the centroids that are linked to more than one centroid are occluded, the performance degrades (though not as strongly as the chain topology). The most efficient method in this category is the star topology, in which a central (root) node is connected to all the centroids [2, 3, 194, 241]. The root node does not correspond to any feature or centroid and is just a virtual node (representing the virtual centroid of the complete object). Thus, this topology is able to deal with occlusion better than the other two topologies and does not need any extra computation for making the topology.

Other methods in which the features are described using transformation methods (like the kernel based methods, PCA, wavelets, etc., discussed in section 0), the independent features can be used to form the object templates. The object templates could be binary vectors that specify if a particular feature is present in an object or not. Such object templates are called bag-of-words, bag of visual words, or bag of features [2, 260, 274, 275, 278, 285-287]. All the possible features are analogous to visual words, and specific combinations of words (in no particular order) together represent the object classes. Such bag of words can also be used for features like colors, textures, intensity, shapes [260], physical features (like eyes, lips, nose for faces, and wheels, headlights, mirrors for cars) etc. [259, 286, 288]. As evident, such bag of words is a simple yet powerful technique for object recognition and detection but may perform poorly for object localization and segmentation. As opposed to them, spatial object templates are more powerful for image localization and segmentation.

In either of the above cases, the object templates can also be in the form of codebooks [2, 3, 202, 205, 240, 241, 287, 289]. A codebook contains a specific code of features for each object class. The code contains the various features that are present in the corresponding class, where the sequence of features may follow a specific order or not. An unordered codebook is in essence similar to the concept of bag of words, where the bag of words may have greater advantage in storing and recalling the features and the object templates. However, codebooks become more powerful if the features in the code are ordered. A code in the order of appearance of spatial templates can help in segmentation [3], while a code in the order of reliability or strength of a feature for a class shall make the object detection and recognition more robust.

Other hierarchical (tree like) object templates may be used to combine the strengths of both the codebooks and bag of words, and to efficiently combine various feature types [189, 203, 240, 245, 250, 255, 258, 266, 273, 284, 287, 290].

Another important method of representing the object templates is based on random forests or fields [256, 282, 291]. In such methods, no explicit object template is defined. Instead, in the feature space (where each feature represents one dimension), clusters of images belonging to same object class are identified [239, 240, 291]. These clusters in the feature space are used as the probabilistic object templates [250]. For every test image, its location in feature space and distance from these clusters determine the decision.

**Matching schemes and decision making**

Once the object templates have been formed, the method should be capable of making decisions (like detecting or recognizing objects in images) for input images (validation and/or test images). We first discuss about the methods of finding a match between the object template and the input image and then discuss about the methods of making the final decision.

Discussion regarding matching schemes is important because of various reasons. While the training dataset can be chosen to meet certain requirements, it cannot be expected that the test images also adhere to those requirements. For example, we may choose that all the training images are of a particular size, illumination condition, contain only single object of interest viewed from a fixed perspective, in uncluttered (white background), etc., such restrictions cannot be imposed on the real test images, which may be of varying size, may contain many objects of interest and may be severely cluttered and occluded and may be taken from various viewpoints.

The problem of clutter and occlusion is largely a matter of feature selection and learning methods. Still, they may lead to wrong inferences if improper matching techniques are used. However, making the matching scheme scale invariant, rotation and pose invariant (at least to some degree), illumination independent, and capable of inferring multiple instances of multiple classes is important and has gained attention of many researchers [3, 246, 248, 284, 292-322].

If the features in the object templates are pixel based (for example patches or edges), the Euclidean distance based measures like Hausdorff distance [310, 321, 323, 324] and Chamfer distance [2, 3, 185, 202, 297, 306, 325] provide quick and efficient matching tools. However, the original forms of both these distances were scale, rotation, and illumination dependent. Chamfer distance has become more popular in this field because of a lot of incremental improvement in Chamfer distance as a matching technique. These improvements include making it scale invariant, illumination independent, rotation invariant, and more robust to pose variations and occlusions [2, 3, 185, 202, 297, 306, 325]. Further, Chamfer distance has also been adapted for hierarchical codebooks [297]. In region based features, concepts like structure entropy [260, 326], mutual information [260, 290], and shape correlation have been used for matching and inference [293, 294]. Worth attention is the work by Wang [260] that proposed a combination of local and global matching scheme for region features. Such scheme can perform matching and similarity evaluation in an efficient manner (also capable of dealing with deformation or pose changes) by incorporating the spatial mutual information with the local entropy in the matching scheme.

Another method of matching or inferring is to use the probabilistic model in order to evaluate the likelihood ratio [193, 240, 245] or expectation in generative models [189, 270]. Otherwise, correlation between the object template and the input image can be computed or probabilistic Hough transform can be used [242, 258, 259, 277]. Each of these measures is linked directly or indirectly with the defining ratio of the generative model (see section 0), $P(\mathbf{x}|(\boldsymbol{\theta}, c))$, which can be computed for an input image and a given class through the learnt hidden variables $\boldsymbol{\theta}$ [199]. For example, in the case of wavelet form of features, $P(\mathbf{x}|(\boldsymbol{\theta}, c))$ will depend upon the wavelet kernel response to the input image for a particular class [199]. Similarly, the posterior probability can be used for inference in the discriminative models. Or else, in the case of classifiers like SVM, k-nearest neighbors based method, binary classifiers, etc., the features are extracted for the input image and the posterior probability (based on the number of features voted into each class) can be used for inference [202, 205, 250]. If two or more classes have the high posterior probability, multiple objects may be inferred [185, 240]. However, if it is known that only one object is present in an image, refined methods based on feature reliability can be used.

If the object class is represented using the feature spaces, the distance of the image from the clusters in feature space is used for inference. Other methods include histograms corresponding to the features (the number of features that were detected) to decide the object category [239, 250, 286, 291].

# References

[1]     G. Griffin, A. Holub, and P. Perona. *Caltech-256 object category database* [http://authors.library.caltech.edu/7694].                    Available: http://authors.library.caltech.edu/7694

[2]     A. Opelt, A. Pinz, and A. Zisserman, "Learning an alphabet of shape and appearance for multi-class object detection," *International Journal of Computer Vision,* vol. 80, pp. 16-44, 2008.

[3]     J. Shotton, A. Blake, and R. Cipolla, "Multiscale categorical object recognition using contour fragments," *IEEE Transactions on Pattern Analysis and Machine Intelligence,* vol. 30, pp. 1270-1281, 2008.

[4]     A. Y.-S. Chia, S. Rahardja, D. Rajan, and M. K. Leung, "Object recognition by discriminative combinations of line segments and ellipses," in *Proceedings of the IEEE Conference on Computer Vision and Pattern Recognition*, San Francisco, USA, 2010, pp. 2225-2232.

[5]     F. O'Gorman, "Edge detection using Walsh functions," *Artificial Intelligence,* vol. 10, pp. 215-223, 1978.

[6]     G. S. Robinson, "Edge detection by compass gradient masks," *Computer Graphics and Image Processing,* vol. 6, pp. 492-501, 1977.

[7]     M. W. Smith and W. A. Davis, "A New Algorithm for Edge Detection," *Computer Graphics and Image Processing,* vol. 4, pp. 55-62, 1975.

[8]     T. Kasvand, "Iterative edge detection," *Computer Graphics and Image Processing,* vol. 4, pp. 279-286, 1975.

[9]     L. S. Davis, "A survey of edge detection techniques," *Computer Graphics and Image Processing,* vol. 4, pp. 248-270, 1975.

[10]    A. Martelli, "Edge detection using heuristic search methods," *Computer Graphics and Image Processing,* vol. 1, pp. 169-182, 1972.

[11]    W. L. Matson, H. A. McKinstry, G. G. Johnson Jr, E. W. White, and R. E. McMillan, "Computer processing of SEM images by contour analyses," *Pattern Recognition,* vol. 2, 1970.

[12]    U. Ramer, "An iterative procedure for the polygonal approximation of plane curves," *Computer Graphics and Image Processing,* vol. 1, pp. 244-256, 1972.

[13]    D. H. Douglas and T. K. Peucker, "Algorithms for the reduction of the number of points required to represent a digitized line or its caricature," *Cartographica: The International Journal for Geographic Information and Geovisualization,* vol. 10, pp. 112-122, 1973.

[14]    R. O. Duda and P. E. Hart, *Pattern Classification and Scene Analysis*. New York: Wiley Publishers, 1973.

[15]    A. Rosenfeld, "Digital straight line segments," *IEEE Transactions on Computers,* vol. C-23, pp. 1264-1269, 1974.

[16]    R. Deriche, "Using Canny's criteria to derive a recursively implemented optimal edge detector," *International Journal of Computer Vision,* vol. 1, pp. 167-187, 1987.

[17]    J. Canny, "A computational approach to edge-detection," *IEEE Transactions on Pattern Analysis and Machine Intelligence,* vol. 8, pp. 679-698, Nov 1986.

[18]    P. Dollar, Z. Tu, and S. Belongie, "Supervised learning of edges and object boundaries," in *Proceedings of the IEEE Conference on Computer Vision and Pattern Recognition*, 2006, pp. 1964-1971.

[19]    I. M. Anderson and J. C. Bezdek, "Curvature and tangential deflection of discrete arcs: a theory based on the commutator of scatter matrix pairs and its application to vertex detection in planar shape data," *IEEE Transactions on Pattern Analysis and Machine Intelligence,* vol. PAMI-6, pp. 27-40, 1984.

[20]    F. De Vieilleville and J. O. Lachaud, "Comparison and improvement of tangent estimators on digital curves," *Pattern Recognition,* vol. 42, pp. 1693-1707, 2009.

[21]    J. O. Lachaud, A. Vialard, and F. De Vieilleville, "Analysis and comparative evaluation of discrete tangent estimators," Poitiers, 2005, pp. 240-251.

[22]    J. O. Lachaud, A. Vialard, and F. de Vieilleville, "Fast, accurate and convergent tangent estimation on digital contours," *Image and Vision Computing,* vol. 25, pp. 1572-1587, 2007.

[23]    T. Lewiner and M. Craizer, "Projective estimators for point/tangent representations of planar curves," in *21st Brazilian Symposium on Computer Graphics and Image Processing, SIBGRAPI 2008*, Campo Grande, BRAZIL, 2008, pp. 223-229.

[24]    K. Otsuka, T. Horikoshi, and S. Suzuki, "Image velocity estimation from trajectory surface in spatiotemporal space," in *Proceedings of the IEEE Conference on Computer Vision and Pattern Recognition*, 1997, pp. 200-205.

[25]    D. K. Prasad and M. K. H. Leung, "Error analysis of geometric ellipse detection methods due to quantization," in *Fourth Pacific-Rim Symposium on Image and Video Technology (PSIVT 2010)*, Singapore, 2010, pp. 58 - 63.

[26]    D. Coeurjolly and R. Klette, "A comparative evaluation of length estimators of digital curves," *IEEE Transactions on Pattern Analysis and Machine Intelligence,* vol. 26, pp. 252-258, 2004.

[27]    M. Worring and A. W. M. Smeulders, "Digital curvature estimation," *Computer Vision and Image Understanding,* vol. 58, pp. 366-382, 1993.

[28]    H. K. Yuen, J. Illingworth, and J. Kittler, "Detecting partially occluded ellipses using the Hough transform," *Image and Vision Computing,* vol. 7, pp. 31-37, 1989.

[29]    A. Faure, L. Buzer, and F. Feschet, "Tangential cover for thick digital curves," *Pattern Recognition,* vol. 42, pp. 2279-2287, 2009.

[30]    D. M. Tsai and M. F. Chen, "Curve fitting approach for tangent angle and curvature measurements," *Pattern Recognition,* vol. 27, pp. 699-711, 1994.

[31]    D. K. Prasad, R. K. Gupta, and M. K. H. Leung, "An Error Bounded Tangent Estimator for Digitized Elliptic Curves," in *Discrete Geometry for Computer Imagery.* vol. 6607, ed: Springer Berlin / Heidelberg, 2011, pp. 272-283.

[32]    Z. Y. Liu and H. Qiao, "Multiple ellipses detection in noisy environments: A hierarchical approach," *Pattern Recognition,* vol. 42, pp. 2421-2433, 2009.

[33]    A. Y. S. Chia, S. Rahardja, D. Rajan, and M. K. Leung, "A Split and Merge Based Ellipse Detector With Self-Correcting Capability," *IEEE Transactions on Image Processing,* vol. 20, pp. 1991-2006, 2011.

[34]    F. Mai, Y. S. Hung, H. Zhong, and W. F. Sze, "A hierarchical approach for fast and robust ellipse extraction," *Pattern Recognition,* vol. 41, pp. 2512-2524, 2008.

[35]    A. Carmona-Poyato, F. J. Madrid-Cuevas, R. Medina-Carnicer, and R. Muñoz-Salinas, "Polygonal approximation of digital planar curves through break point suppression," *Pattern Recognition,* vol. 43, pp. 14-25, 2010.

[36]    A. Masood and S. A. Haq, "A novel approach to polygonal approximation of digital curves," *Journal of Visual Communication and Image Representation,* vol. 18, pp. 264-274, 2007.

[37]    T. P. Nguyen and I. Debled-Rennesson, "A discrete geometry approach for dominant point detection," *Pattern Recognition,* vol. 44, pp. 32-44, 2011.

[38]    S. Lavallee and R. Szeliski, "Recovering the position and orientation of free-form objects from image contours using 3D distance maps," *IEEE Transactions on Pattern Analysis and Machine Intelligence,* vol. 17, pp. 378-390, 1995.

[39]    J. H. Elder and R. M. Goldberg, "Image editing in the contour domain," *IEEE Transactions on Pattern Analysis and Machine Intelligence,* vol. 23, pp. 291-296, 2001.

[40]    A. Kolesnikov and P. Fränti, "Reduced-search dynamic programming for approximation of polygonal curves," *Pattern Recognition Letters,* vol. 24, pp. 2243-2254, 2003.

[41]    R. Yang and Z. Zhang, "Eye gaze correction with stereovision for video-teleconferencing," *IEEE Transactions on Pattern Analysis and Machine Intelligence,* vol. 26, pp. 956-960, 2004.

[42]    A. Kolesnikov and P. Fränti, "Data reduction of large vector graphics," *Pattern Recognition,* vol. 38, pp. 381-394, 2005.

[43]    D. Brunner and P. Soille, "Iterative area filtering of multichannel images," *Image and Vision Computing,* vol. 25, pp. 1352-1364, 2007.

[44]    F. Mokhtarian and A. Mackworth, "Scale-based description and recognition of planar curves and two-dimensional shapes," *IEEE Transactions on Pattern Analysis and Machine Intelligence,* vol. PAMI-8, pp. 34-43, 1986.

[45]     D. K. Prasad, M. K. H. Leung, and S. Y. Cho, "Edge curvature and convexity based ellipse detection method," *Pattern Recognition,* vol. 45, pp. 3204-3221, 2012.

[46]     A. Masood, "Dominant point detection by reverse polygonization of digital curves," *Image and Vision Computing,* vol. 26, pp. 702-715, 2008.

[47]     W. Y. Wu, "An adaptive method for detecting dominant points," *Pattern Recognition,* vol. 36, pp. 2231-2237, 2003.

[48]     A. Kolesnikov and P. Fränti, "Polygonal approximation of closed discrete curves," *Pattern Recognition,* vol. 40, pp. 1282-1293, 2007.

[49]     P. Bhowmick and B. B. Bhattacharya, "Fast polygonal approximation of digital curves using relaxed straightness properties," *IEEE Transactions on Pattern Analysis and Machine Intelligence,* vol. 29, pp. 1590-1602, 2007.

[50]     M. Marji and P. Siy, "Polygonal representation of digital planar curves through dominant point detection - A nonparametric algorithm," *Pattern Recognition,* vol. 37, pp. 2113-2130, 2004.

[51]     C.-H. Teh and R. T. Chin, "On the detection of dominant points on digital curves," *IEEE Transactions on Pattern Analysis and Machine Intelligence,* vol. 11, pp. 859-872, 1989.

[52]     J. C. Perez and E. Vidal, "Optimum polygonal approximation of digitized curves," *Pattern Recognition Letters,* vol. 15, pp. 743-750, 1994.

[53]     D. G. Lowe, "Three-dimensional object recognition from single two-dimensional images," *Artificial Intelligence,* vol. 31, pp. 355-395, 1987.

[54]     L. J. Latecki and R. Lakämper, "Convexity Rule for Shape Decomposition Based on Discrete Contour Evolution," *Computer Vision and Image Understanding,* vol. 73, pp. 441-454, 1999.

[55]    B. K. Ray and K. S. Ray, "An algorithm for detection of dominant points and polygonal approximation of digitized curves," *Pattern Recognition Letters,* vol. 13, pp. 849-856, 1992.

[56]    P. V. Sankar and C. U. Sharma, "A parallel procedure for the detection of dominant points on a digital curve," *Computer Graphics and Image Processing,* vol. 7, pp. 403-412, 1978.

[57]    M. Salotti, "Optimal polygonal approximation of digitized curves using the sum of square deviations criterion," *Pattern Recognition,* vol. 35, pp. 435-443, 2002.

[58]    N. Ansari and K. W. Huang, "Non-parametric dominant point detection," *Pattern Recognition,* vol. 24, pp. 849-862, 1991.

[59]    T. M. Cronin, "A boundary concavity code to support dominant point detection," *Pattern Recognition Letters,* vol. 20, pp. 617-634, 1999.

[60]    B. Sarkar, S. Roy, and D. Sarkar, "Hierarchical representation of digitized curves through dominant point detection," *Pattern Recognition Letters,* vol. 24, pp. 2869-2882, 2003.

[61]    B. K. Ray and K. S. Ray, "Detection of significant points and polygonal approximation of digitized curves," *Pattern Recognition Letters,* vol. 13, pp. 443-452, 1992.

[62]    D. Sarkar, "A simple algorithm for detection of significant vertices for polygonal approximation of chain-coded curves," *Pattern Recognition Letters,* vol. 14, pp. 959-964, 1993.

[63]    T. Lewiner, J. D. Gomes Jr, H. Lopes, and M. Craizer, "Curvature and torsion estimators based on parametric curve fitting," *Computers and Graphics,* vol. 29, pp. 641-655, 2005.

[64]    F. Cazals and M. Pouget, "Estimating differential quantities using polynomial fitting of osculating jets," *Computer Aided Geometric Design,* vol. 22, pp. 121-146, 2005.

[65]    R. A. McLaughlin and M. D. Alder, "The hough transform versus the upwrite," *IEEE Transactions on Pattern Analysis and Machine Intelligence,* vol. 20, pp. 396-400, 1998.

[66]    J. Matas, Z. Shao, and J. Kittler, "Estimation of curvature and tangent direction by median filtered differencing," *Lecture Notes in Computer Science,* vol. 974, pp. 83-88, 1995.

[67]    B. Kerautret and J. O. Lachaud, "Robust estimation of curvature along digital contours with global optimization," in *Lecture Notes in Computer Science*. vol. 4992, ed Berlin, Heidelberg: Springer Verlag, 2008, pp. 334-345.

[68]    E. Kim, M. Haseyama, and H. Kitajima, "Fast and Robust Ellipse Extraction from Complicated Images," in *Proceedings of the International Conference on Information Technology and Applications*, 2002, pp. 357-362.

[69]    D. K. Prasad, M. K. H. Leung, C. Quek, and M. S. Brown, "DEB: Definite error bounded tangent estimator for digital curves," *IEEE Transactions on Image Processing,* 2013(under review).

[70]    R. A. McLaughlin, "Randomized Hough transform: Improved ellipse detection with comparison," *Pattern Recognition Letters,* vol. 19, pp. 299-305, 1998.

[71]    Z. G. Cheng and Y. C. Liu, "Efficient technique for ellipse detection using Restricted Randomized Hough Transform," in *Proceedings of the International Conference on Information Technology*, Las Vegas, NV, 2004, pp. 714-718.

[72]    N. Kiryati, Y. Eldar, and A. M. Bruckstein, "A probabilistic Hough transform," *Pattern Recognition,* vol. 24, pp. 303-316, 1991.

[73]   N. Kiryati, H. Kalviainen, and S. Alaoutinen, "Randomized or probabilistic Hough transform: Unified performance evaluation," *Pattern Recognition Letters,* vol. 21, pp. 1157-1164, 2000.

[74]   J. Illingworth and J. Kittler, "A survey of the hough transform," *Computer Vision, Graphics and Image Processing,* vol. 44, pp. 87-116, 1988.

[75]   P. K. Ser and W. C. Siu, "Novel detection of conics using 2-D Hough planes," in *Proceedings of the IEE Vision, Image and Signal Processing*, 1995, pp. 262-270.

[76]   R. K. K. Yip, P. K. S. Tam, and D. N. K. Leung, "Modification of hough transform for circles and ellipses detection using a 2-dimensional array," *Pattern Recognition,* vol. 25, pp. 1007-1022, 1992.

[77]   A. S. Aguado, M. E. Montiel, and M. S. Nixon, "On using directional information for parameter space decomposition in ellipse detection," *Pattern Recognition,* vol. 29, pp. 369-381, 1996.

[78]   N. Guil and E. L. Zapata, "Lower order circle and ellipse Hough transform," *Pattern Recognition,* vol. 30, pp. 1729-1744, 1997.

[79]   H. X. Li, H. Zheng, and Y. Wang, "Segment Hough transform - a novel Hough-based algorithm for curve detection," in *Proceedings of the International Conference on Image and Graphics*, Chengdu, PEOPLES R CHINA, 2007, pp. 471-477.

[80]   A. Fitzgibbon, M. Pilu, and R. B. Fisher, "Direct least square fitting of ellipses," *IEEE Transactions on Pattern Analysis and Machine Intelligence,* vol. 21, pp. 476-480, May 1999.

[81]    P. L. Rosin and G. A. W. West, "Nonparametric segmentation of curves into various representations," *IEEE Transactions on Pattern Analysis and Machine Intelligence,* vol. 17, pp. 1140-1153, 1995.

[82]    P. L. Rosin, "Ellipse fitting by accumulating five-point fits," *Pattern Recognition Letters,* vol. 14, pp. 661-669, 1993.

[83]    P. L. Rosin, "A note on the least squares fitting of ellipses," *Pattern Recognition Letters,* vol. 14, pp. 799-808, 1993.

[84]    D. Chaudhuri, "A simple least squares method for fitting of ellipses and circles depends on border points of a two-tone image and their 3-D extensions," *Pattern Recognition Letters,* vol. 31, pp. 818-829, Jul 2010.

[85]    J. Cabrera and P. Meer, "Unbiased estimation of ellipses by bootstrapping," *IEEE Transactions on Pattern Analysis and Machine Intelligence,* vol. 18, pp. 752-756, 1996.

[86]    P. L. Rosin, "Analysing error of fit functions for ellipses," *Pattern Recognition Letters,* vol. 17, pp. 1461-1470, 1996.

[87]    P. L. Rosin, "Assessing error of fit functions for ellipses," *Graphical Models and Image Processing,* vol. 58, pp. 494-502, 1996.

[88]    P. L. Rosin, "Ellipse fitting using orthogonal hyperbolae and Stirling's oval," *Graphical Models and Image Processing,* vol. 60, pp. 209-213, 1998.

[89]    P. L. Rosin, "Evaluating Harker and O'Leary's distance approximation for ellipse fitting," *Pattern Recognition Letters,* vol. 28, pp. 1804-1807, Oct 2007.

[90]    S. J. Ahn, W. Rauh, and H. J. Warnecke, "Least-squares orthogonal distances fitting of circle, sphere, ellipse, hyperbola, and parabola," *Pattern Recognition,* vol. 34, pp. 2283-2303, Dec 2001.

[91]    O. Strauss, "Reducing the precision/uncertainty duality in the Hough transform," in *Proceedings of the IEEE International Conference on Image Processing*, 1996, pp. 967-970.

[92]    T. Ellis, A. Abbood, and B. Brillault, "Ellipse detection and matching with uncertainty," *Image and Vision Computing,* vol. 10, pp. 271-276, Jun 1992.

[93]    D. K. Prasad and M. K. H. Leung, "Reliability/Precision Uncertainty in Shape Fitting Problems," in *IEEE International Conference on Image Processing*, Hong Kong, 2010, pp. 4277-4280.

[94]    S. C. Zhang and Z. Q. Liu, "A robust, real-time ellipse detector," *Pattern Recognition,* vol. 38, pp. 273-287, 2005.

[95]    D. K. Prasad and M. K. H. Leung, "An ellipse detection method for real images," in *25th International Conference of Image and Vision Computing New Zealand (IVCNZ 2010)*, Queenstown, New Zealand, 2010, pp. 1-8.

[96]    D. K. Prasad and M. K. H. Leung, "A hybrid approach for ellipse detection in real images," in *2nd International Conference on Digital Image Processing*, Singapore, 2010, pp. 75460I-6.

[97]    Y. Qiao and S. H. Ong, "Connectivity-based multiple-circle fitting," *Pattern Recognition,* vol. 37, pp. 755-765, 2004.

[98]    Y. Qiao and S. H. Ong, "Arc-based evaluation and detection of ellipses," *Pattern Recognition,* vol. 40, pp. 1990-2003, 2007.

[99]    Y. Xie and J. Ohya, "Elliptical object detection by a modified ransac with sampling constraint from boundary Curves' clustering," *IEICE Transactions on Information and Systems,* vol. E93-D, pp. 611-623, 2010.

[100]  M. A. Fischler and R. C. Bolles, "Random sample consensus: a paradigm for model fitting with applications to image analysis and automated cartography," *ACM Communications,* vol. 24, pp. 381-395, 1981.

[101]  A. Y. S. Chia, D. Rajan, M. K. H. Leung, and S. Rahardja, "A split and merge based ellipse detector," in *Proceedings of the IEEE International Conference on Image Processing*, San Diego, CA, 2008, pp. 3212-3215.

[102]  K. Hahn, S. Jung, Y. Han, and H. Hahn, "A new algorithm for ellipse detection by curve segments," *Pattern Recognition Letters,* vol. 29, pp. 1836-1841, 2008.

[103]  T. Kawaguchi and R. Nagata, "Ellipse detection using a genetic algorithm," in *Proceedings of the International Conference on Pattern Recognition*, Brisbane, Australia, 1998, pp. 141-145.

[104]  T. Kawaguchi and R. I. Nagata, "Ellipse detection using grouping of edgels into line-support regions," in *Proceedings of the IEEE International Conference on Image Processing*, 1998, pp. 70-74.

[105]  Q. Ji and R. M. Haralick, "Statistically efficient method for ellipse detection," in *Proceedings of the IEEE International Conference on Image Processing*, 1999, pp. 730-733.

[106]  Y. C. Cheng, "The distinctiveness of a curve in a parameterized neighborhood: Extraction and applications," *IEEE Transactions on Pattern Analysis and Machine Intelligence,* vol. 28, pp. 1215-1222, 2006.

[107]  C. A. Basca, M. Talos, and R. Brad, "Randomized hough transform for ellipse detection with result clustering," in *Proceedings of the International Conference on Computer as a Tool*, Belgrade, SERBIA MONTENEG, 2005, pp. 1397-1400.

[108] J. Princen, J. Illingworth, and J. Kittler, "Hypothesis testing - a framework for analyzing and optimizing Hough transform performance," *IEEE Transactions on Pattern Analysis and Machine Intelligence,* vol. 16, pp. 329-341, Apr 1994.

[109] C. Wang, T. S. Newman, and C. Cao, "New hypothesis distinctiveness measure for better ellipse extraction," in *Lecture Notes in Computer Science* vol. 4633, ed, 2007, pp. 176-186.

[110] Q. Ji and R. M. Haralick, "Error propagation for the Hough transform," *Pattern Recognition Letters,* vol. 22, pp. 813-823, 2001.

[111] O. M. Elmowafy and M. C. Fairburst, "Improving ellipse detection using a fast graphical method," *Electronics Letters,* vol. 35, pp. 135-137, Jan 1999.

[112] E. W. Weisstein, *CRC concise encyclopedia of mathematics* vol. 2. Florida: CRC Press.

[113] P. D. Kovesi. *MATLAB and Octave Functions for Computer Vision and Image Processing (2000 ed.)* [http://www.csse.uwa.edu.au/~pk/Research/MatlabFns/index.html].

[114] X. Z. Bai, C. M. Sun, and F. G. Zhou, "Splitting touching cells based on concave points and ellipse fitting," *Pattern Recognition,* vol. 42, pp. 2434-2446, Nov 2009.

[115] F. Mokhtarian and S. Abbasi, "Affine Curvature Scale Space with Affine Length Parametrisation," *Pattern Analysis & Applications,* vol. 4, pp. 1-8, 2001/03/01 2001.

[116] D. Chetverikov, "A Simple and Efficient Algorithm for Detection of High Curvature Points in Planar Curves," in *Computer Analysis of Images and Patterns.* vol. 2756, N. Petkov and M. Westenberg, Eds., ed: Springer Berlin Heidelberg, 2003, pp. 746-753.

[117] B. Zhong, K. K. Ma, and W. Liao, "Scale-space behavior of planar-curve corners," *IEEE Transactions on Pattern Analysis and Machine Intelligence,* vol. 31, pp. 1517-1524, 2009.

[118] B. Zhong and W. Liao, "Direct Curvature Scale Space in Corner Detection," in *Structural, Syntactic, and Statistical Pattern Recognition.* vol. 4109, D.-Y. Yeung, J. Kwok, A. Fred, F. Roli, and D. Ridder, Eds., ed: Springer Berlin Heidelberg, 2006, pp. 235-242.

[119] D. K. Prasad, "Application of Image Composition Analysis for Image Processing," in *IMI International Workshop on Computational Photography and Aesthetics*, 2009.

[120] D. K. Prasad and M. K. H. Leung, "Clustering of ellipses based on their distinctiveness: An aid to ellipse detection algorithms," in *3rd IEEE International Conference on Computer Science and Information Technology (ICCSIT)*, 2010, pp. 292-297.

[121] D. K. Prasad and J. A. Starzyk, "Object detection and representation in motivated conscious machines," in *Decade of the Mind VI Conference*, 2010.

[122] D. K. Prasad, "High Availability based Migration Analysis to Cloud Computing for High Growth Businesses," *International Journal of Computer Networks (IJCN),* vol. 4, 2012.

[123] D. K. Prasad and C. K. Prasath, "Reconfigurable Virtual Platform for Real Time Kernel," in *9th USENIX Symposium on Operating Systems Design and Implementation (OSDI'10)*, 2010.

[124] C. K. Prasath and D. K. Prasad, "Design and development of a reconfigurable virtual platform for real time kernel," in *2nd International Conference on Software Technology and Engineering (ICSTE)*, 2010, pp. V2-170-V2-174.

[125]  D. K. Prasad and J. A. Starzyk, "A Perspective on Machine Consciousness," in *The Second International Conference on Advanced Cognitive Technologies and Applications, COGNITIVE 2010*, 2010, pp. 109-114.

[126]  J. A. Starzyk and D. K. Prasad, "Machine Consciousness: A Computational Model," in *Third International ICSC Symposium on Models of Consciousness, BICS 2010*, 2010.

[127]  J. A. Starzyk and D. K. Prasad, "A computational model of machine consciousness," *International Journal of Machine Consciousness,* vol. 3, pp. 255-282, 2011.

[128]  D. K. Prasad, C. Quek, and M. K. H. Leung, "A hybrid approach for breast tissue data classification," in *IEEE Region 10 Conference TENCON 2009*, 2009, pp. 1-4.

[129]  D. K. Prasad, "Adaptive traffic signal control system with cloud computing based online learning," in *8th International Conference on Information, Communications, and Signal Processing (ICICS 2011)*, Singapore, 2011.

[130]  D. K. Prasad, "Rise of international schools in India," *International Journal of Education, Economics, and Development,* 2013.

[131]  D. K. Prasad, M. K. H. Leung, and C. Quek, "PRO: A novel Precision and Reliability based optimization method for dominant point detection," *Pattern Analysis and Applications,* 2012(under review).

[132]  D. K. Prasad, M. K. H. Leung, C. Quek, and S.-Y. Cho, "A novel framework for making dominant point detection methods non-parametric," *Image and Vision Computing,* vol. 30, pp. 843-859, 2012.

[133]  D. K. Prasad, C. Quek, and M. K. H. Leung, "A non-heuristic dominant point detection based on suppression of break points," in *Image Analysis and*

*Recognition*. vol. 7324, A. Campilho and M. Kamel, Eds., ed Aveiro, Portugal: Springer Berlin Heidelberg, 2012, pp. 269-276.

[134]   D. K. Prasad and M. K. H. Leung, "Polygonal representation of digital curves," in *Digital Image Processing*, S. G. Stanciu, Ed., ed: InTech, 2012, pp. 71-90.

[135]   D. K. Prasad, C. Quek, M. K. H. Leung, and S. Y. Cho, "A parameter independent line fitting method," in *Asian Conference on Pattern Recognition (ACPR)*, Beijing, China, 2011, pp. 441-445.

[136]   D. K. Prasad, "Assessing error bound for dominant point detection," *International Journal of Image Processing (IJIP),* vol. 6, pp. 326-333, 2012.

[137]   D. K. Prasad, M. K. H. Leung, and C. Quek, "ElliFit: An unconstrained, non-iterative, least squares based geometric Ellipse Fitting method," *Pattern Recognition,* vol. 46, pp. 1449-1465, 2013.

[138]   D. K. Prasad, C. Quek, and M. K. H. Leung, "A precise ellipse fitting method for noisy data," in *Image Analysis and Recognition*. vol. 7324, A. Campilho and M. Kamel, Eds., ed Aveiro, Portugal: Springer Berlin Heidelberg, 2012, pp. 253-260.

[139]   D. K. Prasad and M. K. H. Leung, "Methods for ellipse detection from edge maps of real images," in *Machine Vision - Applications and Systems*, F. Solari, M. Chessa, and S. Sabatini, Eds., ed: InTech, 2012, pp. 135-162.

[140]   D. K. Prasad, C. Quek, and M. K. H. Leung, "Fast segmentation of sub-cellular organelles," *International Journal of Image Processing (IJIP),* vol. 6, pp. 317-325, 2012.

[141]   D. K. Prasad, M. K. H. Leung, and C. Quek, "Object detection using geometric features and heirarchical object template," *in preparation,* 2012.

[142] H. Imai and M. Iri, "Polygonal approximation of a curve (formulations and algorithms)," in *Computational Morphology*, G. T. Toussaint, Ed., ed Amsterdam, 1988, pp. 71-86.

[143] J. Sklansky and V. Gonzalez, "Fast polygonal approximation of digitized curves," *Pattern Recognition,* vol. 12, pp. 327-331, 1980.

[144] K. Wall and P. E. Danielsson, "A fast sequential method for polygonal approximation of digitized curves," *Computer Vision, Graphics, & Image Processing,* vol. 28, pp. 220-227, 1984.

[145] J. G. Dunham, "Optimum uniform piecewise linear approximation of planar curves," *IEEE Transactions on Pattern Analysis and Machine Intelligence,* vol. PAMI-8, pp. 67-75, 1986.

[146] M. K. Leung, "Dynamic two-strip algorithm in curve fitting," *Pattern Recognition,* vol. 23, pp. 69-79, 1990.

[147] A. K. Gupta, S. Chaudhury, and G. Parthasarathy, "A new approach for aggregating edge points into line segments," *Pattern Recognition,* vol. 26, pp. 1069-1086, 1993.

[148] W. Wen and A. Lozzi, "Recognition and inspection of manufactured parts using line moments of their boundaries," *Pattern Recognition,* vol. 26, pp. 1461-1471, 1993.

[149] P. L. Rosin, "Techniques for assessing polygonal approximations of curves," *IEEE Transactions on Pattern Analysis and Machine Intelligence,* vol. 19, pp. 659-666, 1997.

[150] H. Lopes, J. B. Oliveira, and L. H. De Figueiredo, "Robust adaptive polygonal approximation of implicit curves," *Computers and Graphics (Pergamon),* vol. 26, pp. 841-852, 2002.

[151]  P. L. Rosin, "Assessing the behaviour of polygonal approximation algorithms," *Pattern Recognition,* vol. 36, pp. 505-518, 2002.

[152]  N. Barnes, G. Loy, D. Shaw, and A. Robles-Kelly, "Regular polygon detection," in *Proceedings of the IEEE International Conference on Computer Vision*, 2005, pp. 778-785.

[153]  A. Carmona-Poyato, N. L. Fernández-García, R. Medina-Carnicer, and F. J. Madrid-Cuevas, "Dominant point detection: A new proposal," *Image and Vision Computing,* vol. 23, pp. 1226-1236, 2005.

[154]  I. Debled-Rennesson, J. L. Rémy, and J. Rouyer-Degli, "Linear segmentation of discrete curves into blurred segments," *Discrete Applied Mathematics,* vol. 151, pp. 122-137, 2005.

[155]  F. Feschet, "Canonical representations of discrete curves," *Pattern Analysis and Applications,* vol. 8, pp. 84-94, 2005.

[156]  A. Kolesnikov, "Constrained piecewise linear approximation of digital curves," Tampa, FL, 2008.

[157]  L. J. Latecki, M. Sobel, and R. Lakaemper, "Piecewise linear models with guaranteed closeness to the data," *IEEE Transactions on Pattern Analysis and Machine Intelligence,* vol. 31, pp. 1525-1531, 2009.

[158]  N. Barnes, G. Loy, and D. Shaw, "The regular polygon detector," *Pattern Recognition,* vol. 43, pp. 592-602, 2010.

[159]  A. Carmona-Poyato, R. Medina-Carnicer, F. J. Madrid-Cuevas, R. Muoz-Salinas, and N. L. Fernndez-Garca, "A new measurement for assessing polygonal approximation of curves," *Pattern Recognition,* vol. 44, pp. 45-54, 2011.

[160] F. S. Acton, *Analysis of straight-line data*. New York: Peter Smith Publisher, Incorporated, 1984.

[161] R. A. Horn, *Matrix Analysis*. Melbourne, Australia: Cambridge University Press, 1985.

[162] I. Debled-Renesson and J.-P. R´eveill`es, "A linear algorithm for segmentation of discrete curves," *International Journal of Pattern Recognition and Artificial Intelligence,* vol. 9, pp. 635–662, 1995.

[163] Kanungo T., Haralick R. M., Baird H. S., Stuetzle W., and M. D., "Document Degradation Models: Parameter Estimation and Model Validation," in *IAPR Wolkshop on Machine Vision Application*, Kawasaki, Japan, 1994.

[164] G. McCarter and A. Storkey. Air Freight image sequences [Online]. Available: http://homepages.inf.ed.ac.uk/amos/afreightdata.html

[165] D. Martin, C. Fowlkes, D. Tal, and J. Malik, "A database of human segmented natural images and its application to evaluating segmentation algorithms and measuring ecological statistics," in *8th IEEE International Conference on Computer Vision (ICCV 2001)*, Vancouver, BC , Canada, 2001, pp. 416 - 423.

[166] P. Carbonetto, G. Dorko', C. Schmid, H. Kuck, and N. De Freitas, "Learning to recognize objects with little supervision," *International Journal of Computer Vision,* vol. 77, pp. 219-237, 2008.

[167] L. Fei-Fei, R. Fergus, and P. Perona, "Learning generative visual models from few training examples: An incremental Bayesian approach tested on 101 object categories," *Computer Vision and Image Understanding,* vol. 106, pp. 59-70, 2007.

[168] M. Everingham, L. V. Gool, C. K. I. Williams, J. Winn, and A. Zisserman. The PASCAL Visual Object Classes Challenge 2007 (VOC2007) [Online].

Available: http://www.pascal-network.org/challenges/VOC/voc2007/workshop/index.html

[169] M. Everingham, L. V. Gool, C. K. I. Williams, J. Winn, and A. Zisserman. The PASCAL Visual Object Classes Challenge 2008 (VOC2008) [Online]. Available: http://www.pascal-network.org/challenges/VOC/voc2008/workshop/index.html

[170] M. Everingham, L. V. Gool, C. K. I. Williams, J. Winn, and A. Zisserman. The PASCAL Visual Object Classes Challenge 2009 (VOC2009) [Online]. Available: http://www.pascal-network.org/challenges/VOC/voc2009/workshop/index.html

[171] M. Everingham, L. V. Gool, C. K. I. Williams, J. Winn, and A. Zisserman. The PASCAL Visual Object Classes Challenge 2010 (VOC2010) [Online]. Available: http://www.pascal-network.org/challenges/VOC/voc2010/workshop/index.html

[172] R. C. Gonzalez and R. E. Woods, *Digital Image Processing*, 3 ed. Delhi: Pearson Prentice Hall, 2008.

[173] E. S. Maini, "Enhanced direct least square fitting of ellipses," *International Journal of Pattern Recognition and Artificial Intelligence,* vol. 20, pp. 939-953, 2006.

[174] M. Harker, P. O'Leary, and P. Zsombor-Murray, "Direct type-specific conic fitting and eigenvalue bias correction," *Image and Vision Computing,* vol. 26, pp. 372-381, 2008.

[175] R. Halir and J. Flusser, "Numerically stable direct least squares fitting of ellipses," in *Sixth International Conference in Central Europe on Computer Graphics and Visualization*, 1998, pp. 125-132.

[176] D. K. Prasad, M. K. Leung, and C. Quek, "Numerically stable direct least square ellipse fitting method," *Journal of Mathematical Imaging and Vision,* 2012 (under review).

[177] S. M. Rump, "Structured perturbations and symmetric matrices," *Linear Algebra and Its Applications,* vol. 278, pp. 121-132, 1998.

[178] P. N. Tan, M. Steinbach, and V. Kumar, *Introduction to data mining*: Pearson Addison Wesley, 2006.

[179] X. Bai, C. Sun, and F. Zhou, "Splitting touching cells based on concave points and ellipse fitting," *Pattern Recognition,* vol. 42, pp. 2434-2446, 2009.

[180] D. K. Prasad and M. K. H. Leung, "Error analysis of geometric ellipse detection methods due to quantization," in *4th Pacific Rim Symposium on Image and Vision Technology (PSIVT)*, Singapore, 2010.

[181] A. Y. S. Chia, S. Rahardja, D. Rajan, and M. K. H. Leung, "Structural descriptors for category level object detection," *IEEE Transactions on Multimedia,* vol. 11, pp. 1407-1421, 2009.

[182] H. S. Kim, W. S. Kang, J. I. Shin, and S. H. Park, "Face detection using template matching and ellipse fitting," *IEICE Transactions on Information and Systems,* vol. E83-D, pp. 2008-2011, 2000.

[183] X. Feng, C. Fang, X. Ding, and Y. Wu, "Iris localization with dual coarse-to-fine strategy," in *Proceedings of the International Conference on Pattern Recognition*, 2006, pp. 553-556.

[184] X. He and P. Shi, "A novel Iris segmentation method for hand-held capture device," in *Lecture Notes in Computer Science* vol. 3832, ed, 2006, pp. 479-485.

[185] P. F. Felzenszwalb and D. P. Huttenlocher, "Pictorial structures for object recognition," *International Journal of Computer Vision,* vol. 61, pp. 55-79, 2005.

[186] S. Kumar, S. H. Ong, S. Ranganath, and F. T. Chew, "Invariant texture classification for biomedical cell specimens via non-linear polar map filtering," *Computer Vision and Image Understanding,* vol. 114, pp. 44-53, Jan 2010.

[187] G. Dong, N. Ray, and S. T. Acton, "Intravital leukocyte detection using the gradient inverse coefficient of variation," *IEEE Transactions on Medical Imaging,* vol. 24, pp. 910-924, 2005.

[188] T. Peng, G. M. C. Bonamy, E. Glory-Afshar, D. R. Rines, S. K. Chanda, and R. F. Murphy, "Determining the distribution of probes between different subcellular locations through automated unmixing of subcellular patterns," *Proceedings of the National Academy of Sciences of the USA,* vol. 107, pp. 2944-2949, Feb 2010.

[189] G. Bouchard and B. Triggs, "Hierarchical part-based visual object categorization," in *Proceedings of the IEEE Conference on Computer Vision and Pattern Recognition*, 2005, pp. 710-715.

[190] D. K. Prasad, M. K. H. Leung, and C. Quek, "Spectral matching of polygons," vol. in preparation, 2013.

[191] D. K. Prasad, "Survey of the problem of object detection in real images," *International Journal of Image Processing (IJIP),* vol. 6, 2012.

[192] O. C. Hamsici and A. M. Martinez, "Rotation invariant kernels and their application to shape analysis," *IEEE Transactions on Pattern Analysis and Machine Intelligence,* vol. 31, pp. 1985-1999, 2009.

[193] Z. Si, H. Gong, Y. N. Wu, and S. C. Zhu, "Learning mixed templates for object recognition," in *Proceedings of the IEEE Conference on Computer Vision and Pattern Recognition*, 2009, pp. 272-279.

[194] L. Szumilas and H. Wildenauer, "Spatial configuration of local shape features for discriminative object detection," in *Lecture Notes in Computer Science* vol. 5875, ed, 2009, pp. 22-33.

[195] L. Szumilas, H. Wildenauer, and A. Hanbury, "Invariant shape matching for detection of semi-local image structures," in *Lecture Notes in Computer Science* vol. 5627, ed, 2009, pp. 551-562.

[196] M. P. Kumar, P. H. S. Torr, and A. Zisserman, "OBJCUT: Efficient Segmentation Using Top-Down and Bottom-Up Cues," *IEEE Transactions on Pattern Analysis and Machine Intelligence,* vol. 32, pp. 530-545, 2009.

[197] K. Schindler and D. Suter, "Object detection by global contour shape," *Pattern Recognition,* vol. 41, pp. 3736-3748, 2008.

[198] N. Alajlan, M. S. Kamel, and G. H. Freeman, "Geometry-based image retrieval in binary image databases," *IEEE Transactions on Pattern Analysis and Machine Intelligence,* vol. 30, pp. 1003-1013, 2008.

[199] Y. N. Wu, Z. Si, H. Gong, and S. C. Zhu, "Learning Active Basis Model for Object Detection and Recognition," *International Journal of Computer Vision,* pp. 1-38, 2009.

[200] X. Ren, C. C. Fowlkes, and J. Malik, "Learning probabilistic models for contour completion in natural images," *International Journal of Computer Vision,* vol. 77, pp. 47-63, 2008.

[201] J. Winn and J. Shotton, "The layout consistent random field for recognizing and segmenting partially occluded objects," in *Proceedings of the IEEE Conference on Computer Vision and Pattern Recognition*, 2006, pp. 37-44.

[202] V. Ferrari, T. Tuytelaars, and L. Van Gool, "Object detection by contour segment networks," in *Lecture Notes in Computer Science* vol. 3953, ed, 2006, pp. 14-28.

[203] K. Mikolajczyk, B. Leibe, and B. Schiele, "Multiple object class detection with a generative model," in *Proceedings of the IEEE Conference on Computer Vision and Pattern Recognition*, 2006, pp. 26-33.

[204] R. C. Nelson and A. Selinger, "Cubist approach to object recognition," in *Proceedings of the IEEE International Conference on Computer Vision*, 1998, pp. 614-621.

[205] V. Ferrari, L. Fevrier, F. Jurie, and C. Schmid, "Groups of adjacent contour segments for object detection," *IEEE Transactions on Pattern Analysis and Machine Intelligence,* vol. 30, pp. 36-51, 2008.

[206] S. Ali and M. Shah, "A supervised learning framework for generic object detection in images," in *Proceedings of the IEEE International Conference on Computer Vision*, 2005, pp. 1347-1354.

[207] E. Borenstein and S. Ullman, "Learning to segment," in *Lecture Notes in Computer Science* vol. 3023, ed, 2004, pp. 315-328.

[208] E. Borenstein and J. Malik, "Shape guided object segmentation," in *Proceedings of the IEEE Conference on Computer Vision and Pattern Recognition*, 2006, pp. 969-976.

[209]  J. Wang, V. Athitsos, S. Sclaroff, and M. Betke, "Detecting objects of variable shape structure with Hidden State Shape Models," *IEEE Transactions on Pattern Analysis and Machine Intelligence,* vol. 30, pp. 477-492, 2008.

[210]  J. Zhang, M. Marszalek, S. Lazebnik, and C. Schmid, "Local features and kernels for classification of texture and object categories: A comprehensive study," in *Proceedings of the IEEE Conference on Computer Vision and Pattern Recognition Workshops*, 2006, pp. 13-13.

[211]  Y. Amit, D. Geman, and X. Fan, "A coarse-to-fine strategy for multiclass shape detection," *IEEE Transactions on Pattern Analysis and Machine Intelligence,* vol. 26, pp. 1606-1621, 2004.

[212]  J. Shotton, A. Blake, and R. Cipolla, "Contour-based learning for object detection," in *Proceedings of the IEEE International Conference on Computer Vision*, 2005, pp. 503-510.

[213]  A. Opelt, A. Pinz, M. Fussenegger, and P. Auer, "Generic object recognition with boosting," *IEEE Transactions on Pattern Analysis and Machine Intelligence,* vol. 28, pp. 416-431, 2006.

[214]  Y. Freund, "Boosting a Weak Learning Algorithm by Majority," *Information and Computation,* vol. 121, pp. 256-285, 1995.

[215]  A. Mohan, C. Papageorgiou, and T. Poggio, "Example-based object detection in images by components," *IEEE Transactions on Pattern Analysis and Machine Intelligence,* vol. 23, pp. 349-361, 2001.

[216]  A. Demiriz, K. P. Bennett, and J. Shawe-Taylor, "Linear programming boosting via column generation," *Machine Learning,* vol. 46, pp. 225-254, 2002.

[217] S. Agarwal, A. Awan, and D. Roth, "Learning to detect objects in images via a sparse, part-based representation," *IEEE Transactions on Pattern Analysis and Machine Intelligence,* vol. 26, pp. 1475-1490, 2004.

[218] R. Fergus, P. Perona, and A. Zisserman, "A visual category filter for google images," in *Lecture Notes in Computer Science* vol. 3021, ed, 2004, pp. 242-256.

[219] A. Opelt, M. Fussenegger, A. Pinz, and P. Auer, "Weak hypotheses and boosting for generic object detection and recognition," in *Lecture Notes in Computer Science* vol. 3022, ed, 2004, pp. 71-84.

[220] A. Torralba, K. P. Murphy, and W. T. Freeman, "Sharing features: Efficient boosting procedures for multiclass object detection," in *Proceedings of the IEEE Conference on Computer Vision and Pattern Recognition*, 2004, pp. 762-769.

[221] A. Bar-Hillel, T. Hertz, and D. Weinshall, "Object class recognition by boosting a part-based model," in *Proceedings of the IEEE Conference on Computer Vision and Pattern Recognition*, 2005, pp. 702-709.

[222] E. Bart and S. Ullman, "Cross-generalization: Learning novel classes from a single example by feature replacement," in *Proceedings of the IEEE Conference on Computer Vision and Pattern Recognition*, 2005, pp. 672-679.

[223] R. Fergus, L. Fei-Fei, P. Perona, and A. Zisserman, "Learning object categories from Google's image search," in *Proceedings of the IEEE International Conference on Computer Vision*, 2005, pp. 1816-1823.

[224] F. Jurie and B. Triggs, "Creating efficient codebooks for visual recognition," in *Proceedings of the IEEE International Conference on Computer Vision*, 2005, pp. 604-610.

[225]  Z. Tu, "Probabilistic boosting-tree: Learning discriminative models for classification, recognition, and clustering," in *Proceedings of the IEEE International Conference on Computer Vision*, 2005, pp. 1589-1596.

[226]  W. Zhang, B. Yu, G. J. Zelinsky, and D. Samaras, "Object class recognition using multiple layer boosting with heterogeneous features," in *Proceedings of the IEEE Conference on Computer Vision and Pattern Recognition*, 2005, pp. 323-330.

[227]  A. Opelt, A. Pinz, and A. Zisserman, "Incremental learning of object detectors using a visual shape alphabet," in *Proceedings of the IEEE Conference on Computer Vision and Pattern Recognition*, 2006, pp. 3-10.

[228]  D. D. Le and S. Satoh, "Ent-Boost: Boosting using entropy measures for robust object detection," *Pattern Recognition Letters,* vol. 28, pp. 1083-1090, 2007.

[229]  A. Bar-Hillel and D. Weinshall, "Efficient learning of relational object class models," *International Journal of Computer Vision,* vol. 77, pp. 175-198, 2008.

[230]  L. Fürst, S. Fidler, and A. Leonardis, "Selecting features for object detection using an AdaBoost-compatible evaluation function," *Pattern Recognition Letters,* vol. 29, pp. 1603-1612, 2008.

[231]  X. Li, B. Yang, F. Zhu, and A. Men, "Real-time object detection based on the improved boosted features," in *Proceedings of SPIE - The International Society for Optical Engineering*, 2009.

[232]  J. J. Yokono and T. Poggio, "Object recognition using boosted oriented filter based local descriptors," *IEEJ Transactions on Electronics, Information and Systems,* vol. 129, 2009.

[233] Y. Chi and M. K. H. Leung, "Part-based object retrieval in cluttered environment," *IEEE Transactions on Pattern Analysis and Machine Intelligence,* vol. 29, pp. 890-895, May 2007.

[234] H. P. Moravec, "Rover visual obstacle avoidance," in *Proceedings of the International Joint Conference on Artificial Intelligence,* Vancouver, CANADA, 1981, pp. 785-790.

[235] C. Harris and M. Stephens, "A combined corner and edge detector," presented at the Alvey Vision Conference, 1988.

[236] J. Li and N. M. Allinson, "A comprehensive review of current local features for computer vision," *Neurocomputing,* vol. 71, pp. 1771-1787, 2008.

[237] K. Mikolajczyk and H. Uemura, "Action recognition with motion-appearance vocabulary forest," in *Proceedings of the IEEE Conference on Computer Vision and Pattern Recognition*, 2008, pp. 1-8.

[238] K. Mikolajczyk and C. Schmid, "A performance evaluation of local descriptors," in *Proceedings of the IEEE Conference on Computer Vision and Pattern Recognition*, 2003, pp. 1615-1630.

[239] D. G. Lowe, "Distinctive image features from scale-invariant keypoints," *International Journal of Computer Vision,* vol. 60, pp. 91-110, 2004.

[240] B. Ommer and J. Buhmann, "Learning the Compositional Nature of Visual Object Categories for Recognition," *IEEE Transactions on Pattern Analysis and Machine Intelligence,* 2010.

[241] B. Leibe, A. Leonardis, and B. Schiele, "Robust object detection with interleaved categorization and segmentation," *International Journal of Computer Vision,* vol. 77, pp. 259-289, 2008.

[242] M. Varma and A. Zisserman, "A statistical approach to material classification using image patch exemplars," *IEEE Transactions on Pattern Analysis and Machine Intelligence,* vol. 31, pp. 2032-2047, 2009.

[243] P. M. Roth, S. Sternig, H. Grabner, and H. Bischof, "Classifier grids for robust adaptive object detection," in *Proceedings of the IEEE Computer Vision and Pattern Recognition*, Miami, FL, 2009, pp. 2727-2734.

[244] J. Matas, O. Chum, M. Urban, and T. Pajdla, "Robust wide-baseline stereo from maximally stable extremal regions," *Image and Vision Computing,* vol. 22, pp. 761-767, 2004.

[245] Y. Chen, L. Zhu, A. Yuille, and H. J. Zhang, "Unsupervised learning of probabilistic object models (POMs) for object classification, segmentation, and recognition using knowledge propagation," *IEEE Transactions on Pattern Analysis and Machine Intelligence,* vol. 31, pp. 1747-1774, 2009.

[246] G. Carneiro and A. D. Jepson, "The quantitative characterization of the distinctiveness and robustness of local image descriptors," *Image and Vision Computing,* vol. 27, pp. 1143-1156, 2009.

[247] W. T. Lee and H. T. Chen, "Histogram-based interest point detectors," in *Proceedings of the IEEE Conference on Computer Vision and Pattern Recognition*, 2009, pp. 1590-1596.

[248] H. T. Comer and B. A. Draper, "Interest Point Stability Prediction," in *Proceedings of the International Conference on Computer Vision Systems*, Liege, 2009.

[249] Y. Ke and R. Sukthankar, "PCA-SIFT: A more distinctive representation for local image descriptors," in *Proceedings of the IEEE Conference on Computer Vision and Pattern Recognition*, 2004, pp. 506-513.

[250] H. Zhang, W. Gao, X. Chen, and D. Zhao, "Object detection using spatial histogram features," *Image and Vision Computing,* vol. 24, pp. 327-341, 2006.

[251] R. Sandler and M. Lindenbaum, "Optimizing gabor filter design for texture edge detection and classification," *International Journal of Computer Vision,* vol. 84, pp. 308-324, 2009.

[252] H. Bischof, H. Wildenauer, and A. Leonardis, "Illumination insensitive recognition using eigenspaces," *Computer Vision and Image Understanding,* vol. 95, pp. 86-104, 2004.

[253] C. H. Lampert and J. Peters, "Active structured learning for high-speed object detection," in *Lecture Notes in Computer Science* vol. 5748, ed, 2009, pp. 221-231.

[254] C. Wallraven, B. Caputo, and A. Graf, "Recognition with local features: The kernel recipe," in *Proceedings of the IEEE International Conference on Computer Vision*, 2003, pp. 257-264.

[255] A. Zalesny, V. Ferrari, G. Caenen, and L. Van Gool, "Composite texture synthesis," *International Journal of Computer Vision,* vol. 62, pp. 161-176, 2005.

[256] J. Shotton, J. Winn, C. Rother, and A. Criminisi, "TextonBoost for image understanding: Multi-class object recognition and segmentation by jointly modeling texture, layout, and context," *International Journal of Computer Vision,* vol. 81, pp. 2-23, 2009.

[257] M. V. Rohith, G. Somanath, D. Metaxas, and C. Kambhamettu, "D - Clutter: Building object model library from unsupervised segmentation of cluttered scenes," in *Proceedings of the IEEE Conference on Computer Vision and Pattern Recognition*, 2009, pp. 2783-2789.

[258] C. Gu, J. J. Lim, P. Arbeláez, and J. Malik, "Recognition using regions," in *Proceedings of the IEEE Conference on Computer Vision and Pattern Recognition*, 2009, pp. 1030-1037.

[259] A. Bosch, A. Zisserman, and X. Muñoz, "Scene classification using a hybrid generative/discriminative approach," *IEEE Transactions on Pattern Analysis and Machine Intelligence,* vol. 30, pp. 712-727, 2008.

[260] H. Wang and J. Oliensis, "Rigid shape matching by segmentation averaging," *IEEE Transactions on Pattern Analysis and Machine Intelligence,* vol. 32, pp. 619-635, 2010.

[261] K. Mikolajczyk and C. Schmid, "A performance evaluation of local descriptors," *IEEE Transactions on Pattern Analysis and Machine Intelligence,* vol. 27, pp. 1615-1630, 2005.

[262] T. Tuytelaars and K. Mikolajczyk, "Local invariant feature detectors: A survey," *Foundations and Trends in Computer Graphics and Vision,* vol. 3, pp. 177-280, 2007.

[263] N. Adluru and L. J. Latecki, "Contour grouping based on contour-skeleton duality," *International Journal of Computer Vision,* vol. 83, pp. 12-29, 2009.

[264] D. Cailliere, F. Denis, D. Pele, and A. Baskurt, "3D mirror symmetry detection using Hough transform," in *Proceedings of the IEEE International Conference on Image Processing*, San Diego, CA, 2008, pp. 1772-1775.

[265] B. Leibe and B. Schiele, "Analyzing appearance and contour based methods for object categorization," in *Proceedings of the IEEE Conference on Computer Vision and Pattern Recognition*, 2003, pp. 409-415.

[266] P. Schnitzspan, M. Fritz, S. Roth, and B. Schiele, "Discriminative structure learning of hierarchical representations for object detection," in *Proceedings of*

*the IEEE Conference on Computer Vision and Pattern Recognition*, 2009, pp. 2238-2245.

[267]   J. Shotton, "Contour and texture for visual recognition of object categories," Doctoral of Philosphy, Queen's College, University of Cambridge, Cambridge, 2007.

[268]   V. Ferrari, T. Tuytelaars, and L. Van Gool, "Simultaneous object recognition and segmentation by image exploration," in *Lecture Notes in Computer Science* vol. 3021, ed, 2004, pp. 40-54.

[269]   V. Ferrari, T. Tuytelaars, and L. Van Gool, "Simultaneous object recognition and segmentation from single or multiple model views," *International Journal of Computer Vision,* vol. 67, pp. 159-188, 2006.

[270]   A. R. Pope and D. G. Lowe, "Probabilistic models of appearance for 3-D object recognition," *International Journal of Computer Vision,* vol. 40, pp. 149-167, 2000.

[271]   J. A. Lasserre, C. M. Bishop, and T. P. Minka, "Principled hybrids of generative and discriminative models," in *Proceedings of the IEEE Conference on Computer Vision and Pattern Recognition*, 2006, pp. 87-94.

[272]   A. E. C. Pece, "On the computational rationale for generative models," *Computer Vision and Image Understanding,* vol. 106, pp. 130-143, 2007.

[273]   D. Parikh, C. L. Zitnick, and T. Chen, "Unsupervised learning of hierarchical spatial structures in images," in *Proceedings of the IEEE Conference on Computer Vision and Pattern Recognition*, 2009, pp. 2743-2750.

[274]   C. H. Lampert, H. Nickisch, and S. Harmeling, "Learning to detect unseen object classes by between-class attribute transfer," in *Proceedings of the IEEE Computer Vision and Pattern Recognition Workshop*, 2009, pp. 951-958.

[275] T. Yeh, J. J. Lee, and T. Darrell, "Fast concurrent object localization and recognition," in *Proceedings of the IEEE Conference on Computer Vision and Pattern Recognition*, 2009, pp. 280-287.

[276] A. J. Joshi, F. Porikli, and N. Papanikolopoulos, "Multi-class active learning for image classification," in *Proceedings of the IEEE Conference on Computer Vision and Pattern Recognition*, 2009, pp. 2372-2379.

[277] S. Maji and J. Malik, "Object detection using a max-margin hough transform," in *Proceedings of the IEEE Computer Vision and Pattern Recognition*, Miami, FL, 2009, pp. 1038-1045.

[278] L. Wu, Y. Hu, M. Li, N. Yu, and X. S. Hua, "Scale-invariant visual language modeling for object categorization," *IEEE Transactions on Multimedia*, vol. 11, pp. 286-294, 2009.

[279] P. Jain and A. Kapoor, "Active learning for large Multi-class problems," in *Proceedings of the IEEE Conference on Computer Vision and Pattern Recognition*, 2009, pp. 762-769.

[280] A. Stefan, V. Athitsos, Q. Yuan, and S. Sclaroff, "Reducing JointBoost-Based Multiclass Classification to Proximity Search," in *Proceedings of the IEEE Conference on Computer Vision and Pattern Recognition*, 2009, pp. 589-596.

[281] M. Fritz, B. Leibe, B. Caputo, and B. Schiele, "Integrating representative and discriminant models for object category detection," in *Proceedings of the IEEE International Conference on Computer Vision*, 2005, pp. 1363-1370.

[282] Y. Li, L. Gu, and T. Kanade, "A robust shape model for multi-view car alignment," in *Proceedings of the IEEE Computer Vision and Pattern Recognition Workshop*, 2009, pp. 2466-2473.

[283] M. F. Demirci, A. Shokoufandeh, and S. J. Dickinson, "Skeletal shape abstraction from examples," *IEEE Transactions on Pattern Analysis and Machine Intelligence,* vol. 31, pp. 944-952, 2009.

[284] M. Bergtholdt, J. Kappes, S. Schmidt, and C. Schnörr, "A study of parts-based object class detection using complete graphs," *International Journal of Computer Vision,* vol. 87, pp. 93-117, 2010.

[285] J. R. R. Uijlings, A. W. M. Smeulders, and R. J. H. Scha, "What is the spatial extent of an object?," in *Proceedings of the IEEE Conference on Computer Vision and Pattern Recognition*, 2009, pp. 770-777.

[286] F. Perronnin, "Universal and adapted vocabularies for generic visual categorization," *IEEE Transactions on Pattern Analysis and Machine Intelligence,* vol. 30, pp. 1243-1256, 2008.

[287] S. Lazebnik, C. Schmid, and J. Ponce, "Beyond bags of features: Spatial pyramid matching for recognizing natural scene categories," in *Proceedings of the IEEE Conference on Computer Vision and Pattern Recognition*, 2006, pp. 2169-2178.

[288] D. A. Ross and R. S. Zemel, "Learning parts-based representations of data," *Journal of Machine Learning Research,* vol. 7, pp. 2369-2397, 2006.

[289] S. Lazebnik and M. Raginsky, "Supervised learning of quantizer codebooks by information loss minimization," *IEEE Transactions on Pattern Analysis and Machine Intelligence,* vol. 31, pp. 1294-1309, 2009.

[290] B. Epshtein and S. Ullman, "Feature hierarchies for object classification," in *Proceedings of the IEEE International Conference on Computer Vision*, 2005, pp. 220-227.

[291] J. Gall and V. Lempitsky, "Class-specific hough forests for object detection," in *Proceedings of the IEEE Conference on Computer Vision and Pattern Recognition Workshops*, 2009, pp. 1022-1029.

[292] S. Basalamah, A. Bharath, and D. McRobbie, "Contrast marginalised gradient template matching," in *Lecture Notes in Computer Science* vol. 3023, ed, 2004, pp. 417-429.

[293] S. Belongie, J. Malik, and J. Puzicha, "Matching shapes," in *Proceedings of the IEEE International Conference on Computer Vision*, 2001, pp. 454-461.

[294] S. Belongie, J. Malik, and J. Puzicha, "Shape matching and object recognition using shape contexts," *IEEE Transactions on Pattern Analysis and Machine Intelligence,* vol. 24, pp. 509-522, 2002.

[295] A. C. Berg, T. L. Berg, and J. Malik, "Shape matching and object recognition using low distortion correspondences," in *Proceedings of the IEEE Conference on Computer Vision and Pattern Recognition*, 2005, pp. 26-33.

[296] S. Biswas, G. Aggarwal, and R. Chellappa, "Robust estimation of albedo for illumination-invariant matching and shape recovery," *IEEE Transactions on Pattern Analysis and Machine Intelligence,* vol. 31, pp. 884-899, 2009.

[297] G. Borgefors, "Hierarchical Chamfer matching: A parametric edge matching algorithm," *IEEE Transactions on Pattern Analysis and Machine Intelligence,* vol. 10, pp. 849-865, 1988.

[298] A. M. Bronstein, M. M. Bronstein, A. M. Bruckstein, and R. Kimmel, "Partial similarity of objects, or how to compare a centaur to a horse," *International Journal of Computer Vision,* vol. 84, pp. 163-183, 2009.

[299] R. Brunelli and T. Poggio, "Template matching: Matched spatial filters and beyond," *Pattern Recognition,* vol. 30, pp. 751-768, 1997.

[300]   G. J. Burghouts and J. M. Geusebroek, "Performance evaluation of local colour invariants," *Computer Vision and Image Understanding,* vol. 113, pp. 48-62, 2009.

[301]   J. R. Burrill, S. X. Wang, A. Barrow, M. Friedman, and M. Soffen, "Model-based matching using elliptical features," in *Proceedings of SPIE - The International Society for Optical Engineering*, 1996, pp. 87-97.

[302]   M. Ceccarelli and A. Petrosino, "The orientation matching approach to circular object detection," in *Proceedings of the IEEE International Conference on Image Processing*, 2001, pp. 712-715.

[303]   S. H. Chang, F. H. Cheng, W. H. Hsu, and G. Z. Wu, "Fast algorithm for point pattern matching: Invariant to translations, rotations and scale changes," *Pattern Recognition,* vol. 30, pp. 311-320, 1997.

[304]   F. H. Cheng, "Multi-stroke relaxation matching method for handwritten Chinese character recognition," *Pattern Recognition,* vol. 31, pp. 401-410, 1998.

[305]   Y. Chi and M. K. H. Leung, "A local structure matching approach for large image database retrieval," in *Proceedings of the International Conference on Image Analysis and Recognition*, Oporto, PORTUGAL, 2004, pp. 761-768.

[306]   T. H. Cho, "Object matching using generalized hough transform and chamfer matching," in *Lecture Notes in Computer Science* vol. 4099, ed, 2006, pp. 1253-1257.

[307]   O. Choi and I. S. Kweon, "Robust feature point matching by preserving local geometric consistency," *Computer Vision and Image Understanding,* vol. 113, pp. 726-742, 2009.

[308] P. F. Felzenszwalb, "Representation and detection of deformable shapes," *IEEE Transactions on Pattern Analysis and Machine Intelligence,* vol. 27, pp. 208-220, 2005.

[309] P. F. Felzenszwalb and J. D. Schwartz, "Hierarchical matching of deformable shapes," in *Proceedings of the IEEE Conference on Computer Vision and Pattern Recognition*, 2007, pp. 1-8.

[310] Y. S. Gao and M. K. H. Leung, "Line segment Hausdorff distance on face matching," *Pattern Recognition,* vol. 35, pp. 361-371, Feb 2002.

[311] H. Hakalahti, D. Harwood, and L. S. Davis, "Two-dimensional object recognition by matching local properties of contour points," *Pattern Recognition Letters,* vol. 2, pp. 227-234, 1984.

[312] F. Li, M. K. H. Leung, and X. Z. Yu, "A two-level matching scheme for speedy and accurate palmprint identification," in *Proceedings of the International Multimedia Modeling Conference*, Singapore, SINGAPORE, 2007, pp. 323-332.

[313] X. Lin, Z. Zhu, and W. Deng, "Stereo matching algorithm based on shape similarity for indoor environment model building," in *Proceedings of the IEEE International Conference on Robotics and Automation*, 1996, pp. 765-770.

[314] Z. Lin and L. S. Davis, "Shape-based human detection and segmentation via hierarchical part-template matching," *IEEE Transactions on Pattern Analysis and Machine Intelligence,* vol. 32, pp. 604-618, 2010.

[315] H.-C. Liu and M. D. Srinath, "Partial shape classification using contour matching in distance transformation," *IEEE Transactions on Pattern Analysis and Machine Intelligence,* vol. 12, pp. 1072-1079, 1990.

[316] G. Mori, S. Belongie, and J. Malik, "Efficient shape matching using shape contexts," *IEEE Transactions on Pattern Analysis and Machine Intelligence*, vol. 27, pp. 1832-1837, 2005.

[317] C. F. Olson, "A general method for geometric feature matching and model extraction," *International Journal of Computer Vision*, vol. 45, pp. 39-54, 2001.

[318] C. F. Olson and D. P. Huttenlocher, "Automatic target recognition by matching oriented edge pixels," *IEEE Transactions on Image Processing*, vol. 6, pp. 103-113, 1997.

[319] F. C. D. Tsai, "Robust affine invariant matching with application to line features," in *Proceedings of the IEEE Conference on Computer Vision and Pattern Recognition*, 1993, pp. 393-399.

[320] C. Xu, J. Liu, and X. Tang, "2D shape matching by contour flexibility," *IEEE Transactions on Pattern Analysis and Machine Intelligence*, vol. 31, pp. 180-186, 2009.

[321] X. Z. Yu and M. K. H. Leung, "Shape recognition using curve segment Hausdorff distance," in *Proceedings of the International Conference on Pattern Recognition*, Hong Kong, PEOPLES R CHINA, 2006, pp. 441-444.

[322] X. Z. Yu, M. K. H. Leung, and Y. S. Gao, "Hausdorff distance for shape matching," in *Proceedings of the IASTED International Conference on Visualization, Imaging, and Image Processing*, Marbella, SPAIN, 2004, pp. 819-824.

[323] F. Li and M. K. H. Leung, "Two-stage approach for palmprint identification using Hough transform and Hausdorff distance," in *Proceedings of the*

*International Conference on Control, Automation, Robotics and Vision*, Singapore, SINGAPORE, 2006, pp. 1302-1307.

[324] D. G. Sim and R. H. Park, "Two-dimensional object alignment based on the robust oriented Hausdorff similarity measure," *IEEE Transactions on Image Processing,* vol. 10, pp. 475-483, 2001.

[325] P. F. Felzenszwalb, "Learning models for object recognition," in *Proceedings of the IEEE Conference on Computer Vision and Pattern Recognition*, 2001, pp. 1056-1062.

[326] S. Lazebnik, C. Schmid, and J. Ponce, "A maximum entropy framework for part-based texture and object recognition," in *Proceedings of the IEEE International Conference on Computer Vision*, 2005, pp. 832-838.

# List of publications

## Publications incorporated in the thesis

### Journals

1. Dilip K. Prasad, Maylor K. H. Leung and Siu-Yeung Cho, "Edge curvature and convexity based ellipse detection method", *Pattern Recognition*, Vol. 45 issue 9, 2012.

2. Dilip K. Prasad, Maylor K. H. Leung, Chai Quek and Siu-Yeung Cho, "A novel framework for making dominant point detection methods non-parametric," *Image and Vision Computing*, vol. 30, pp. 843-859, 2012.

3. Dilip K. Prasad, Maylor K. H. Leung, and Chai Quek, "ElliFit: An unconstrained, non-iterative, least squares based geometric Ellipse Fitting method", *Pattern Recognition,* vol. 46, pp. 1449-1465, 2013.

4. Dilip K. Prasad, Maylor K. H. Leung, and Chai Quek, "DEB: Definite error bounded tangent estimator for digital curves", *IEEE Transactions on Image Processing*, 2013.(in review)

5. Dilip K. Prasad, Maylor K. H. Leung, and Chai Quek, "PRO: A novel Precision and Reliability based optimization method for dominant point detection", *Pattern Analysis and Applications.*(in review)

6. Dilip K. Prasad, C. Quek, and M. K. H. Leung, "Fast segmentation of sub-cellular organelles," *International Journal of Image Processing (IJIP)*, vol. 6, pp. 317-325, 2012.

7. Dilip K. Prasad, Maylor K. H. Leung, and Chai Quek, "Numerical stability and ellipticity bias in least square fitting of ellipses", in preparation, 2012.

8. Dilip K. Prasad, M. K. H. Leung, and C. Quek, "Object detection using geometric features and heirarchical object template," in preparation, 2012.

9. Dilip K. Prasad and Chai Quek, "Comparison of error bounds for non-parametric dominant point detection," *Pattern Recognition Letters*, 2012. (in review)

10. Dilip K. Prasad, "Assessing error bound for dominant point detection," *International Journal of Image Processing (IJIP)*, vol. 6, pp. 326-333, 2012.

11. D. K. Prasad, "Survey of the problem of object detection in real images," *International Journal of Image Processing (IJIP)*, vol. 6, 2012.

**Book Chapters**

12. Dilip K. Prasad and Maylor K.H. Leung, "Polygonal representation of digital curves", *Digital Image Processing*, InTech, Jan. 2012.

13. Dilip K. Prasad and Maylor K.H. Leung, "Methods for ellipse detection from edge maps of real images", *Machine Vision - Applications and Systems*, InTech, Mar. 2012.

**International Conferences**

14. Dilip K. Prasad, Chai Quek, and Maylor K.H. Leung, "A precise ellipse fitting method for noisy data," *International Conference on Image Analysis and Recognition (ICIAR 2012)*, Aveiro, Portugal, 25-27 June, 2012.

15. Dilip K. Prasad, Chai Quek, and Maylor K.H. Leung, "A non-heuristic dominant point detection based on suppression of break points," *International Conference on Image Analysis and Recognition (ICIAR 2012)*, Aveiro, Portugal, 25-27 June, 2012.

16. Dilip K. Prasad, Chai Quek, Maylor K.H. Leung, and Siu-Yeung Cho, "A parameter independent line fitting method," *First Asian Conference on Pattern Recognition (ACPR 2011)*, Beijing, China, 28-30 November 2011.

17. Dilip K. Prasad, Raj K. Gupta and Maylor K.H. Leung, "An error bounded tangent estimator for digitized elliptic curves," *16th IAPR International Conference on Discrete Geometry for Computer Imagery (DGCI 2011)*, Nancy, France, 6-8 April, 2011.

18. Dilip K. Prasad and Maylor K.H. Leung, "Error Analysis of Geometric Ellipse Detection Methods due to Quantization," *Fourth Pacific-Rim Symposium on Image and Video Technology (PSIVT 2010)*, Singapore, 14-17 November, 2010.

19. Dilip Kumar Prasad and Maylor K.H. Leung, "An ellipse detection method for real images," *25th International Conference of Image and Vision Computing New Zealand (IVCNZ 2010)*, Queenstown, New Zealand, 8-9 November, 2010.

20. Dilip K. Prasad and Maylor K.H. Leung, "An error bounded tangent estimator for digital curves," *25th International Conference of Image and Vision Computing New Zealand (IVCNZ 2010)*, Queenstown, New Zealand, 8-9 November, 2010.

21. Dilip K. Prasad and Maylor K.H. Leung, "Reliability/Precision Uncertainty in Shape Fitting Problems," *International Conference on Image Processing (ICIP 2010)*, Hong Kong, 26-29 September, 2010.

## Publications related to but not included in the thesis

### International Conferences

22. Dilip K. Prasad and Janusz A. Starzyk, "Object detection and representation in motivated conscious machines," *Conference on Decade of the Mind – VI*, Singapore, 18-20 October, 2010.

23. Dilip K. Prasad and Maylor K.H. Leung, "Clustering of Ellipses based on their Distinctiveness: An aid to Ellipse Detection Algorithms," *IEEE International Conference on Computer Science and Information Technology (ICCSIT 2010)*, Chengdu, China, 9-11 July, 2010.

24. Dilip K. Prasad and Maylor K.H. Leung, "A Hybrid Approach for Ellipse Detection in Real Images," *Second International Conference on Digital Image Processing (ICDIP 2010)*, Singapore, 26-28 February, 2010.

25. Dilip K. Prasad, "Application of Image Composition Analysis for Image Processing," *IMI International Workshop on Computational Photography and Aesthetics*, Singapore, 12-13 December, 2009.

## Other publications

### Journals

26. Dilip K. Prasad, "High Availability based Migration Analysis to Cloud Computing for High Growth Businesses," *International Journal of Computer Networks (IJCN)*, Vol. 4 issue 2, 2012.

27. Janusz A. Starzyk and Dilip K. Prasad, "A Computational Model on Machine Consciousness", *International Journal of Machine Consciousness*, Vol. 3 issue 2, 2011.

28. Dilip K. Prasad, "Rise of international schools in India," *International Journal of Education, Economics, and Development*, 2013.

### International Conferences

29. Dilip K. Prasad, "Adaptive traffic signal control system with cloud computing based online learning," *8th International Conference on Information, Communications, and Signal Processing (ICICS 2011)*, Singapore, 13-16 December, 2011.

30. Dilip K. Prasad and C. Krishna Prasath, "Reconfigurable Virtual Platform for Real Time Kernel" , *9th USENIX Symposium on Operating System Design and Implementation (OSDI 2010)*, Vancouver, BC, Canada, 4-6 October, 2010.

31. Dilip K. Prasad and Janusz A. Starzyk, "A Perspective on Machine Consciousness," *Second International Conference on Advanced Cognitive Technologies and Applications, COGNITIVE 2010*, Lisbon, Portugal, 21-26 November, 2010.

32. C. Krishna Prasath and Dilip K. Prasad, "Design and Development of a Reconfigurable Virtual Platform for Real Time Kernel," *Second International Conference on Software Technology and Engineering (ICSTE 2010)*, San Juan, Puerto Rico, USA, 3-5 October, 2010.

33. Janusz A. Starzyk and Dilip K. Prasad, "Machine Consciousness: A Computational Model," *Third International ICSC Symposium on Models of Consciousness, BICS 2010*, Madrid, Spain, 14-16 July, 2010.

34. Dilip K. Prasad, Chai H. Quek, and Maylor K.H. Leung, "A Hybrid Approach for Breast Tissue Data Classification," *IEEE TENCON 2009*, Singapore, 23-26 November, 2009.